\documentclass[nohyperref]{article}

\usepackage{microtype}
\usepackage{graphicx}
\usepackage{booktabs} 

\usepackage{epstopdf}
\usepackage{epsfig} 

\usepackage{hyperref}

\usepackage[accepted]{icml2023}

\usepackage{amsmath}
\usepackage{amssymb}
\usepackage{mathtools}
\usepackage{amsthm}

\usepackage[capitalize,noabbrev]{cleveref}

\theoremstyle{plain}

\theoremstyle{definition}

\theoremstyle{remark}

\usepackage[textsize=tiny]{todonotes}

\usepackage{subcaption}     
\usepackage{multirow}
\usepackage{bbm}
\usepackage{tikz}
\usepackage{pifont}
\newcommand{\xmark}{\ding{55}} 
\usepackage{hhline}

\usepackage[capitalize]{cleveref}
\Crefname{section}{Section}{Sections}
\Crefname{table}{Table}{Tables}
\Crefname{figure}{Figure}{Figures}

\usepackage{xcolor}

\usepackage{xspace}
\makeatletter
\DeclareRobustCommand\onedot{\futurelet\@let@token\@onedot}
\def\@onedot{\ifx\@let@token.\else.\null\fi\xspace}
\def\eg{\emph{e.g}\onedot} 
\def\ie{\emph{i.e}\onedot}

\makeatother
\usepackage{wrapfig}

\usepackage{enumitem}

\icmltitlerunning{A Closer Look at the Intervention Procedure of Concept Bottleneck Models}

\begin{document}

\twocolumn[
\icmltitle{A Closer Look at the Intervention Procedure of Concept Bottleneck Models}



\icmlsetsymbol{equal}{*}
\icmlsetsymbol{norelation}{*}

\begin{icmlauthorlist}
\icmlauthor{Sungbin Shin}{postech}
\icmlauthor{Yohan Jo}{amazon,norelation}
\icmlauthor{Sungsoo Ahn}{postech}
\icmlauthor{Namhoon Lee}{postech}
\end{icmlauthorlist}

\icmlaffiliation{postech}{POSTECH, South Korea}
\icmlaffiliation{amazon}{Amazon -- Alexa AI, USA}

\icmlcorrespondingauthor{Sungbin Shin}{ssbin4@postech.ac.kr}

\icmlkeywords{Machine Learning, ICML}

\vskip 0.3in
]



\newcommand{\Authorfootnote}{\textsuperscript{*}This work is not associated with Amazon.}
\printAffiliationsAndNotice{\Authorfootnote} 

\begin{abstract}
  Concept bottleneck models (CBMs) are a class of interpretable neural network models that predict the target response of a given input based on its high-level concepts.
  Unlike the standard end-to-end models, CBMs enable domain experts to intervene on the predicted concepts and rectify any mistakes at test time, so that more accurate task predictions can be made at the end.
  While such intervenability provides a powerful avenue of control, many aspects of the intervention procedure remain rather unexplored.
  In this work, we develop various ways of selecting intervening concepts to improve the intervention effectiveness and conduct an array of in-depth analyses as to how they evolve under different circumstances.
  Specifically, we find that an informed intervention strategy can reduce the task error more than ten times compared to the current baseline under the same amount of intervention counts in realistic settings, and yet, this can vary quite significantly when taking into account different intervention granularity.
  We verify our findings through comprehensive evaluations, not only on the standard real datasets, but also on synthetic datasets that we generate based on a set of different causal graphs.
  We further discover some major pitfalls of the current practices which, without a proper addressing, raise concerns on reliability and fairness of the intervention procedure.
\end{abstract}

\section{Introduction}
\begin{figure}[!th]
  \centering
  \begin{subfigure}[b]{0.45\linewidth}
    \centering
    \includegraphics[width=\linewidth]{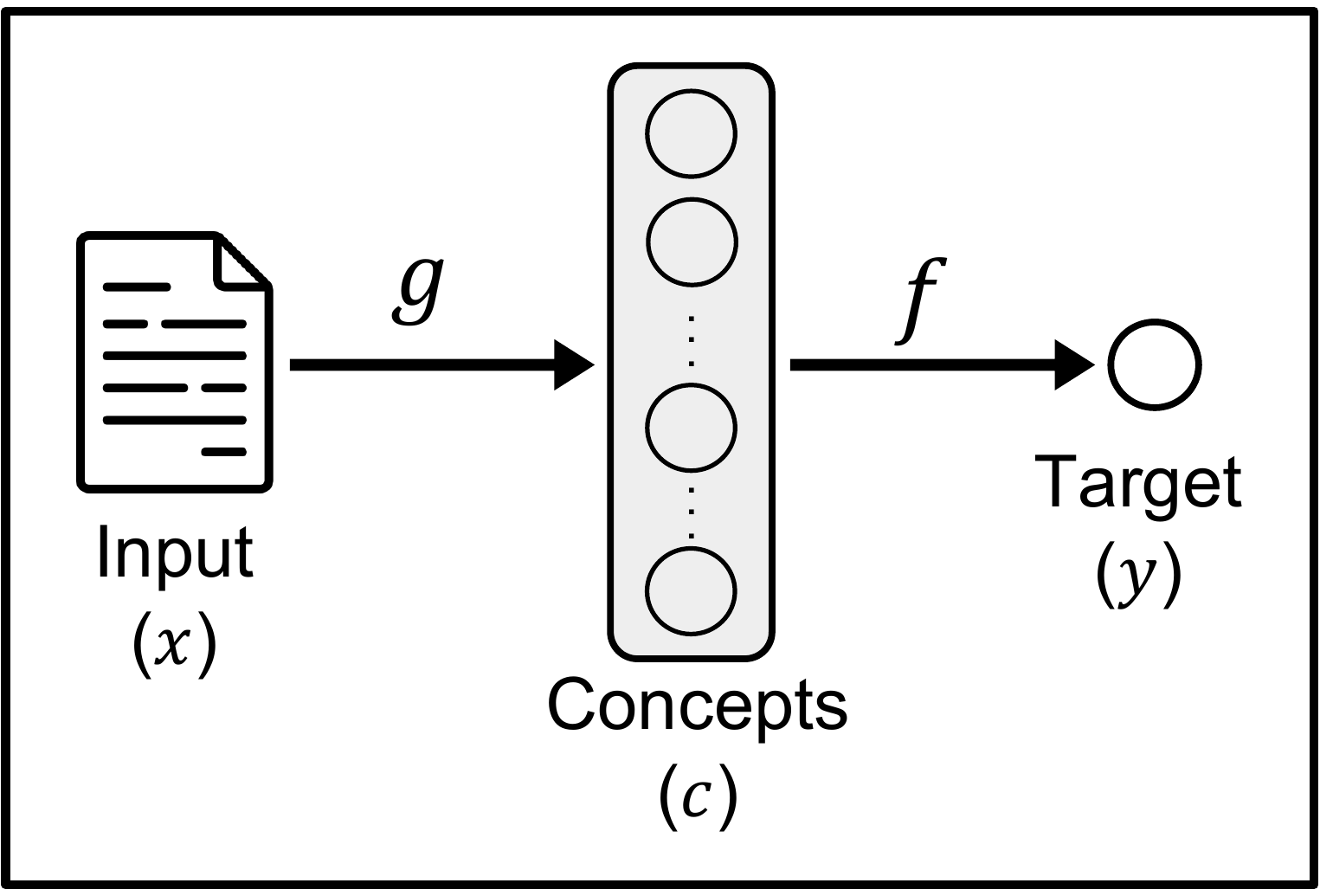}
    \vspace{-0.1em}
    \caption{Diagram of CBMs}
    \label{fig:cbm_fig}
  \end{subfigure}
  \hspace*{1em}
  \begin{subfigure}[b]{0.45\linewidth}
    \centering
    \includegraphics[width=\linewidth]{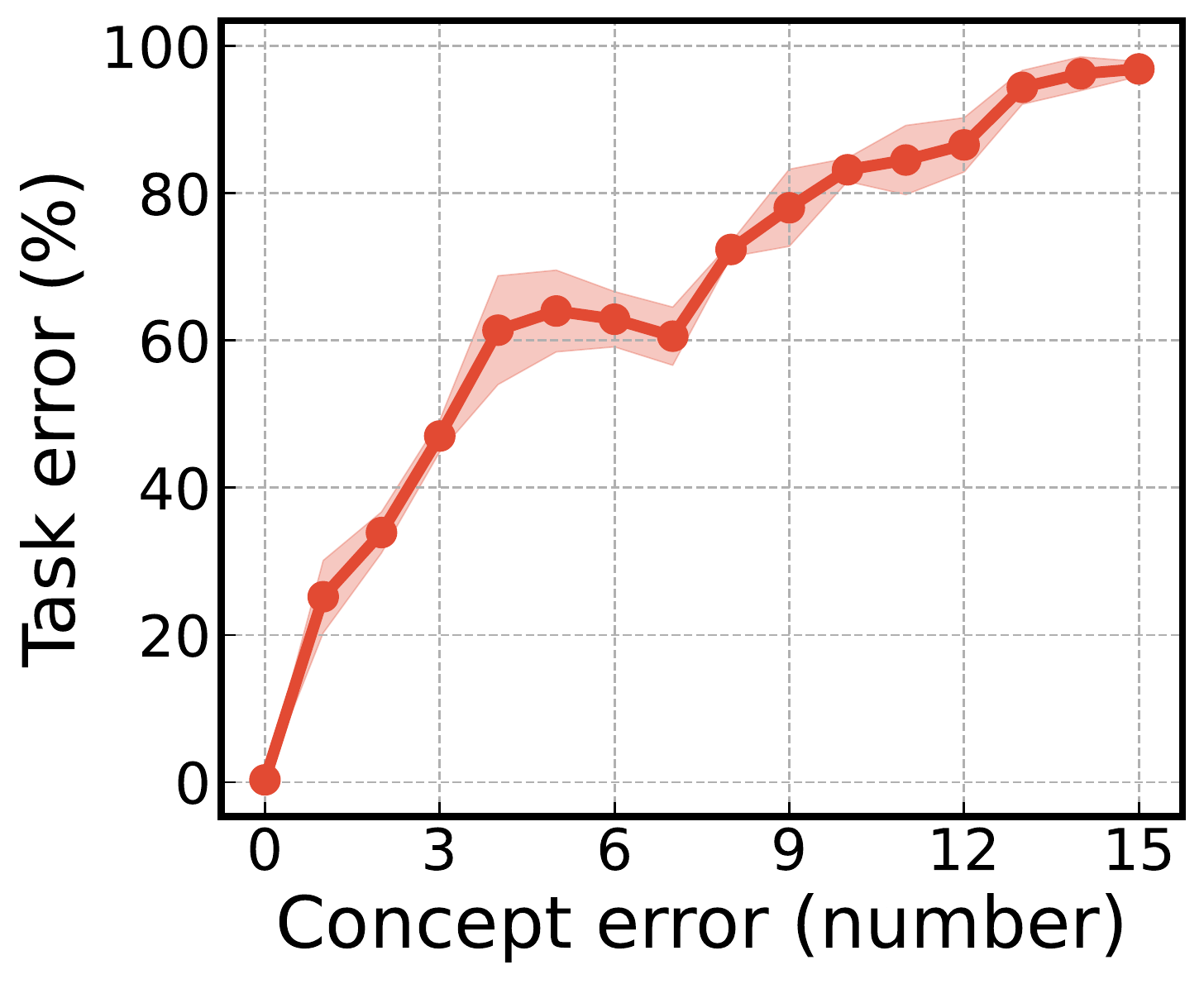}
    \caption{Task vs. Concept errors}
    \label{fig:yerr_by_cerr}
  \end{subfigure}
  \caption{
      (a)
      Given input data CBMs first predict its concepts ($g: x \rightarrow c$), and then based on which it makes a subsequent prediction for the target response ($f: c \rightarrow y$).
      (b)
      Average task error (mis-classification rate) vs. the number of incorrectly predicted concepts (on the CUB dataset).
      The task error increases rapidly as more mistakes are made in concept prediction;
      \eg, making a single mistake yields $25\%$ increase in task error.
  }
  \label{fig:cbm_intro}
\end{figure}

While deep learning has made rapid strides in recent years \citep{lecun2015deep,jordan2015machine}, the standard neural network models are not quite explainable, in that their decision-making process is neither straightforward to account for nor easy to control.
To tackle this issue, various interpretable models have been proposed including, for example, those using concept activation vectors \citep{kim2018interpretability,ghorbani2019towards}, relating pixel contributions to image classification \citep{zhou2016learning,selvaraju2017grad}, or building intrinsically interpretable architectures \citep{alvarez2018towards}.

Concept bottleneck models (CBMs) are among these to empower interpretability \citep{koh2020concept,bahadori2020debiasing,margeloiu2021concept,mahinpei2021promises,sawada2022concept,zarlenga2022concept}.
Unlike standard end-to-end models, CBMs work in two steps:
they first predict human-interpretable properties of a given input called \emph{concepts}, and based on which, they subsequently make the final prediction for the given task.
For instance, CBMs may classify the species of a bird based on its wing pattern or leg color rather than straight from the raw pixel values (see \cref{fig:cbm_fig}).

Revisited recently by \citet{koh2020concept}, this classic idea further facilitates human-model interaction in addition to plain interpretability, in that it allows one to \emph{intervene} on the predicted concepts at test time, such that the subsequent prediction is made based on the rectified concept values.
Notably, such intervention must be treated attentively as we find that correcting only a small number of mistakes on mis-predicted concepts can lead to a significant increase in the task performance (see \cref{fig:yerr_by_cerr}).
Considering the high cost of intervention, \ie, having domain experts go over each concept requires tremendous effort, this result further indicates the necessity of efficient intervention procedures to ensure the utility of CBMs.

\begin{table}[!t]
    \centering
    \resizebox{\linewidth}{!}{
    \begin{tabular}{l c c c c c c}
      \toprule
      Work & Selection & Cost & Level & Imp. & Data & Rel. \\
      \midrule
      \citet{koh2020concept}          & \xmark   & \xmark   & $\triangle$ & $\triangle$   & \xmark  & \xmark \\    
      \citet{chauhan2022interactive}  & \checkmark & $\triangle$ & $\triangle$ & $\triangle$   & \xmark  & \xmark\\
      \citet{sheth2022learning}       & \checkmark & \xmark & $\triangle$ & \xmark        & \xmark  & \xmark \\  
      Ours & \checkmark & \checkmark & \checkmark & \checkmark & \checkmark & \checkmark\\
      \bottomrule
  \end{tabular}
  }
  \caption{
    Comparison between the studies on intervention strategy of CBMs.
    $\triangle$ represents that the corresponding work provides only partial evaluations.
    \emph{Selection} and \emph{Cost} represent concept selection criteria and their analysis in terms of theoretical cost as will be discussed in \cref{sec:results-criteria}.
    We study the effects of \emph{Level, Implementation} and \emph{Data} on intervention effectiveness in \cref{sec:restuls-levels,sec:exp_train_inf_strategies,sec:ablation-dataset}.
    \emph{Reliability} of intervention practice is discussed in \cref{sec:pitfalls}.
  }
  \label{tab:comparison_works}
\end{table}

Despite the great potential, the intervention procedure of CBMs has not been studied much in the literature, quite surprisingly. 
For example, previous works tend to focus on increasing task performance \citep{sawada2022concept, zarlenga2022concept} and addressing the problem of confounding factors \citep{bahadori2020debiasing} or information leakage \citep{margeloiu2021concept,mahinpei2021promises,havasi2022addressing,marconato2022glancenets}.
While a few concurrent works suggest new intervention methods \citep{chauhan2022interactive, sheth2022learning}, we find that many critical aspects of the intervention procedure still remain unexplored (see \cref{tab:comparison_works}).

Our contributions are summarized as follows.
First of all, we develop various concept selection criteria as new intervention strategies, improving the intervention performance of CBMs quite dramatically given the same amount of intervention counts.
We also provide extensive evaluations to analyze these criteria under a wide variety of experimental settings considering the theoretical cost of each criterion, levels of intervention related to test-time environments, and how to train these models or conceptualize the concept predictions.
We further develop a new framework to generate synthetic data using diverse causal graphs and conduct fully controlled experiments to verify the effectiveness of intervention on varying data.
These results reveal that data characteristics as well as intervention granularity can affect the intervention procedure quite significantly.
Finally, we identify some pitfalls of the current intervention practices, which helps to take a step toward building trustworthy and responsible interpretable models.
\section{Related Work}

Since the seminal work of \citet{koh2020concept}, CBMs have evolved in many different ways.
\citet{bahadori2020debiasing} develop a debiased CBM to remove the impact of confounding information to secure causality.
\citet{sawada2022concept} augment CBMs with unsupervised concepts to improve task performance.
\citet{mahinpei2021promises,margeloiu2021concept} suggest addressing the information leakage problem in CBMs to improve interpretability of learned concepts, while \citet{marconato2022glancenets, havasi2022addressing} design new CBMs based on disentangled representations or autoregressive models.
\citet{zarlenga2022concept} proposes to learn semantically meaningful concepts using concept embedding models to push the accuracy-interpretability trade-off.
Both \citet{chauhan2022interactive} and \citet{sheth2022learning} present uncertainty based intervention methods to determine which concepts to intervene on.
We remark that previous work is mostly focused on developing CBM variants for high task performance from model-centric perspectives, whereas our work provides in-depth analyses and comprehensive evaluations on the intervention procedure of the standard CBMs in greater granularity.
\section{Intervention Strategies}
\label{sec:int_strategies}

\subsection{Preliminary}
\label{subsec:preliminary}

Let $x \in \mathbb{R}^{d}$, $c \in \{0,1\}^k$, $y \in \mathcal{Y}$ be input data, binary concepts, and target responses, respectively;
here, $d$ and $k$ denote the dimensionality of input data and cardinality of concepts, and we assume $\mathcal{Y}$ encodes categorical distribution for classification tasks.
Given some input data (\eg, an image), a CBM first predicts its concepts (\eg, existing attributes in the given image) using a concept predictor $g$ and subsequently target response (\eg, class of the image) using a target predictor $f$:
\ie, first $\hat{c} = g(x)$ then $\hat{y} = f(\hat{c})$, where $\hat{c}$ and $\hat{y}$ are predictions of concepts and target response. 

In this process, one can intervene on a set of concepts $\mathcal{S} \subseteq \{1, \cdots, k\}$ so that the final prediction can be made based on rectified concept values, \ie, $\hat{y} = f(\tilde{c})$ where $\tilde{c} = \{\hat{c}_{\setminus\mathcal{S}}, c_{\mathcal{S}}\}$ denotes the updated concept values partly rectified on $\mathcal{S}$ with $\hat{c}_{\setminus\mathcal{S}}$ referring to the predicted concept values excluding $\mathcal{S}$.

\subsection{Concept Selection Criteria}
\label{sec:concept_selection_criteria}

How should one select which concepts to intervene on?
This is a fundamental question to be answered in order to legitimize CBMs in practice since intervention incurs the cost of employing experts, which would increase as with the number of intervening concepts $|\mathcal{S}|$.
In principle, one would select a concept by which it leads to the largest increase in the task performance.
To address this question and investigate the effectiveness of intervention procedure in current practice, we develop various concept selection criteria for which a selection score $s_i$ for $i$-th concept is defined.
Then, intervening concepts will be done based on the decreasing order of these scores.

\textbf{Random} (\textsc{rand})\quad
It selects concepts uniformly at random as in \citet{koh2020concept}.
We can treat this method as assigning a random score for each concept, \ie, $s_i \sim \mathcal{U}_{[0, 1]}$.
It will serve as a baseline to study the effectiveness of concept selection criteria.

\textbf{Uncertainty of concept prediction} (\textsc{ucp})\quad
It selects concepts with the highest uncertainty of concept prediction.
Specifically, it defines $s_i = \mathcal{H} (\hat{c_i})$ where $\mathcal{H}$ is the entropy function.
When the concepts are binary, it follows that $s_i = 1/|\hat{c_i} - 0.5|$ as in \citet{lewis1994heterogeneous,lewis1995sequential}.
Intuitively, uncertain concepts may have an adverse influence on making the correct target prediction, and thus, they are fixed first by this criterion.

\textbf{Loss on concept prediction} (\textsc{lcp})\quad
It selects concepts with the largest loss on concept prediction compared to the ground-truth.
Specifically, it defines $s_i = \lvert \hat{c}_i - c_i \rvert$.
This scheme can be advantageous to increasing task performance since a low concept prediction error is likely to lead to a correct target prediction.
Nonetheless, this score is unavailable in practice as the ground-truth is unknown at test time. 

\begin{table}[!t]
  \centering
  \footnotesize
  \begin{tabular}{l c c c}
      \toprule
      Criteria & $N_g$ & $N_f$ & Cost in complexity\\
      \midrule
      \textsc{rand}  & $1$ & $1$   & $\mathcal{O} \big(\tau_g + \tau_f + n \tau_i \big)$\\ 
      \textsc{ucp}   & $1$ & $1$  & $\mathcal{O} \big( \tau_g + \tau_f + n \tau_i \big)$\\  
      \textsc{lcp}   & $1$ & $1$ & $\mathcal{O} \big( \tau_g + \tau_f + n \tau_i \big)$\\ 
      \textsc{cctp}  & $1$ & $3$ & $\mathcal{O} \big( \tau_g + 3 \tau_f + n \tau_i \big)$\\  
      \textsc{ectp}  & $1$ & $2k + 2$ & $\mathcal{O} \big( \tau_g + (2k + 2) \tau_f + n \tau_i \big)$\\  
      \textsc{eudtp} & $1$ & $2k + 2$ & $\mathcal{O} \big( \tau_g + (2k + 2) \tau_f + n \tau_i \big)$\\  
      \bottomrule
  \end{tabular}
  \caption{
      Theoretical cost of employing concept selection criteria to make final prediction with $n$ number of intervened concepts.
      $N_g$ and $N_f$ refer to the number of forward/backward passes to run $g$ and $f$, respectively.
  }
  \label{tab:intervention_methods_cost}
\end{table}

\textbf{Contribution of concept on target prediction} (\textsc{cctp})\quad
It selects concepts with the highest contribution on target prediction.
Specifically, it sums up the contribution as $s_i = \sum _{j=1} ^ {M} \big\lvert \hat{c}_i \frac{\partial f_j}{\partial \hat{c}_i} \big\rvert$ where $f_j$ is the output related to $j$-th target class and $M$ is the number of classes.
This scheme is inspired by methods to explain neural network predictions \citep{selvaraju2017grad}.

\textbf{Expected change in target prediction} (\textsc{ectp})\quad
It selects concepts with the highest expected change in the target predictive distribution with respect to intervention.
Specifically, it defines $s_i = (1 - \hat{c}_i) D_{\text{KL}}(\hat{y}_{\hat{c}_i = 0} \Vert \hat{y}) + \hat{c}_i D_{\text{KL}} (\hat{y}_{\hat{c}_i = 1} \Vert \hat{y})$ where $D_{\text{KL}}$ refers to the Kullback-Leibler divergence, and $\hat{y}_{\hat{c}_i = 0}$ and $\hat{y}_{\hat{c}_i = 1}$ refer to the new target prediction with $\hat{c}_i$ being intervened to be $0$ and $1$, respectively.
The intuition behind this scheme is that it would be better to intervene on those concepts whose rectification leads to a large expected change in target prediction \citep{settles2007multiple}.

\textbf{Expected uncertainty decrease in target prediction} (\textsc{eudtp})\quad
It selects concepts with the largest expected entropy decrease in target predictive distribution with respect to intervention.
Specifically, it defines $s_i = (1 - \hat{c}_i) \mathcal{H}(\hat{y}_{\hat{c}_i=0}) + \hat{c}_i\mathcal{H}(\hat{y}_{\hat{c}_i=1}) - \mathcal{H}(\hat{y})$.
Intuitively, it penalizes the concepts whose expected decrease in the target prediction entropy is low when intervened \citep{guo2007optimistic}.

\subsubsection{Cost of Intervention}
\label{sec:cost}

Note that the cost of intervention may differ by the choice of concept selection criteria.
Specifically, let the theoretical cost of intervening on a concept be $\tau_i$ (\eg, the time for an expert to look at the input and fix its attribute), and the theoretical cost of making inference on $g$ and $f$ be $\tau_g$ and $\tau_f$, respectively.
Then, the total cost of utilizing \textsc{cctp} needed up to making the final prediction with $n$ number of intervened concepts, for example, would be $\mathcal{O} (\tau_g + 3\tau_f + n\tau_i)$; here we assume that the cost of the backward pass on $f$ is the same as $\tau_f$.
We summarize the cost of all concept selection criteria in \cref{tab:intervention_methods_cost}.

\subsection{Levels of Intervention}

\begin{figure}[!t]
  \begin{subfigure}{0.43\linewidth}
      \centering
      \includegraphics[width=\linewidth]{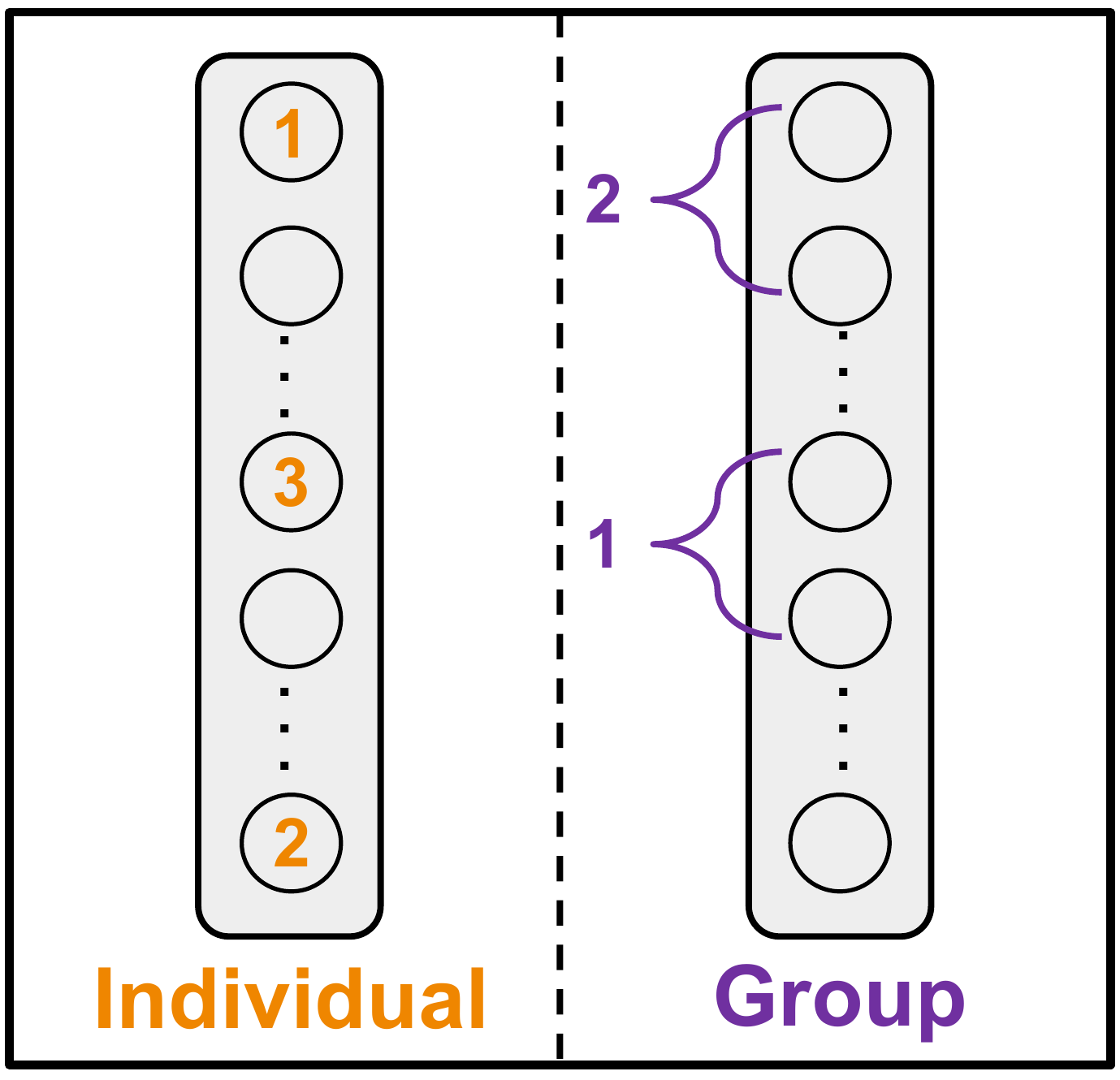}
      \caption{Individual vs. Group}
      \label{fig:level_ind_group}
  \end{subfigure}
  \hspace*{\fill}
  \begin{subfigure}{0.53\linewidth}
      \centering
      \includegraphics[width=\linewidth]{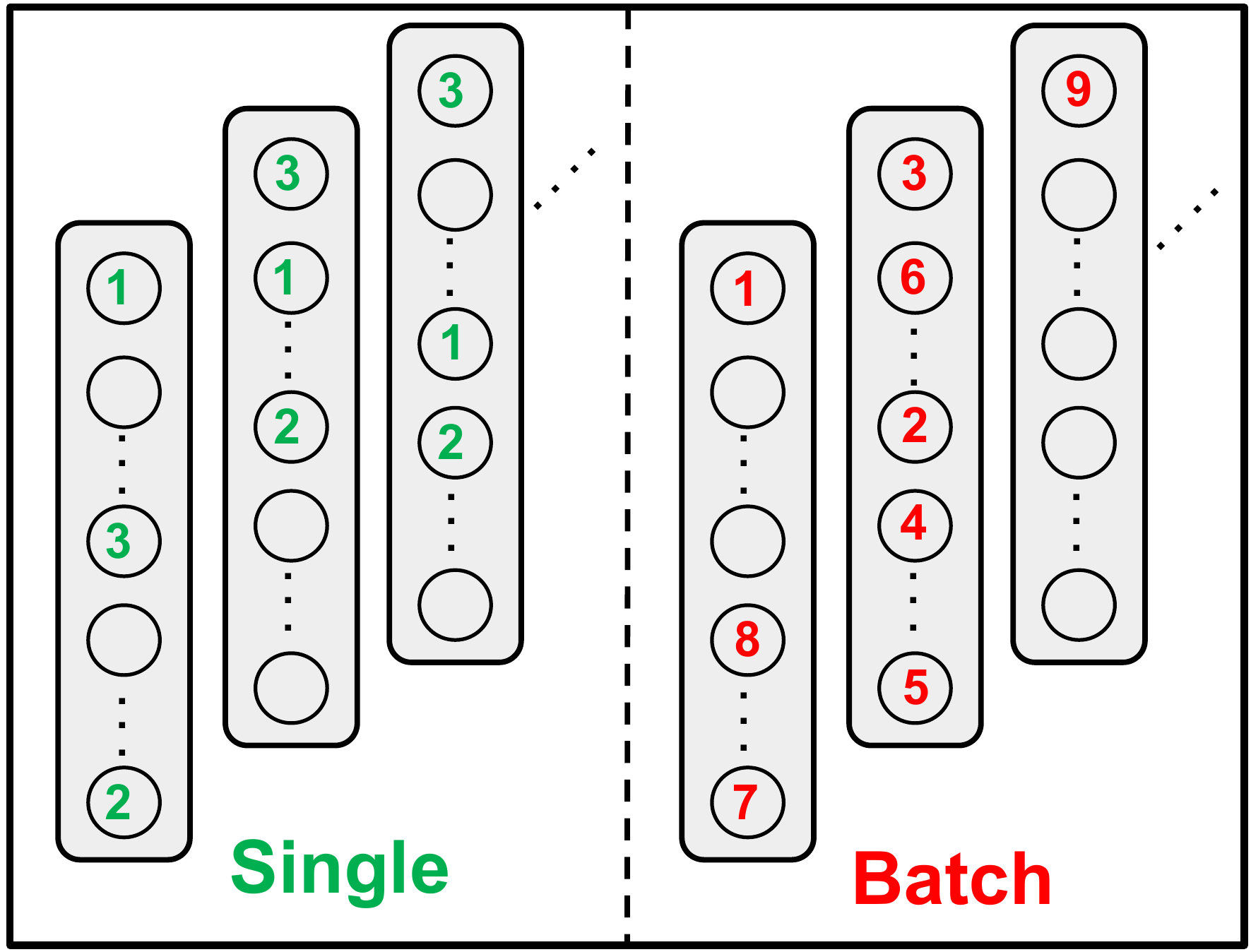}
      \caption{Single vs. Batch}
      \label{fig:level_single_batch}
  \end{subfigure}
  \caption{
      Different levels of intervention conducted on concepts.
      Each number represents the order of intervention.
  }
  \label{fig:intervention_levels}
\end{figure}

We find that intervention can be done at different levels given some auxiliary information about the structure of concepts or economic constraints put on practitioners.
For example, it is often the case that datasets used to train CBMs have the grouping information for related concepts \citep{wah2011caltech}.
Another situation worth consideration is where one has access to a batch of data to process with a budget constraint, and the goal is to maximize the overall task performance while minimizing the intervention effort (\eg, examining medical images in a hospital).
Taking into account these scenarios, we extend the intervention procedure at various levels to study the effectiveness of concept selection criteria.

\textbf{Individual vs. Group intervention}\quad
Intervention can be done depending on concept association (see \cref{fig:level_ind_group}):
\begin{itemize}[noitemsep, leftmargin=10pt, topsep=0pt]
    \item Individual (\textsc{i}):
        Concepts are assumed to be independent of each other and thus selected individually one at a time.
    \item Group (\textsc{g}):
        A group of related concepts is selected at once whose association information is subject to datasets.
        The selection score is computed by taking the average of selection scores of individual concepts within group.
\end{itemize}

\begin{figure*}[t!]
  \begin{subfigure}{0.3\linewidth}
    \centering
    \includegraphics[width=\linewidth]{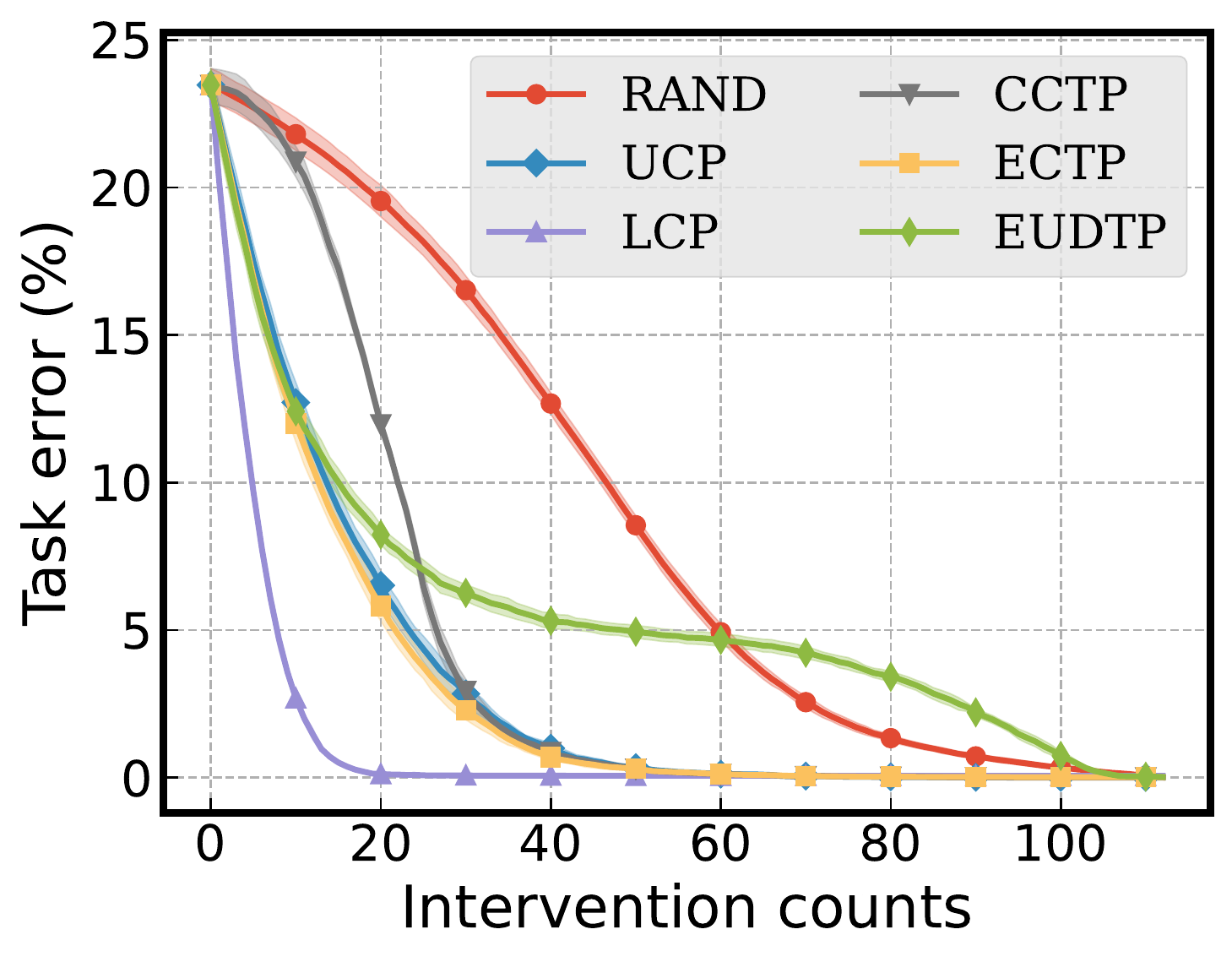}
    \caption{CUB}
    \label{fig:cub_result_main}
  \end{subfigure}
  \hspace*{\fill}
  \begin{subfigure}{0.3\linewidth}
    \centering
    \includegraphics[width=\linewidth]{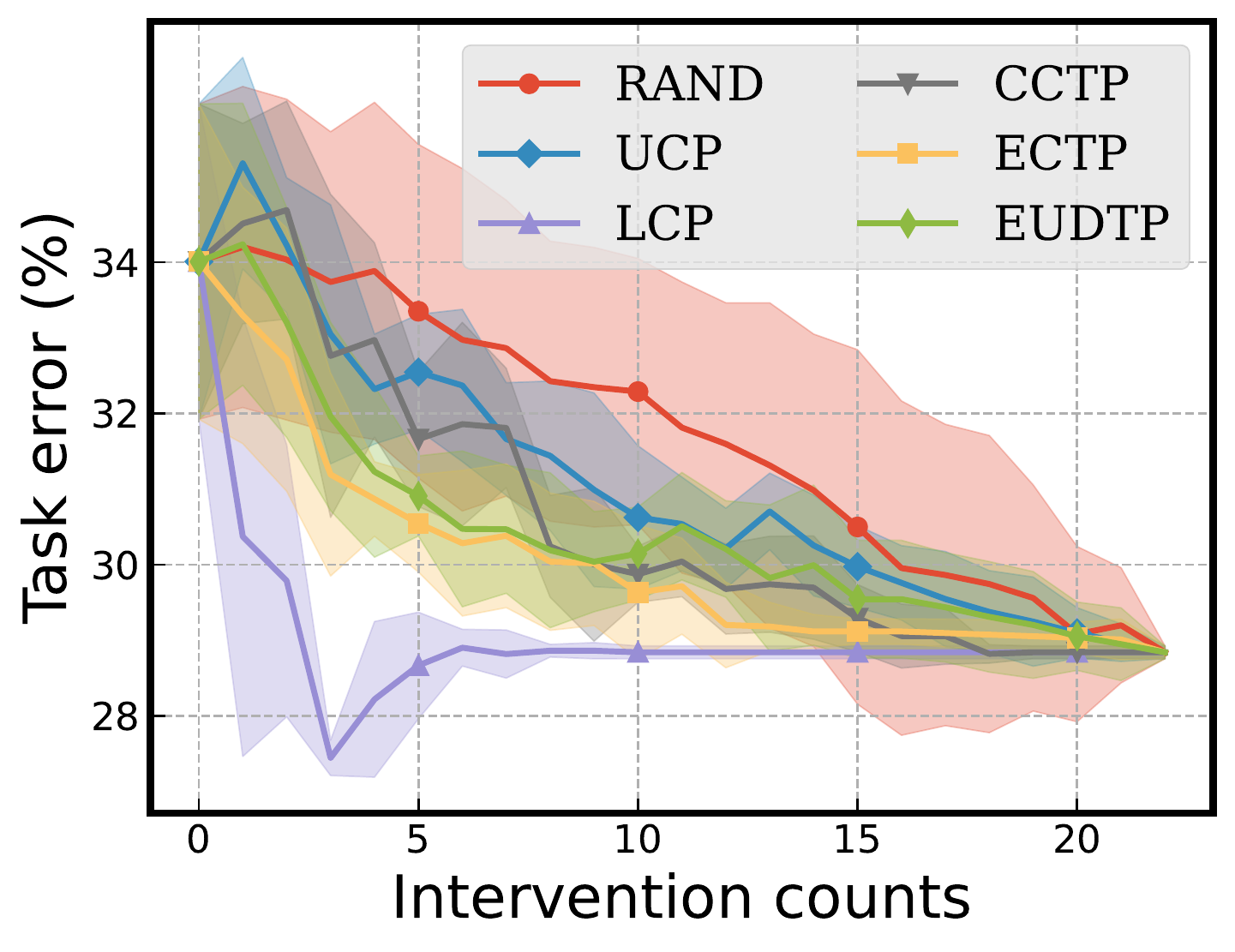}
    \caption{SkinCon}
    \label{fig:skincon_result_main}
  \end{subfigure}
  \hspace*{\fill}
  \begin{subfigure}{0.3\linewidth}
    \centering
    \includegraphics[width=\linewidth]{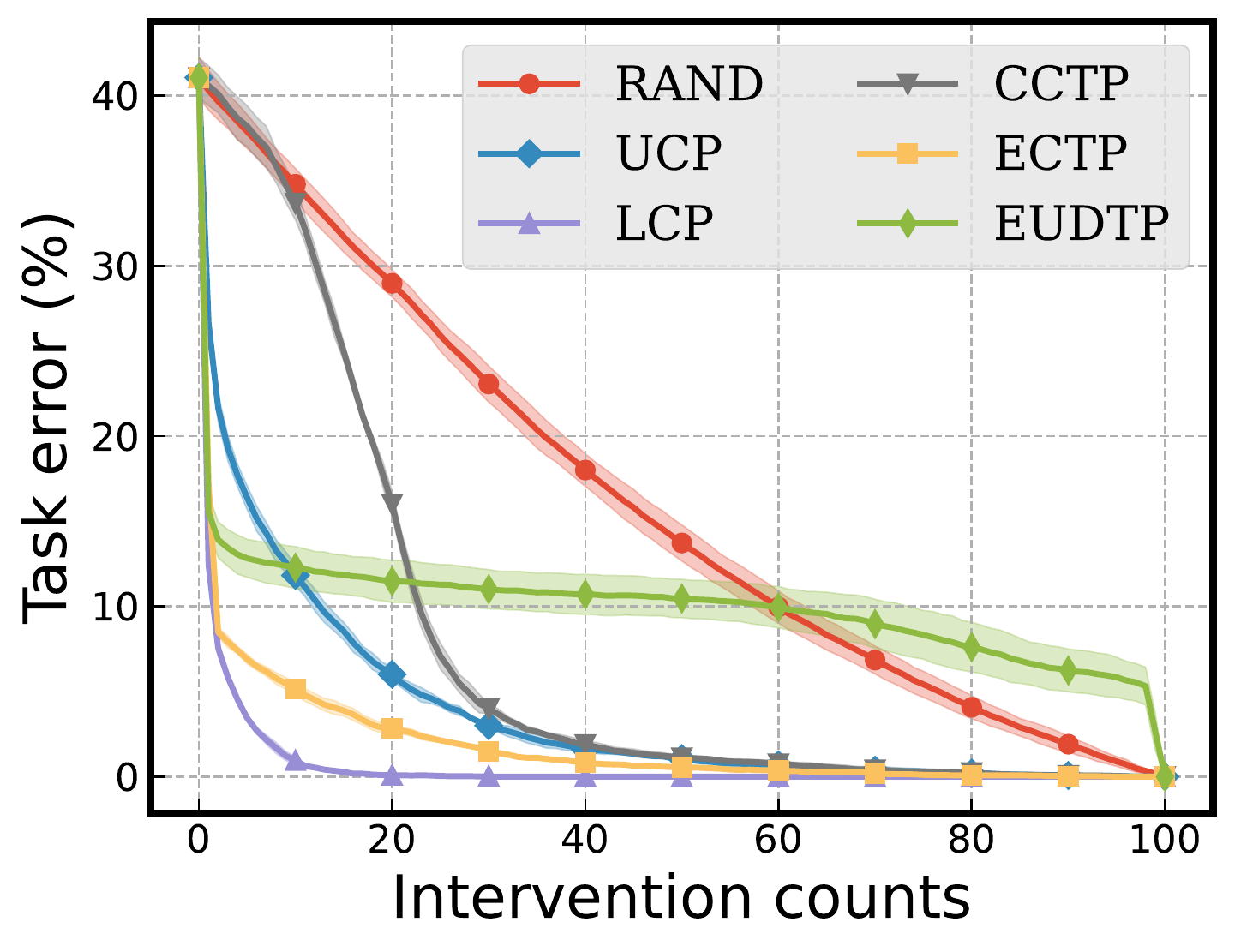}
    \caption{Synthetic}
    \label{fig:synthetic_result_main}
  \end{subfigure}
  \caption{
    Intervention effectiveness of concept selection criteria (task error vs. number of concepts corrected by intervention) measured on \textsc{I+S} level.
    A more effective method would reduce the error more for the same number of concepts intervened.
  }
  \label{fig:result-main}
\end{figure*}

\textbf{Single vs. Batch intervention}\quad
Intervention can be done depending on data accessibility (see \cref{fig:level_single_batch}):
\begin{itemize}[noitemsep, leftmargin=10pt, topsep=0pt]
    \item Single (\textsc{s}):
        Every test case is allocated with the same amount of intervention budget (\eg, intervention counts).
        This could be useful for online systems where each test data comes in sequentially, and experts need to process as many cases as possible under a budget constraint.
    \item Batch (\textsc{b}):
        A batch of test cases shares a total intervention budget.
        This scheme could be particularly useful when the concept prediction is imbalanced toward easy cases, and one wants to focus on intervening on hard cases so as to maximize the overall task performance.
\end{itemize}
\section{Evaluating Intervention Strategies}
\label{sec:results}

\subsection{Experiment Settings}

\textbf{Dataset}\quad
We experiment with three datasets: (1) CUB \citep{wah2011caltech} -- the standard dataset used to study CBMs, (2) SkinCon \citep{daneshjouskincon} -- a medical dataset used to build interpretable models, and (3) Synthetic -- the synthetic datasets we generate based on different causal graphs to conduct a wide range of controlled experiments.
Extensive details of these datasets including preprocessing, label characteristics, data splits, and the generation process are provided in \cref{app:datasets}.

\textbf{Implementation}\quad
We follow the standard implementation protocols as in previous works.
The full details including model architectures and optimization hyperparameters are provided in \cref{app:implementation}.
Our code is available at \url{https://github.com/ssbin4/Closer-Intervention-CBM}.

\textbf{Training}\quad

We consider the following training strategies similarly to \citet{koh2020concept}:
\begin{itemize}[noitemsep, leftmargin=10pt, topsep=0pt]
    \item \textsc{ind}:
        $g$ and $f$ are trained \underline{ind}ependently of each other.
        $f$ always takes ground-truth concept values as input.
    \item \textsc{seq}:
        $g$ and $f$ are trained \underline{seq}uentially, $g$ first and $f$ next.
        $f$ takes predicted concept values as input from trained $g$.
    \item \textsc{jnt}:
        $g$ and $f$ are trained \underline{j}oi\underline{nt}ly at the same time as a multi-objective. This results in increased initial task accuracy but comes with the price of decreased intervention effectiveness \citep{koh2020concept}.
    \item \textsc{jnt+p}:
        similar to \textsc{jnt} but the input to $f$ is sigmoid-activated \underline{p}robability distribution rather than logits.
\end{itemize}
\textbf{Conceptualization}\quad
We consider different forms of concept predictions as input to the target predictor at inference:
\begin{itemize}[noitemsep, leftmargin=10pt, topsep=0pt]
    \item \textsc{soft}:
        $f$ takes real values of $\hat{c} \in [0, 1]^k$ as \underline{soft} representation of concepts \citep{koh2020concept}.
    \item \textsc{hard}:
        $f$ takes binary values of $\hat{c} \in \{0, 1\}^k$ as \underline{hard} representation of concepts based on $\mathbbm{1}[\hat{c} \geq 0.5]$ \citep{mahinpei2021promises}.
        This prevents information leakage \citep{havasi2022addressing} in exchange for decreased prediction performance.
    \item \textsc{samp}:
        $m$ random \underline{samp}les are drawn by treating the soft concept prediction scores as a probability distribution, and the target prediction is made as an ensemble, \ie, $\hat{y} = \frac{1}{m} \sum_{i=1}^{m} f(\hat{c})$ where $\hat{c}$ is binarized concept prediction \citep{havasi2022addressing}.
        We use $m=5$ for the experiments.
\end{itemize}

\begin{figure*}[!t]
  \begin{subfigure}{0.16\linewidth}
    \centering
    \includegraphics[width=\linewidth]{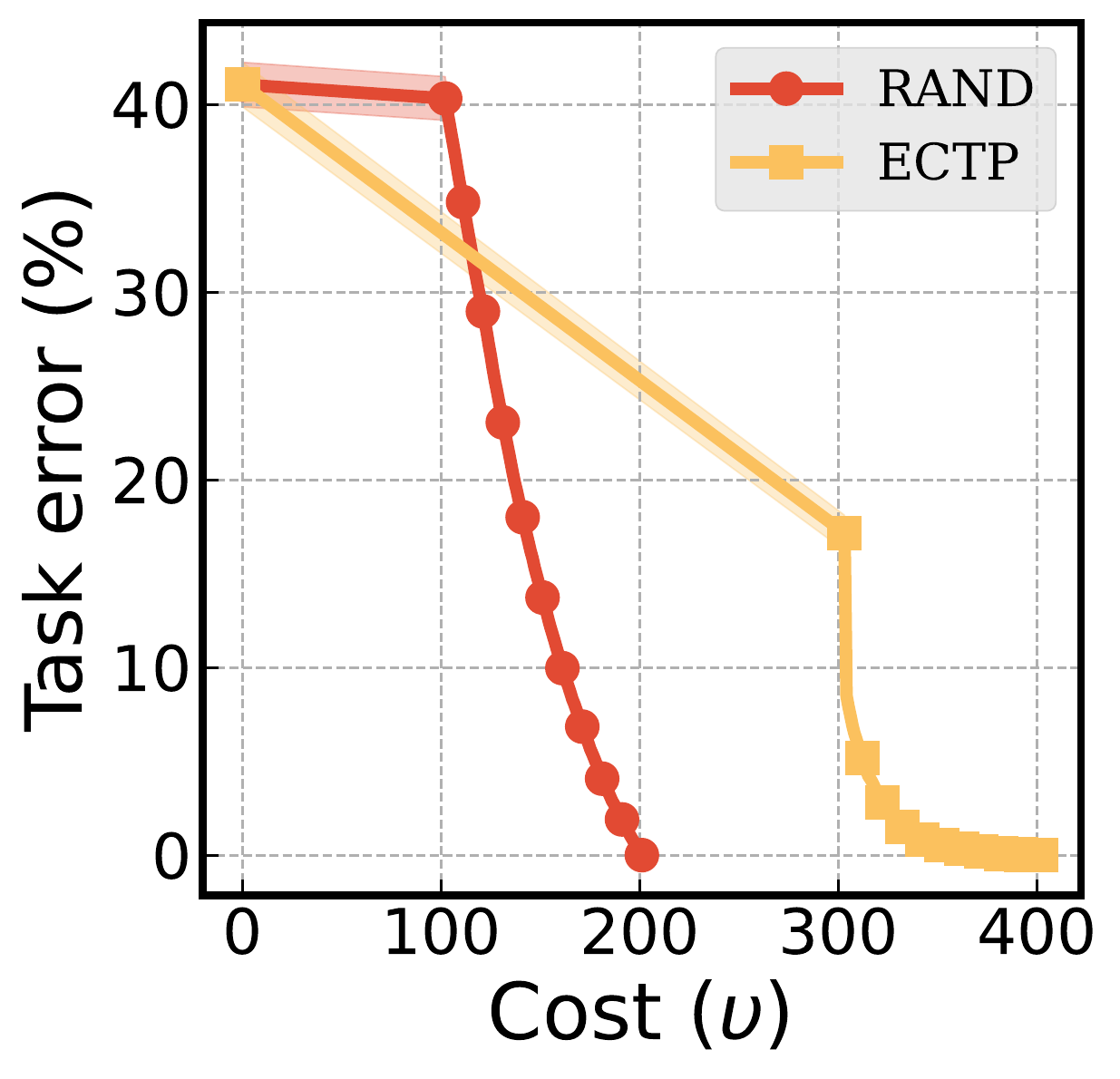}
    \caption{$\alpha = 0.01$}
    \label{fig:cost_alpha0.01}
  \end{subfigure}%
  \hspace*{\fill}
  \begin{subfigure}{0.16\linewidth}
    \centering
    \includegraphics[width=\linewidth]{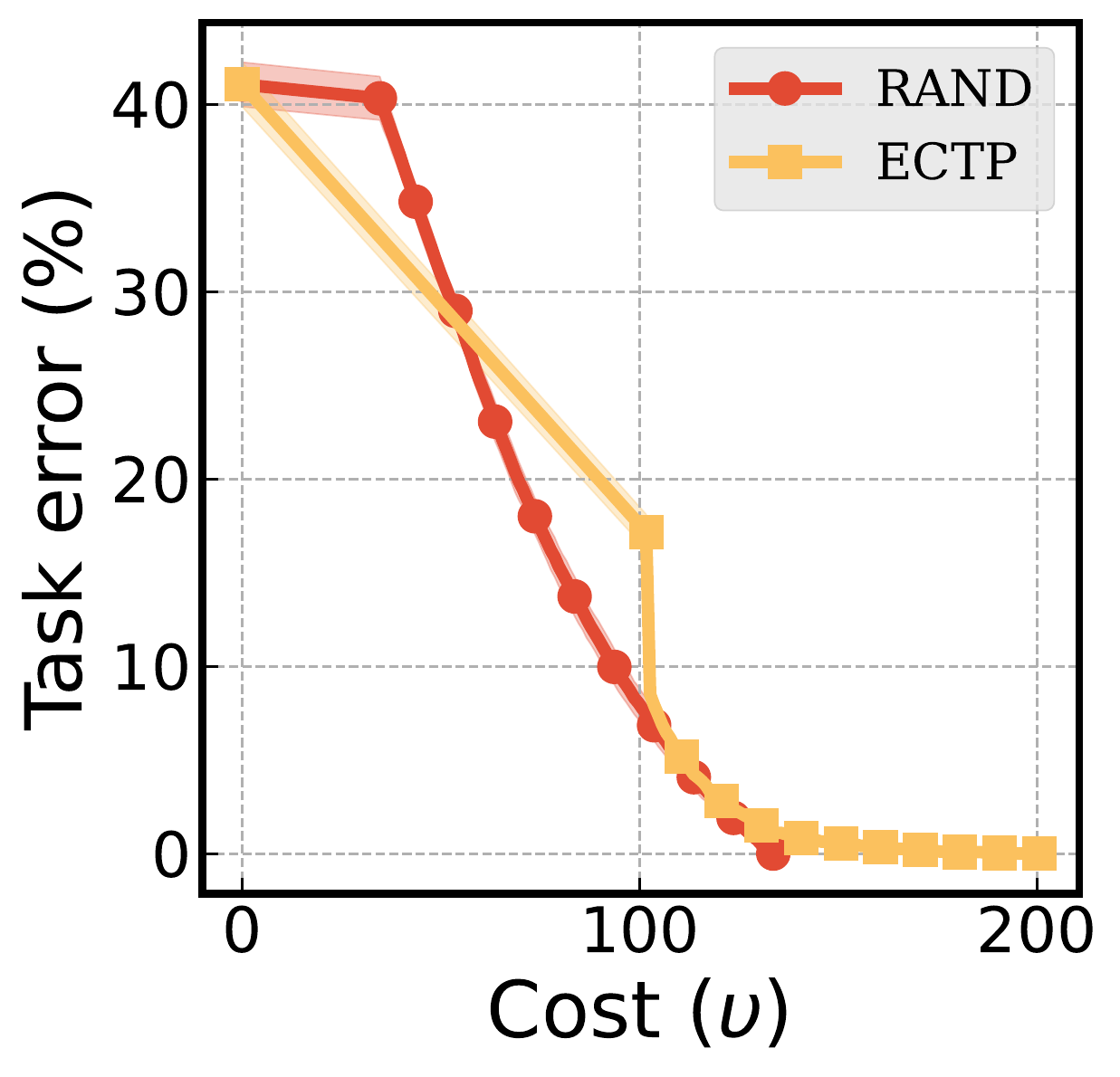}
    \caption{$\alpha = 0.03$}
    \label{fig:cost_alpha0.03}
  \end{subfigure}%
  \hspace*{\fill}
  \begin{subfigure}{0.16\linewidth}
    \centering
    \includegraphics[width=\linewidth]{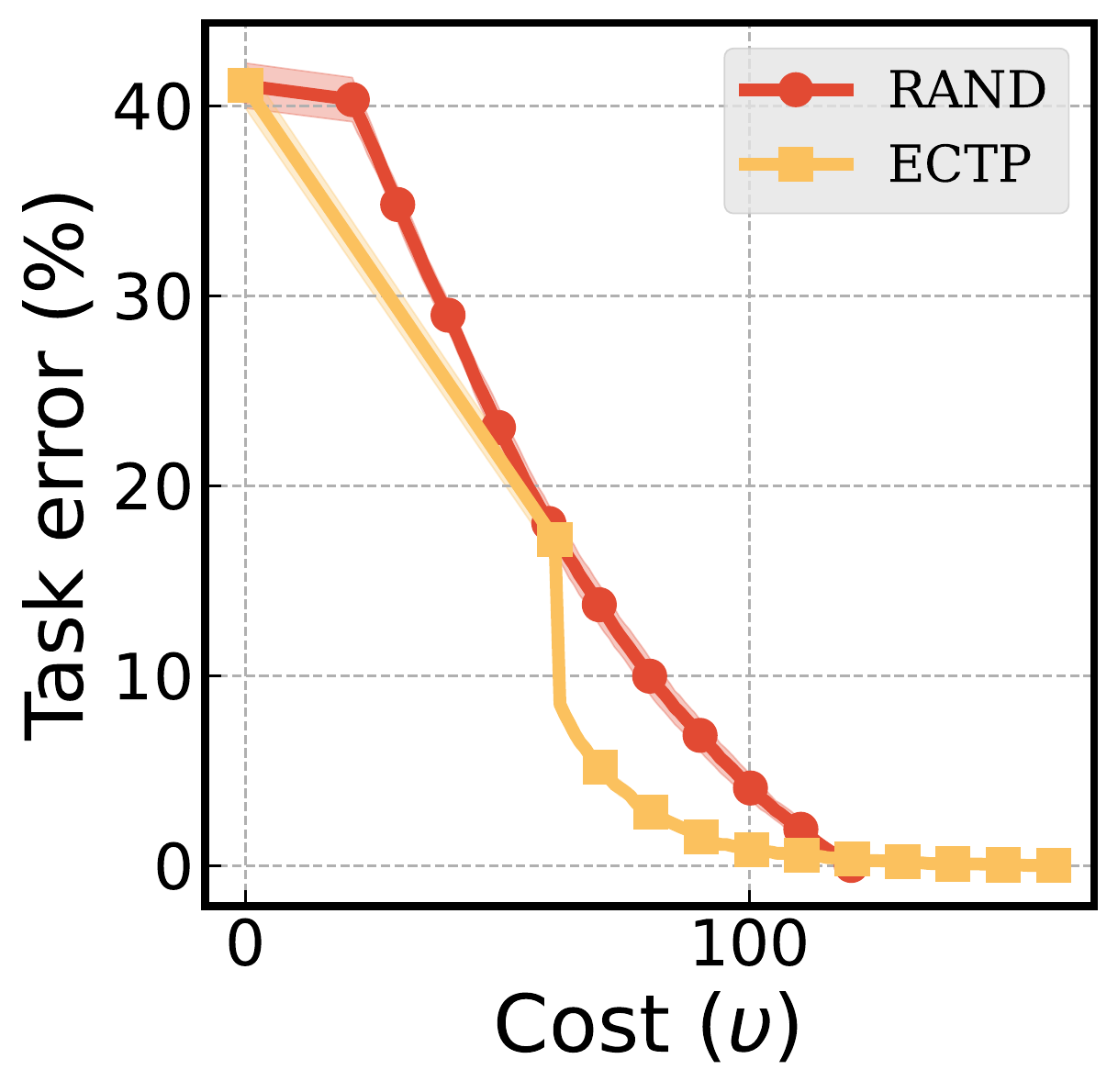}
    \caption{$\alpha = 0.05$}
    \label{fig:cost_alpha0.05}
  \end{subfigure}%
  \hspace*{\fill}
  \begin{subfigure}{0.16\linewidth}
    \centering
    \includegraphics[width=\linewidth]{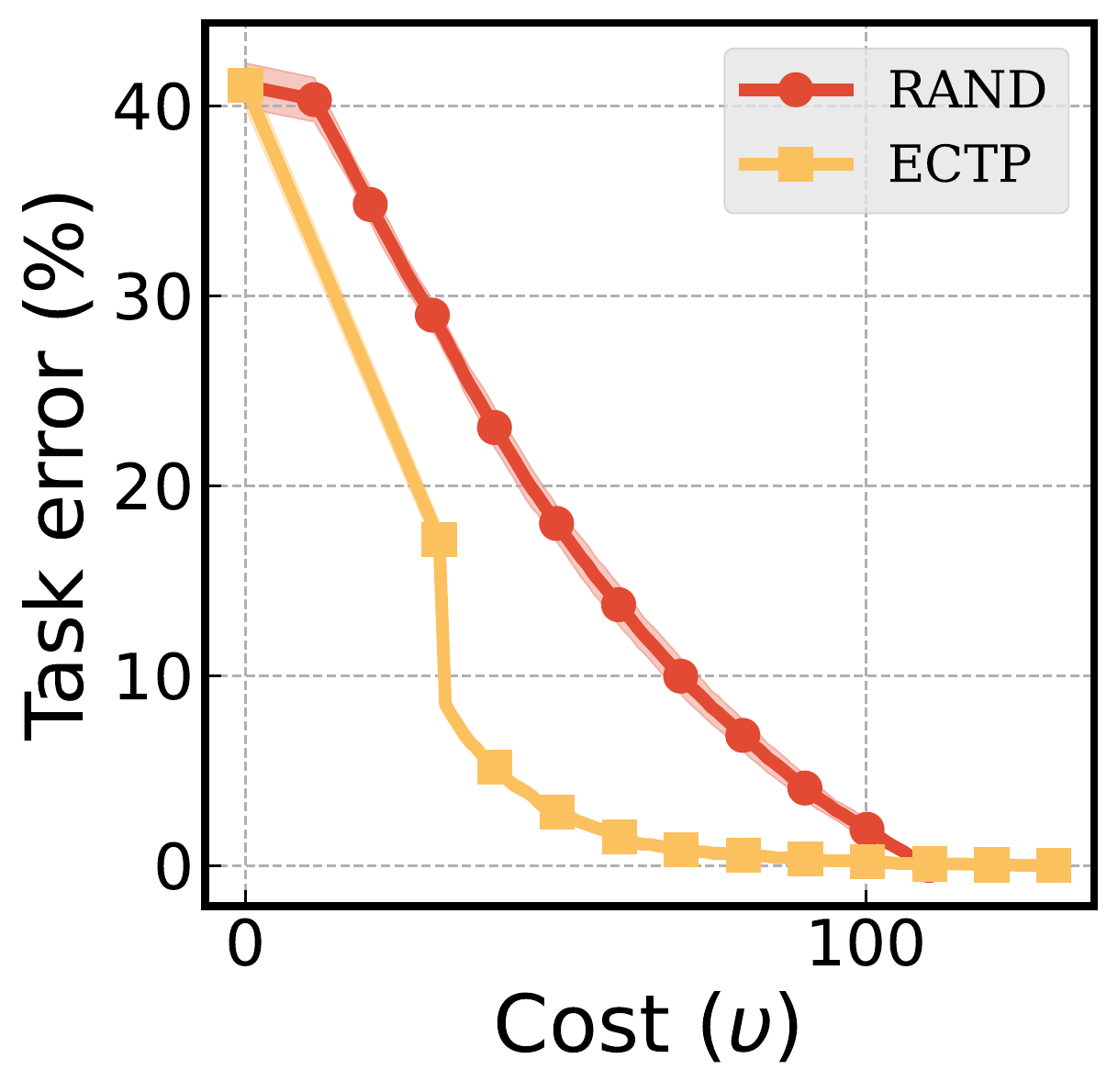}
    \caption{$\alpha = 0.1$}
    \label{fig:cost_alpha0.1}
  \end{subfigure}%
  \hspace*{\fill}
  \begin{subfigure}{0.16\linewidth}
    \centering
    \includegraphics[width=\linewidth]{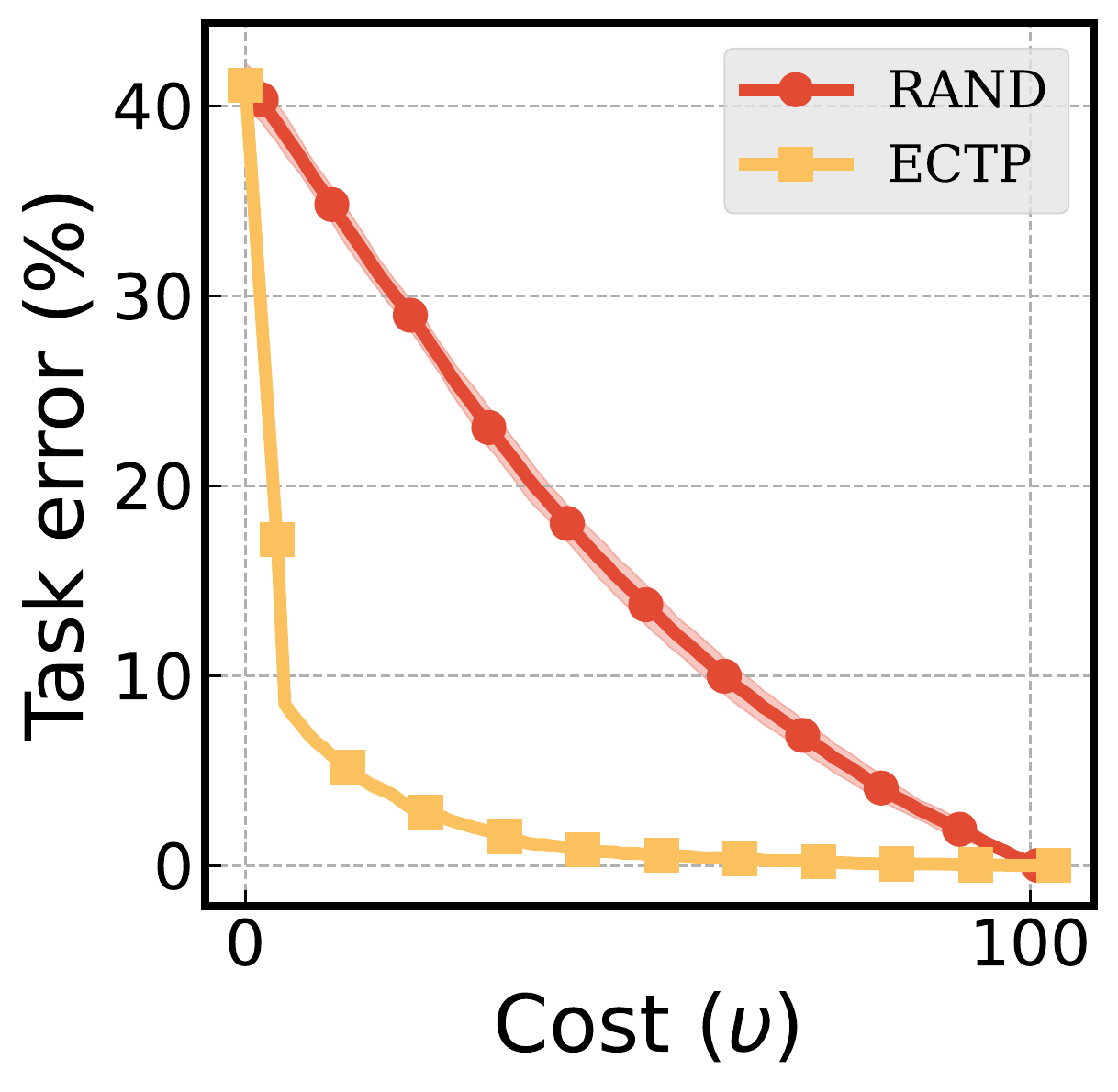}
    \caption{$\alpha = 1.0$}
    \label{fig:cost_alpha1}
  \end{subfigure}%
  \hspace*{\fill}
  \begin{subfigure}{0.16\linewidth}
    \centering
    \includegraphics[width=\linewidth]{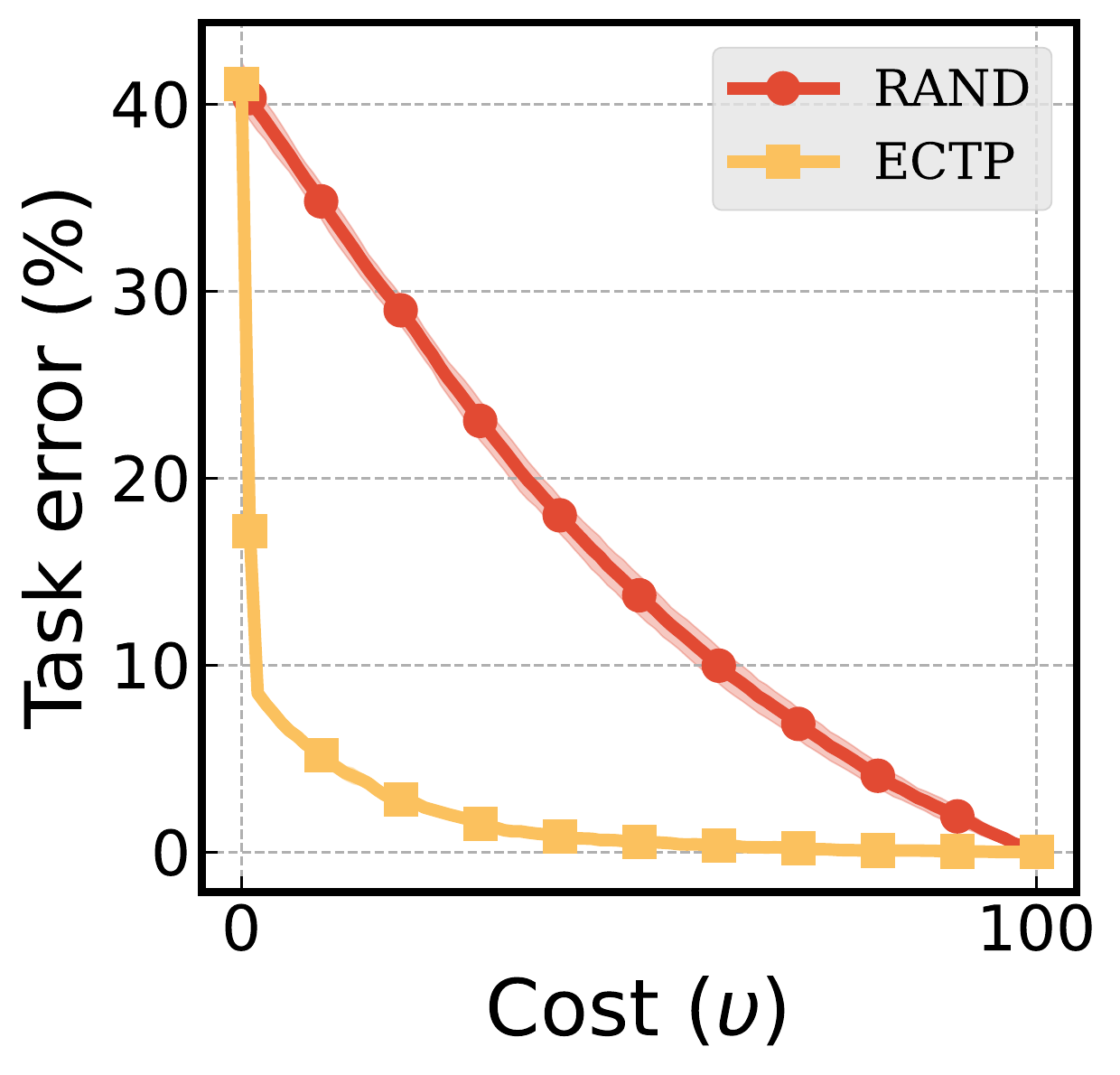}
    \caption{$\alpha = 100.0$}
    \label{fig:cost_alpha100}
  \end{subfigure}%
  \caption{
  Effect of $\alpha$ on intervention (on Synthetic).
  We fix $\tau_i = 1, \beta = 100, k=100$.
  \textsc{ectp}, the intervention method strongly evaluated previously, becomes less effective as $\alpha$ decreases.
  Here, kinked shapes are due to the relatively high initial cost on the first intervention before $n$ becomes large.
  }
  \label{fig:cost_result_alpha}
\end{figure*}

\subsection{Evaluating Concept Selection Criteria}
\label{sec:results-criteria}

We first evaluate the intervention effectiveness of concept selection criteria and present the results in \cref{fig:result-main}.
Across all datasets, we find that the current practice of random intervention (\textsc{rand}) is easily outperformed by the other alternatives in almost all cases with a significant margin.
Specifically, in the CUB experiment, correcting $20$ concepts by random intervention reduces the task error less than $4\%$ whereas correcting the same amount based on the uncertainty of concept predictions (\textsc{ucp}) leads to more than $16\%$ error reduction.
To put it differently, \textsc{rand} requires to intervene on $43$ concepts in order to reduce the error by half, while it is only $12$ concepts to fix for \textsc{ucp} to achieve the same reduction.
In the SkinCon experiment, selecting concepts based on the expected change in target prediction (\textsc{ectp}) leads the way among others, and yet, the scale of improvements over \textsc{rand} is not as large.
Note also that the strategy of fixing concepts with the largest loss first (\textsc{lcp}) performs exceptionally well in all cases.
This is however due to the help of the ground-truth knowledge on concepts which is unavailable in practice.
Nonetheless, we believe this can serve as an indicator to guide a better intervention strategy which we defer to future work.

\subsubsection{Reflecting cost of intervention}

\label{sec:results-cost}

As we discussed in \cref{sec:cost}, the cost of intervention may differ by concept selection criteria.
Taking into account this aspect, we set up experiments where we can evaluate the intervention effectiveness in terms of the theoretical cost.
Specifically, we model the relationships between $\tau_i$, $\tau_g$, $\tau_f$ as $\tau_i = \alpha \tau_g$ and $\tau_g = \beta \tau_f$, which means that the cost of intervention (\eg, time to fix a concept) is $\alpha$-proportional to the cost of making inference on $g$, and likewise, $\tau_g$ is $\beta$-proportional to $\tau_f$.
Then we can evaluate the cost-reflected intervention effectiveness with respect to arbitrary unit ($\upsilon$), and from which, we can further show how it transforms by controlling $\alpha$ and $\beta$.

First, the result of changing $\alpha$ is plotted in \cref{fig:cost_result_alpha}.
As $\alpha$ becomes smaller \textsc{rand} becomes very effective compared to \textsc{ectp}.
This makes sense because with small $\alpha$, $\tau_i$ becomes relatively small and the other terms related to $\tau_g$ or $\tau_f$ dominate the cost of \textsc{ectp} which is $\mathcal{O} \big( \tau_g + (2k + 2) \tau_f + n \tau_i \big)$ as seen in \cref{tab:intervention_methods_cost}.
\textsc{ectp} thus becomes penalized when it comes to the intervention effectiveness in the small $\alpha$ regime.
In contrast, when $\alpha$ becomes larger, $\tau_i$ dominates the cost of \textsc{ectp} as with increasing $n$, which in turn recovers the effectiveness of \textsc{ectp}.
The former can happen in extreme circumstances, for example, when using very large models (\ie, large $\tau_g$) or in places with a tight labor marker (\ie, small $\tau_i$ in terms of monetized value).
We clearly remark, however, that this can be seen as a hypothetical case and $\alpha$ will be much greater than $1$ in realistic settings as summoning a domain expert for intervention would require more cost than a forward pass of neural networks.

We also experiment on changing $\beta$ to control the relative cost between $\tau_g$ and $\tau_f$.
As a result, we find that when $\beta$ is small \textsc{ectp} can perform poorly while \textsc{rand} can be effective as it only requires a single forward pass of $f$ to make the final prediction.
Furthermore, we extend this analysis to the CUB experiment with more realistic settings where $\tau_g$ and $\tau_f$ are set based on the wall-clock times of running each model, and $\tau_i$ is set based on the actual concept annotation time provided in the dataset.
All of these results are put in \cref{app:cost} with detailed analysis for space reasons.

\subsection{Analyzing Intervention Levels}

\label{sec:restuls-levels}

\begin{figure*}[!t]
  \begin{subfigure}{0.22\linewidth}
    \centering
    \includegraphics[width=\linewidth]{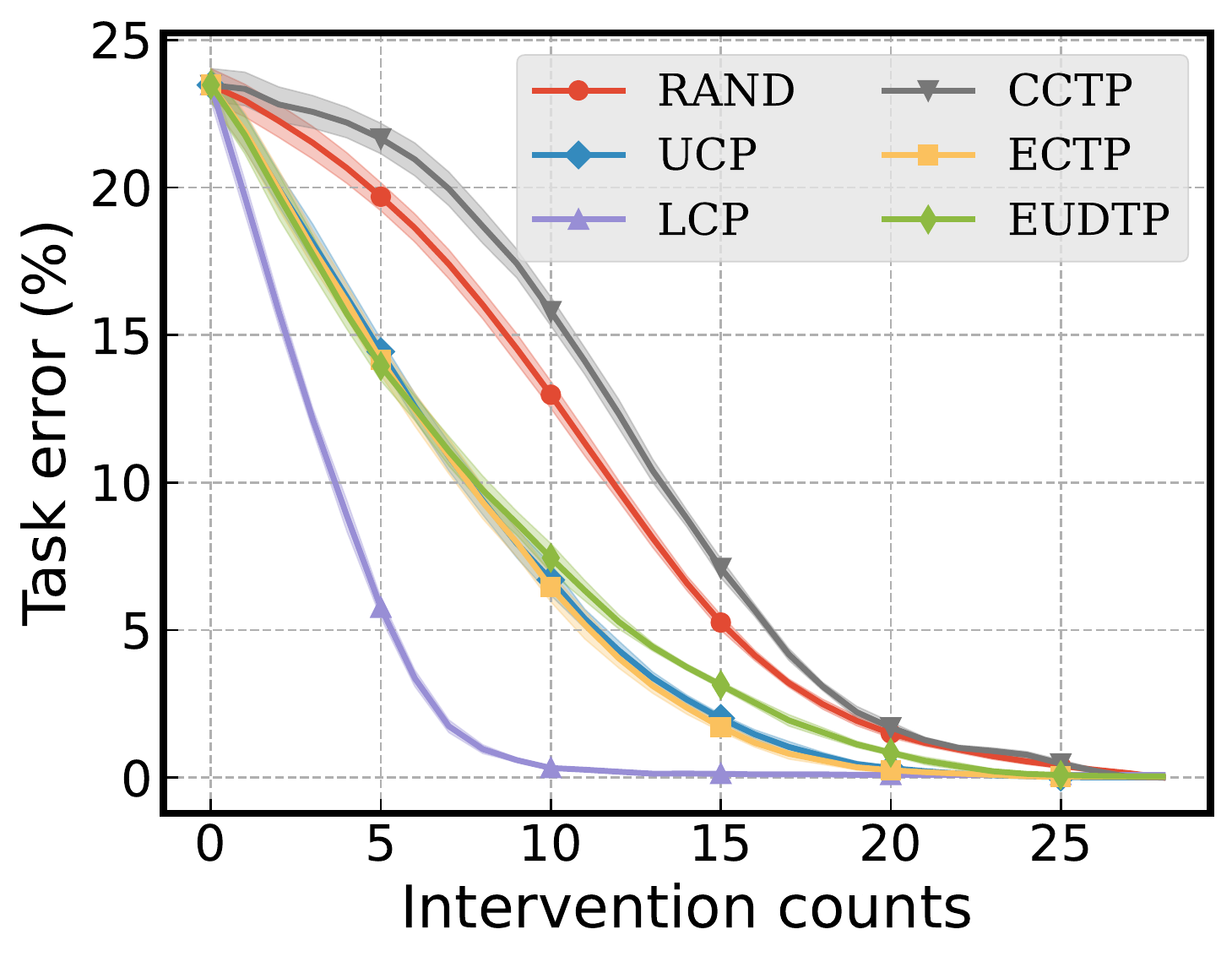}
    \caption{\textsc{G+S} level}
    \label{fig:cub_result_level_gs_main}
  \end{subfigure}
  \hspace*{\fill}
  \begin{subfigure}{0.22\linewidth}
    \centering
    \includegraphics[width=\linewidth]{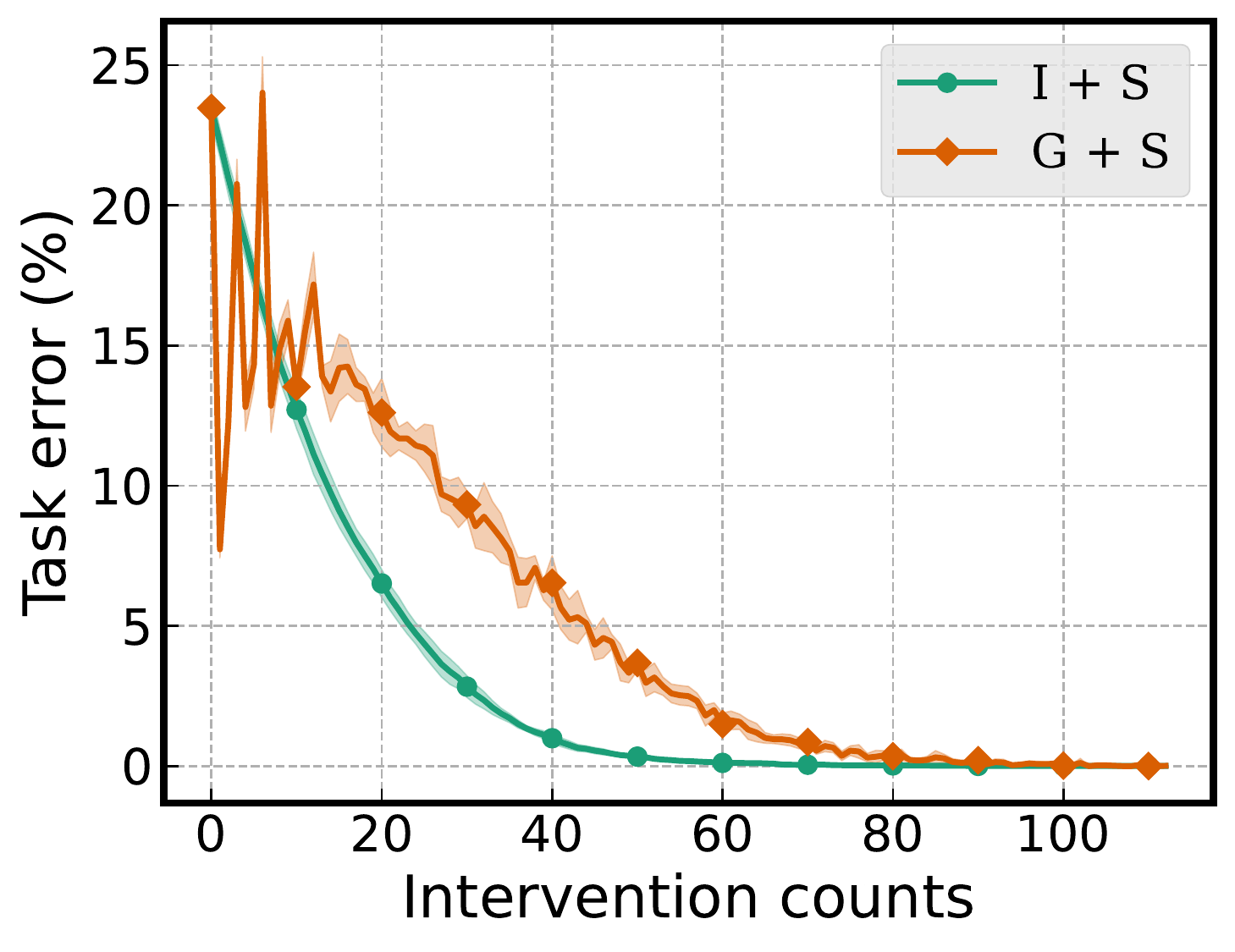}
    \caption{(\textsc{I+S}) vs. (\textsc{G+S})}
    \label{fig:cub_level_is_gs_ucp_main}
  \end{subfigure}
  \hspace{12mm}
  \begin{subfigure}{0.22\linewidth}
    \includegraphics[width=\linewidth]{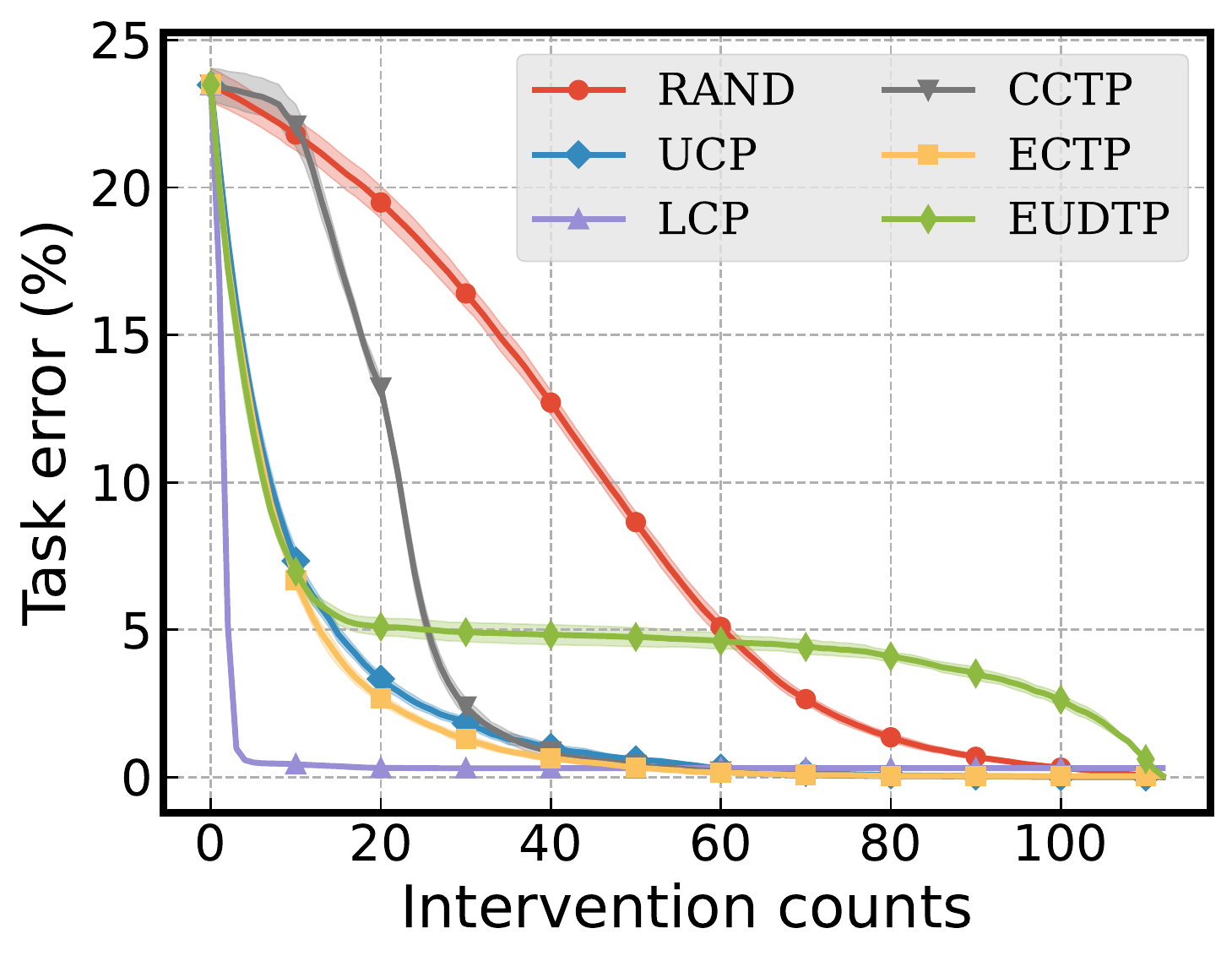}
    \caption{\textsc{I+B} level}
    \label{fig:cub_result_level_ib_main}
  \end{subfigure}
  \hspace*{\fill}
  \begin{subfigure}{0.22\linewidth}
    \centering
    \includegraphics[width=\linewidth]{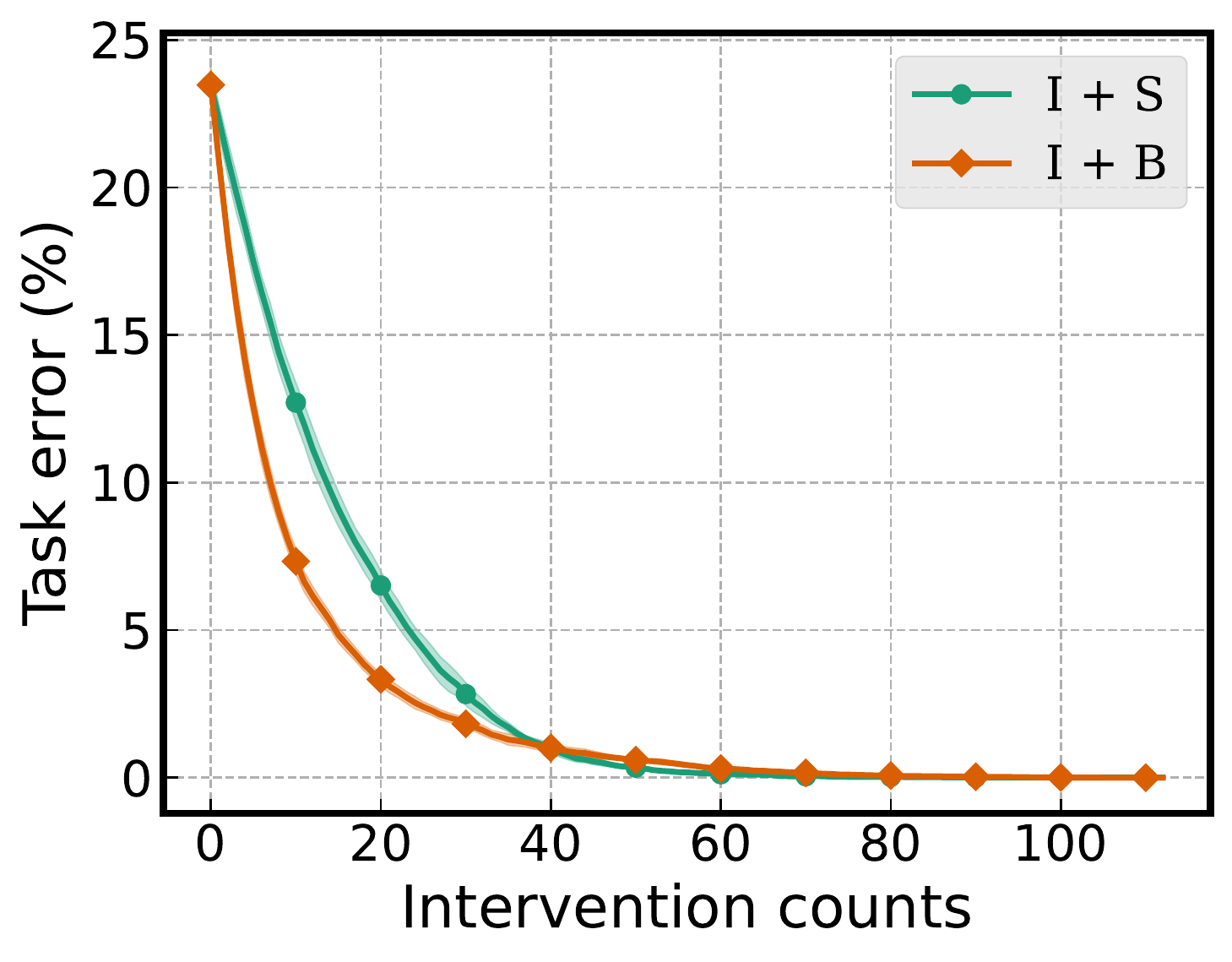}
    \caption{(\textsc{I+S}) vs. (\textsc{I+B})}
    \label{fig:cub_level_is_ib_ucp_main}
  \end{subfigure}
  \caption{
    Comparing the effects of different intervention levels using the CUB dataset.
    Here, intervention counts denote the number of intervened groups and average number of intervened concepts for \textsc{g} and \textsc{b}, respectively.
    We fix the selection criterion to be \textsc{ucp} in (b) and (d) while all other cases are provided in \cref{sec:results-others-levels}.
  }
  \label{fig:cub_level}
\end{figure*}

As seen in \cref{fig:cub_result_level_gs_main}, most criteria still remain more effective than \textsc{rand} in group-wise single (\textsc{G + S}) intervention.
Specifically, \textsc{rand} needs $39.3\%$ ($11$ out of $28$), while \textsc{ucp} needs $25.0\%$ ($7$ out of $28$) of the groups to be intervened to decrease the task error by half.
However, \textsc{cctp} does not outperform \textsc{rand} this time.
We also find a similar pattern for the batch case \textsc{G + B} (see \cref{fig:cub_result_level} in \cref{sec:results-others-levels}).
We suspect that calculating the mean of the scores loses some discriminative information in some selection criteria and perhaps a different surrogate needs to be designed.

In addition, we find that group-wise intervention is in general less effective than individual counterpart with the same budget of intervention expense (see \cref{fig:cub_level_is_gs_ucp_main}).
Intuitively, correcting concepts within the same group may not provide rich information as opposed to selecting concepts across different groups with the same intervention counts.
Nonetheless, we remark that group-wise intervention can potentially be cost-effective when concepts within the same group are mutually exclusive, which depends on how the concepts are annotated during the creation of datasets.

The proposed concept selection criteria also remain effective for batch intervention (\textsc{b}) as seen in \cref{fig:cub_result_level_ib_main}.
Interestingly, batch intervention turns out to be more effective when compared to single (\textsc{s}) as well as seen in \cref{fig:cub_level_is_ib_ucp_main}.
This trend holds true for other criteria besides \textsc{ucp} except for \textsc{cctp} and extends to group-wise batch (\textsc{G+B}) intervention (see \cref{sec:results-others-levels} for full results).

\begin{figure*}[!t]
  \begin{subfigure}{0.22\linewidth}
    \centering
    \includegraphics[width=\linewidth]{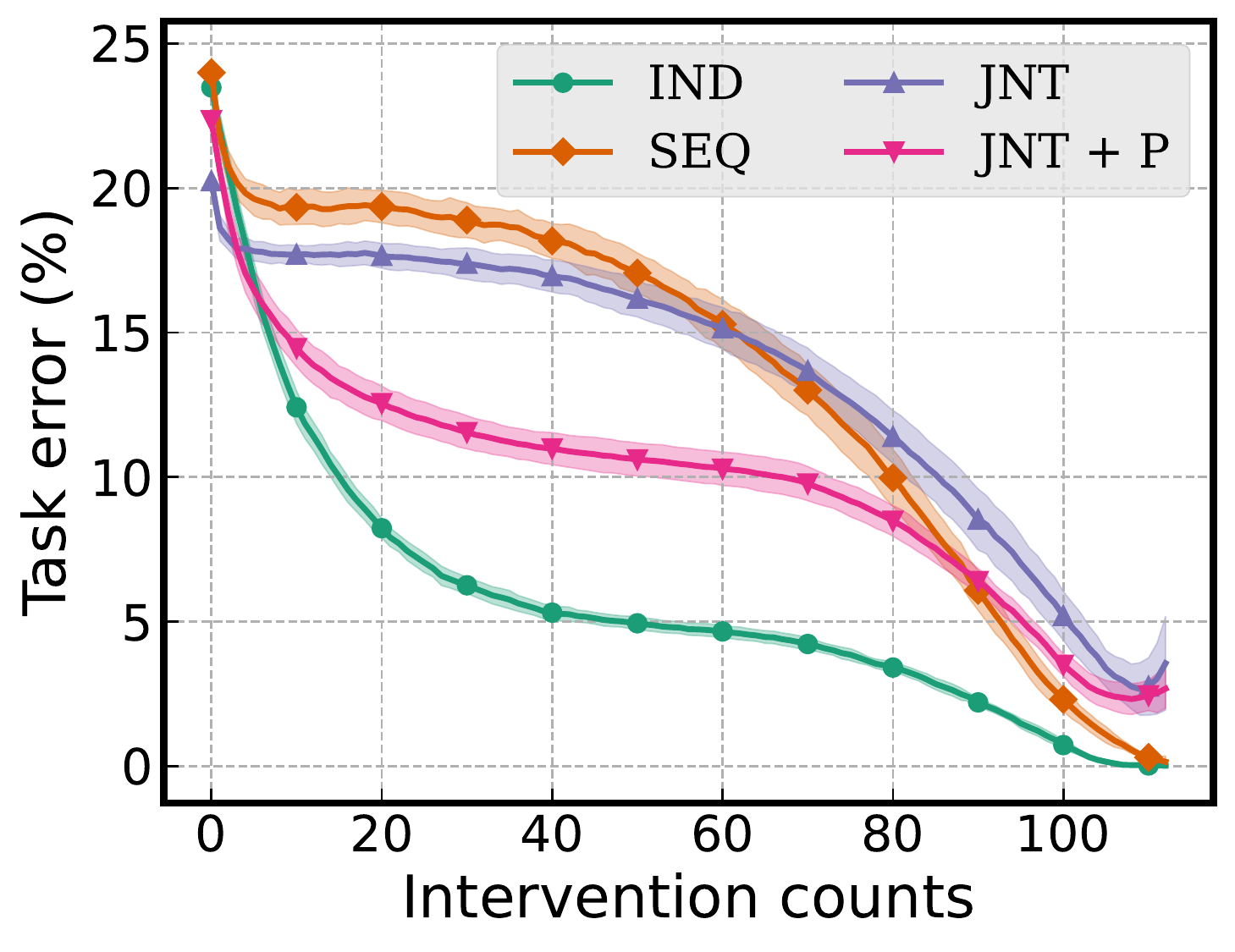}
    \caption{Training on CUB}
    \label{fig:cub_training_eudtp_main}
  \end{subfigure}%
  \hspace*{\fill}
  \begin{subfigure}{0.22\linewidth}
    \centering
    \includegraphics[width=\linewidth]{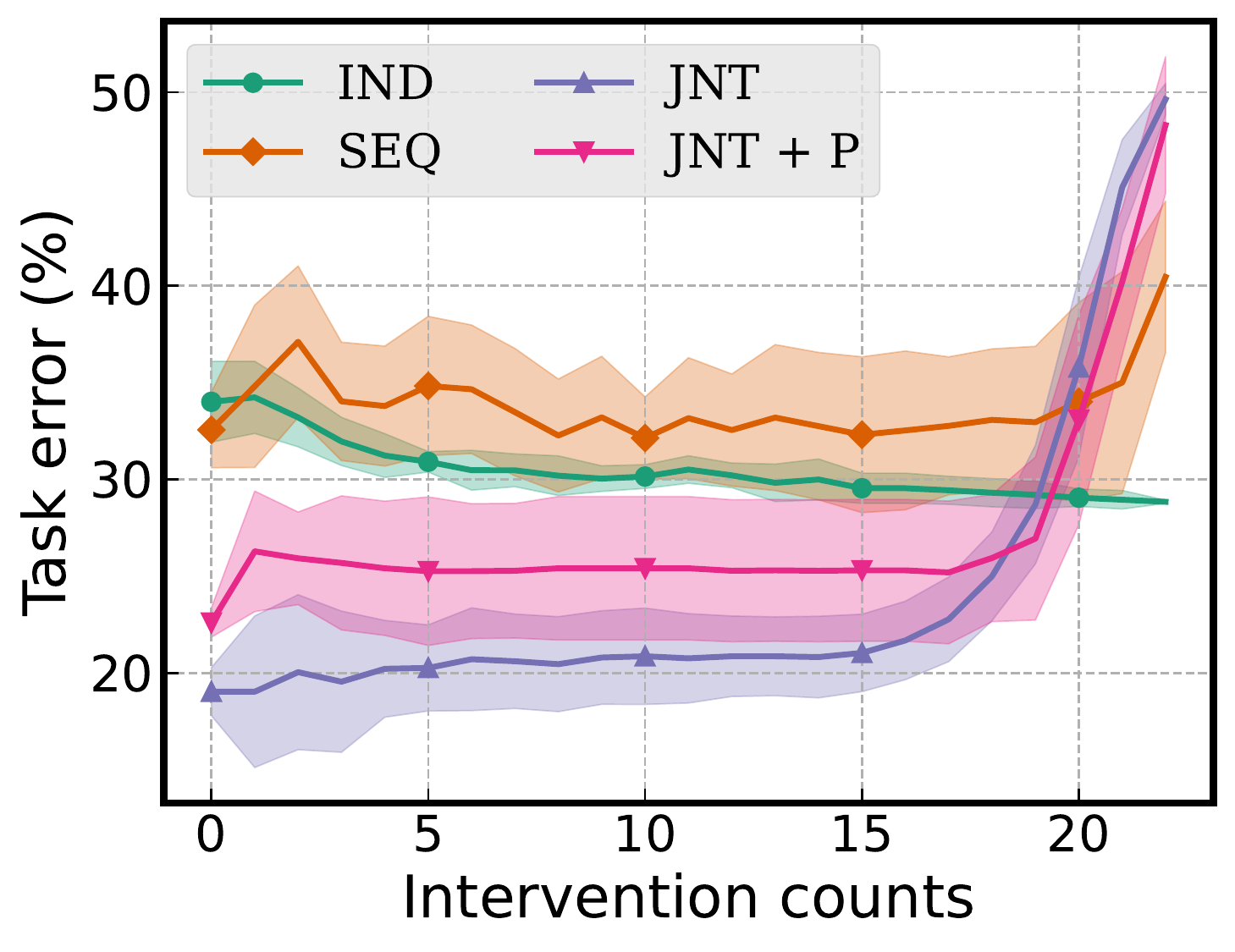}
    \caption{Training on SkinCon}
    \label{fig:skincon_training_eudtp_main}
  \end{subfigure}%
  \hspace{12mm}
  \begin{subfigure}{0.22\linewidth}
    \centering
    \includegraphics[width=\linewidth]{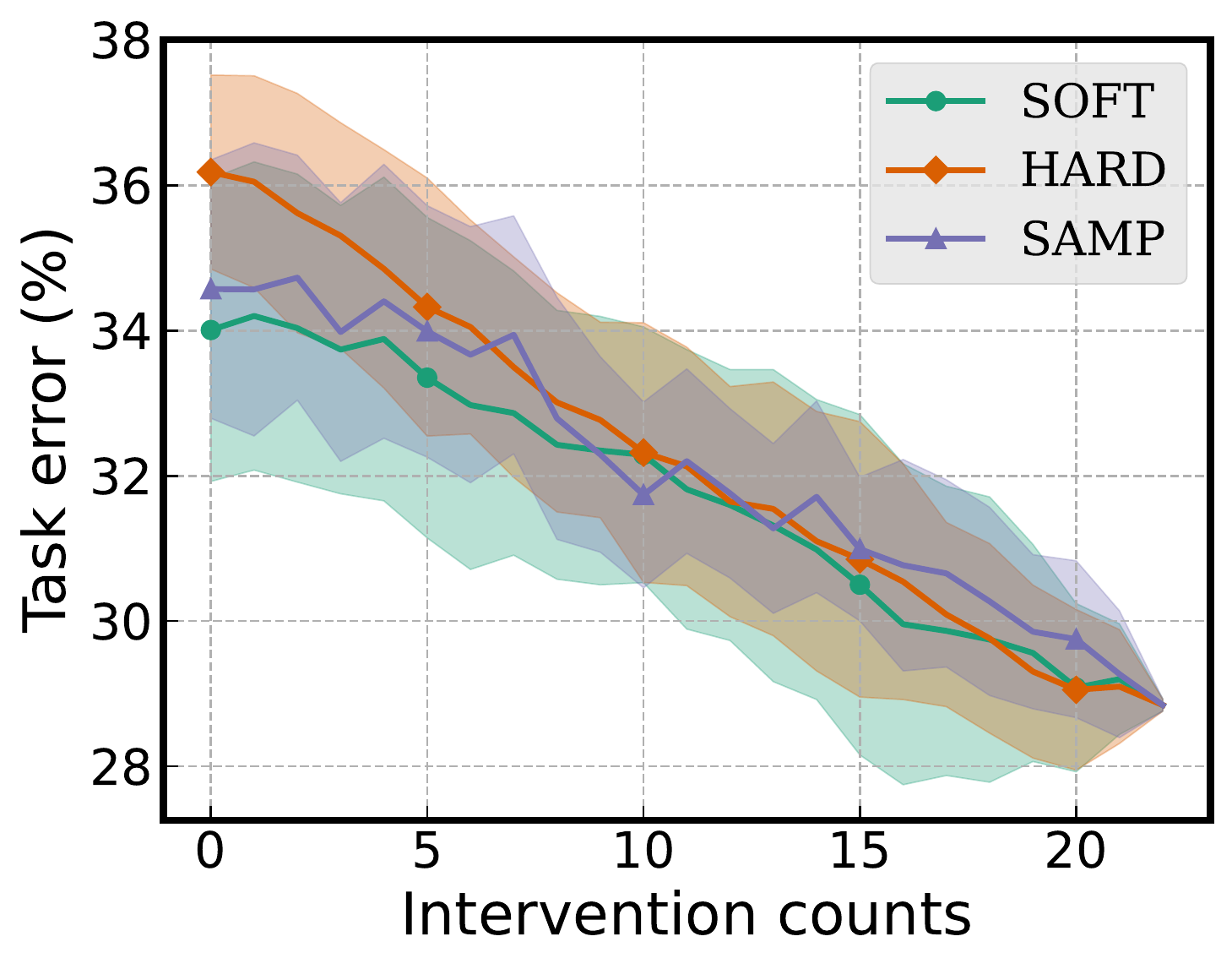}
    \caption{Conceptualization: \textsc{rand}}
    \label{fig:skincon_conceptualization_rand_main}
  \end{subfigure}%
  \hspace*{\fill}
  \begin{subfigure}{0.22\linewidth}
    \centering
    \includegraphics[width=\linewidth]{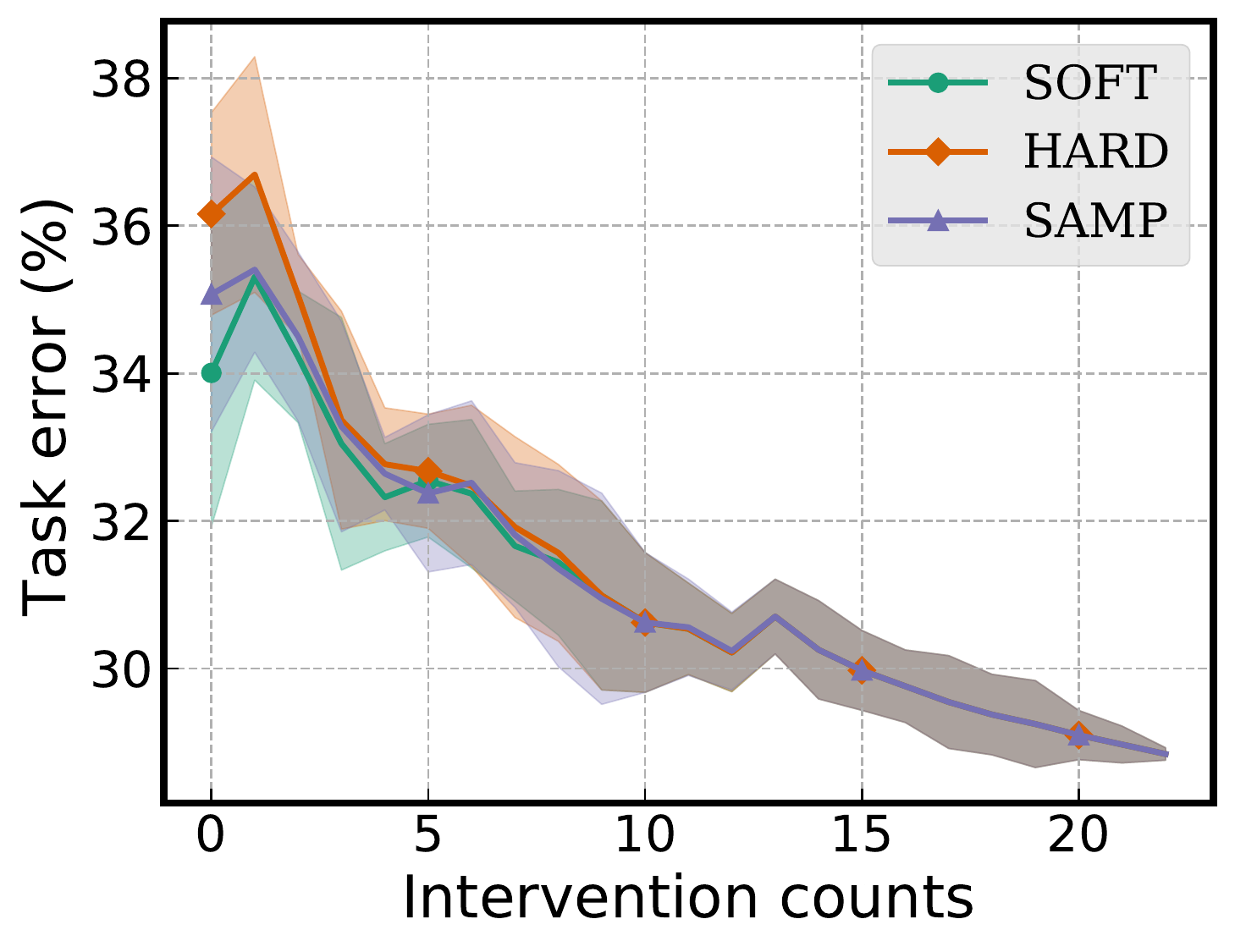}
    \caption{Conceptualization: \textsc{ucp}}
    \label{fig:skincon_conceptualization_ucp_main}
  \end{subfigure}%
  \caption{
    Comparing the effects of different training strategies (a,b) and conceptualization methods (c, d).
    We choose \textsc{eudtp} as the concept criterion for (a,b) and SkinCon as the dataset for (c, d).
    We provide all other results in \cref{sec:results-others-training,sec:results-others-conceptualization}.
    } 
  \label{fig:training_and_conceptualization}
\end{figure*}

\subsection{Considering Training and Conceptualization}
\label{sec:exp_train_inf_strategies}

\textbf{Effect of training scheme}\quad
As seen in \cref{fig:cub_training_eudtp_main}, intervention is in general the most effective under the \textsc{ind} training scheme.
We believe that this is because $f$ is not trained with the ground-truth concept labels in the case of \textsc{seq} and \textsc{jnt}(+\textsc{p}), and fixing concept predictions for these schemes may not work as well.
We also find that \textsc{eudtp} becomes much less effective under \textsc{seq} or \textsc{jnt} than other alternatives and actually underperforms \textsc{rand} (see \cref{sec:results-others-training}).
Hence, the effectiveness of a criterion can depend on which training strategy to use, implying the need of comprehensive evaluations for newly developed criteria.

For the SkinCon dataset, however, intervening on the concepts under \textsc{seq, jnt, jnt + p} strategies rather increases the average task error regardless of the concept selection criteria.
Specifically, training under \textsc{jnt} already achieves low task error and applying intervention does not help reduce it further (see \cref{fig:skincon_training_eudtp_main}).
We hypothesize that this is due to some inherent characteristics of the dataset as well as limited concepts provided in the bottleneck, resulting in the negative influence on making correct task predictions with binarized concepts.
This can potentially correspond to the known issue of information leakage in CBMs \citep{mahinpei2021promises, havasi2022addressing}.

\textbf{Effect of conceptualization}\quad
We find that \textsc{hard} and \textsc{samp} may begin with high task error compared to \textsc{soft} as expected.
However, when making use of the developed concept selection criteria such as \textsc{ucp}, the gap between these conceptualization methods decreases much faster with more intervention compared to \textsc{rand} as seen in \cref{fig:skincon_conceptualization_rand_main,fig:skincon_conceptualization_ucp_main}.
This result is consistent across different training strategies and datasets (see \cref{sec:results-others-conceptualization}).
\section{Analyzing Intervention with Synthetic Data}
\label{sec:ablation-dataset}

We have observed that intervention can often yield different results over datasets.
Precisely, intervening on all concepts decreases the task error down to $0\%$ on CUB, whereas the amount of decrease is much less and the average task error remains still high around $29\%$ on SkinCon.
Also, the relative order of effectiveness between concept selection criteria can vary.
We find that it is difficult to unravel these findings if only experimenting on real datasets as in previous work \citep{koh2020concept,chauhan2022interactive, sheth2022learning, zarlenga2022concept}.
To provide an in-depth analysis, we develop a framework to generate synthetic datasets based on three different causal graphs that control the followings: input noise, hidden concepts, and concept diversity.

\subsection{Generating Synthetic Data}

\begin{figure}
  \centering
  \begin{subfigure}{0.29\linewidth}
    \includegraphics[width=\linewidth]{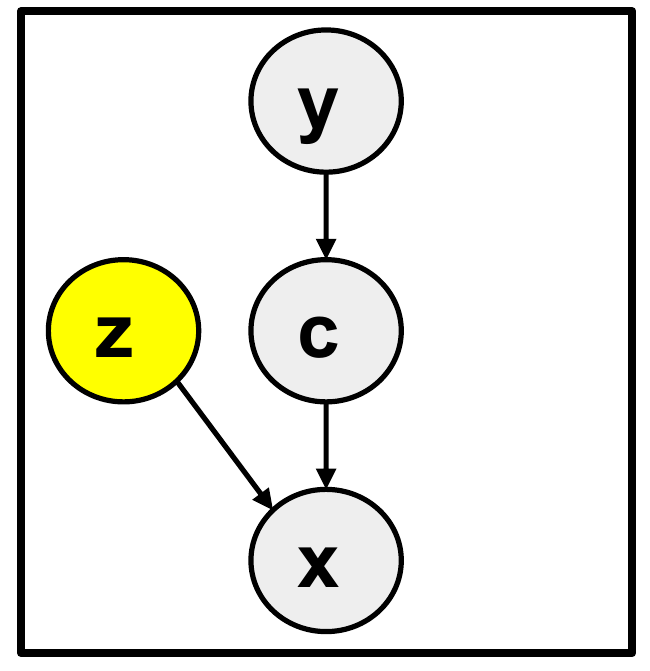}
    \caption{Noisy input}
    \label{fig:synthetic_graph_noisyinput}
  \end{subfigure}%
  \hspace*{\fill}
  \begin{subfigure}{0.29\linewidth}
    \includegraphics[width=\linewidth]{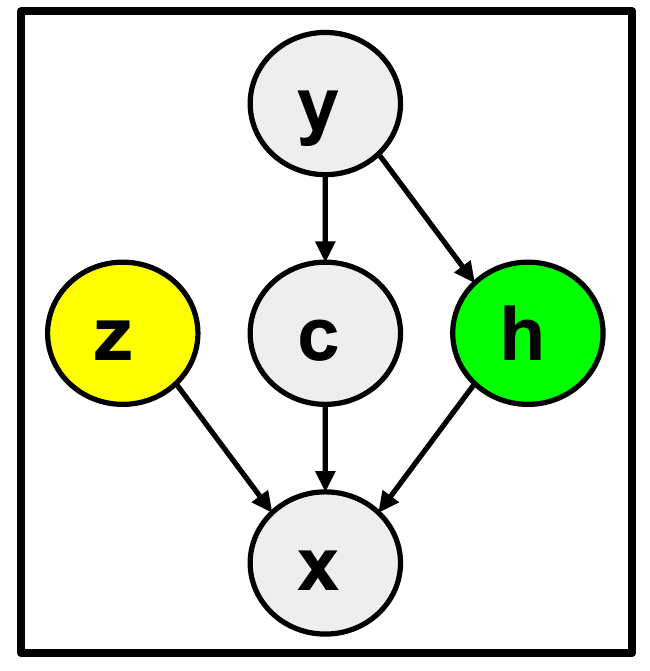}
    \caption{Hidden concept}
    \label{fig:synthetic_graph_hidden}
  \end{subfigure}
  \hspace*{\fill}
  \begin{subfigure}{0.29\linewidth}
    \includegraphics[width=\linewidth]{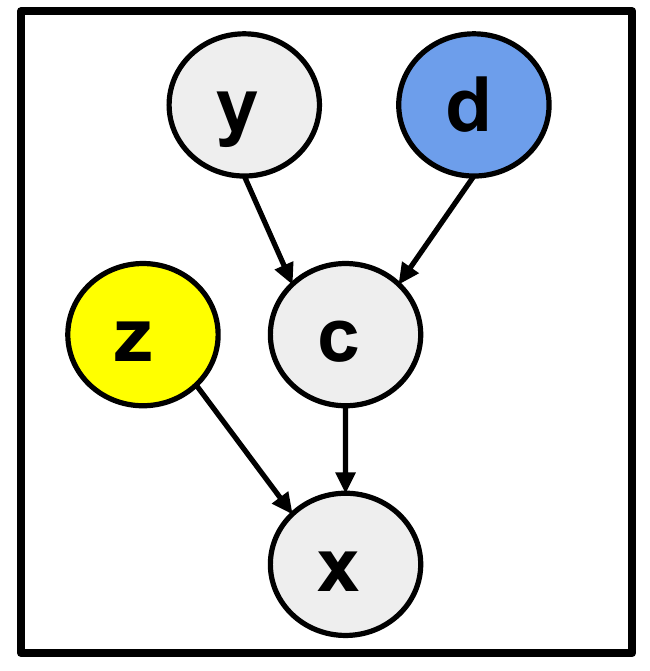}
    \caption{Diverse concept}
    \label{fig:synthetic_graph_diverseconcept}
  \end{subfigure}
  \caption{
    Causal graphs for generating synthetic datasets.
    $z$, $h$, and $d$ represent factors of input noise, hidden concepts, and concept diversity, respectively.
    The full details of the data generation process are provided in \cref{sec:synthetic_dataset}.
    }
  \label{fig:synthetic_graphs}
\end{figure}

\begin{figure}[!t]
  \begin{subfigure}{0.32\linewidth}
    \centering
    \includegraphics[width=\linewidth]{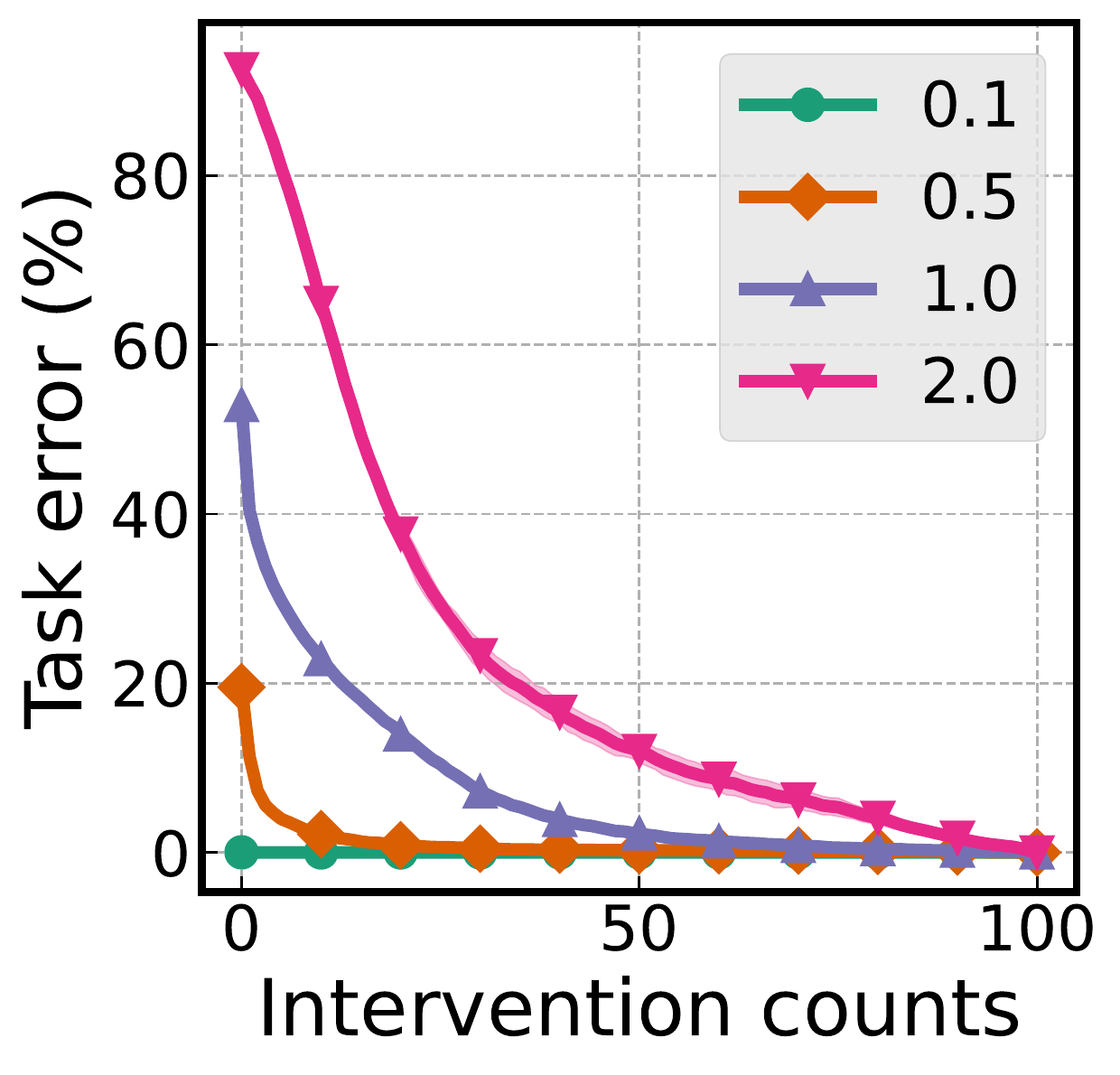}
    \caption{Noisy input}
    \label{fig:synthetic_inputnoise_ucp}
  \end{subfigure}
  \hspace*{\fill}
  \begin{subfigure}{0.32\linewidth}
    \centering
    \includegraphics[width=\linewidth]{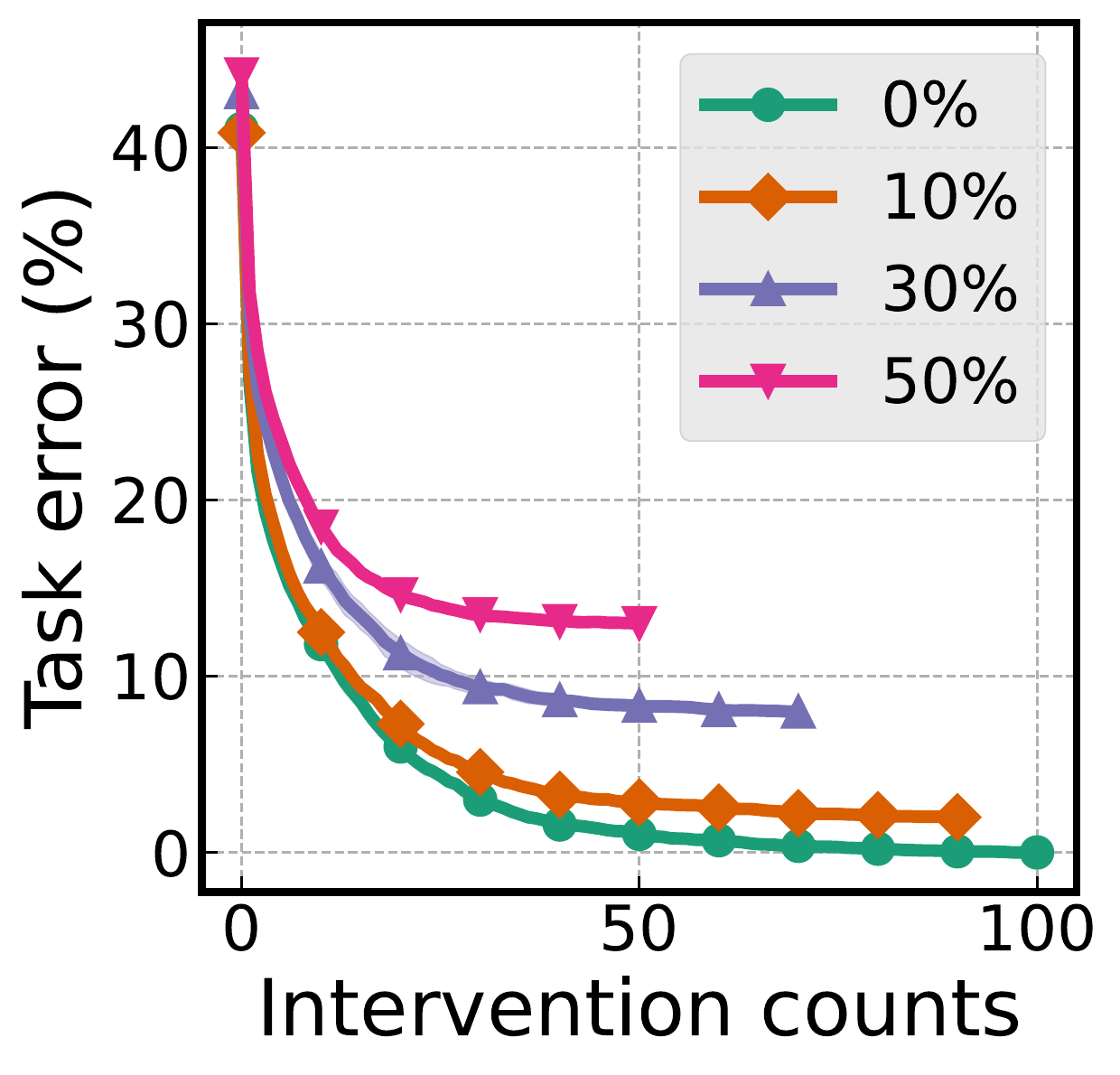}
    \caption{Hidden concept}
    \label{fig:synthetic_hidden_ucp}
  \end{subfigure}
  \hspace*{\fill}
  \begin{subfigure}{0.32\linewidth}
    \centering
    \includegraphics[width=\linewidth]{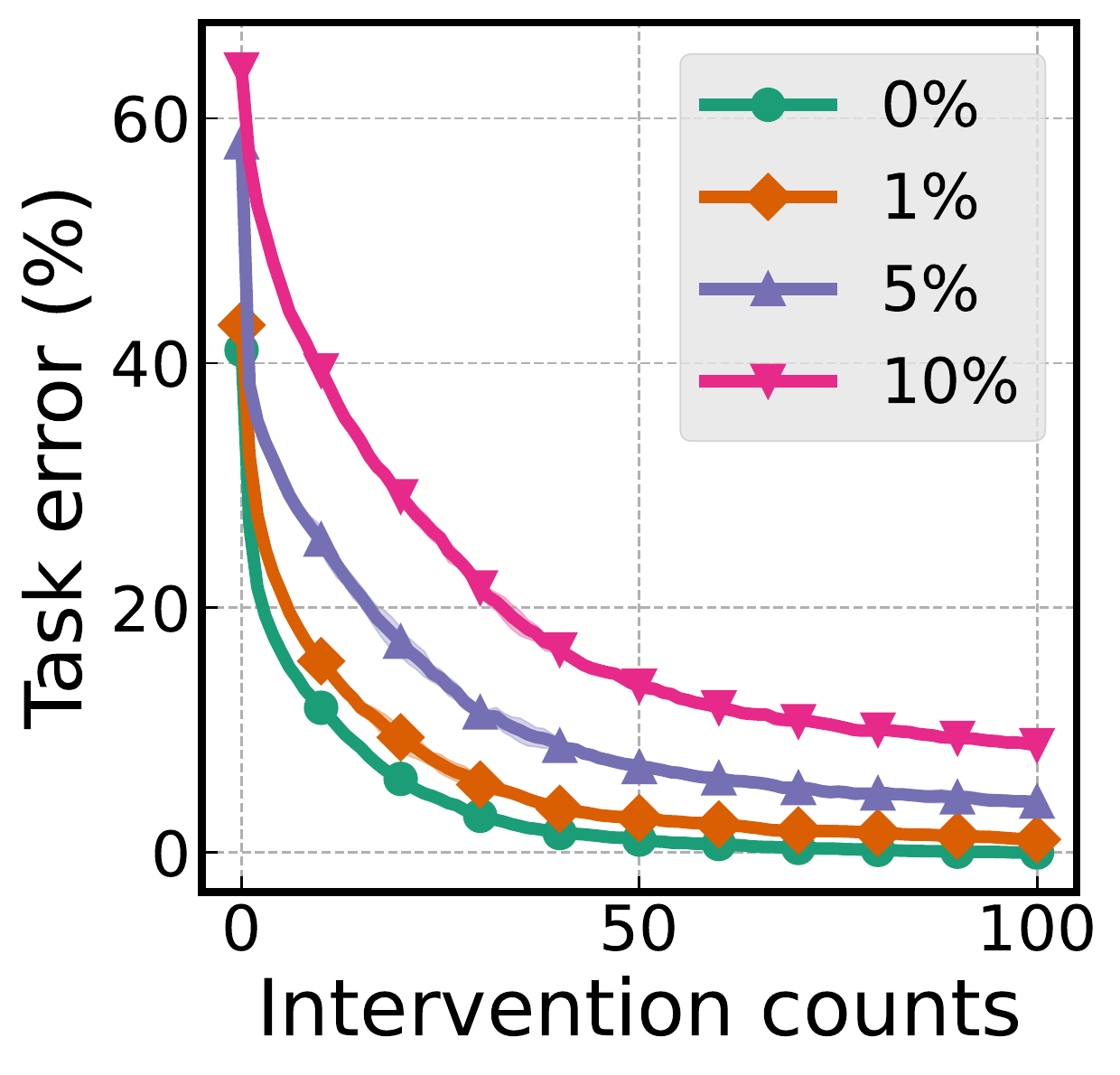}
    \caption{Diverse concept}
    \label{fig:synthetic_diversity_ucp}
  \end{subfigure}
  \caption{
      Effects of data on intervention with \textsc{ucp}.
      Each plot is with different values of the variance of noise ($z$), the ratio of hidden concepts ($h$), and the probability to perturb the concept values ($d$), respectively.
  }
  \label{fig:synthetic_dataset_characteristics}
\end{figure}

\textbf{\textsc{Case 1}: Noisy input}\quad
Real-world data contains a lot of random noise coming from various sources (\eg, lighting).
We construct a causal graph to consider this case where the Gaussian noise is added on input data (see \cref{fig:synthetic_graph_noisyinput}).

\textbf{\textsc{Case 2}: Hidden concept}\quad
When a subset of concepts is unknown or hidden, the target prediction is made incomplete with only available concepts as deep representations are not fully captured in the bottleneck layer.
We design a causal graph for this case and generate synthetic data for which some concepts that are necessary to make correct target predictions are hidden on purpose (see \cref{fig:synthetic_graph_hidden}).

\textbf{\textsc{Case 3}: Diverse concept}\quad
Examples within the same class can have different values for the same concept in realistic settings.
For instance, simple concept-level noise or fine-grained sub-classes (\eg, \texttt{`black swan'} and \texttt{`white swan'} for \texttt{`swan'} class) can make such diverse concept values.
We construct a causal graph to generate such data for which concept values can vary probabilistically and inputs are produced according to these concepts (see \cref{fig:synthetic_graph_diverseconcept}).

\begin{figure}[!t]
  \begin{subfigure}{0.48\linewidth}
    \centering
    \includegraphics[width=\linewidth]{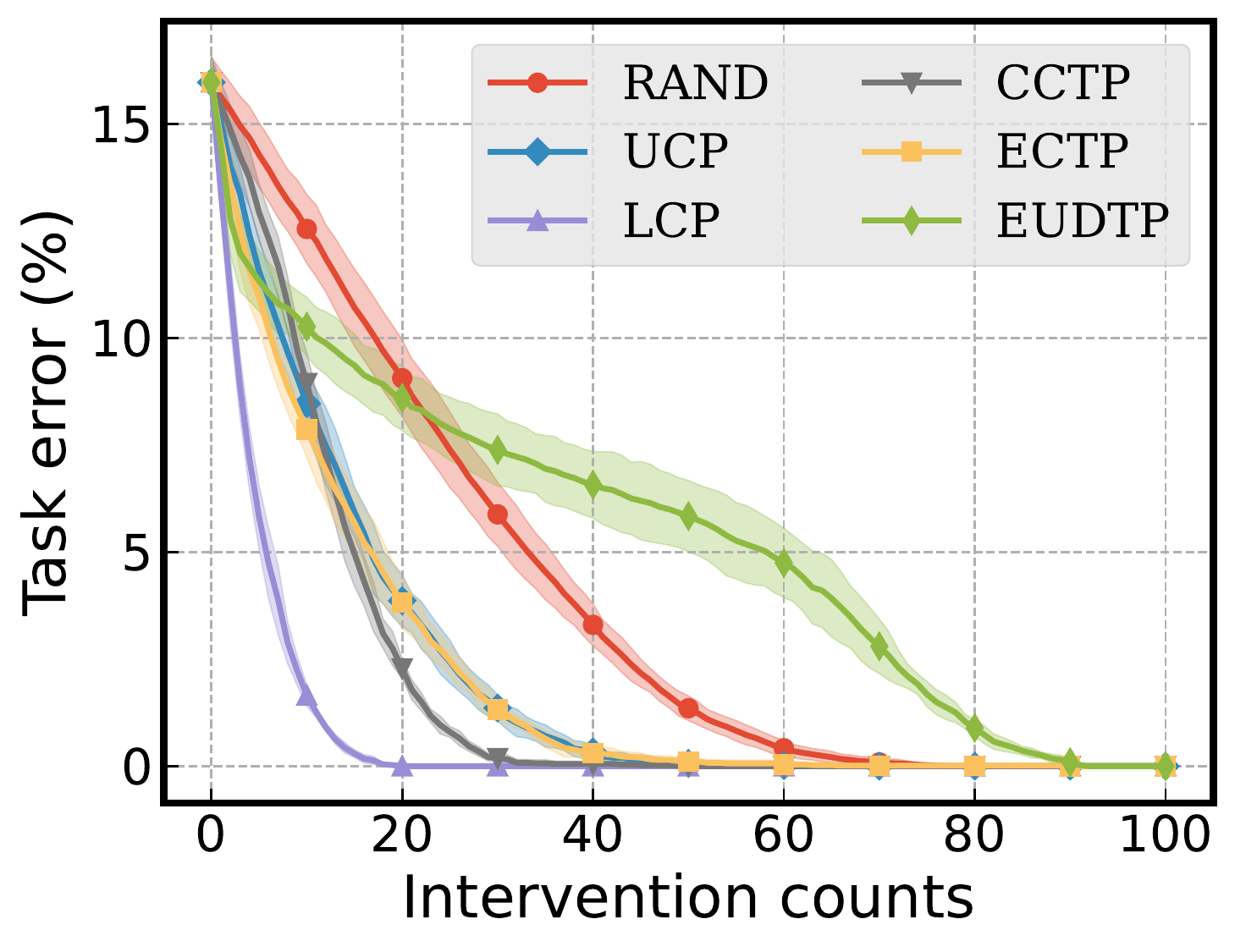}
    \caption{$\gamma = 1$}
    \label{fig:synthetic_subgroupsize_1}
  \end{subfigure}
  \begin{subfigure}{0.48\linewidth}
    \centering
    \includegraphics[width=\linewidth]{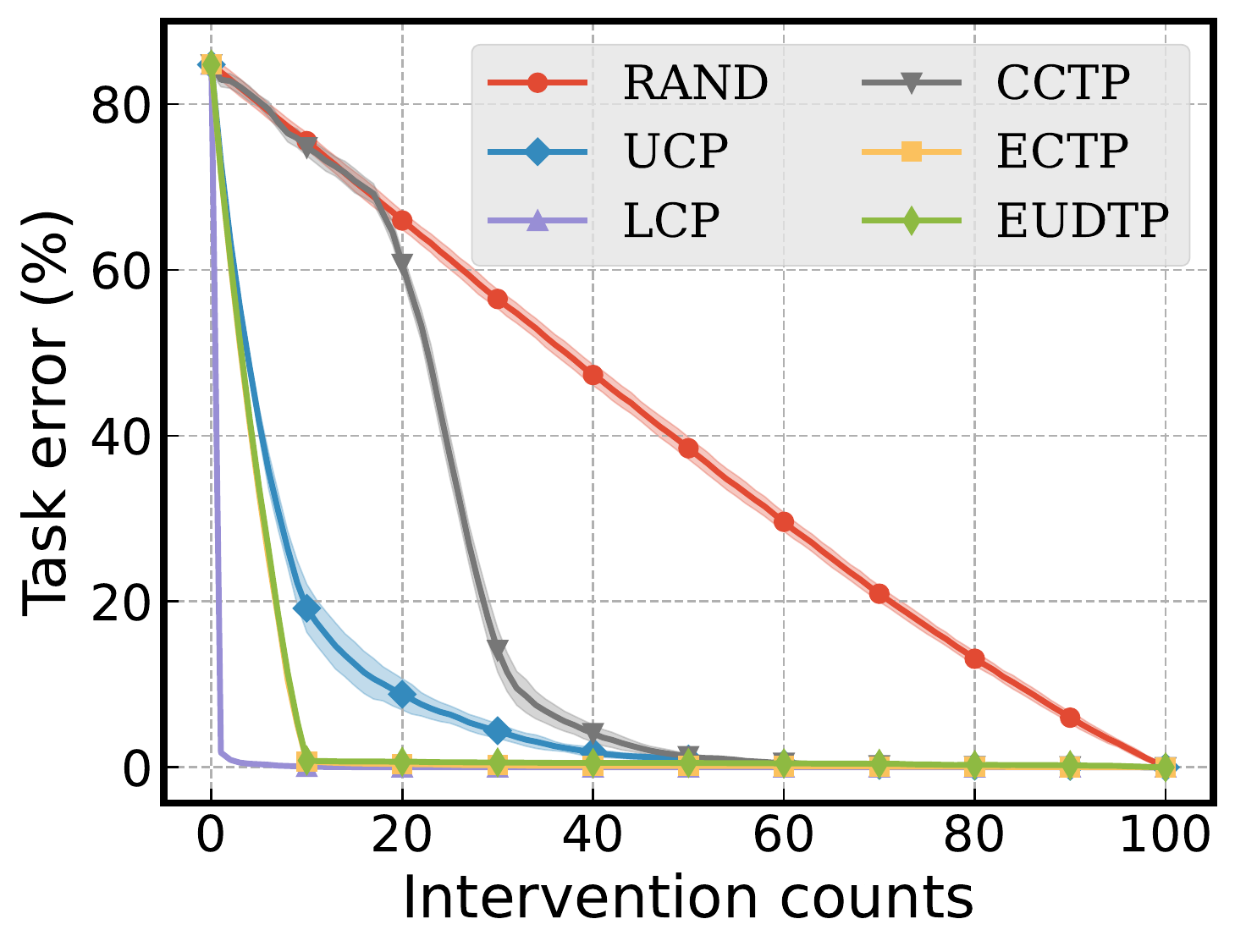}
    \caption{$\gamma = 10$}
    \label{fig:synthetic_subgroupsize_10}
  \end{subfigure}
  \caption{
    Intervention effectiveness with different sub-group size $\gamma$.
    The relative order of effectiveness between selection criteria changes significantly according to $\gamma$.
  }
  \label{fig:synthetic_subgroupsize}
  \vspace{-1em}
\end{figure}

\subsection{Results}
\label{sec:synthetic-result}

First, we display the effect of input noise in \cref{fig:synthetic_inputnoise_ucp}.
The initial task error increases with a level of noise ($z$) due to the poor performance on concept prediction.
Specifically, we need $17$ intervention counts to decrease the task error by half with extremely noisy data ($z=2.0$) while correcting only $2$ concepts yields the same effect for a moderate level of noise case ($z=0.5$).
In contrast, the initial task error is already near $0\%$ with an extremely small level of noise ($z=0.1$) where we do not need intervention at all.

Next, we evaluate the effect of hidden concepts in \cref{fig:synthetic_hidden_ucp}.
The final task error increases with more hidden concepts, and thus, intervention becomes less effective.
Specifically, the error is still high around $13\%$ when half of the concepts are hidden ($h=50\%$) while it reaches zero error without hidden concepts ($h=0\%$).
This is due to the fact that the target prediction cannot be made with complete information when there exist hidden concepts, which is often the case for constructing CBMs in realistic settings.

We also find that generating more diverse concept values within the same class increases both initial and final task errors, making intervention less effective (see \cref{fig:synthetic_diversity_ucp}).
This is because learning discriminative representations for target prediction would be a lot more difficult.
To circumvent this issue, many previous works \citep{koh2020concept,zarlenga2022concept,havasi2022addressing} attempt to preprocess the data so as to force concepts within the same class have the same value.
However, this may have an adverse effect on model fairness as we discuss in \cref{sec:pitfalls}.

Furthermore, we discover that different sub-group sizes can change the relative ordering of intervention effectiveness between concept selection criteria.
Here, we define a sub-group as classes with similar concept values and denote its size as $\gamma$.
Interestingly, \textsc{eudtp} becomes less effective with a small group size ($\gamma = 1$) even compared to \textsc{rand} whereas it becomes the most effective when $\gamma=10$ except for \textsc{lcp} as seen in \cref{fig:synthetic_subgroupsize}.
We believe that it is because classes within the same sub-group are classified more easily by decreasing uncertainty in target prediction using \textsc{eudtp} when $\gamma$ is large.
The result indicates that the behavior of a criterion can vary significantly across different datasets and again demonstrate the necessity of a comprehensive evaluation of the newly developed criteria.
We refer to \cref{sec:results-others-dataset} for results on the effect of some other factors on intervention.
\section{Pitfalls of Intervention Practices}
\label{sec:pitfalls}

So far we have focused on analyzing the effectiveness of intervention procedure in many aspects.
In this section, we add another dimension, namely, reliability and fairness of the current intervention practices, to help advance toward trustworthy and responsible machine learning models.

\subsection{Nullifying Void Concepts Increases Task Error}
\label{sec:nvc}

\begin{figure}[!t]
  \begin{subfigure}{0.31\linewidth}
    \centering
    \includegraphics[width=\linewidth]{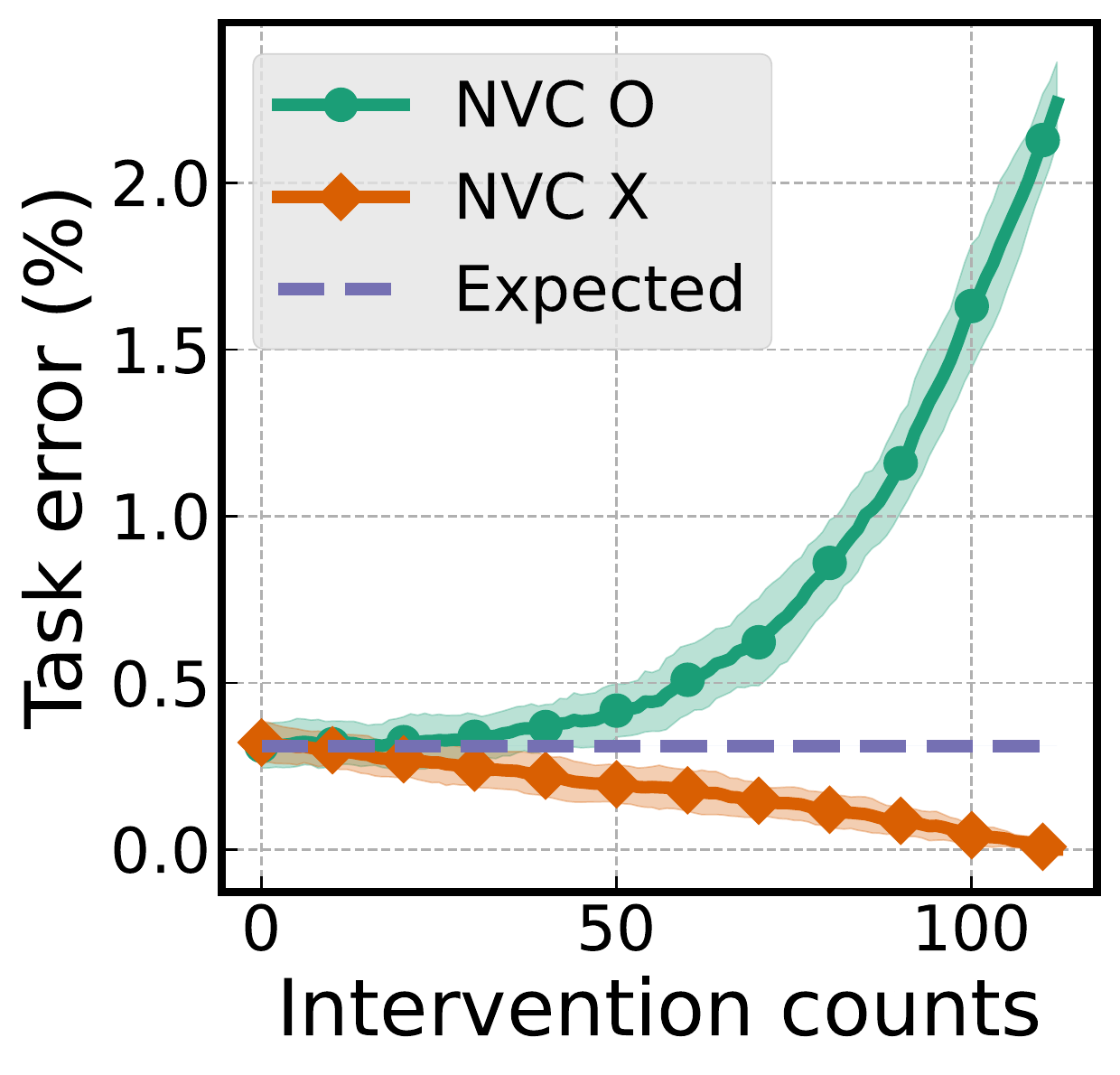}
    \caption{\textsc{rand}}
    \label{fig:cub_nvc_rand_main}
  \end{subfigure}
  \hspace*{\fill}
  \begin{subfigure}{0.31\linewidth}
    \centering
    \includegraphics[width=\linewidth]{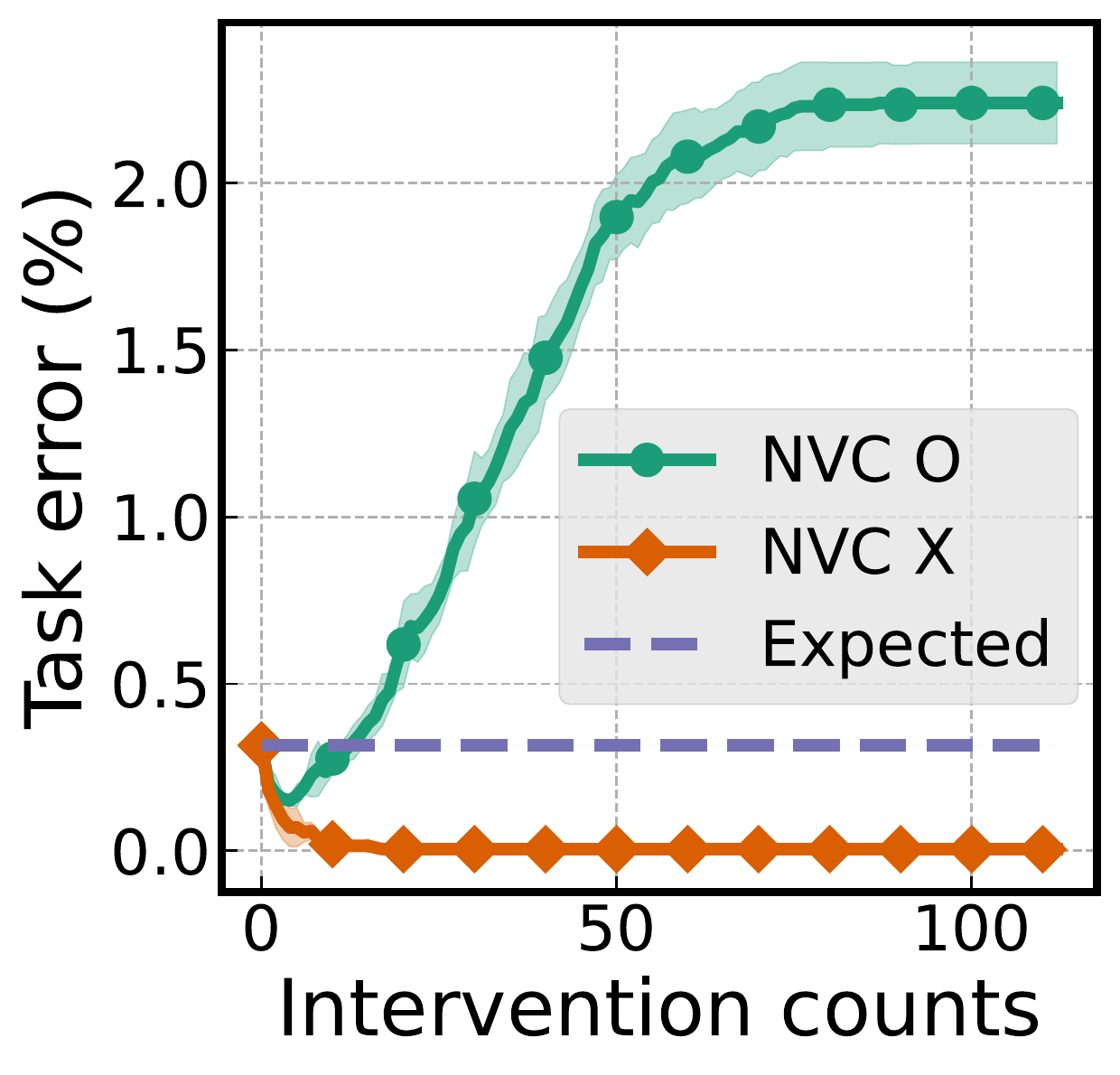}
    \caption{\textsc{ucp}}
    \label{fig:cub_nvc_ucp}
  \end{subfigure}
  \hspace*{\fill}
  \begin{subfigure}{0.31\linewidth}
    \centering
    \includegraphics[width=\linewidth]{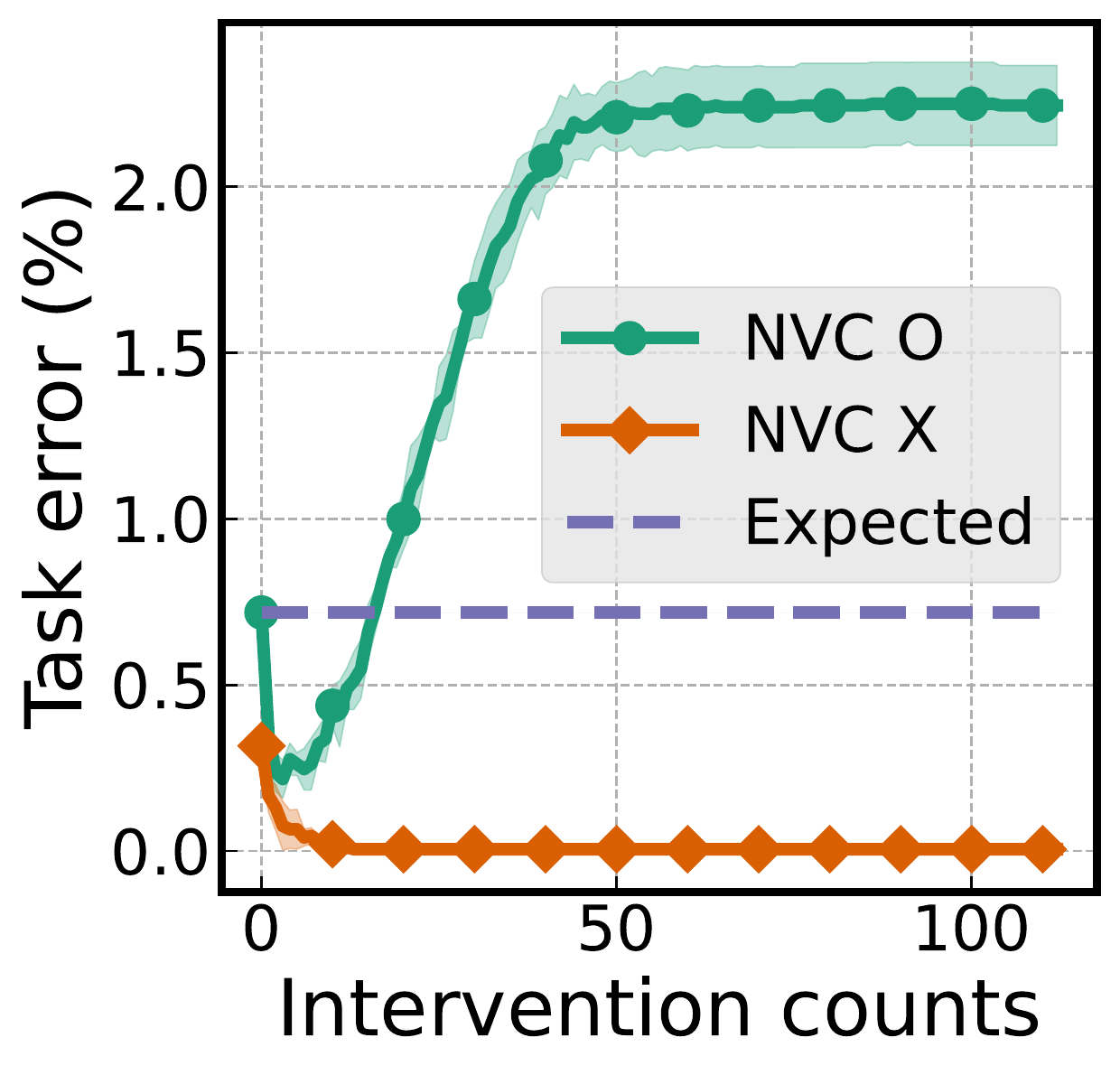}
    \caption{\textsc{ectp}}
    \label{fig:cub_nvc_ectp}
  \end{subfigure}
  \hspace*{\fill}
  \caption{
    Effect of \textsc{nvc} on task error.
    Intervention is done on the CUB images for which concept prediction is $100\%$ accurate, and yet, \textsc{nvc} keeps on increasing the task error.
    \textsc{nvc o} and \textsc{nvc x} each correspond to the result with and without \textsc{nvc}.
  }
  \label{fig:cub_nvc}
\end{figure}

Does intervention always help target prediction?
Contrary to expectation, we find that the answer is no, and in fact, intervention can rather increase the task error.
To verify this, we set up an ablation experiment using the CUB dataset where intervention is conducted only on the cases for which all concepts are predicted correctly with zero error; ideally intervention should have no effect in this case.
The results are quite the opposite as presented in \cref{fig:cub_nvc}.
The task error keeps on increasing as with more intervention, and the prediction error reaches to more than seven times as much as that with no intervention.

It turns out that it is due to nullifying void concepts (\textsc{nvc}), a common practice of treating unsure concepts by setting them to be simply zero, which leads to this catastrophic failure.
For example, just because the wing part of a bird species is invisible does not necessarily mean that the concept `\texttt{wing color:black}' should be zero valued;
this bird can fall in the class of `\texttt{Black\_Tern}' whose wing color is actually black.
We identify that this seemingly plausible tactic can in fact mistreat invalid concepts, and therefore, for invalid cases applying \textsc{nvc} intervention should be avoided.

\subsection{Majority Voting Neglects Minorities}
\label{sec:fairness}
\begin{figure}[!t]
  \begin{subfigure}{0.38\linewidth}
    \includegraphics[width=\linewidth]{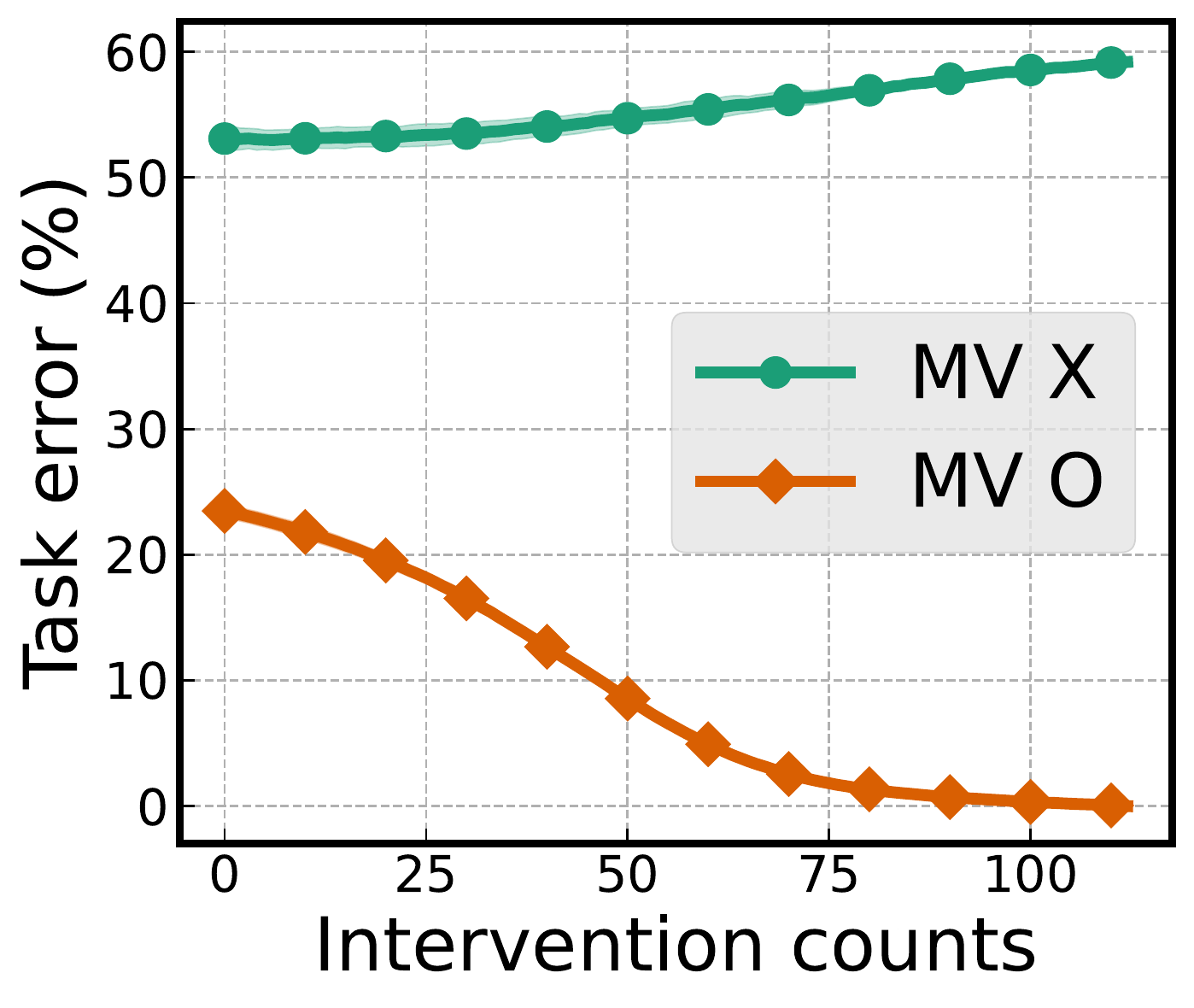}
    \caption{Advantage of \textsc{mv}}
    \label{fig:cub_mv_rand_main}
  \end{subfigure}%
  \hspace*{\fill}
  \begin{subfigure}{0.55\linewidth}
    \includegraphics[width=\linewidth]{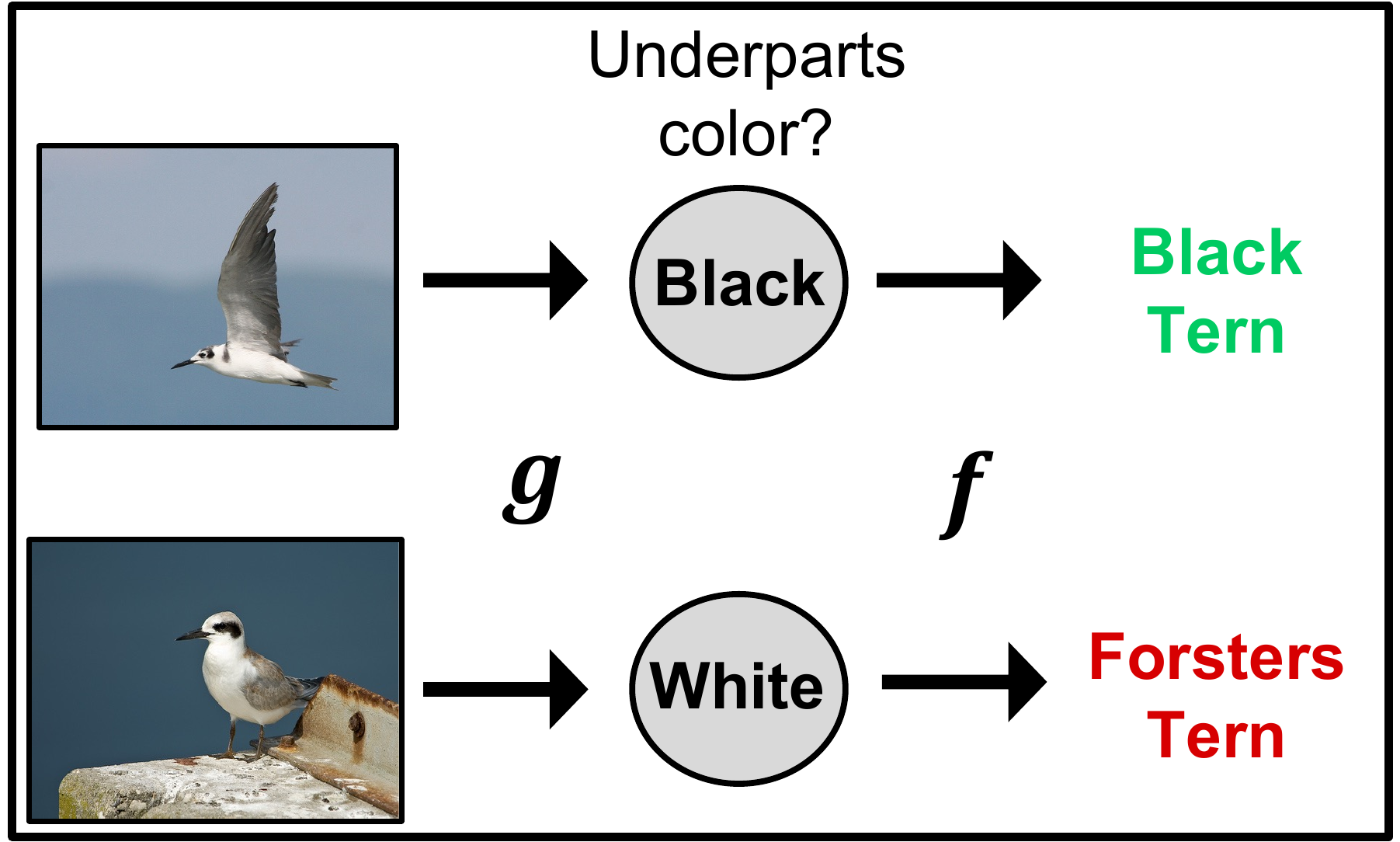}
    \caption{Disadvantage of \textsc{mv}}
    \label{fig:fairness_prediction}
  \end{subfigure}
  \caption{
  Effects of majority voting (\textsc{mv}) on target prediction.
  \textsc{mv o} and \textsc{mv x} each correspond to the result with and without \textsc{mv}.
  (a)
  While it helps decrease task error on intervention,
  (b)
  it yields biased predictions against minorities.
    }
  \label{fig:fairness_example}
\end{figure}

Another common practice often taken by the community \citep{koh2020concept,zarlenga2022concept,havasi2022addressing} is to coalesce concept values among the same class by forcing them to have their majority votes (\textsc{mv}).
As a preprocessing, this tactic can dramatically improve the task performance as we demonstrate in \cref{fig:cub_mv_rand_main}.
This is quite obvious by now as with our Synthetic experiment results in \cref{sec:synthetic-result} where we show that high concept diversity can deteriorate the target prediction performance.

However, it turns out that \textsc{mv} can have a negative impact on model fairness by ignoring minority samples.
As a concrete example, consider the CUB dataset in which the majority of images of `\texttt{black tern}' class have black underparts while some minority samples have white underparts.
When \textsc{mv} is used in this case, we find that the underparts color predictions for the minorities are mis-guided to be black, which correspond to the majority-voted values, so as to yield the correct target prediction;
if the minorities follow their own concept values before \textsc{mv} otherwise, it can lead to an incorrect target prediction (see \cref{fig:fairness_prediction}).
Intervention can even aggravate the situation since it can decrease the task error for the minorities only when the predicted concept value is changed to the majority-voted value (black).
In this sense, target predictions become biased toward the majority when \textsc{mv} is used.

This scenario can be problematic in the real world when the dataset contains sensitive concepts, e.g., gender or race. 
Consider the case where the target task is to predict the occupation of a person based on his/her look and ‘race’ is included in the concepts.
When most ‘physicians’ are Caucasians and if we apply MV in this case, then an ‘Asian physician’ can be correctly classified only when he is predicted as a Caucasian; otherwise, it would lead to an incorrect target prediction. 
While this might be somewhat exaggerated, we remark that this kind of situation can happen in practice.
Besides, \textsc{mv} also forces to misconduct intervention at test time with the majority votes, which is neither available in practice nor considered fair.
We defer addressing the trade-off between performance and fairness to future work.
\section{Discussion and future work}

\vspace{-0.5em}

In this section, we discuss our key findings, their potential implications to the community, and possible future research directions.

\textbf{In-depth analysis of intervention procedure}\quad
We design and conduct a wide variety of new experiments from scratch to investigate the effectiveness of the current intervention procedure of CBMs.
In a nutshell, our results reveal that not only is it the specific way of selecting which concept to intervene, but also how to intervene on what data under which environments matters to the degree of drastically changing results.
Future works can extend our analysis to theoretically investigate the intervention strategies in more detail.

\textbf{Benchmark for evaluating concept selection methods}\quad
Our evaluation protocol can serve as a way to evaluate any newly developed concept selection methods for their effectiveness.
We also provide a framework to generate synthetic data based on which the effectiveness of proposed methods can be tested under various circumstances.

\textbf{Analyzing the cost of intervention}\quad
The effectiveness of concept selection criteria can change when reflecting the cost of intervention (see \cref{sec:results-cost}).
Specifically, we find that a strongly evaluated criterion can become less effective in hypothetical cases considering the size of the models or the status of the labor markets.
This indicates that choosing the concept selection criterion should reflect the available budgets and environments at test time, especially in some extreme environments.

\textbf{Identifying the effect of data on intervention}\quad
The effectiveness of the intervention procedure can vary quite significantly depending on some unknown characteristics of the real-world datasets (see \cref{sec:ablation-dataset}).
For example, intervention becomes less effective on datasets containing more hidden concepts or more diverse concept values within the same class.
Practitioners should take into account this aspect when developing and deploying CBMs since intervention may not work effective as expected.

\textbf{Reliability and fairness of intervention}\quad
While the current trend is mostly focused on developing new intervention methods, we discovered somewhat unexpected and previously unknown issues, which can be critical for ensuring reliability and fairness of the intervention procedure (see \cref{sec:pitfalls}).
To be more specific, intervention can sometimes increase the task error contrary to the expectation and have a negative impact on model fairness by making the predictions biased toward the majority.
We call for future work to address these problems before blindly adopting CBMs in practice.

\textbf{Extension of our work to other settings}\quad
We remark that we have only focused on the classification tasks, considering the characteristics of the real-world datasets used in the literature \citep{koh2020concept,zarlenga2022concept,havasi2022addressing} \footnote{Concept and target variables in the OAI dataset \citep{nevitt2006osteoarthritis} take $4$ integer values and thus the tasks can be easily converted into the classification problem.}.
Extension of the intervention strategies to the regression problems with real-valued concepts or targets can be a promising avenue for future works.
Analyzing intervention under more diverse settings could also be interesting, such as introducing architectural variations with hard autoregressive models \citep{havasi2022addressing} or concept embedding models \citep{zarlenga2022concept}.
\section{Conclusion}

The intervention procedure of CBMs has been unattended in previous work despite its critical impact on practitioners.
In this work, we study a wide range of aspects regarding the procedure and provide an in-depth analysis for the first time in the literature.
Specifically, we develop various concept selection criteria that can be used for intervention and demonstrate that their behaviors can vary quite significantly based on an array of factors including intervention levels, cost, training, conceptualization, and data characteristics.
We also find several pitfalls in the current practices that need a careful addressing to be deployed in realistic settings.
We plan to investigate further on developing more effective and reliable intervention strategies in future work.

\section*{Acknowledgement}

This work was partly supported by Institute of Information \& communications Technology Planning \& Evaluation (IITP) grant funded by the Korea government (MSIT) (No.2019-0-01906, Artificial Intelligence Graduate School Program (POSTECH) and No.2022-0-00959, (part2) Few-Shot learning of Causal Inference in Vision and Language for Decision Making) and National Research Foundation of Korea (NRF) grant funded by the Korea government (MSIT) (2022R1C1C1013366, 2022R1F1A1064569, RS-2023-00210466).

\nocite{langley00}

\bibliography{references}
\bibliographystyle{icml2023}

\newpage
\appendix
\onecolumn
\section{Datasets}
\label{app:datasets}

\subsection{CUB}

CUB \citep{wah2011caltech} is the standard dataset used to study CBMs in the previous works \citep{koh2020concept, zarlenga2022concept,havasi2022addressing, sawada2022concept}.
There are $5994$ and $5794$ examples for train and test sets in total, in which each example consists of the triplet of (image $x$, concepts $c$, label $y$) of a bird species.
All the concepts have binary values; for example, the `\texttt{wing color:black}' for a given bird image can be either $1$ (for true) or $0$ (for false).
Following previous works \citep{koh2020concept,sawada2022concept,zarlenga2022concept}, we perform so-called majority voting as pre-processing so that images of the same class always have the same concept values; for example, if more than half of the crow images have true value for the concept `\texttt{wing color:black}' then this process converts all concept labels for images belonging to the crow class to have the same true value.
Since the original concept labels are too noisy, this procedure helps to increase the overall performance.
However, it can be potentially harmful to model fairness in some cases as we address in \cref{sec:fairness}.
We also remove concepts that are too sparse (\ie, concepts that are present in less than $10$ classes) which results in $112$ out of $312$ concepts remaining.
It is suggested in \citet{koh2020concept} that including these sparse concepts in the concept layer makes it hard to predict their values as the positive training examples are too scarce.

\subsection{SkinCon}

SkinCon \citep{daneshjouskincon} is a medical dataset which can be used to build interpretable machine learning models.
The dataset provides densely annotated concepts for $3230$ images from Fitzpatrick 17k skin disease dataset \citep{groh2021evaluating}, which makes a triplet of (image $x$, concepts $c$, disease label $y$) of a skin lesion for each example.
Since training and test sets are not specified in the SkinCon dataset, we randomly split the dataset into $70\%, 15\%, 15\%$ of training, validation, and test sets respectively.
The dataset provides various levels of class labels ranging from individual disease labels with $114$ classes to binary labels representing if the skin is benign or malignant.
Following the experiments with Post-hoc CBM \citep{yuksekgonul2022post} introduced in \citet{daneshjouskincon}, we use the binary labels for the target task and only use $22$ concepts which are present in at least $50$ images.
Since the binary class labels are highly imbalanced ($87\%$ vs. $13\%$), we train the target predictor $f$ with weighted loss and use the average of per-class error as the metric instead of overall error for a fair comparison.

\subsection{Synthetic dataset}
\label{sec:synthetic_dataset}

\begin{algorithm}[H]
  \begin{algorithmic}[1]
  \STATE Sample $p_i \sim \mathcal{N} (\mu_\alpha, \sigma_\alpha)$ for $i = \{1, 2, \cdots, k\}$
  \FOR{group $\ell = 0, 1, \cdots, k/\gamma - 1$} 
      \STATE Sample $\zeta_i \sim \mathcal{U}_{[0,1]}$ and set $\ell_i = \mathbbm{1} [\zeta_i \geq p_i]$ for $i = \{1, 2, \cdots, k\}$
      \FOR{$y = 1, \cdots, \gamma$}
      \STATE Sample $i_y \in \{1, 2, \cdots, k\}$ uniformly at random without replacement
      \STATE Set $c_i^j = \neg \ell_i$ if $i = i_y$ and $c_i^j = \ell_i$ otherwise (class index $j = \gamma * \ell + y$)
      \ENDFOR
  \ENDFOR
  \STATE Generate  $W_x \in \mathbb{R}^{k \times k}$ with each element distributed according to the unit normal distribution $\mathcal{N}(0, \sigma_w)$
  \FOR{class $j = 1, \cdots, k$}
  \STATE Generate $\nu$ samples for class $j$ as $x = W_x \cdot c^j + z$ where $z \sim \mathcal{N} (0, \sigma_z)$
  \ENDFOR
  \end{algorithmic}
  \caption{Generating synthetic data}
  \label{alg:synthetic_full_algorithm}
\end{algorithm}

We generate the synthetic data following \cref{alg:synthetic_full_algorithm} to test the effect of dataset characteristics on intervention.
Here, we first assume that all examples within the same class share the same concept values and denote the $i$-th concept value of $j$-th class as $c_i^j$.
We also assume for simplicity that the dimensionality of inputs and the number of target classes are the same as the number of concepts $k$, following \citet{bahadori2020debiasing}.
In line $1$,  $\mu_\alpha$ and $p_i = P(c_i = 0)$ each represent the overall sparsity level of the concepts (proportion of concepts with value $0$) and the probability of $i$-th concept taking value $0$, respectively.
We set $\mu_\alpha$ to be $0.8$ considering that $80\%$ of the concepts have value $0$ in the CUB dataset.
We then divide classes into $k/\gamma$ sub-groups of size $\gamma$ to make those within the same group have similar concept values.
Note that the classes within each sub-group only differ by two concept values as seen in line $6$.
We set $\gamma=2, k = 100, \nu = 100, \sigma_\alpha = 0.1, \sigma_w = 0.1, z_\alpha = 0.8$ unless stated otherwise.
We randomly divide the generated examples into $70\%$ of training sets, $15\%$ of validation sets, and $15\%$ of test sets.

To generate the data with hidden concepts, we randomly pick $h\%$ of the concepts and remove them from the concept layer of CBMs. 
For training the models and intervention experiments, we only consider the remaining concepts.
In addition, a new dataset with diverse concepts can be easily produced by introducing a single variable $d$ and reversing the value of each concept from the previously generated dataset with probability $d$.
In other words, $d$ stands for a factor to give variations to concept-target pairs that can exist in real world datasets, and it differs from the role of $z$ which controls the noise level to the input.

\section{Architectures and Training}
\label{app:implementation}

\paragraph{CUB}
For the CUB dataset, we use Inception-v3 \citep{szegedy2016rethinking} pretrained on Imagenet \citep{deng2009imagenet} for the concept predictor $g$ and $1$-layer MLP for the target predictor $f$ respectively following the standard setup as in \citet{koh2020concept}.
Here, both $g$ and $f$ are trained with the same training hyperparameters as in \citet{koh2020concept}.
We used $\lambda=0.01$ for \textsc{jnt} and \textsc{jnt+p} whose values were directly taken from \citet{koh2020concept}.
For the experiments without majority voting (\cref{fig:cub_mv} in \cref{sec:results-others-fairness}), we use Inceptionv3 pretrained on the Imagenet for $g$ and 2-layer MLP for $f$ with the dimensionality of $200$ so that it can describe more complex functions.
We searched the best hyperparameters for both $g$ and $f$ over the same sets of values as in \citet{koh2020concept}.
Specifically, we tried initial learning rates of $[0.01, 0.001]$, constant learning rate and decaying the learning rate by $0.1$ every $[10, 15, 20]$ epoch, and the weight decay of $[0.0004, 0.00004]$.
After finding the optimal values of hyperparameters whose validation accuracy is the best, we trained the networks with the same values again over 5 different random seeds on both training and validation sets.

\paragraph{SkinCon}
For the SkinCon dataset, we fine-tune Deepderm \citep{daneshjou2022disparities} for the concept predictor $g$, which is the Inception-v3 network trained on the data in \citet{esteva2017dermatologist}, and train $1$-layer MLP for the target predictor $f$.
We select hyperparameters that achieve the best performance (in terms of overall accuracy and average per-class accuracy for $g$ and $f$ respectively) in the validation set.
Specifically, we tried initial learning rates of $[0.0005, 0.001, 0.005]$, and constant learning rate and decaying the learning rate by $0.1$ every $50$ epoch.
Here, we did not use the weight decay factor.
For \textsc{jnt} and \textsc{jnt+p} training strategies, we tried concept loss weight $\lambda$ of $[0.01, 0.1, 1.0, 5.0]$, but all of the values failed to decrease the task error at intervention.
As in the CUB dataset, we trained the networks with the best hyperparameters over $5$ different random seeds on the both training and validation sets.

\paragraph{Synthetic}
For the synthetic datasets, we use $3$-layer MLP of hidden layer size $\{100, 100\}$ for $g$ and a single linear layer for $f$, as similar to \citet{zarlenga2022concept}.
For all the experiments, we tried constant learning rates of $[0.01, 0.1, 1.0]$ without learning rate decay or weight decay factor and trained the networks with the best hyperparameters over $5$ different random seeds on the training sets.
We used $\lambda=0.1$ for \textsc{jnt} and \textsc{jnt+p} whose values were determined by grid search over $[0.01, 0.1, 1.0]$.

\vspace{2em}

\section{More on Reflecting Cost of Intervention}
\label{app:cost}

\begin{figure}[!t]  
  \begin{subfigure}{0.19\linewidth}
    \centering
    \includegraphics[width=\linewidth]{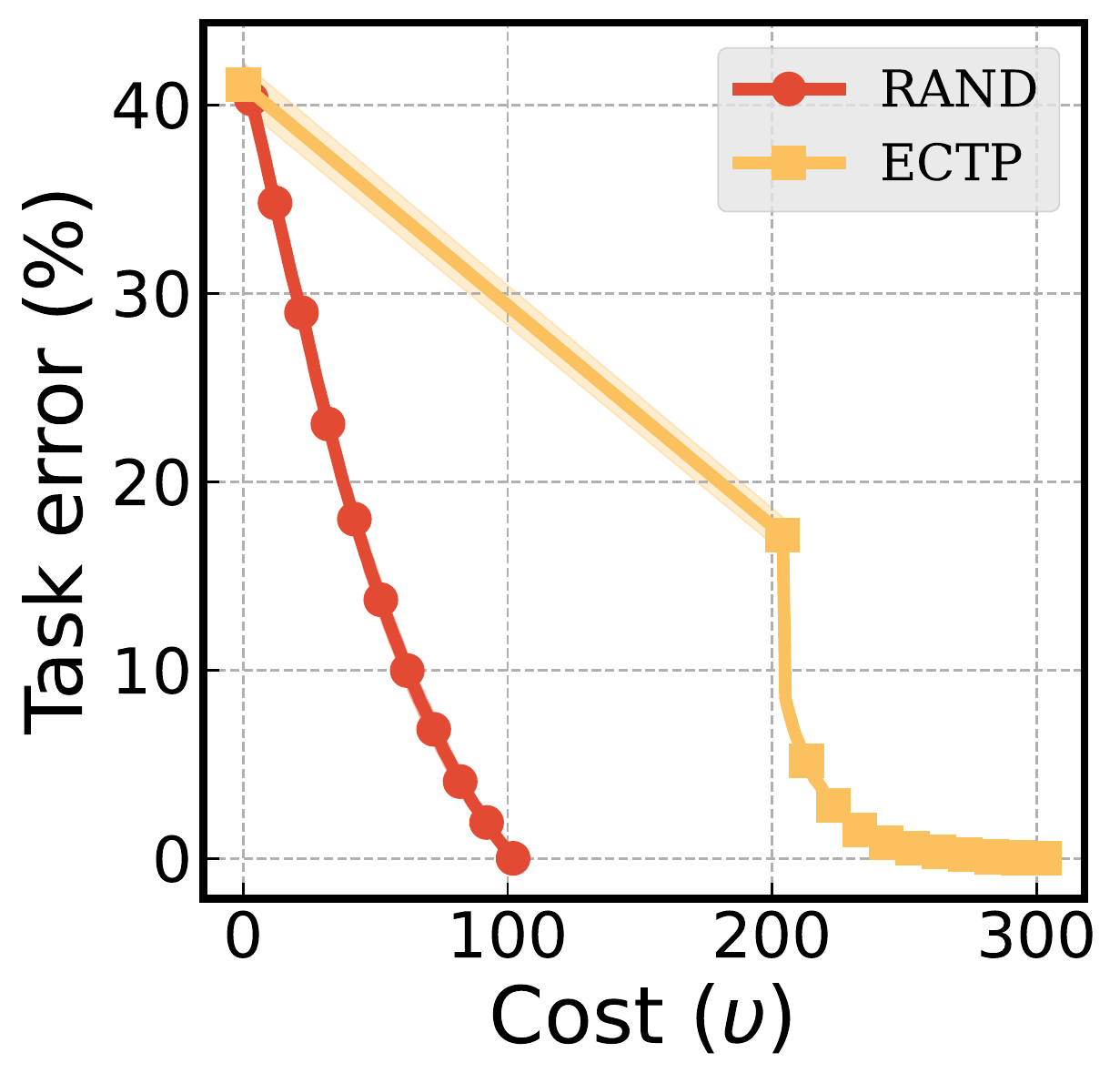}
    \caption{$\beta = 1$}
    \label{fig:cost_beta1}
  \end{subfigure}%
  \hspace*{\fill}
    \begin{subfigure}{0.19\linewidth}
    \centering
    \includegraphics[width=\linewidth]{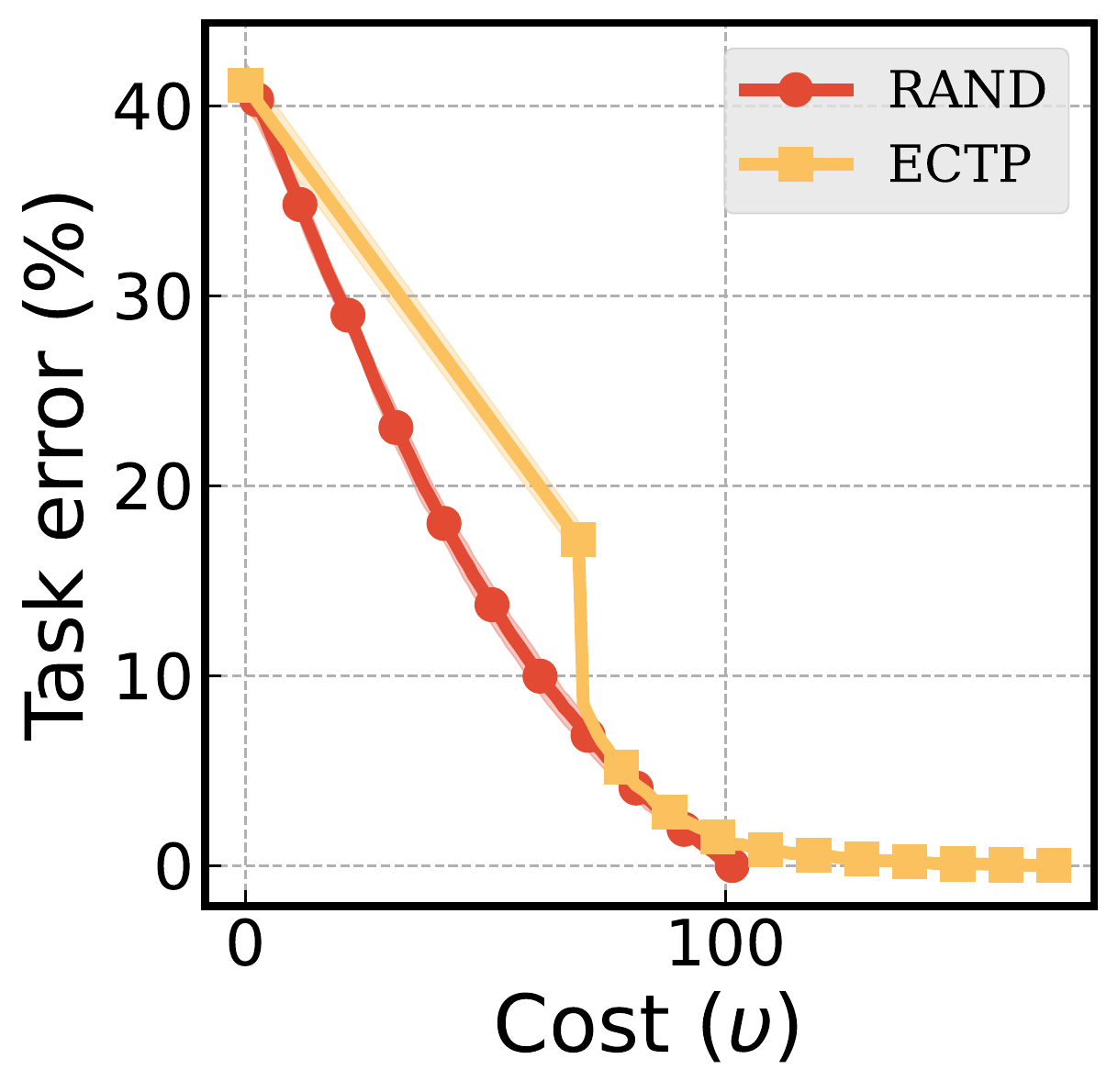}
    \caption{$\beta = 3$}
    \label{fig:cost_beta3}
  \end{subfigure}%
  \hspace*{\fill}
  \begin{subfigure}{0.19\linewidth}
    \centering
    \includegraphics[width=\linewidth]{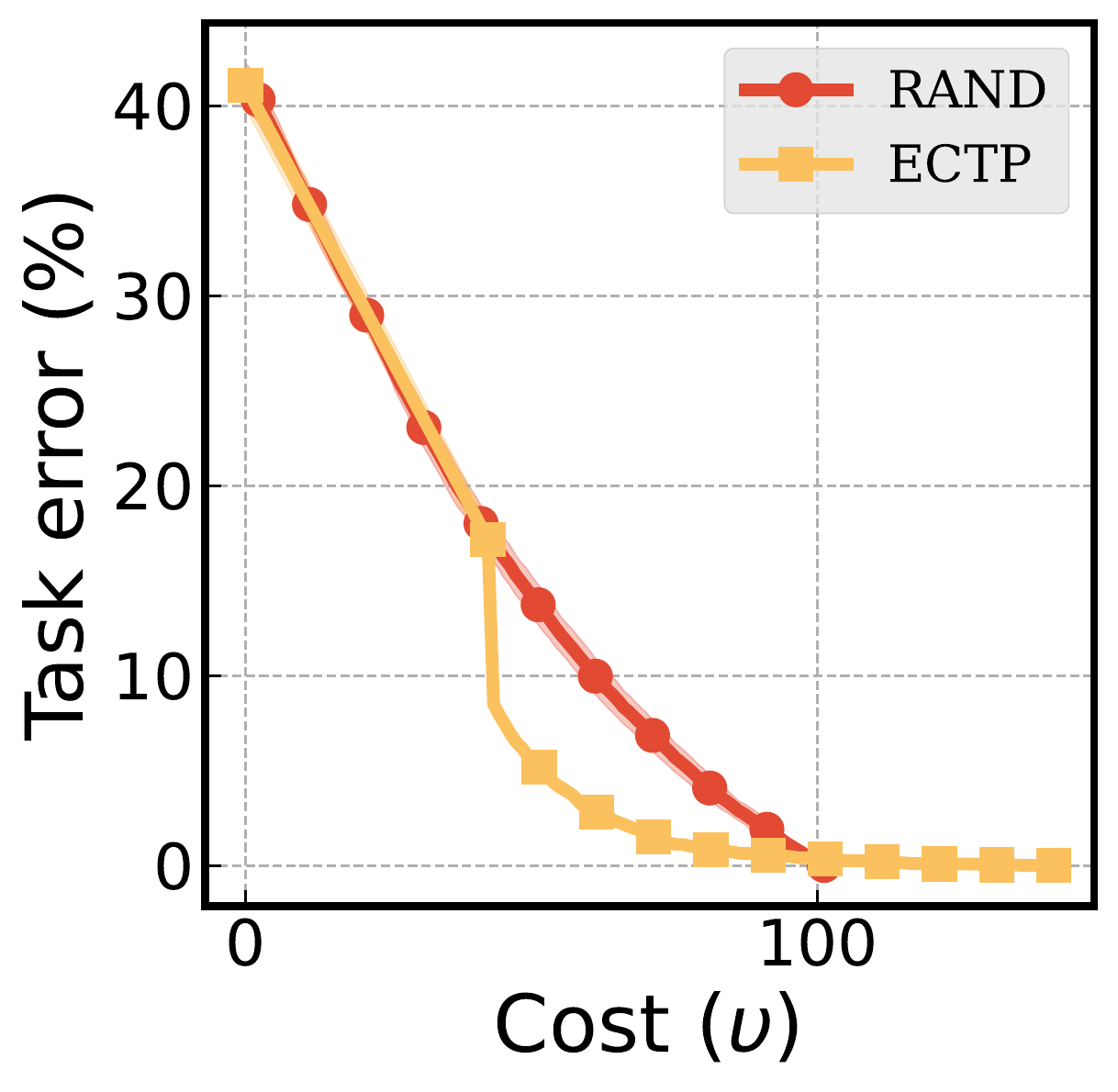}
    \caption{$\beta = 5$}
    \label{fig:cost_beta5}
  \end{subfigure}%
  \hspace*{\fill}
  \begin{subfigure}{0.19\linewidth}
    \centering
    \includegraphics[width=\linewidth]{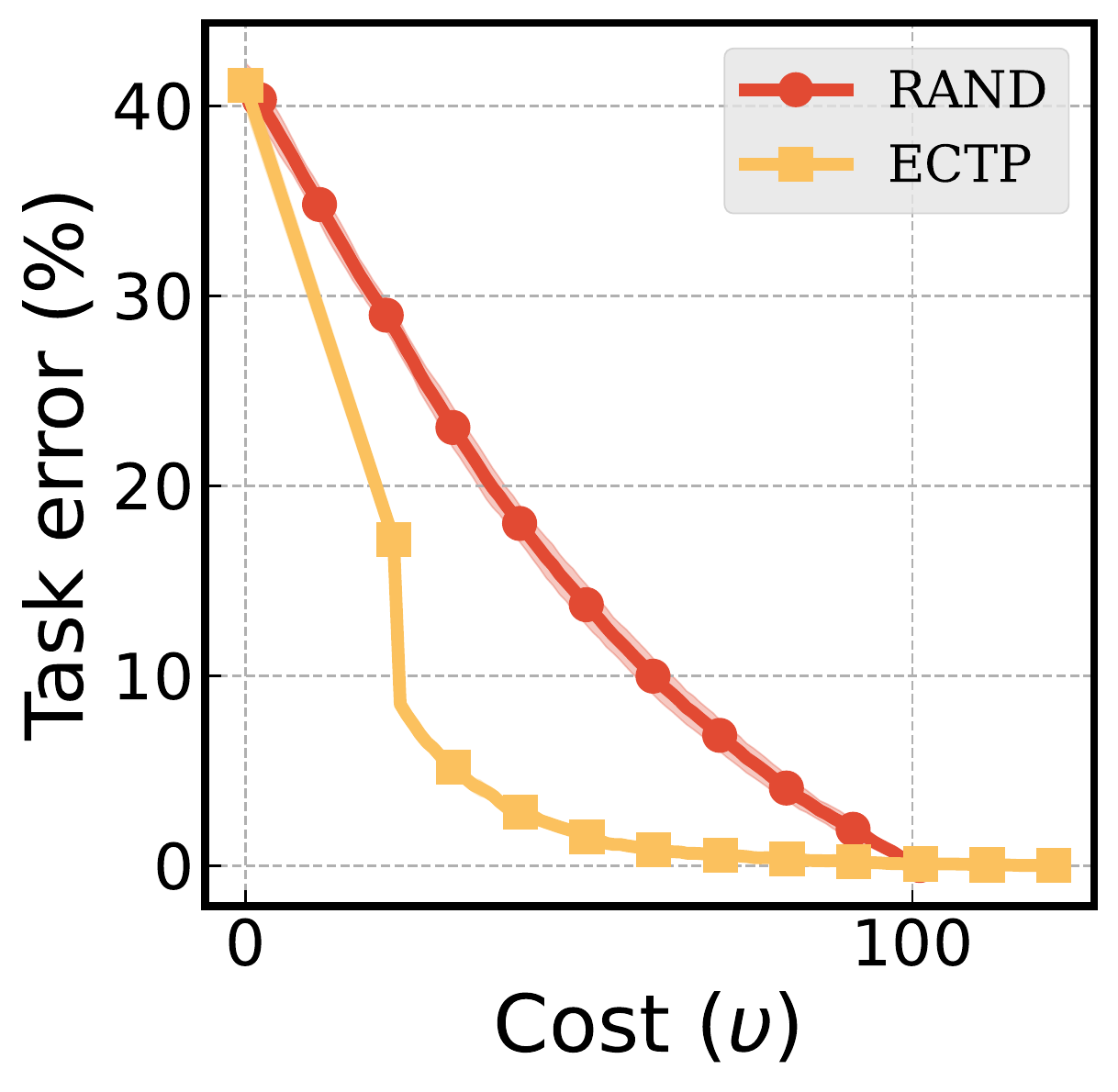}
    \caption{$\beta = 10$}
    \label{fig:cost_beta10}
  \end{subfigure}%
  \hspace*{\fill}
  \begin{subfigure}{0.19\linewidth}
    \centering
    \includegraphics[width=\linewidth]{figures/synthetic/cost_a1.0_b100.pdf}
    \caption{$\beta = 100$}
    \label{fig:cost_beta100}
  \end{subfigure}%
  \caption{
  Effect of $\beta$ on intervention.
  We fix $\tau_i = 1, \alpha = 1, k=100$.
  \textsc{ectp}, the concept selection criteria strongly evaluated previously, becomes less effective as $\beta$ decreases.
  }
  \label{fig:cost_result_beta}
\end{figure}

\begin{figure}[!t]
\centering
\begin{subfigure}{0.2\linewidth}
  \centering
  \includegraphics[width=\linewidth]{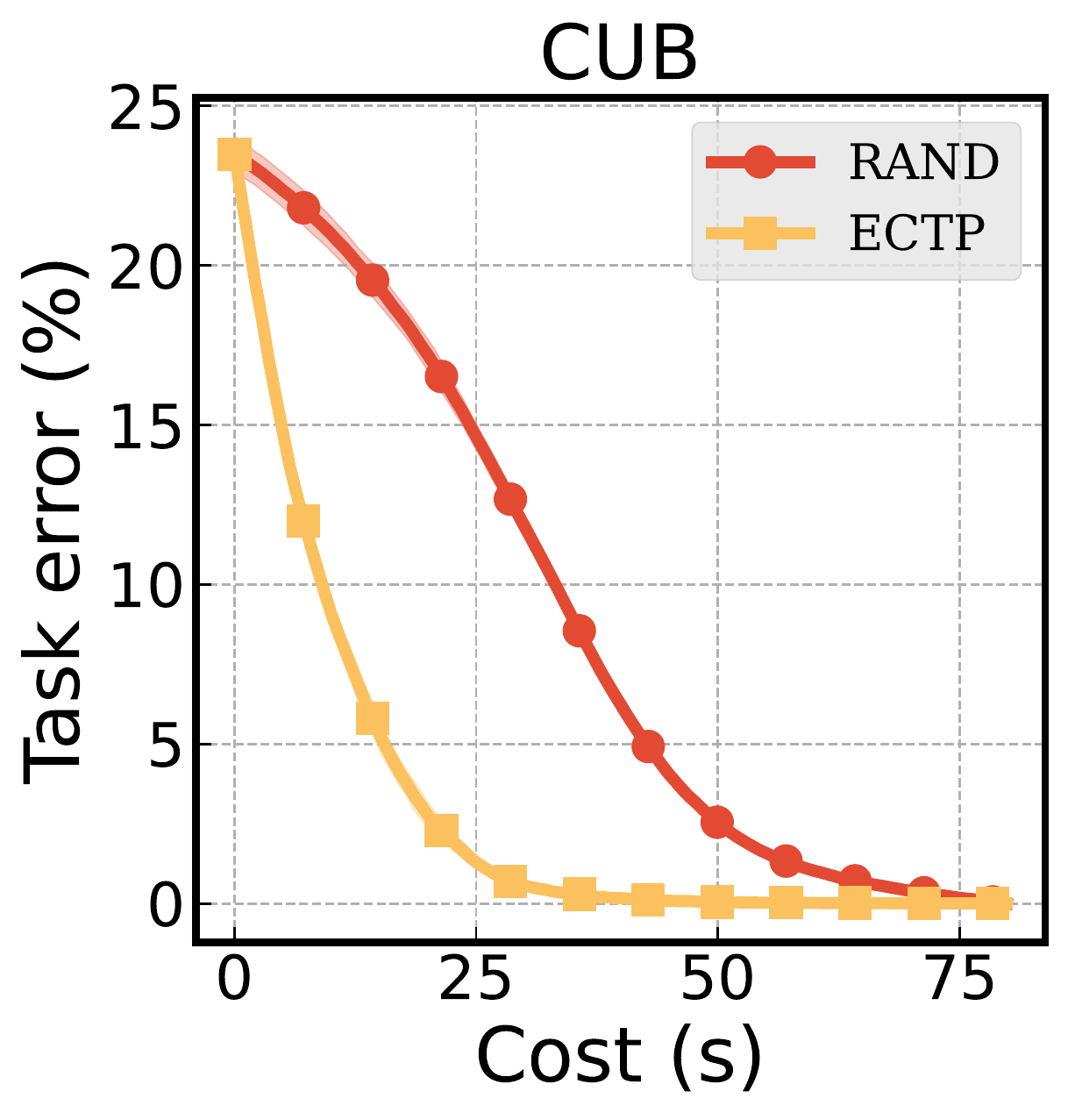}
  \caption{$\tau_i$ set as the average among all concepts}
  \label{fig:cost_cub1}
\end{subfigure}%
\hspace*{3em}
\begin{subfigure}{0.2\linewidth}
  \centering
  \includegraphics[width=\linewidth]{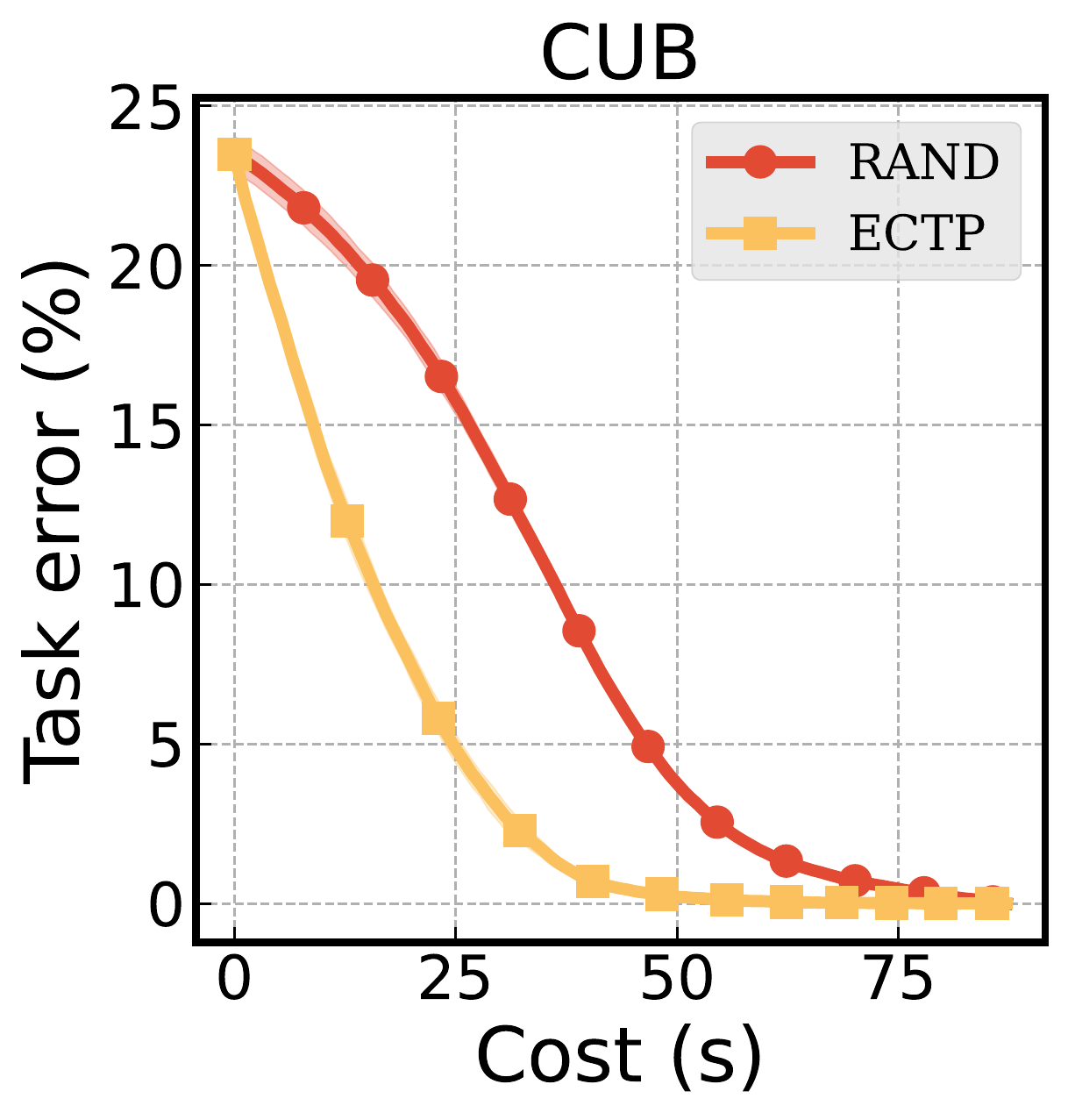}
  \caption{$\tau_i$ set as the average per each intervention step}
  \label{fig:cost_cub2}
\end{subfigure}%
\caption{
Comparison between concept selection criteria in terms of the intervention cost for the CUB.
Here, cost represents the seconds for concept annotation time and inference times for $g, f$.
}
\label{fig:cost_cub}
\end{figure}

As $\beta$ becomes smaller \textsc{rand} becomes more effective compared to \textsc{ectp} (see \cref{fig:cost_result_beta}).
This is because with small $\beta$, $\tau_g$ becomes marginalized in the cost of \textsc{ectp} which is $\mathcal{O}(\tau_g + (2k+2)\tau_f + n \tau_i)$, and therefore, the intervention effectiveness of \textsc{ectp} is penalized as with increasing $k$ compared to \textsc{rand} which only requires a single forward pass of $f$.

In addition, we experiment with more realistic settings for the CUB where we set $\tau_i$ as the concept annotation time (seconds) provided in the dataset and $\tau_g, \tau_f$ as the wall-clock times for the inference.
Specifically, we set $\tau_i \approx 0.7$ by dividing the annotation time into the number of concepts within the group and taking the average.
In addition, $\tau_g \approx 18.7 * 10^{-3}$ and $\tau_f \approx 0.03 * 10^{-3}$ are acquired by measuring the inference time with RTX 3090 GPU and taking the average of $300$ repetitions.
In this setting, $\tau_i$ dominates the others, \ie, $\alpha$ is large, and the relative effectiveness between the criteria remains the same as seen in \cref{fig:cost_cub}.
Nonetheless, we remark that the result can change with different model sizes or GPU environments in extreme cases.
We also considered a more detailed case where we do not directly take the average of $\tau_i$'s (concept annotation time) at once but rather take the average per intervention step, reflecting differences of intervention costs between different concepts.
The relative rankings between \textsc{rand} and \textsc{ectp} do not change but interestingly we have found that \textsc{ectp} first selects the concepts which require more intervention costs (\ie, more concept annotation time).

\section{More Results on the Effect of Intervention Levels on Intervention}
\label{sec:results-others-levels}

\begin{figure}[!th]
  \begin{subfigure}{0.24\linewidth}
    \centering
    \includegraphics[width=\linewidth]{figures/cub/result_main.pdf}
    \caption{\textsc{i+s}}
    \label{fig:cub_result_level_is}
  \end{subfigure}
  \begin{subfigure}{0.24\linewidth}
    \centering
    \includegraphics[width=\linewidth]{figures/cub/result_level_gs.pdf}
    \caption{\textsc{g+s}}
    \label{fig:cub_result_level_gs}
  \end{subfigure}
  \begin{subfigure}{0.24\linewidth}
    \centering
    \includegraphics[width=\linewidth]{figures/cub/result_level_ib.pdf}
    \caption{\textsc{i+b}}
    \label{fig:cub_result_level_ib}
  \end{subfigure}
  \begin{subfigure}{0.24\linewidth}
    \centering
    \includegraphics[width=\linewidth]{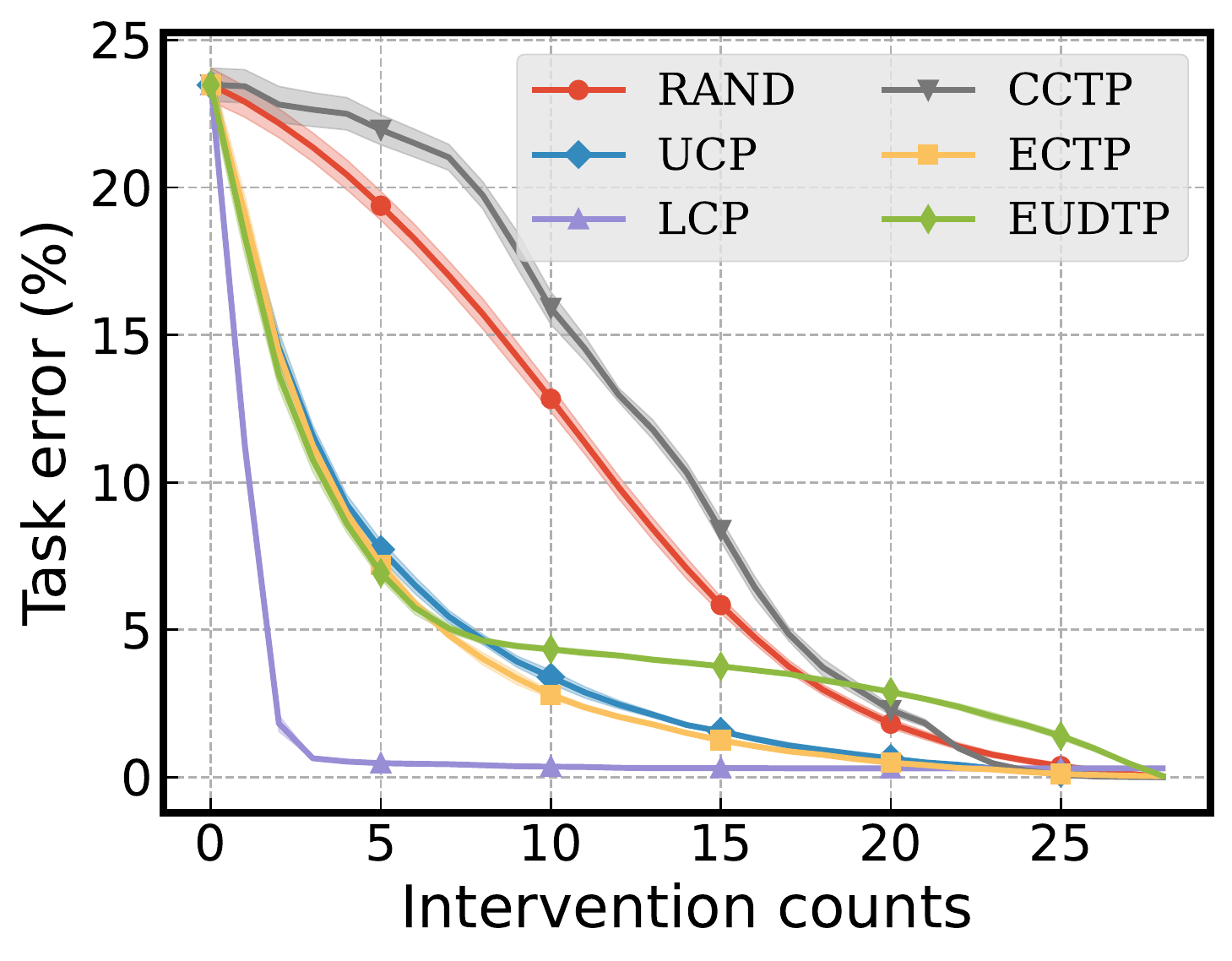}
    \caption{\textsc{g+b}}
    \label{fig:cub_result_level_gb}
  \end{subfigure}
  \caption{Comparison between intervention criteria under different levels for the CUB.}
  \label{fig:cub_result_level}
\end{figure}

\begin{figure}[!th]
  \begin{subfigure}{0.16\linewidth}
    \centering
    \includegraphics[width=\linewidth]{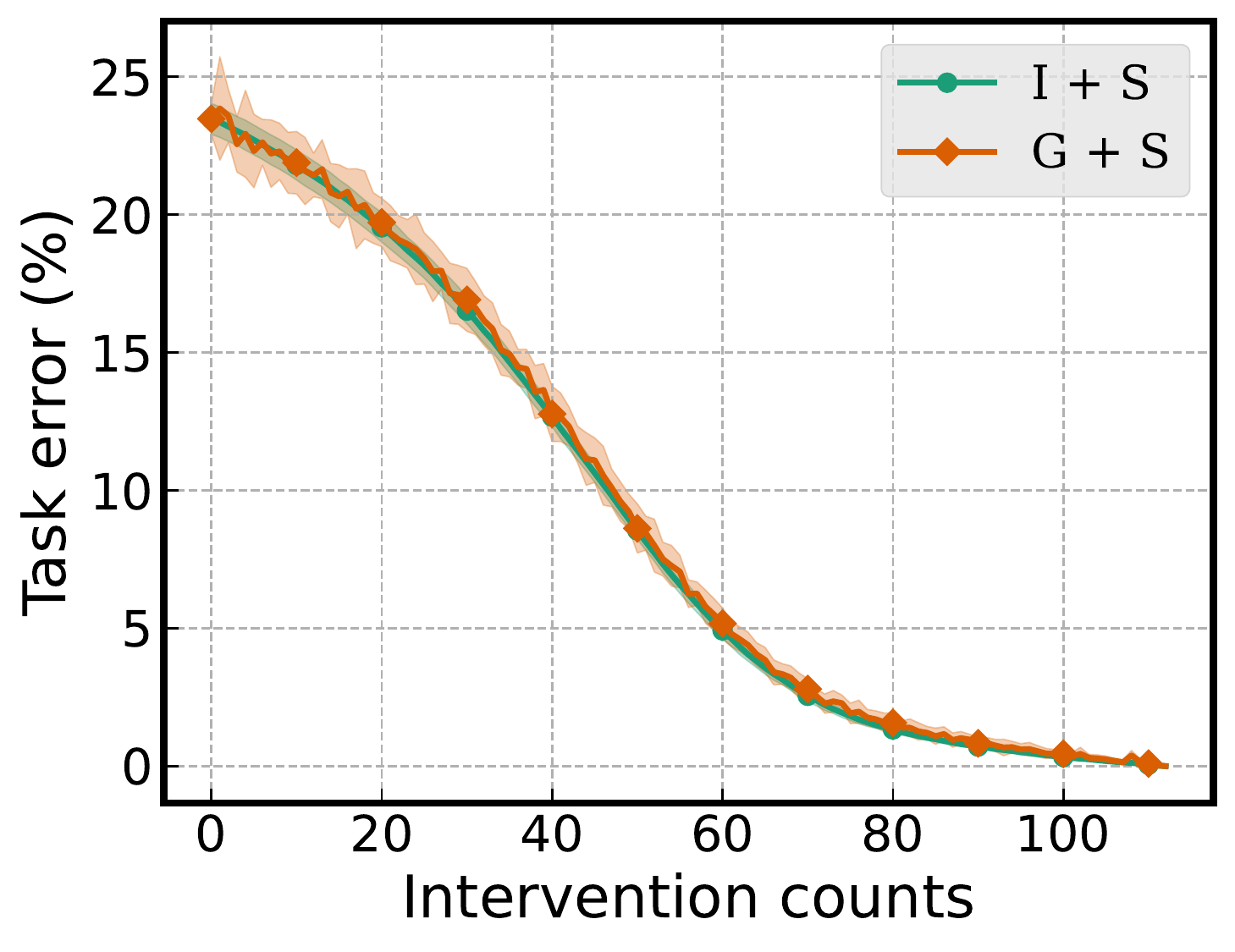}
    \caption{\textsc{rand}}
    \label{fig:cub_level_is_gs_rand}
  \end{subfigure}
  \begin{subfigure}{0.16\linewidth}
    \centering
    \includegraphics[width=\linewidth]{figures/cub/level_is_gs_ucp.pdf}
    \caption{\textsc{ucp}}
    \label{fig:cub_level_is_gs_ucp}
  \end{subfigure}
  \begin{subfigure}{0.16\linewidth}
    \centering
    \includegraphics[width=\linewidth]{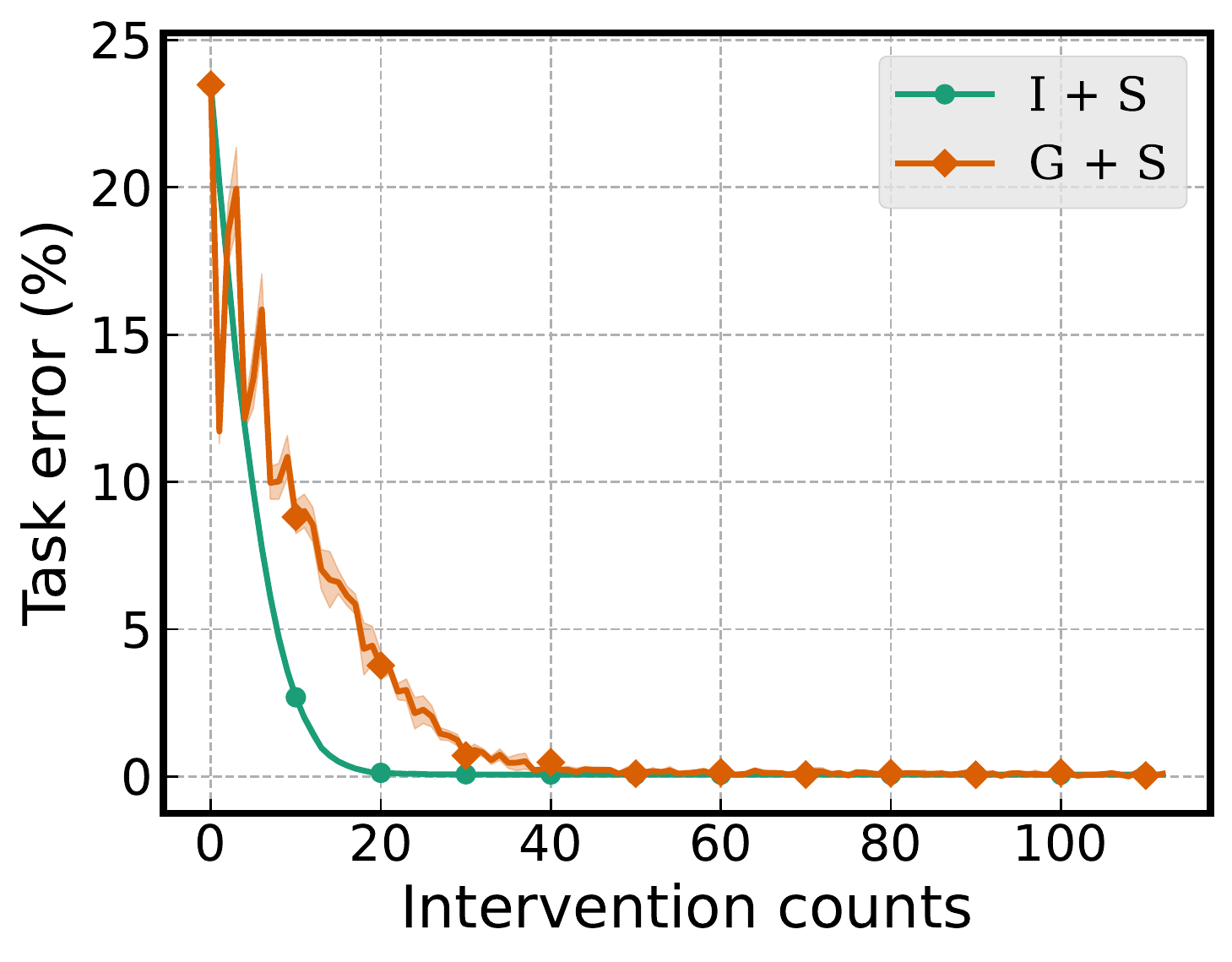}
    \caption{\textsc{lcp}}
    \label{fig:cub_level_is_gs_lcp}
  \end{subfigure}
  \begin{subfigure}{0.16\linewidth}
    \centering
    \includegraphics[width=\linewidth]{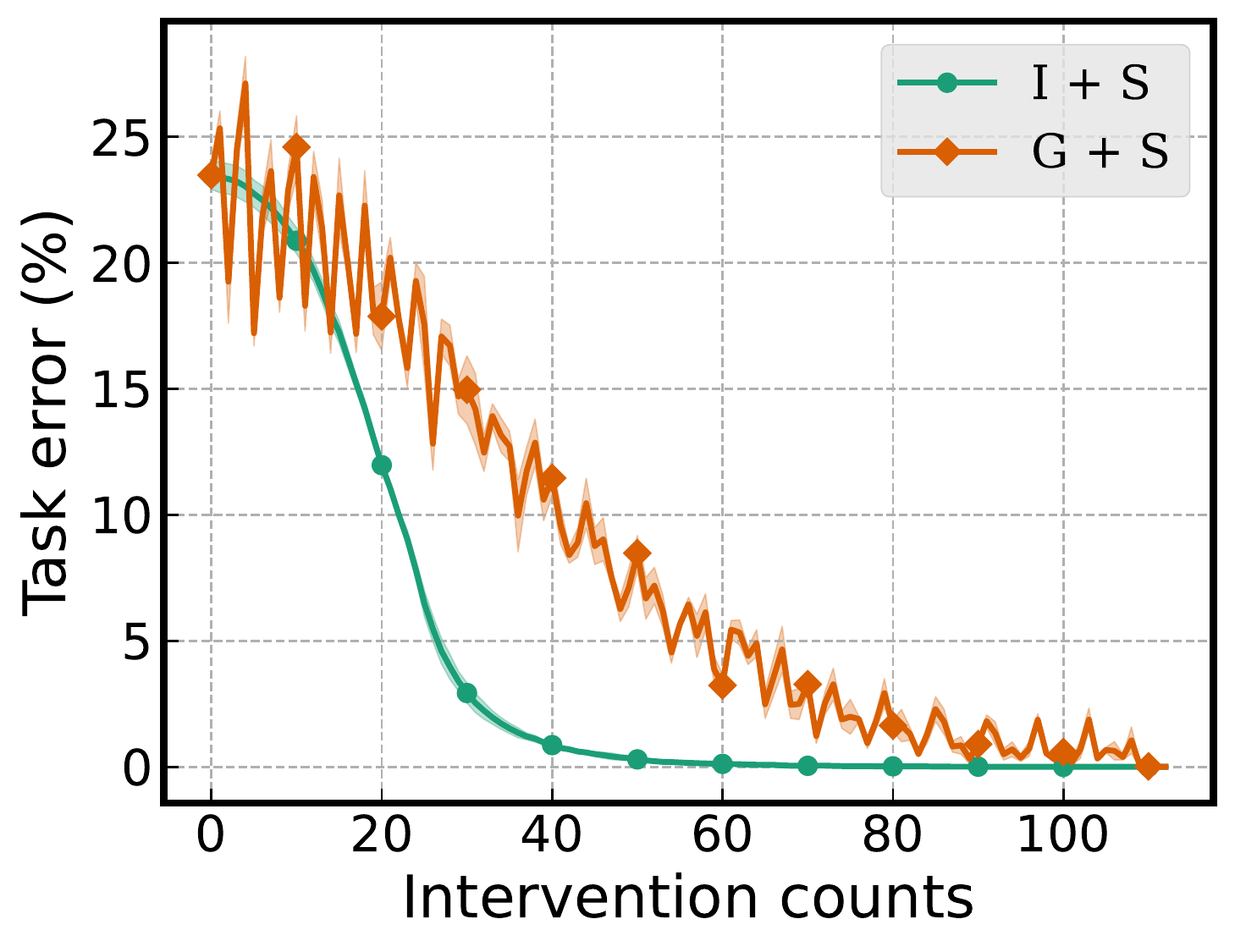}
    \caption{\textsc{cctp}}
    \label{fig:cub_level_is_gs_cctp}
  \end{subfigure}
  \begin{subfigure}{0.16\linewidth}
    \centering
    \includegraphics[width=\linewidth]{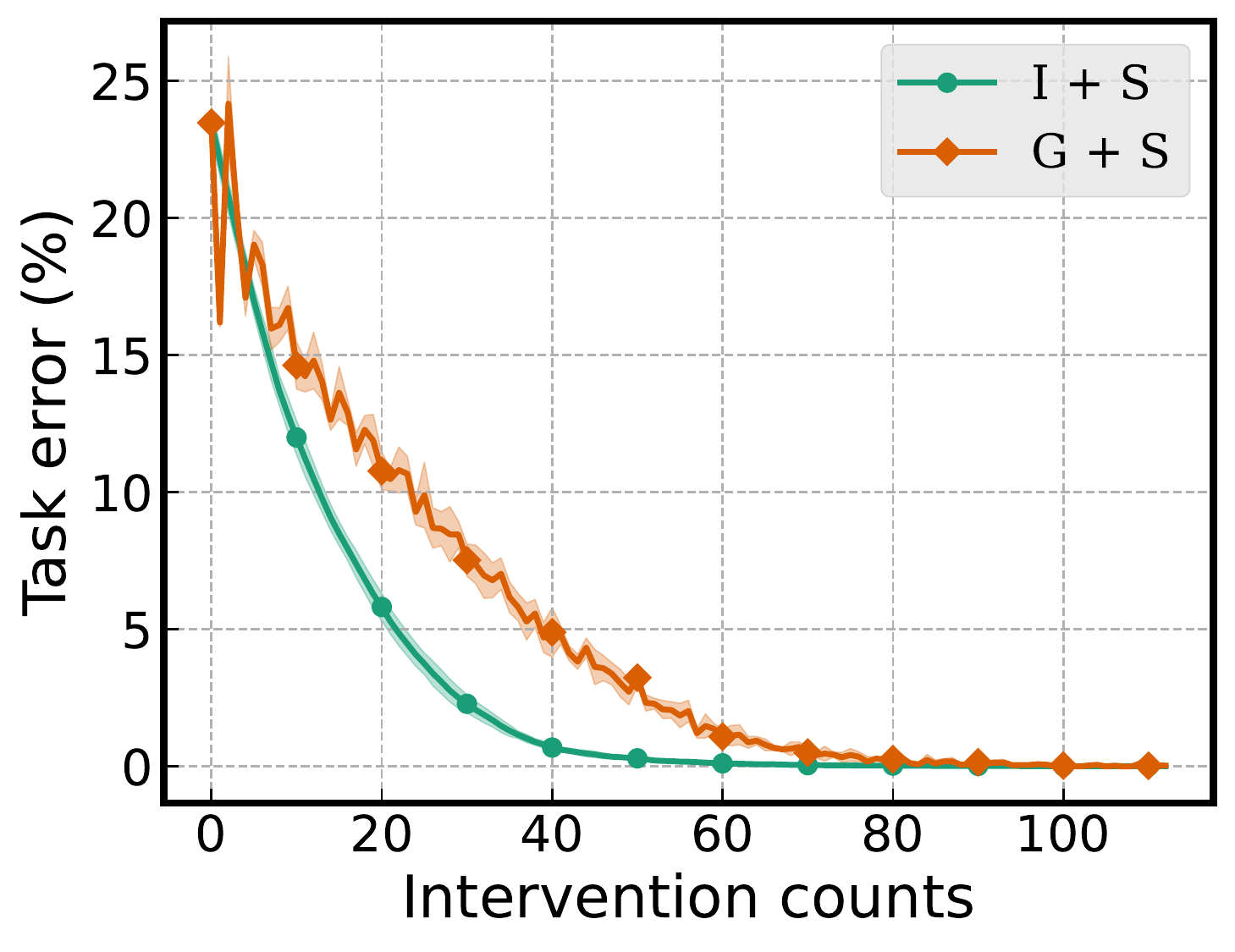}
    \caption{\textsc{ectp}}
    \label{fig:cub_level_is_gs_ectp}
  \end{subfigure}
  \begin{subfigure}{0.16\linewidth}
    \centering
    \includegraphics[width=\linewidth]{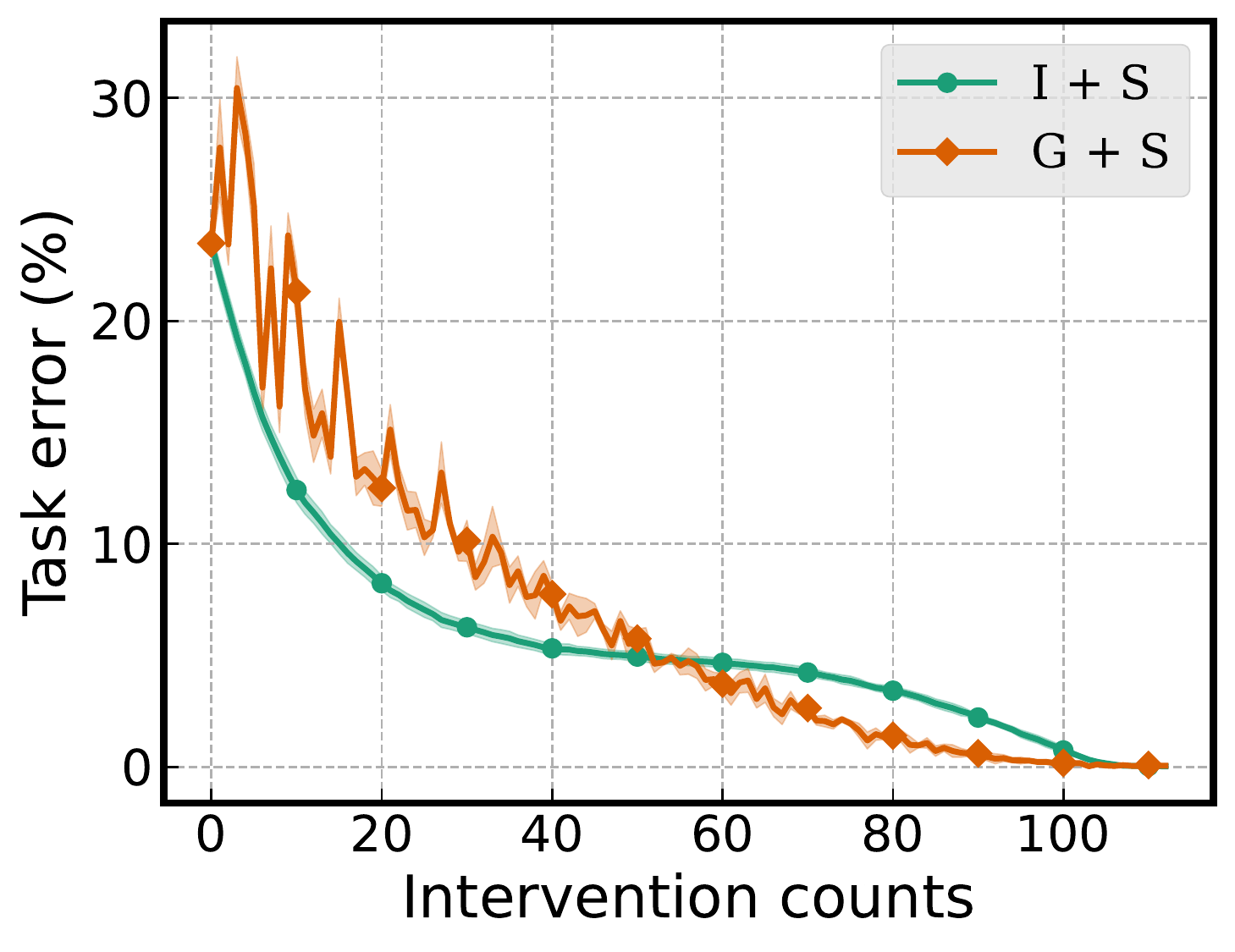}
    \caption{\textsc{eudtp}}
    \label{fig:cub_level_is_gs_eudtp}
  \end{subfigure}
  \caption{
    Comparison between \textsc{i+s} vs. \textsc{g+s} for the CUB.
  }
  \label{fig:cub_level_is_gs_all}
\end{figure}

\begin{figure}[!th]
  \begin{subfigure}{0.16\linewidth}
    \centering
    \includegraphics[width=\linewidth]{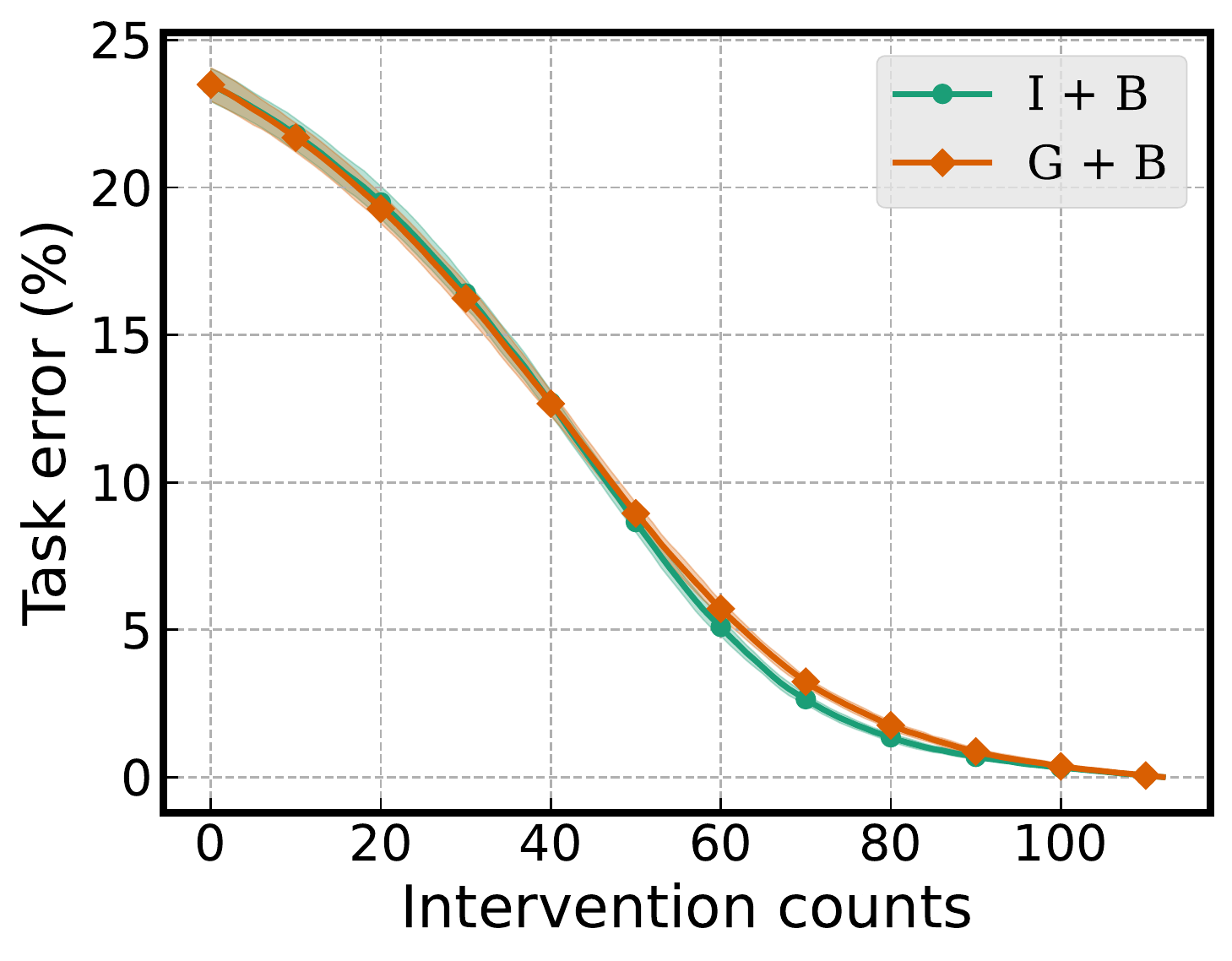}
    \caption{\textsc{rand}}
    \label{fig:cub_level_ib_gb_rand}
  \end{subfigure}
  \begin{subfigure}{0.16\linewidth}
    \centering
    \includegraphics[width=\linewidth]{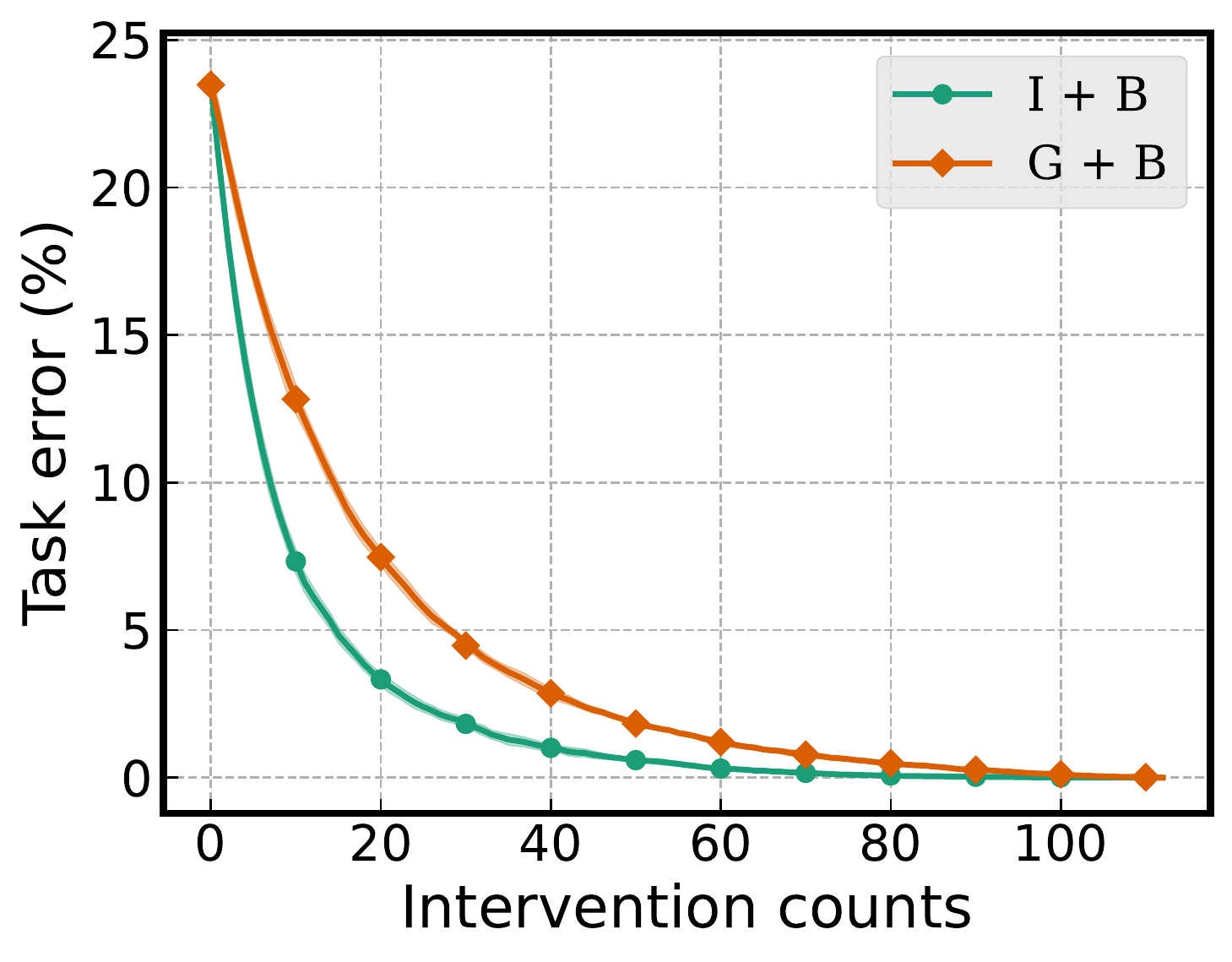}
    \caption{\textsc{ucp}}
    \label{fig:cub_level_ib_gb_ucp}
  \end{subfigure}
  \begin{subfigure}{0.16\linewidth}
    \centering
    \includegraphics[width=\linewidth]{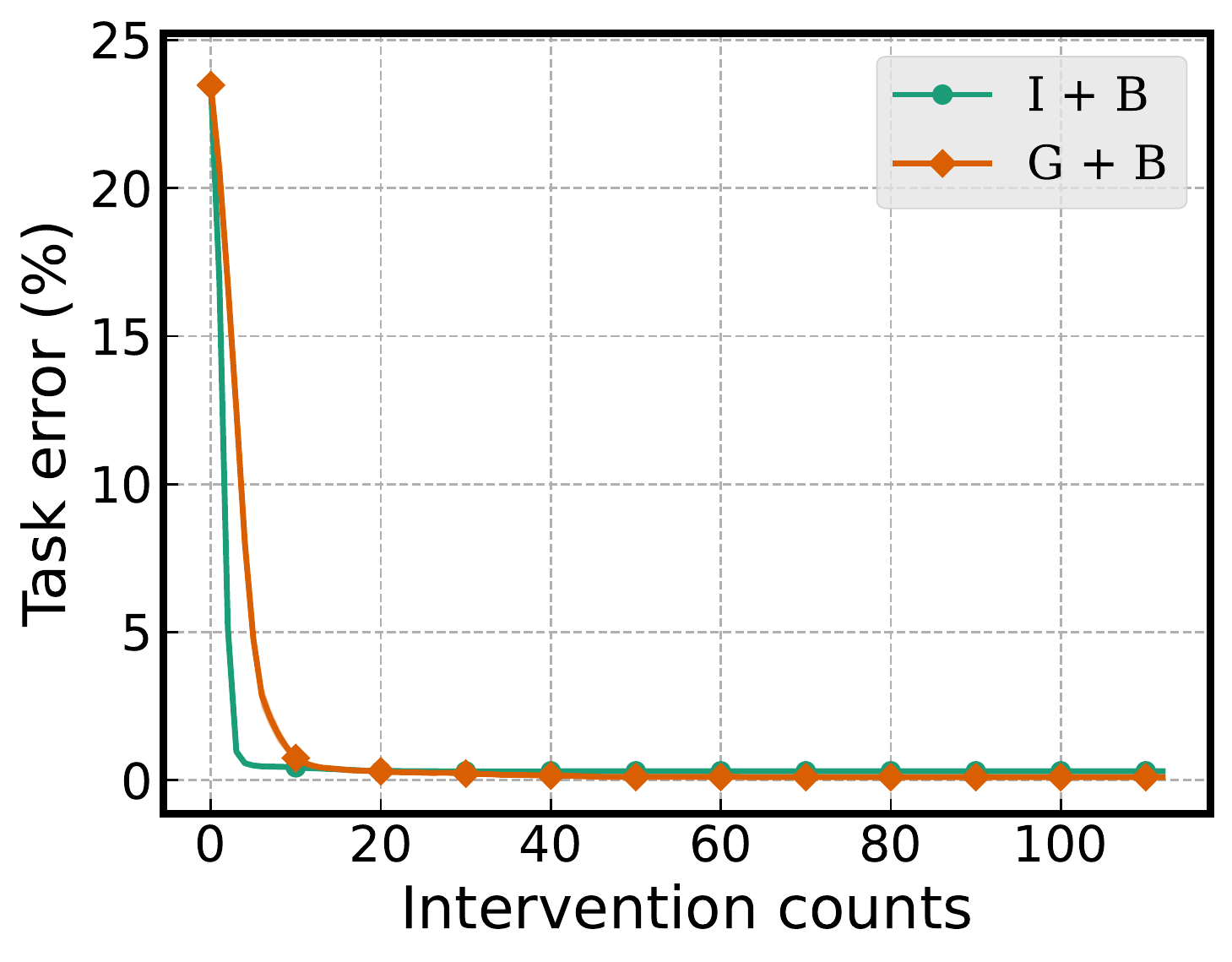}
    \caption{\textsc{lcp}}
    \label{fig:cub_level_ib_gb_lcp}
  \end{subfigure}
  \begin{subfigure}{0.16\linewidth}
    \centering
    \includegraphics[width=\linewidth]{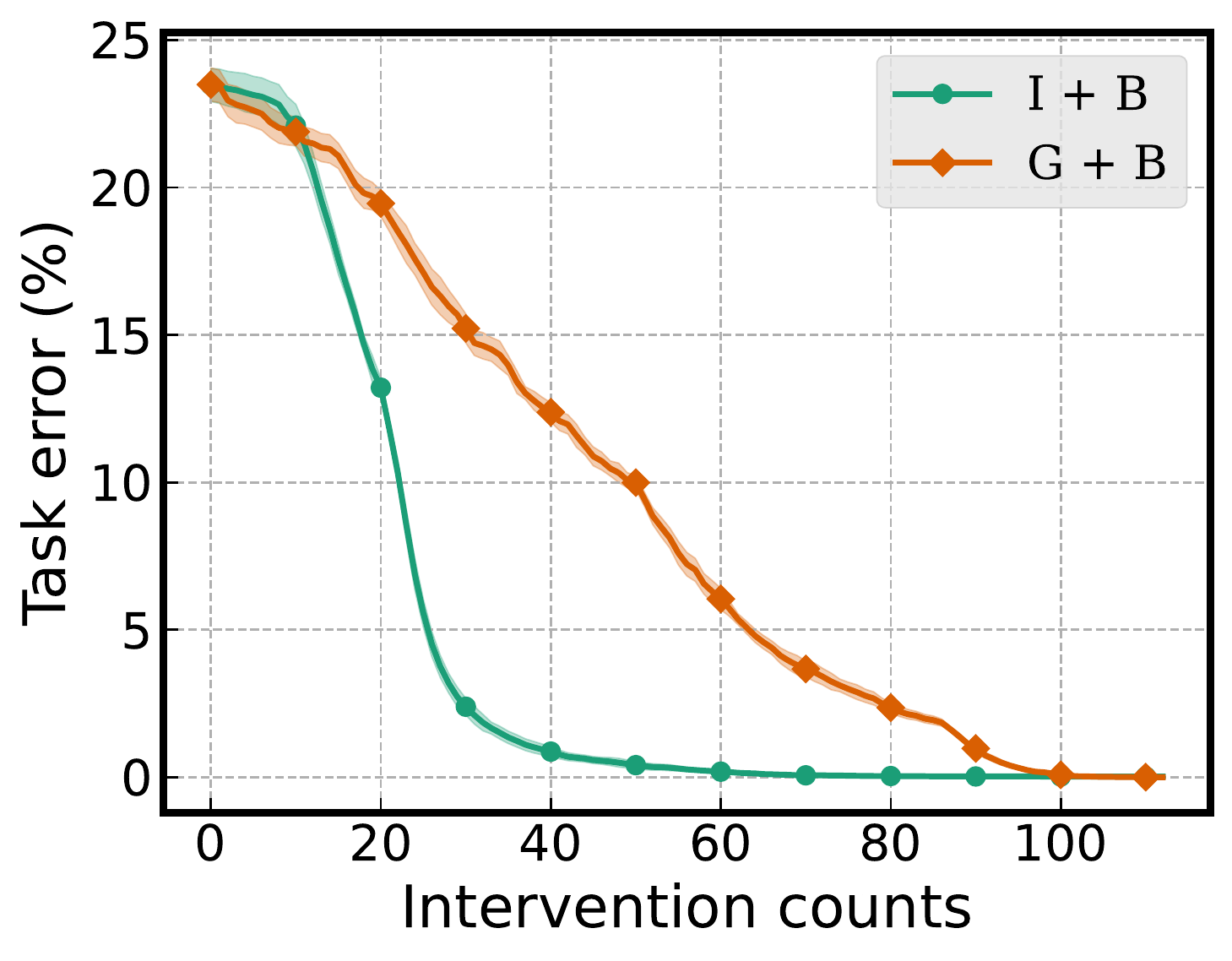}
    \caption{\textsc{cctp}}
    \label{fig:cub_level_ib_gb_cctp}
  \end{subfigure}
  \begin{subfigure}{0.16\linewidth}
    \centering
    \includegraphics[width=\linewidth]{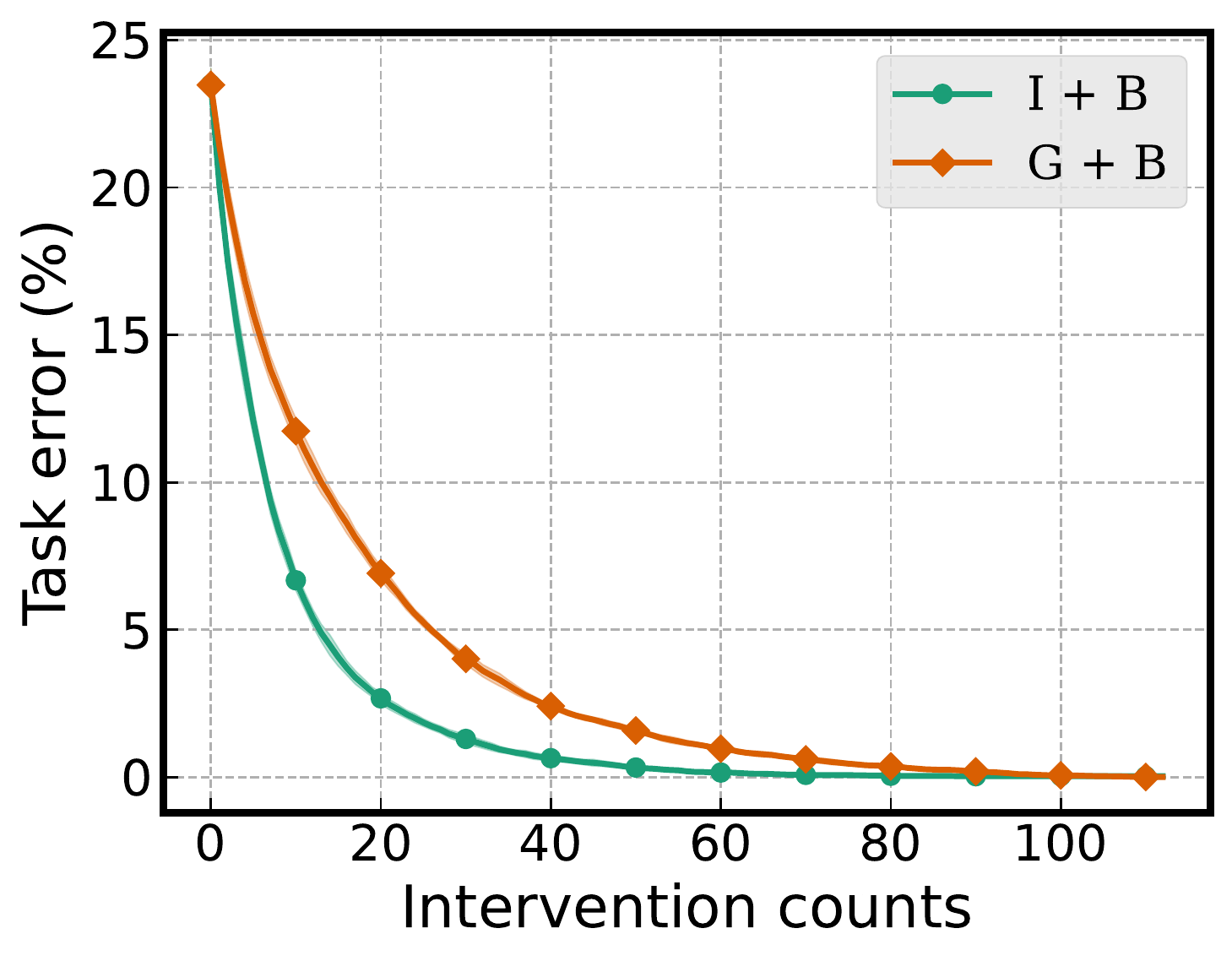}
    \caption{\textsc{ectp}}
    \label{fig:cub_level_ib_gb_ectp}
  \end{subfigure}
  \begin{subfigure}{0.16\linewidth}
    \centering
    \includegraphics[width=\linewidth]{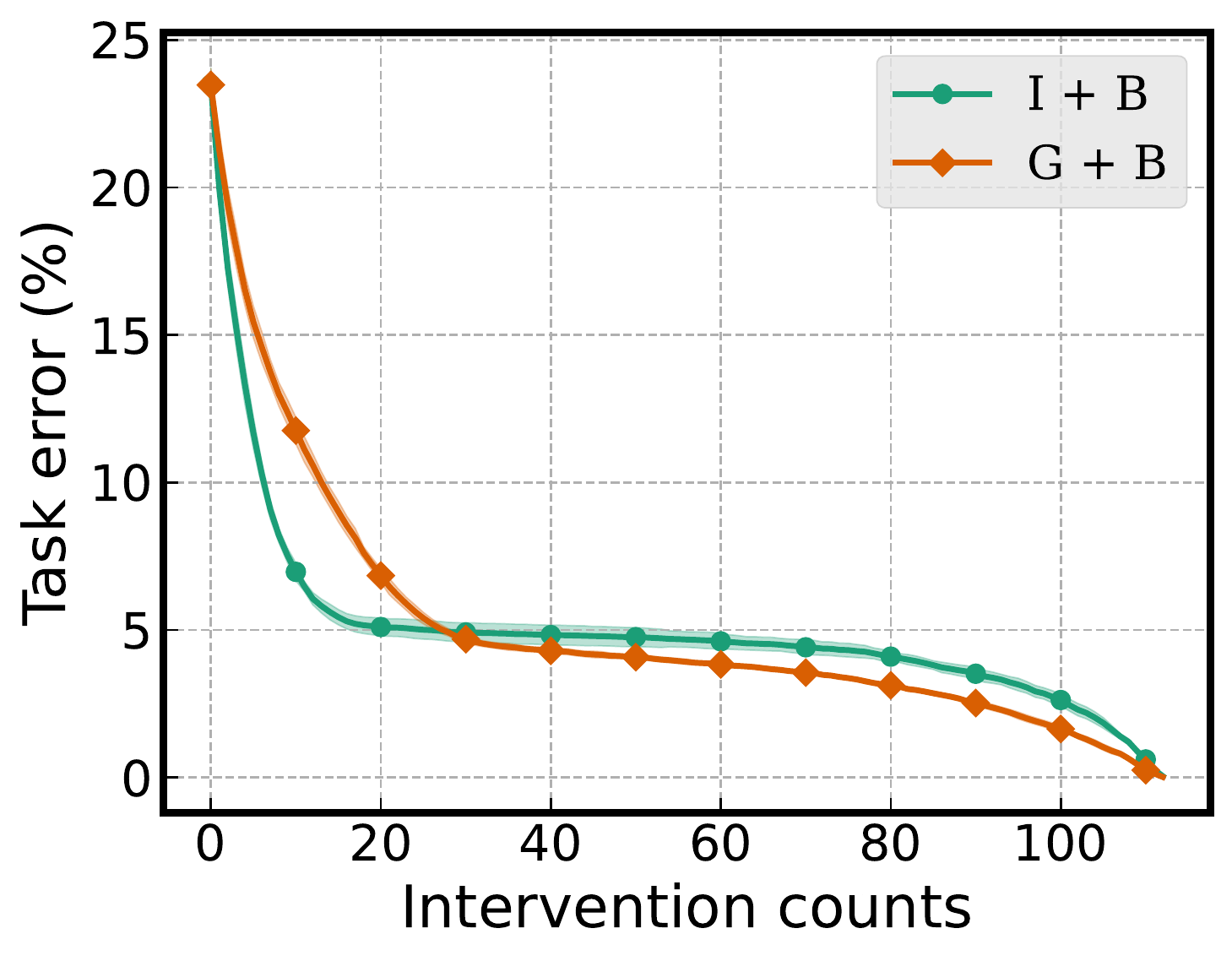}
    \caption{\textsc{eudtp}}
    \label{fig:cub_level_ib_gb_eudtp}
  \end{subfigure}
  \caption{
    Comparison between \textsc{i+b} vs. \textsc{g+b} for the CUB.
  }
  \label{fig:cub_level_ib_gb_all}
\end{figure}

\begin{figure}[!th]
  \begin{subfigure}{0.16\linewidth}
    \centering
    \includegraphics[width=\linewidth]{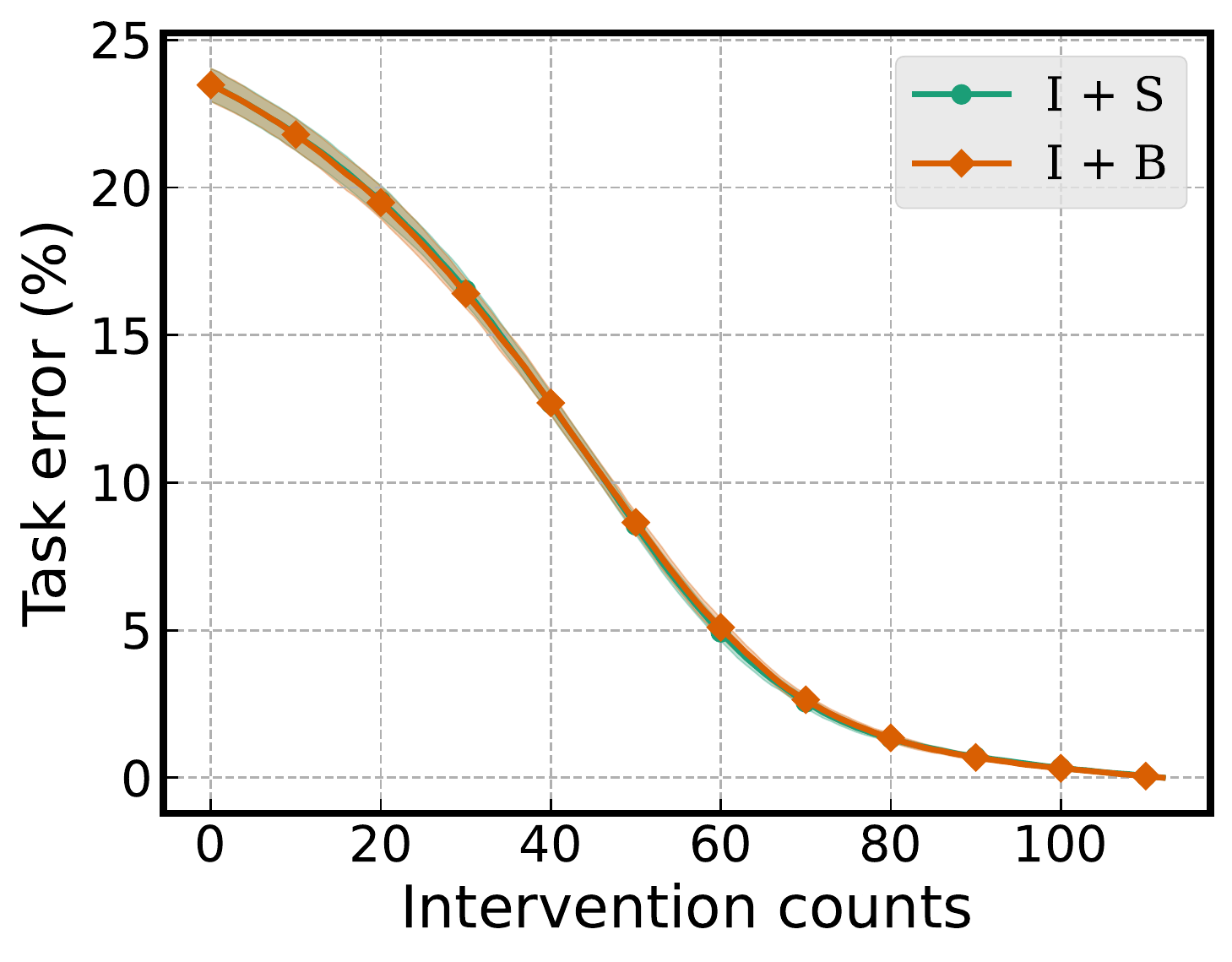}
    \caption{\textsc{rand}}
    \label{fig:cub_level_is_ib_rand}
  \end{subfigure}
  \begin{subfigure}{0.16\linewidth}
    \centering
    \includegraphics[width=\linewidth]{figures/cub/level_is_ib_ucp.pdf}
    \caption{\textsc{ucp}}
    \label{fig:cub_level_is_ib_ucp}
  \end{subfigure}
  \begin{subfigure}{0.16\linewidth}
    \centering
    \includegraphics[width=\linewidth]{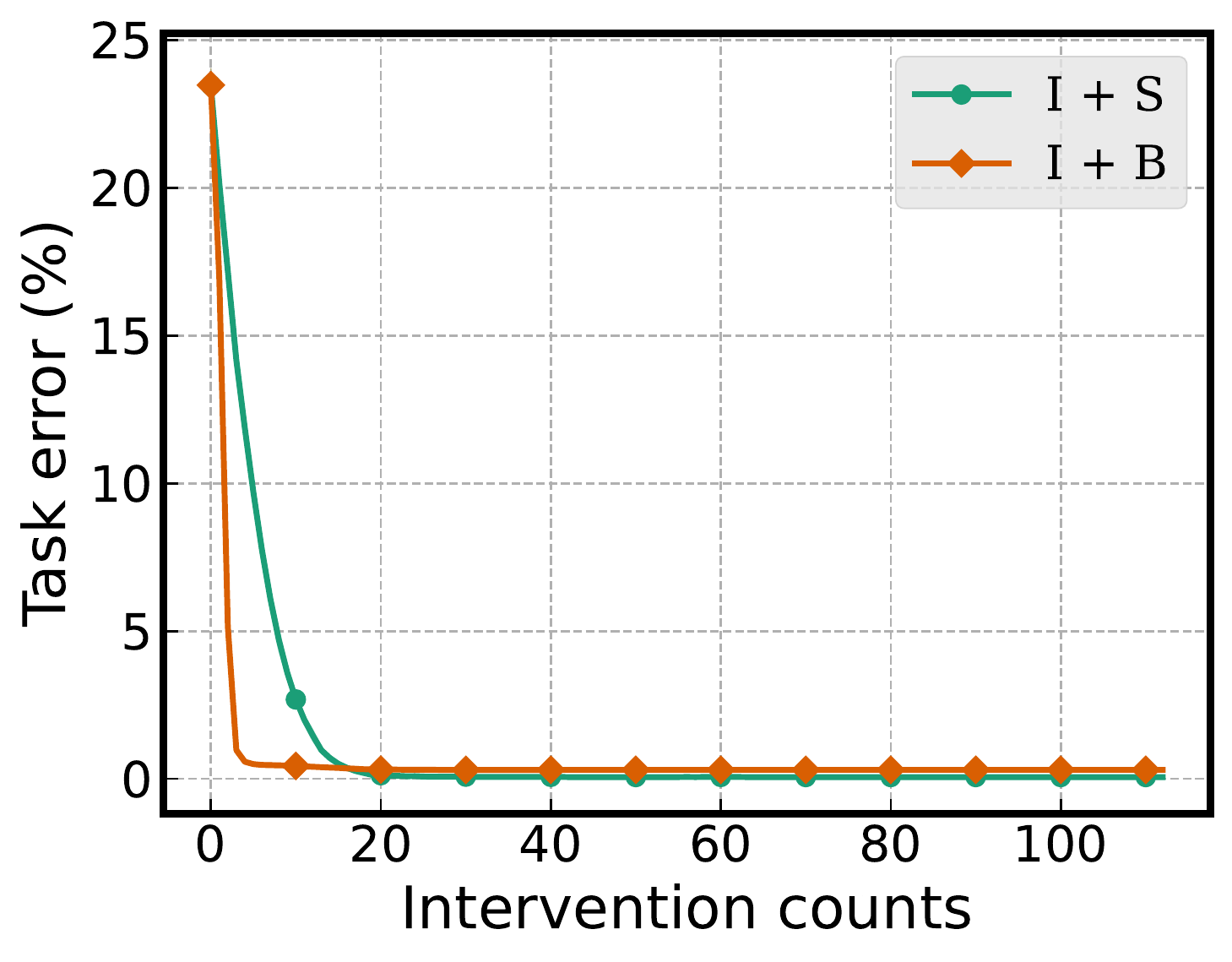}
    \caption{\textsc{lcp}}
    \label{fig:cub_level_is_ib_lcp}
  \end{subfigure}
  \begin{subfigure}{0.16\linewidth}
    \centering
    \includegraphics[width=\linewidth]{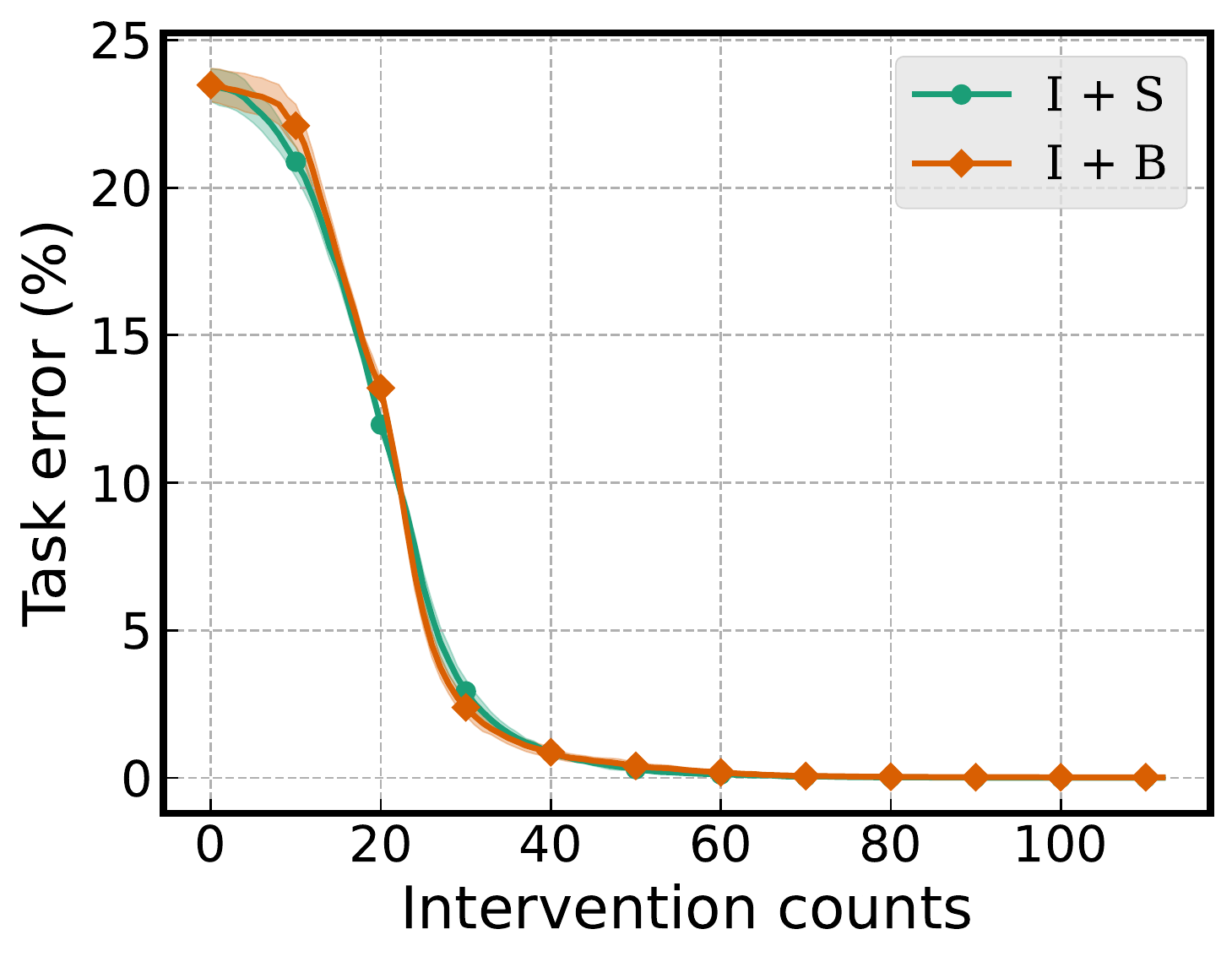}
    \caption{\textsc{cctp}}
    \label{fig:cub_level_is_ib_cctp}
  \end{subfigure}
  \begin{subfigure}{0.16\linewidth}
    \centering
    \includegraphics[width=\linewidth]{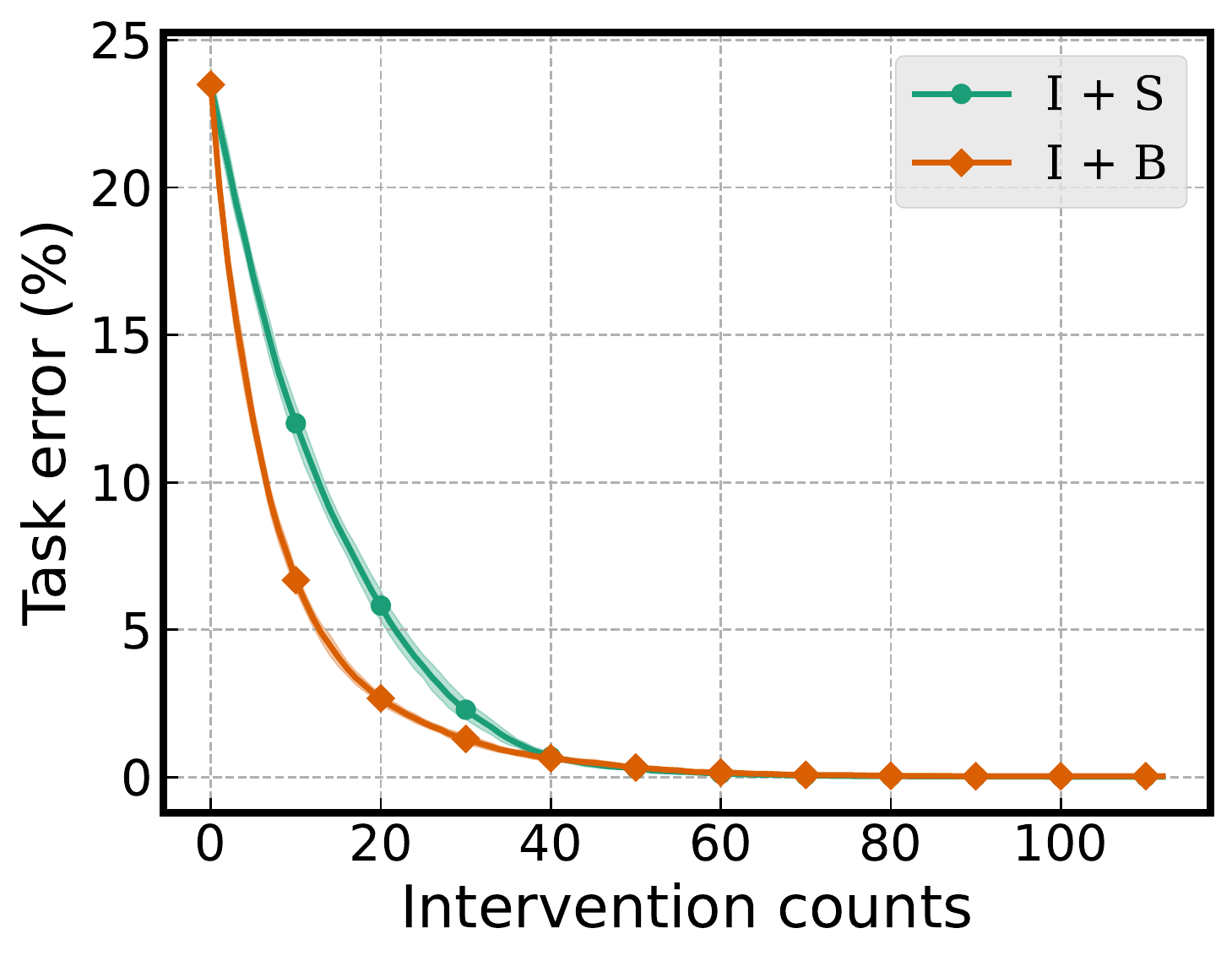}
    \caption{\textsc{ectp}}
    \label{fig:cub_level_is_ib_ectp}
  \end{subfigure}
  \begin{subfigure}{0.16\linewidth}
    \centering
    \includegraphics[width=\linewidth]{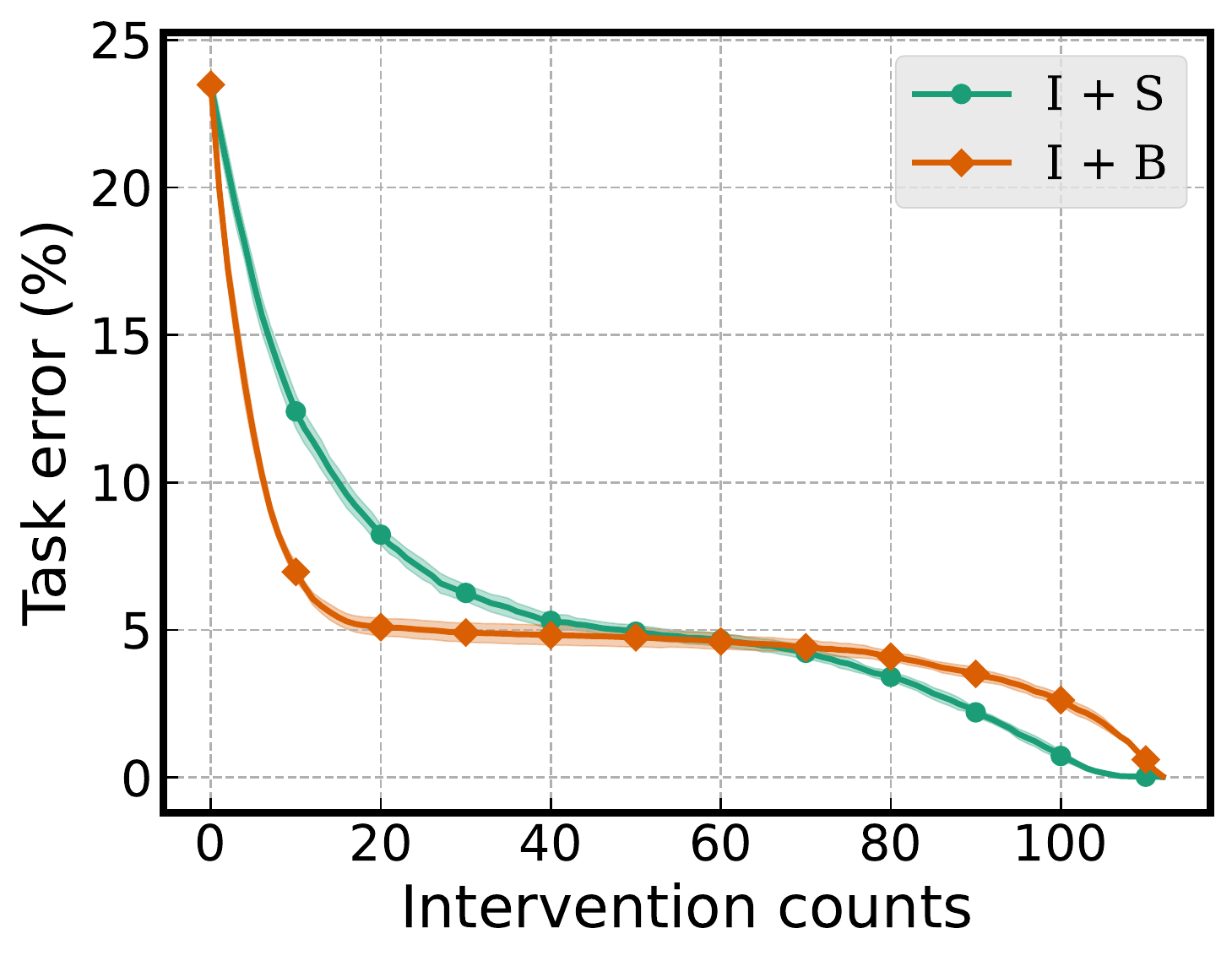}
    \caption{\textsc{eudtp}}
    \label{fig:cub_level_is_ib_eudtp}
  \end{subfigure}
  \caption{
    Comparison between \textsc{i+s} vs. \textsc{i+b} for the CUB.
  }
  \label{fig:cub_level_is_ib_all}
\end{figure}

\begin{figure}[!th]
  \begin{subfigure}{0.16\linewidth}
    \centering
    \includegraphics[width=\linewidth]{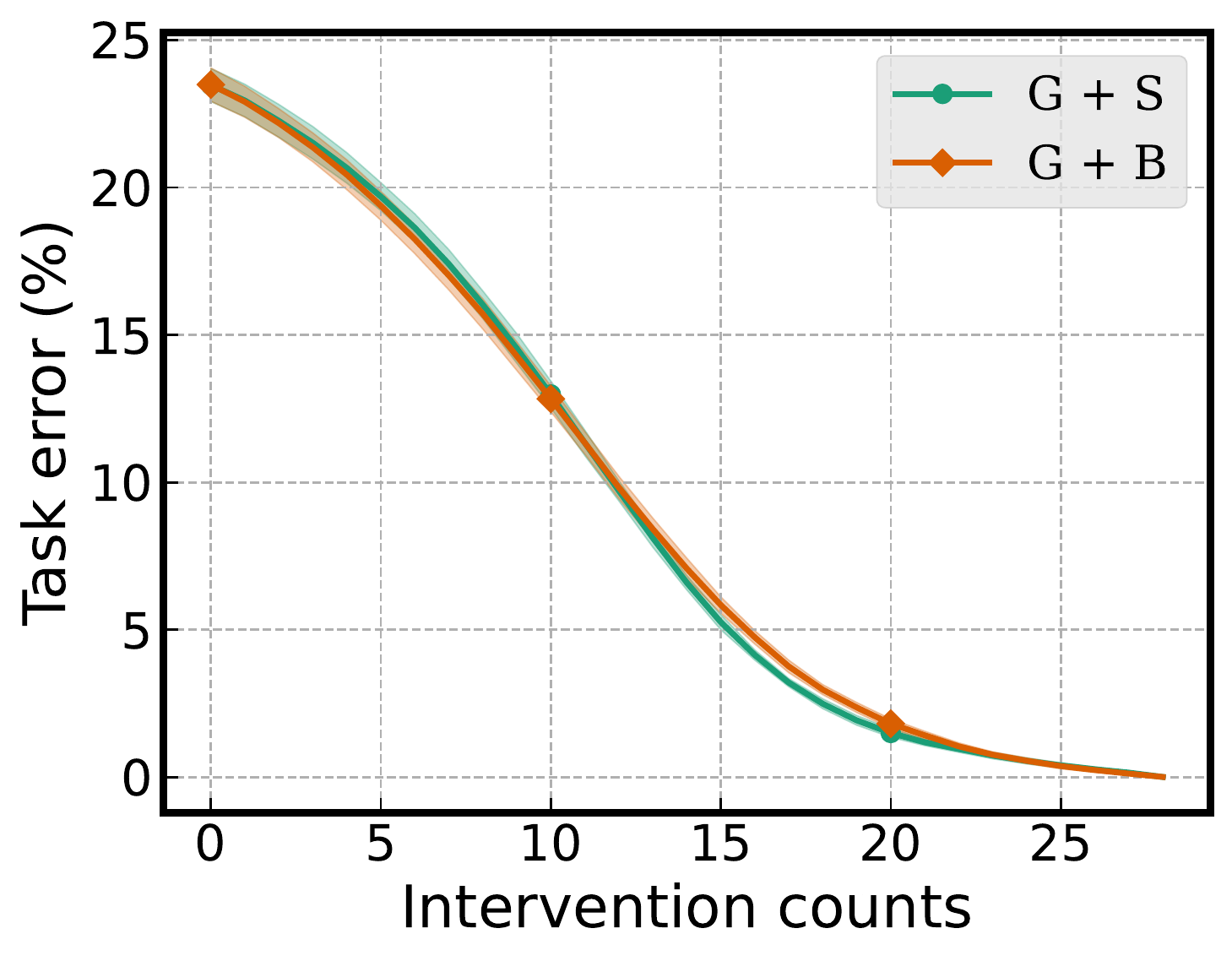}
    \caption{\textsc{rand}}
    \label{fig:cub_level_gs_gb_rand}
  \end{subfigure}
  \begin{subfigure}{0.16\linewidth}
    \centering
    \includegraphics[width=\linewidth]{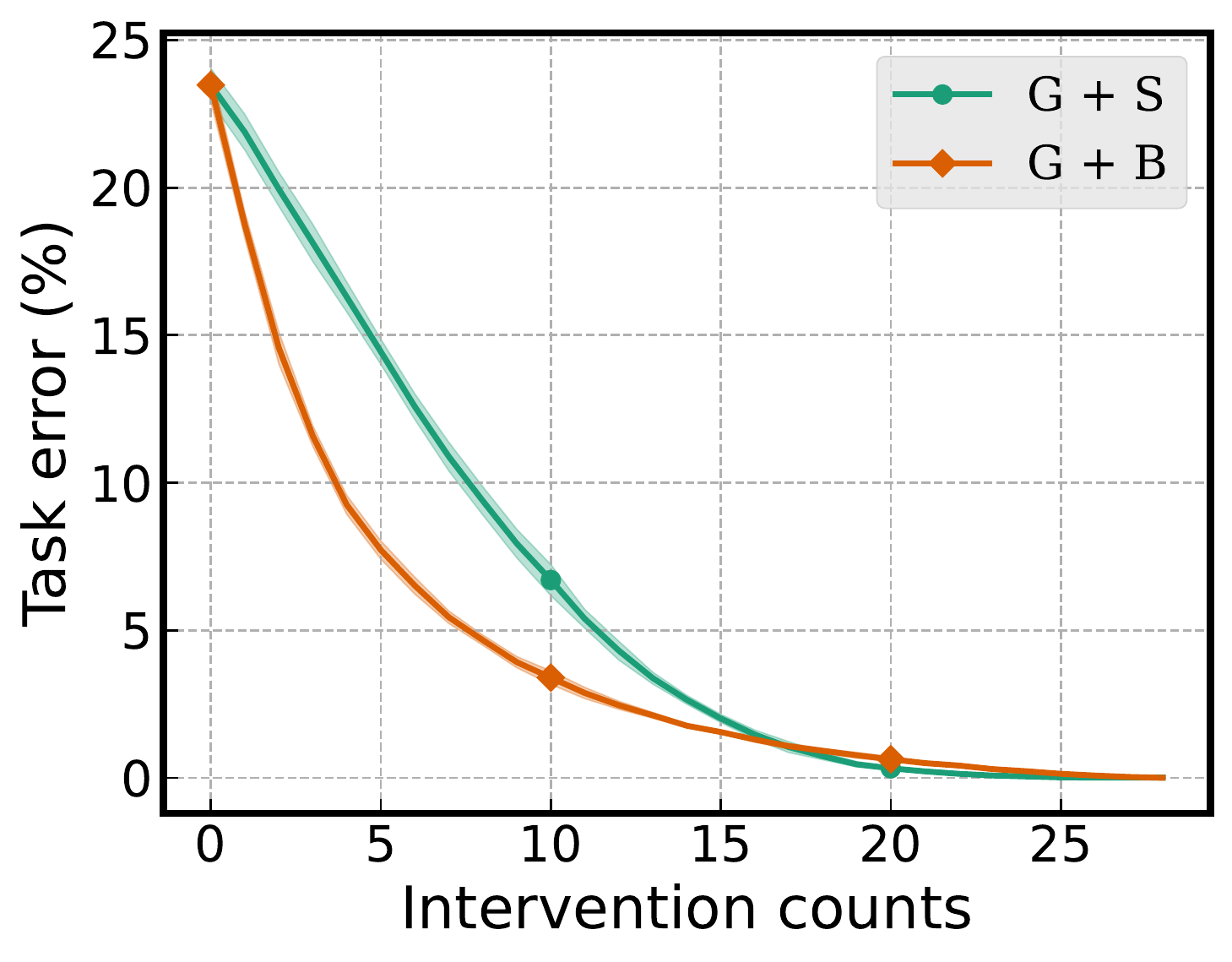}
    \caption{\textsc{ucp}}
    \label{fig:cub_level_gs_gb_ucp}
  \end{subfigure}
  \begin{subfigure}{0.16\linewidth}
    \centering
    \includegraphics[width=\linewidth]{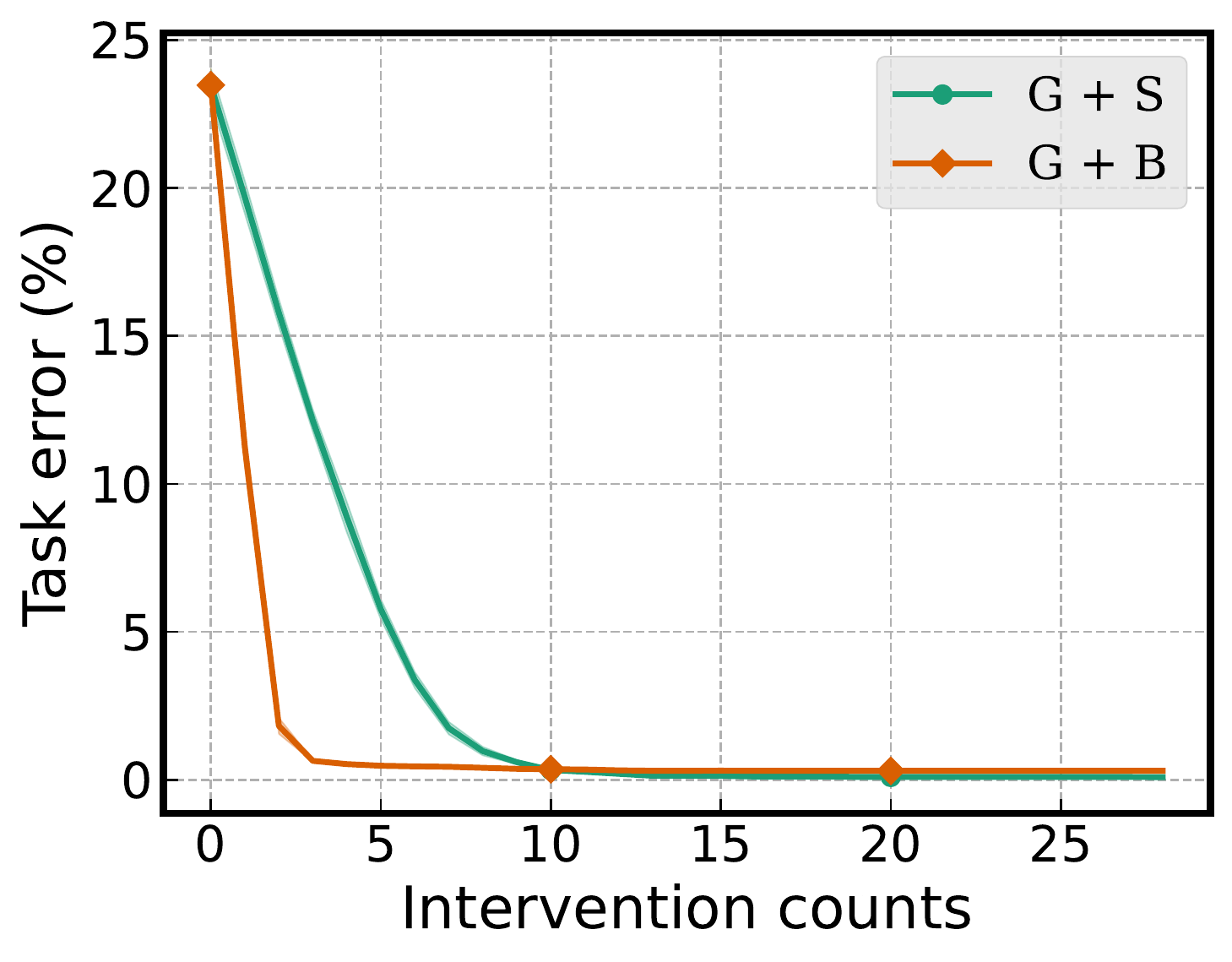}
    \caption{\textsc{lcp}}
    \label{fig:cub_level_gs_gb_lcp}
  \end{subfigure}
  \begin{subfigure}{0.16\linewidth}
    \centering
    \includegraphics[width=\linewidth]{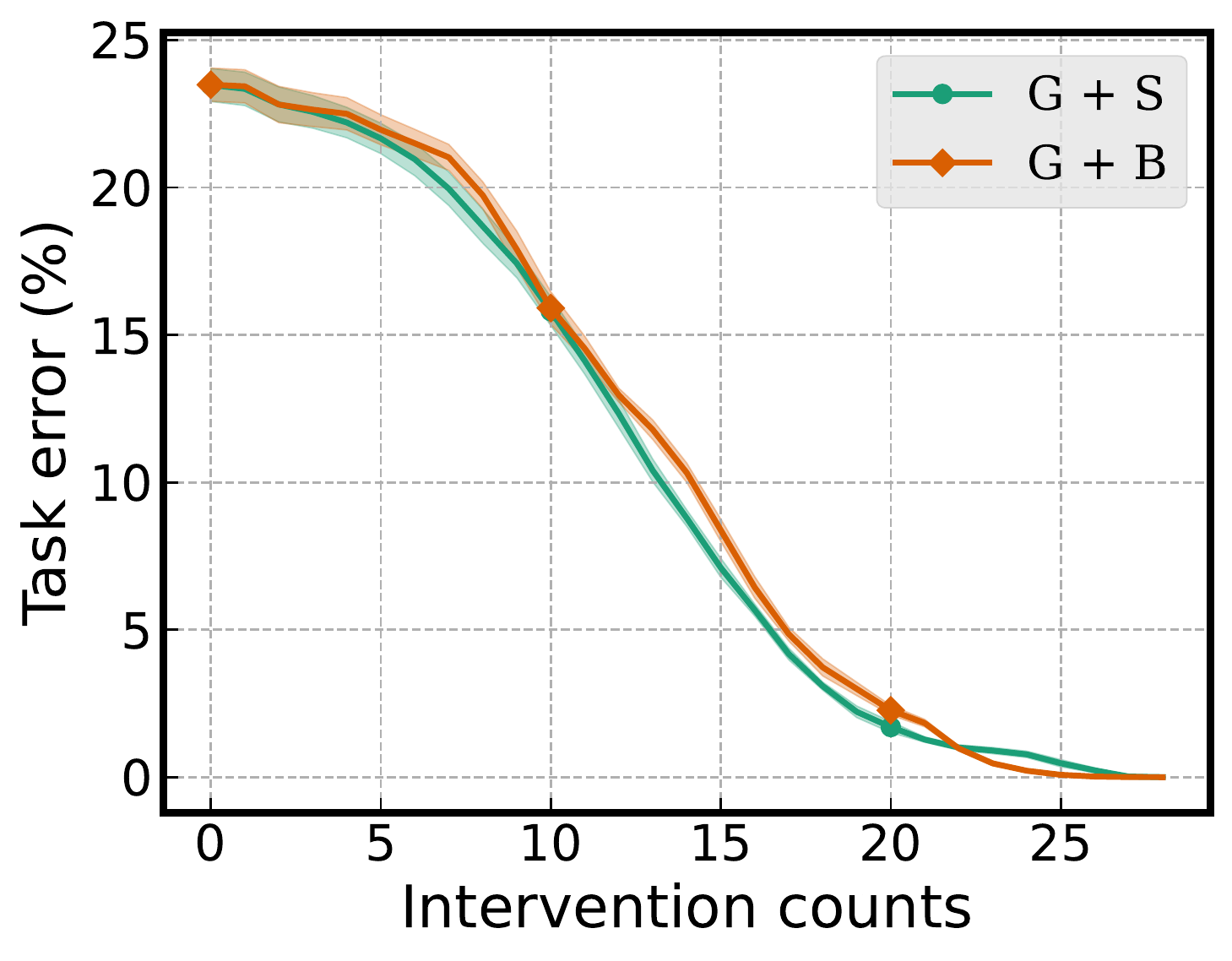}
    \caption{\textsc{cctp}}
    \label{fig:cub_level_gs_gb_cctp}
  \end{subfigure}
  \begin{subfigure}{0.16\linewidth}
    \centering
    \includegraphics[width=\linewidth]{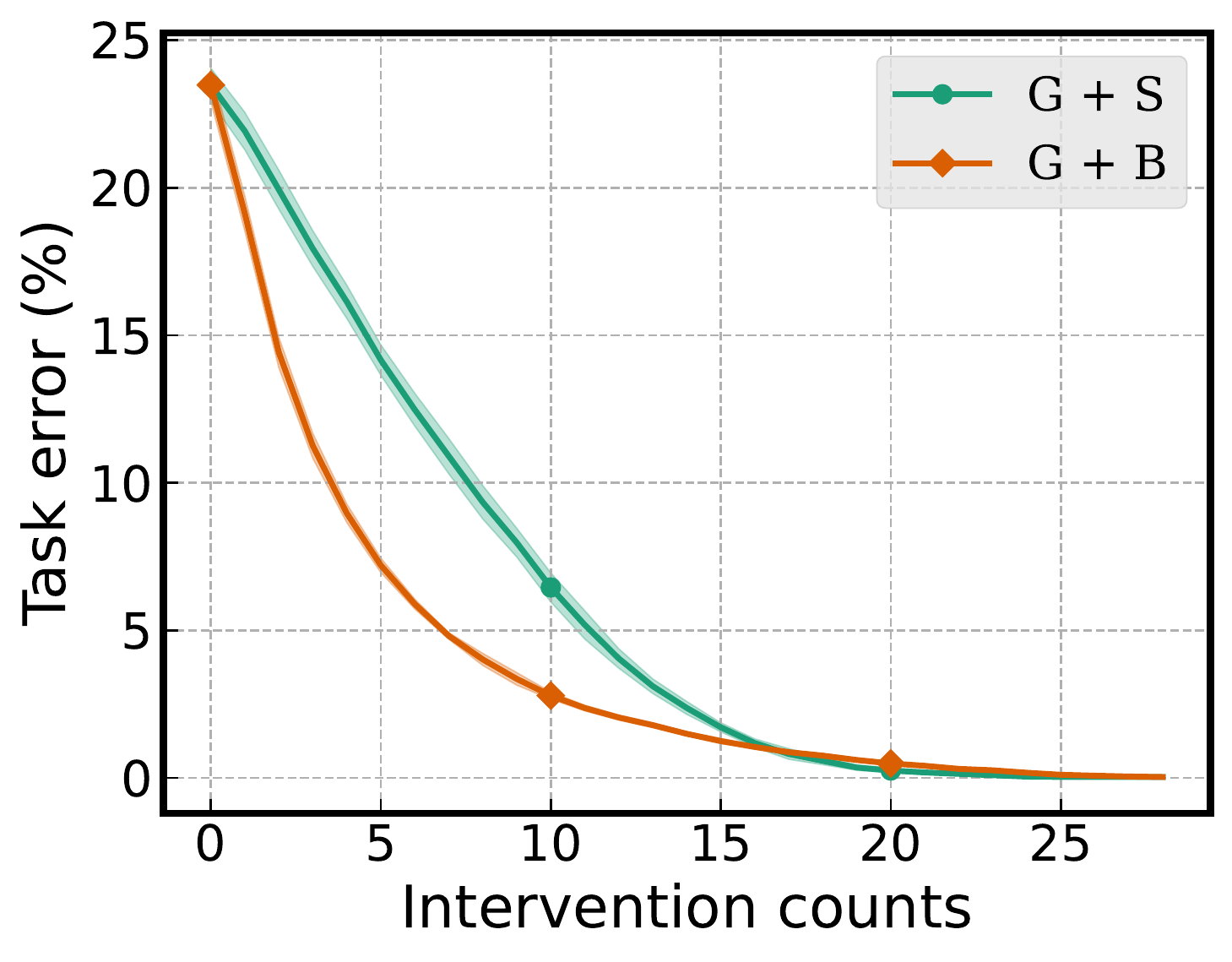}
    \caption{\textsc{ectp}}
    \label{fig:cub_level_gs_gb_ectp}
  \end{subfigure}
  \begin{subfigure}{0.16\linewidth}
    \centering
    \includegraphics[width=\linewidth]{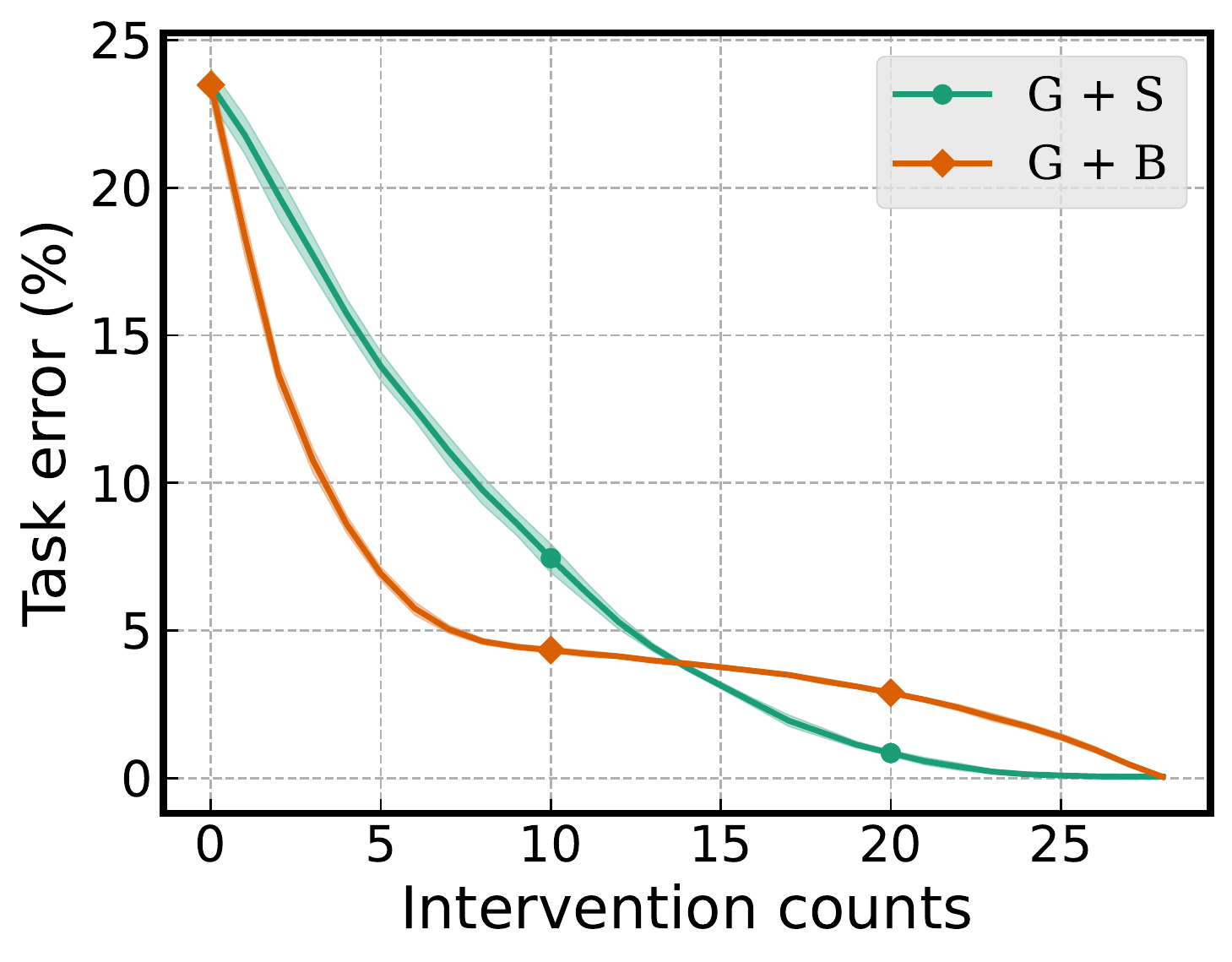}
    \caption{\textsc{eudtp}}
    \label{fig:cub_level_gs_gb_eudtp}
  \end{subfigure}
  \caption{
    Comparison between \textsc{g+s} vs. \textsc{g+b} for the CUB.
    For \textsc{g+b}, each point is plotted when the average number of intervened concepts per image first exceeds each integer value.
  }
  \label{fig:cub_level_gs_gb_all}
\end{figure}

The comparison between \textsc{i+s} and \textsc{g+s} using different concept selection criteria is presented in \cref{fig:cub_level_is_gs_all}.
Individual intervention is in general more effective than group-wise intervention except for \textsc{rand} criterion.
We find similar results for the comparison between \textsc{i+b} and \textsc{g+b} (see \cref{fig:cub_level_ib_gb_all}).
We also note that \textsc{cctp} becomes less effective in \textsc{g} levels as seen in \cref{fig:cub_result_level}.

Batch intervention is either more effective or at least as competitive as single intervention across different concept selection criteria as seen in \cref{fig:cub_level_is_ib_all}.
In \cref{fig:cub_level_gs_gb_all}, we observe that \textsc{g+b} are also more effective than \textsc{g+s} level.
\textsc{cctp} does not show much difference between \textsc{s} and \textsc{b}.
It is because the target predictor $f$ is a simple linear layer for our experiments and thus $\frac{\partial f_j}{\partial \hat{c}_i} = w_{ij} $ is fixed for all examples where $w_{ij}$ is the weight of $i$-th concept to $j$-th class in $f$.

\section{More Results on the Effect of Training Strategies on Intervention}

\label{sec:results-others-training}

\begin{figure*}[!th]
\centering
  \begin{subfigure}{0.24\linewidth}
    \centering
    \includegraphics[width=\linewidth]{figures/cub/result_main.pdf}
    \caption{\textsc{ind}}
    \label{fig:cub_result_training_ind}
  \end{subfigure}%
  \begin{subfigure}{0.24\linewidth}
    \centering
    \includegraphics[width=\linewidth]{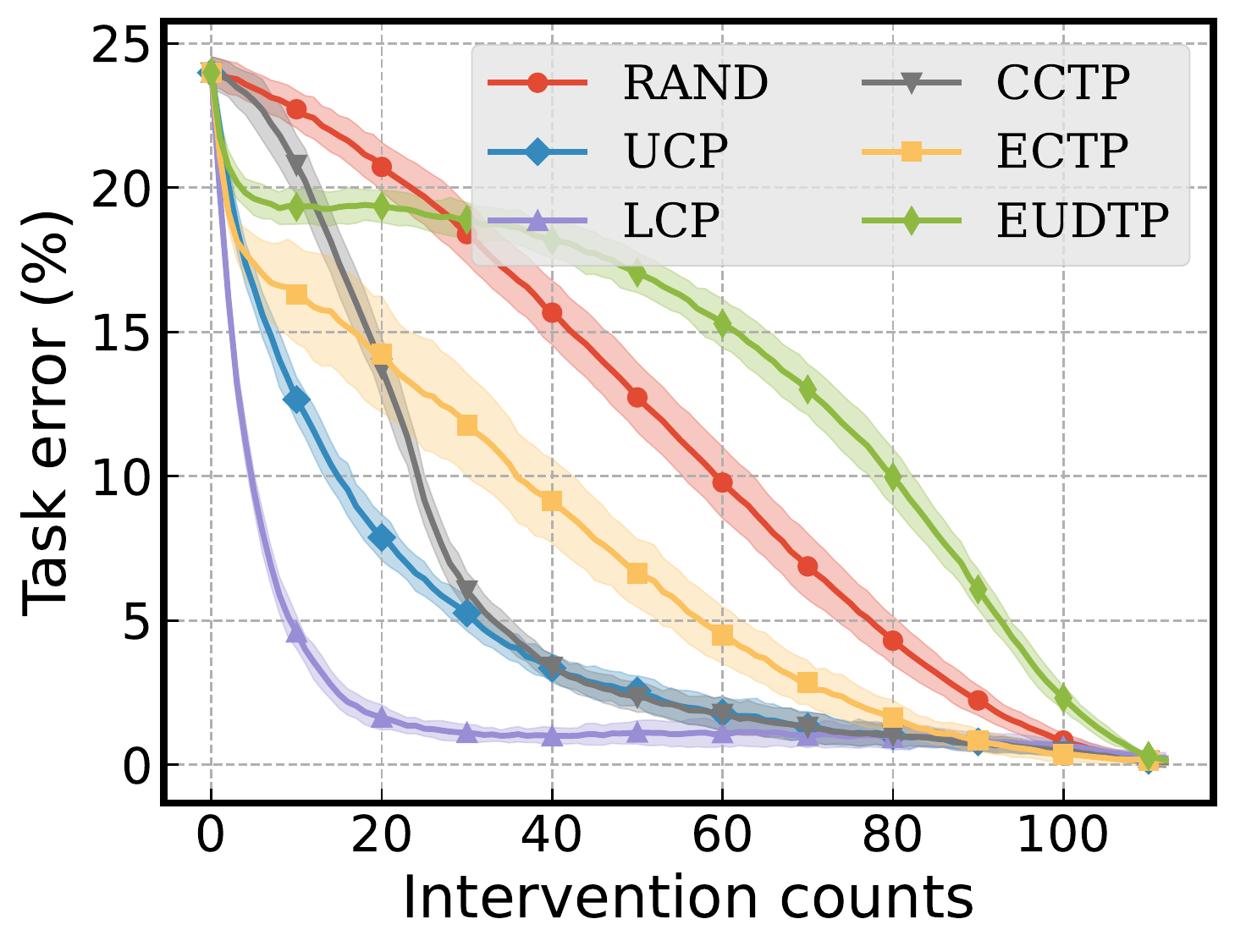}
    \caption{\textsc{seq}}
    \label{fig:cub_result_training_seq}
  \end{subfigure}%
  \begin{subfigure}{0.24\linewidth}
    \centering
    \includegraphics[width=\linewidth]{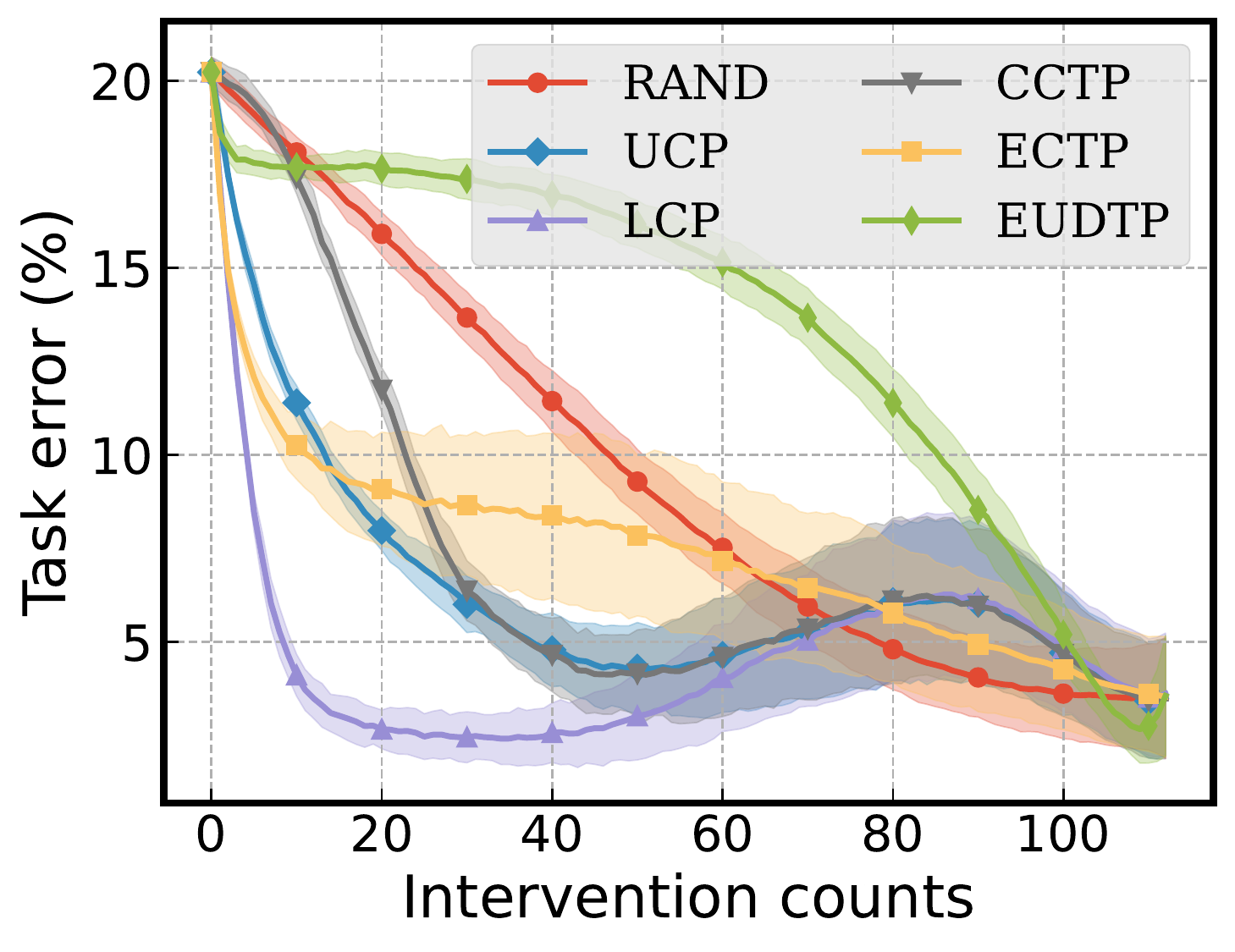}
    \caption{\textsc{jnt}}
    \label{fig:cub_result_training_jnt}
  \end{subfigure}%
  \begin{subfigure}{0.24\linewidth}
    \centering
    \includegraphics[width=\linewidth]{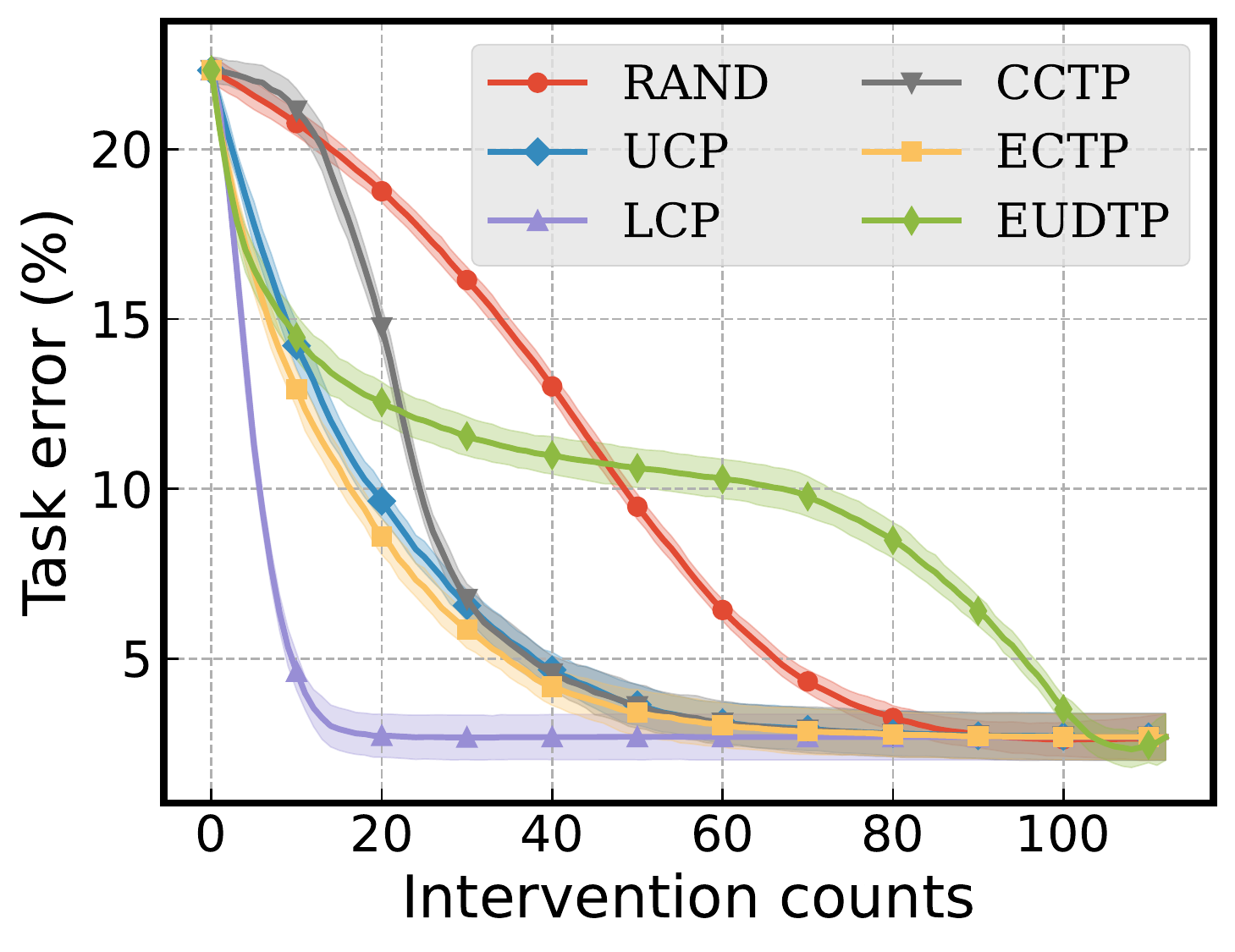}
    \caption{\textsc{jnt + p}}
    \label{fig:cub_result_training_jntp}
  \end{subfigure}%
  \caption{
  Comparison between concept selection criteria using different training strategies for the CUB.
  For \textsc{jnt, jnt + p}, we present the results when $\lambda=0.01$.
  }
  \label{fig:cub_result_training}
\end{figure*}

\begin{figure*}[!th]
  \hspace*{\fill}
  \begin{subfigure}{0.16\linewidth}
    \centering
    \includegraphics[width=\linewidth]{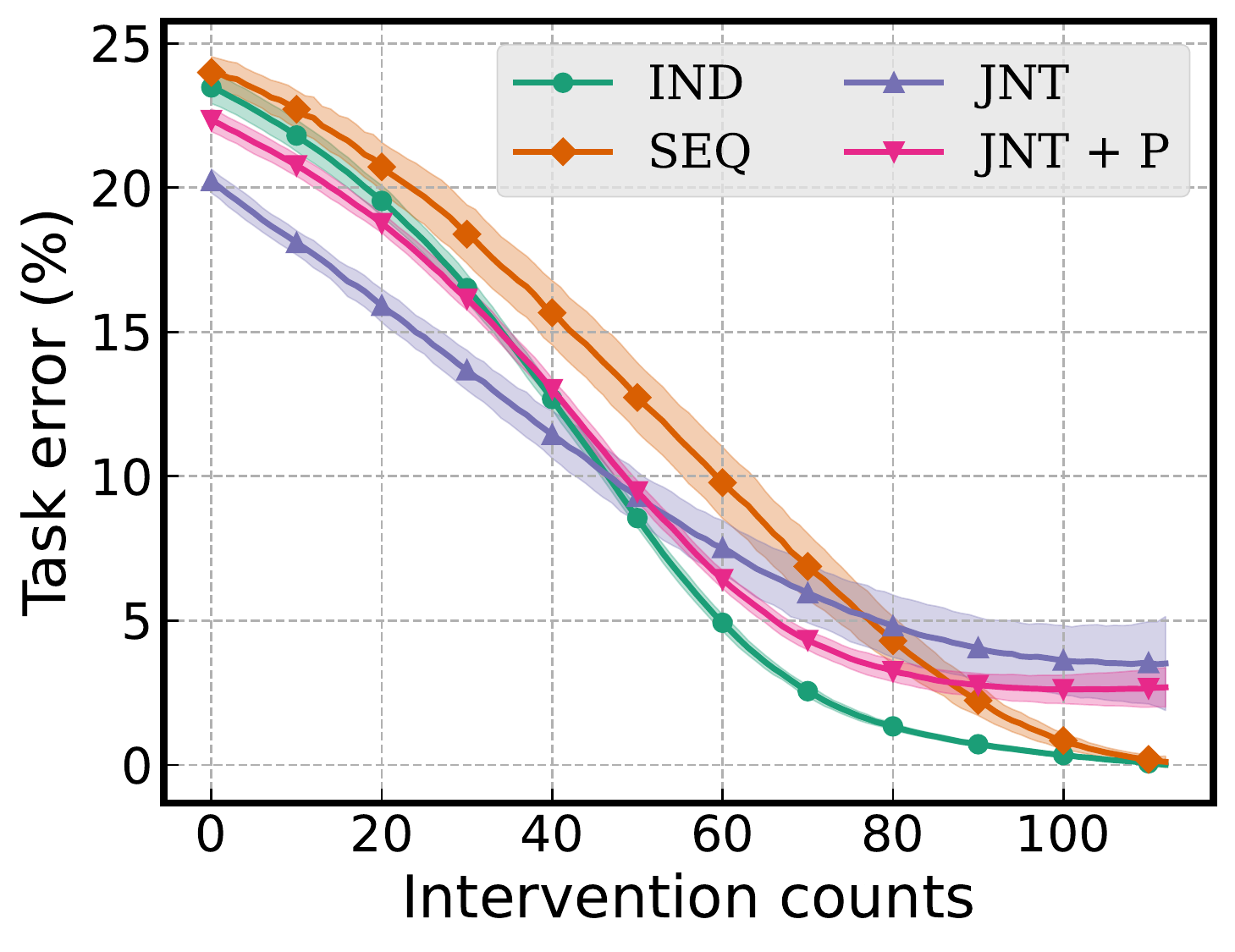}
    \caption{\textsc{rand}}
    \label{fig:cub_training_rand}
  \end{subfigure}%
  \begin{subfigure}{0.16\linewidth}
    \centering
    \includegraphics[width=\linewidth]{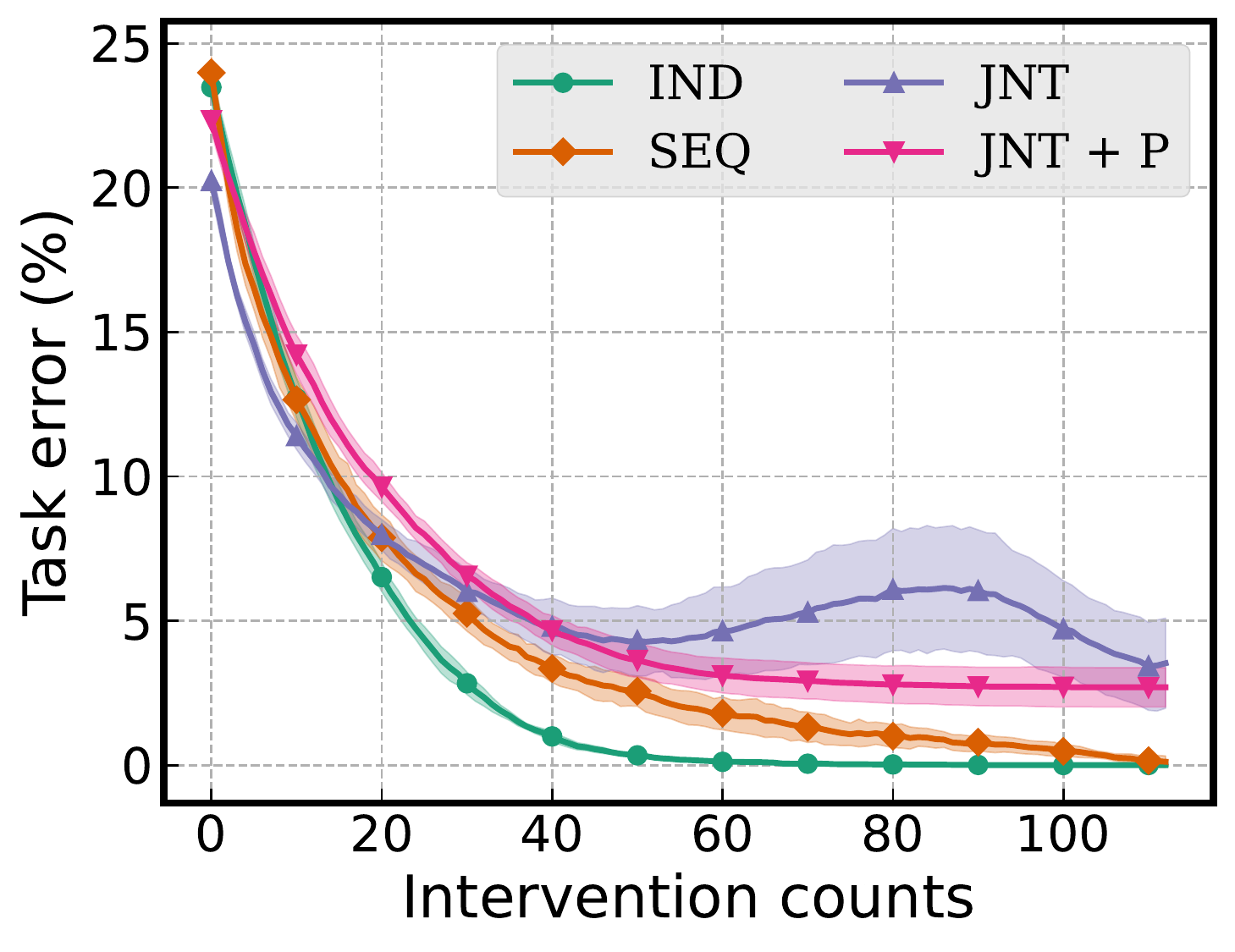}
    \caption{\textsc{ucp}}
    \label{fig:cub_training_ucp}
  \end{subfigure}%
  \hspace*{\fill}
  \begin{subfigure}{0.16\linewidth}
    \centering
    \includegraphics[width=\linewidth]{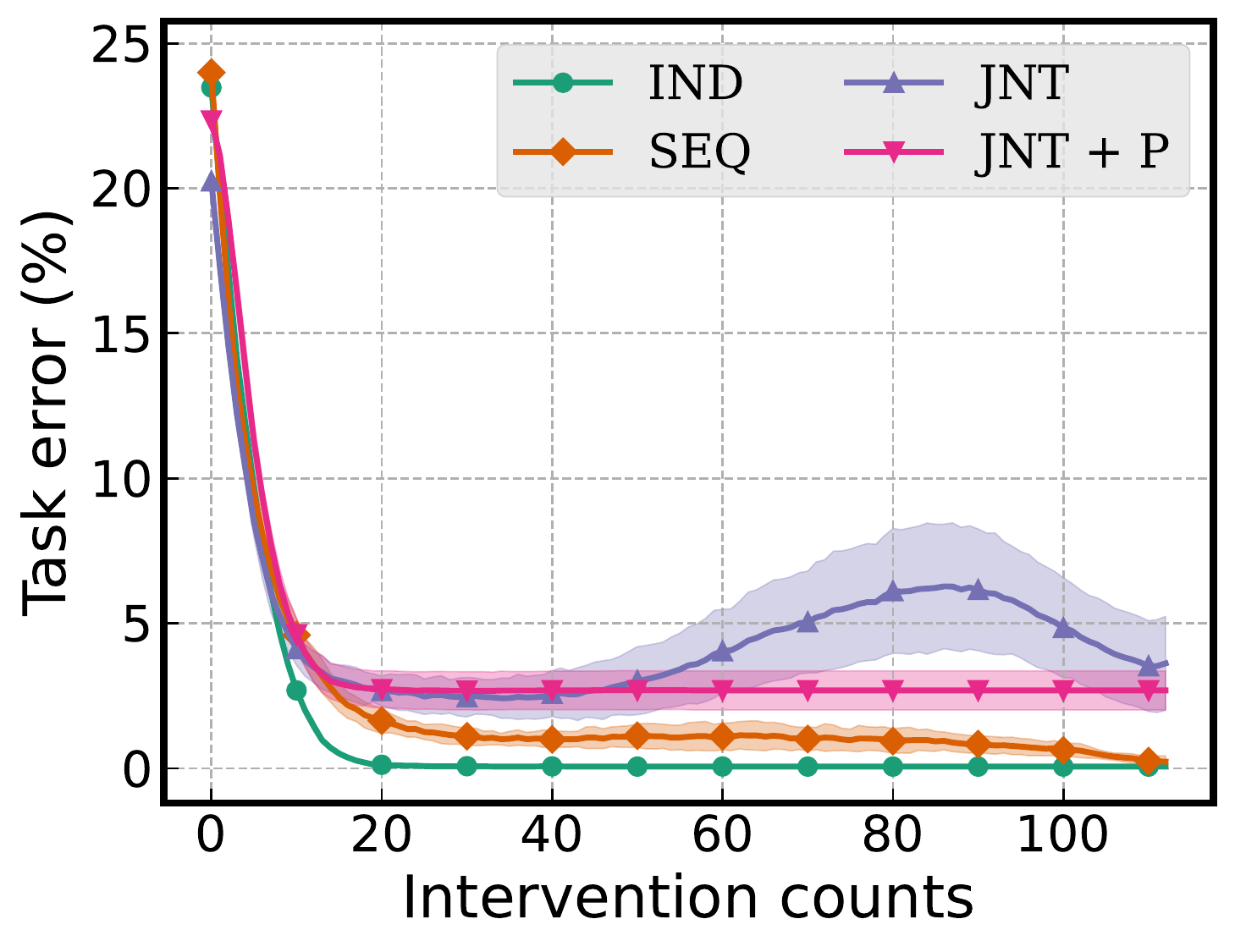}
    \caption{\textsc{lcp}}
    \label{fig:cub_training_lcp}
  \end{subfigure}%
  \hspace*{\fill}
  \begin{subfigure}{0.16\linewidth}
    \centering
    \includegraphics[width=\linewidth]{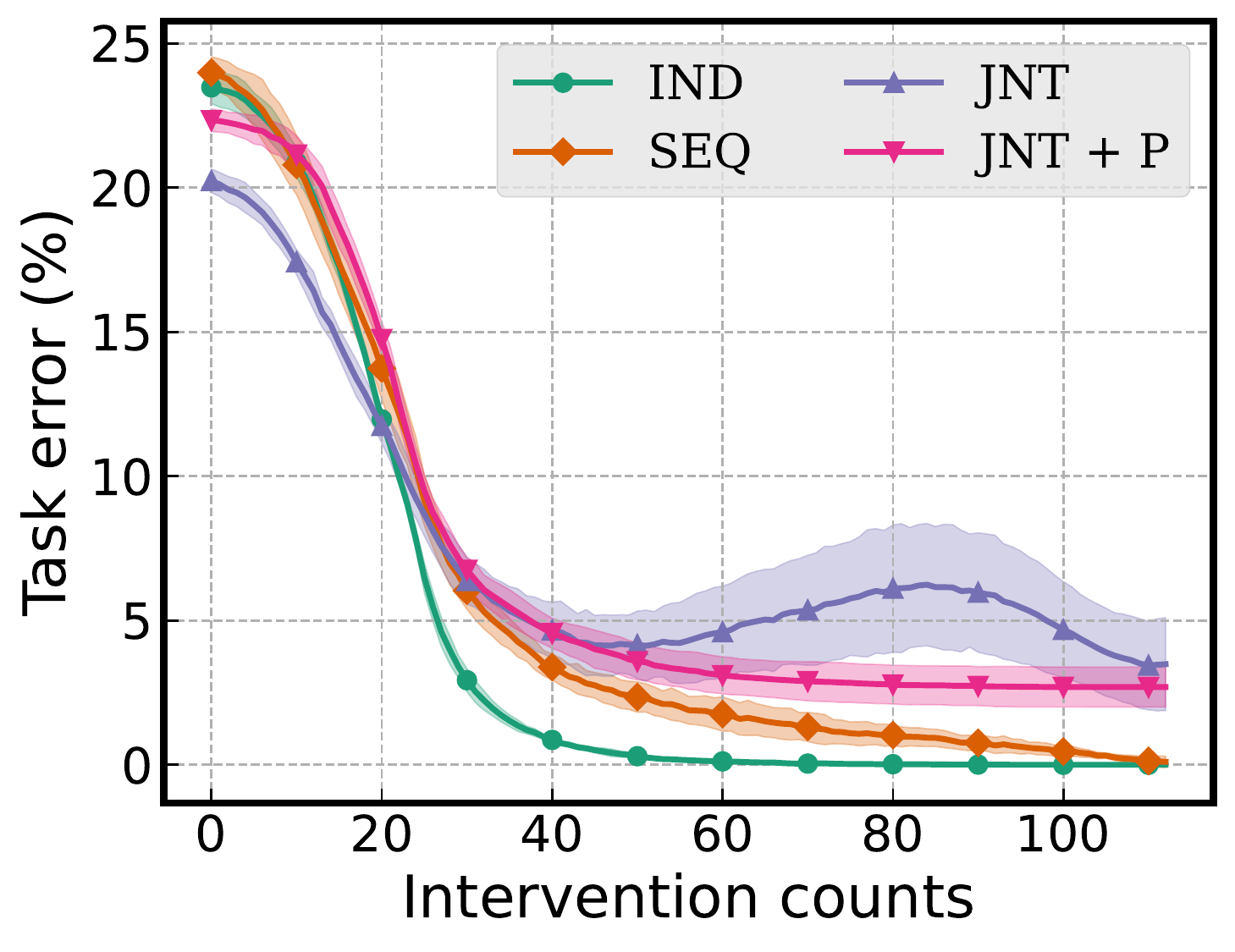}
    \caption{\textsc{cctp}}
    \label{fig:cub_training_cctp}
  \end{subfigure}%
  \hspace*{\fill}
  \begin{subfigure}{0.16\linewidth}
    \centering
    \includegraphics[width=\linewidth]{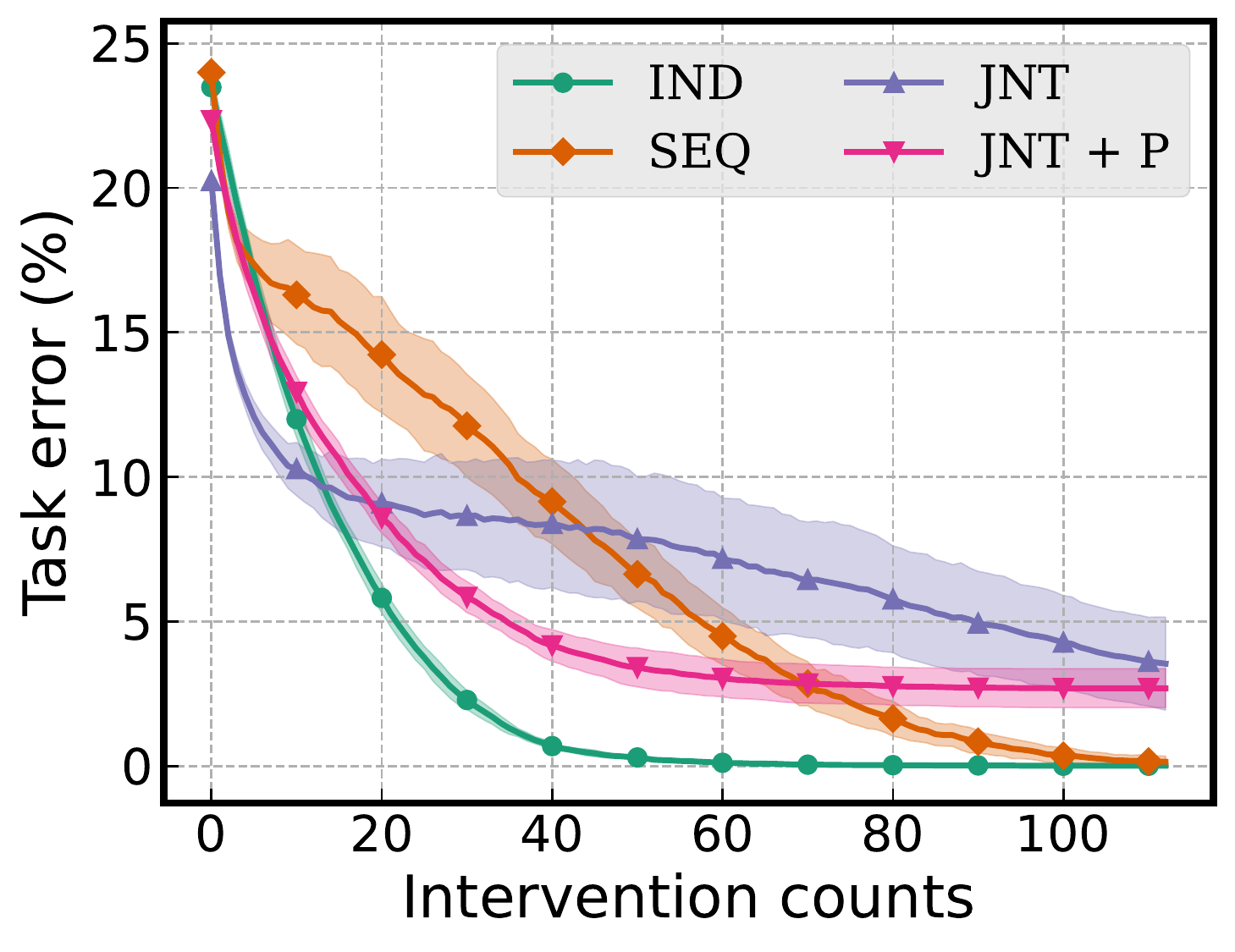}
    \caption{\textsc{ectp}}
    \label{fig:cub_training_ectp}
  \end{subfigure}%
  \begin{subfigure}{0.16\linewidth}
    \centering
    \includegraphics[width=\linewidth]{figures/cub/training_eudtp.pdf}
    \caption{\textsc{eudtp}}
    \label{fig:cub_training_eudtp}
  \end{subfigure}%
  \caption{
  Comparison between different training strategies for a fixed concept selection criterion for the CUB.
    }
  \label{fig:cub_training_comparison}
\end{figure*}

\begin{figure*}[!th]
\centering
  \begin{subfigure}{0.24\linewidth}
    \centering
    \includegraphics[width=\linewidth]{figures/synthetic/result_main.pdf}
    \caption{\textsc{ind}}
    \label{fig:synthetic_result_training_ind}
  \end{subfigure}%
  \begin{subfigure}{0.24\linewidth}
    \centering
    \includegraphics[width=\linewidth]{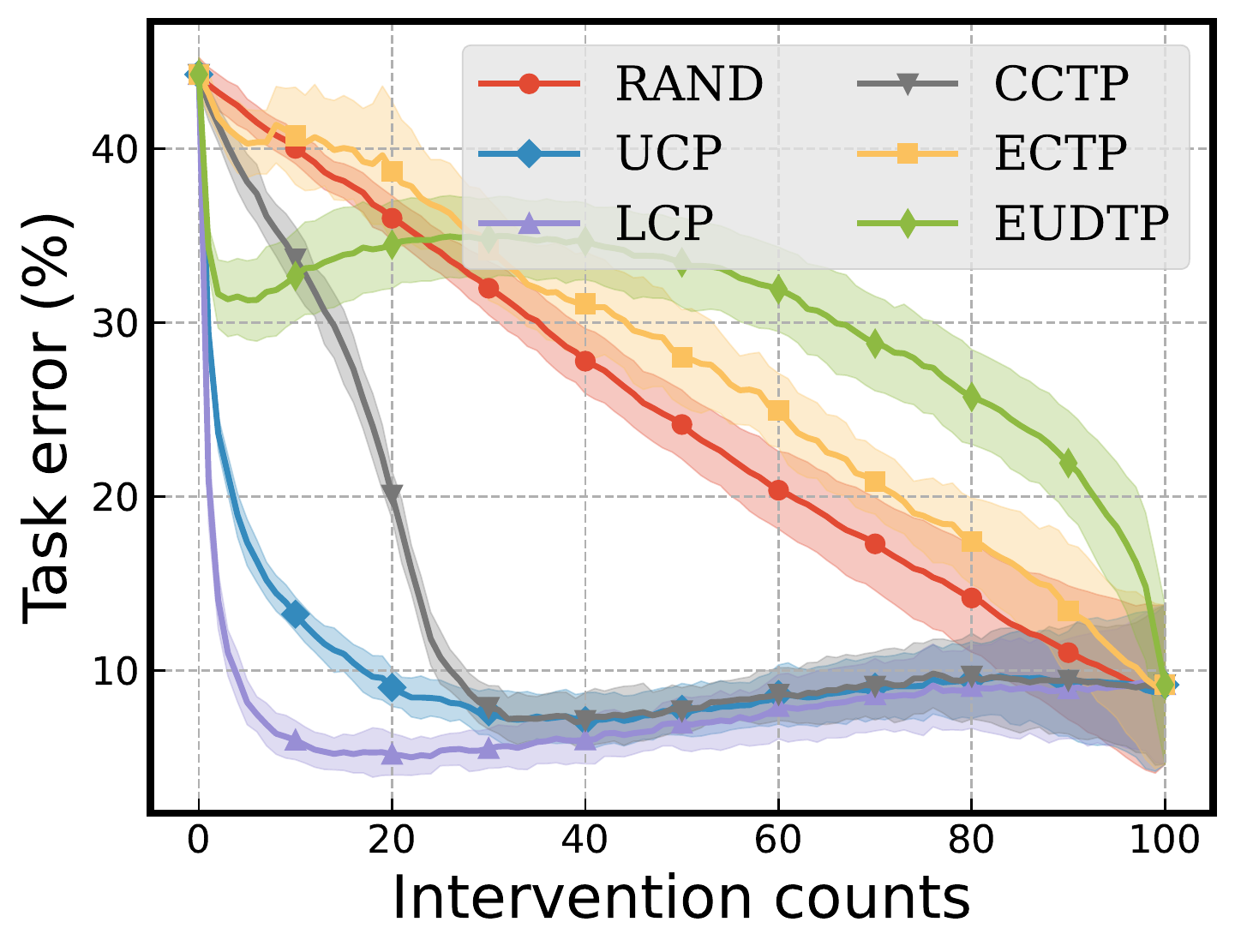}
    \caption{\textsc{seq}}
    \label{fig:synthetic_result_training_seq}
  \end{subfigure}%
  \begin{subfigure}{0.24\linewidth}
    \centering
    \includegraphics[width=\linewidth]{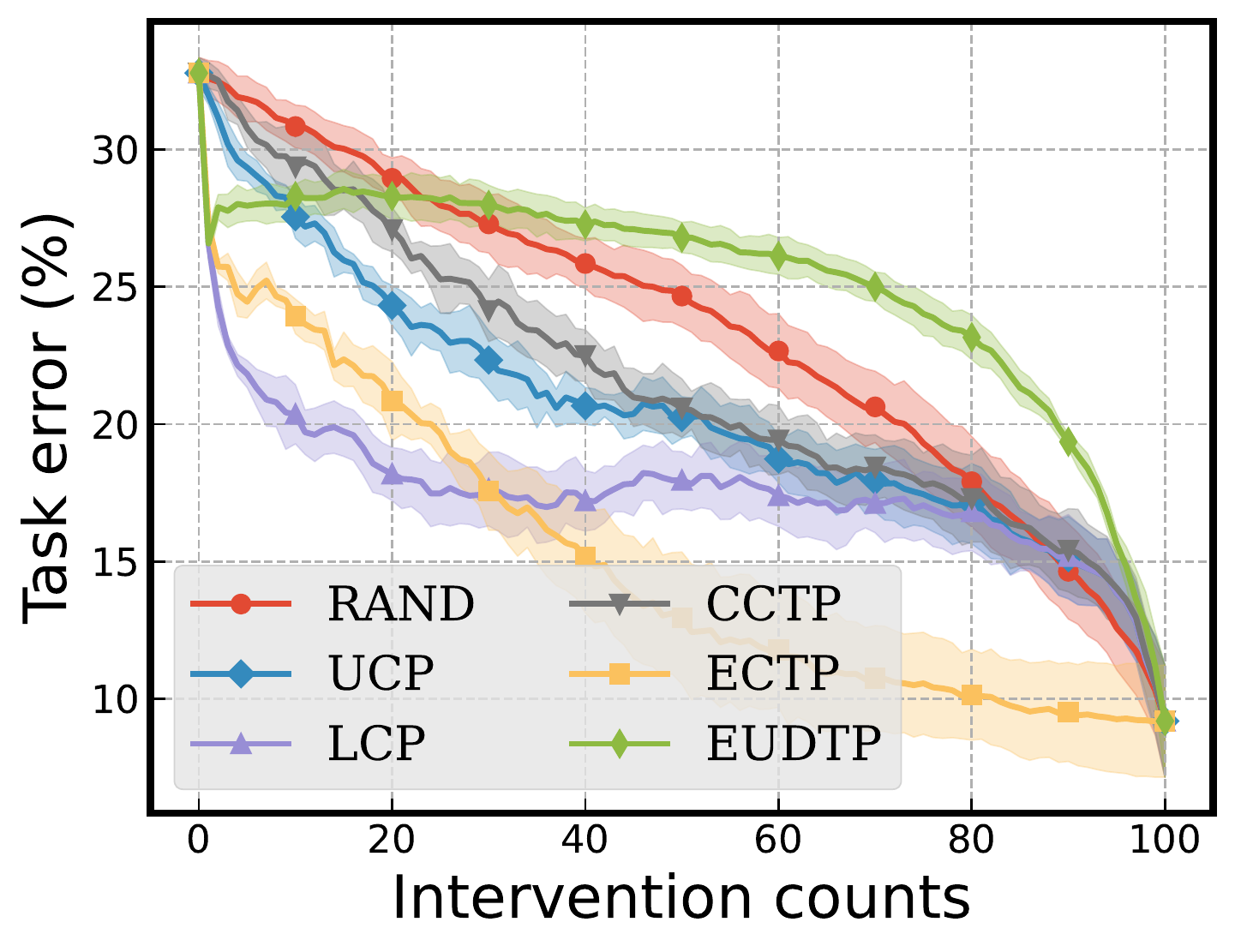}
    \caption{\textsc{jnt}}
    \label{fig:synthetic_result_training_jnt}
  \end{subfigure}%
  \begin{subfigure}{0.24\linewidth}
    \centering
    \includegraphics[width=\linewidth]{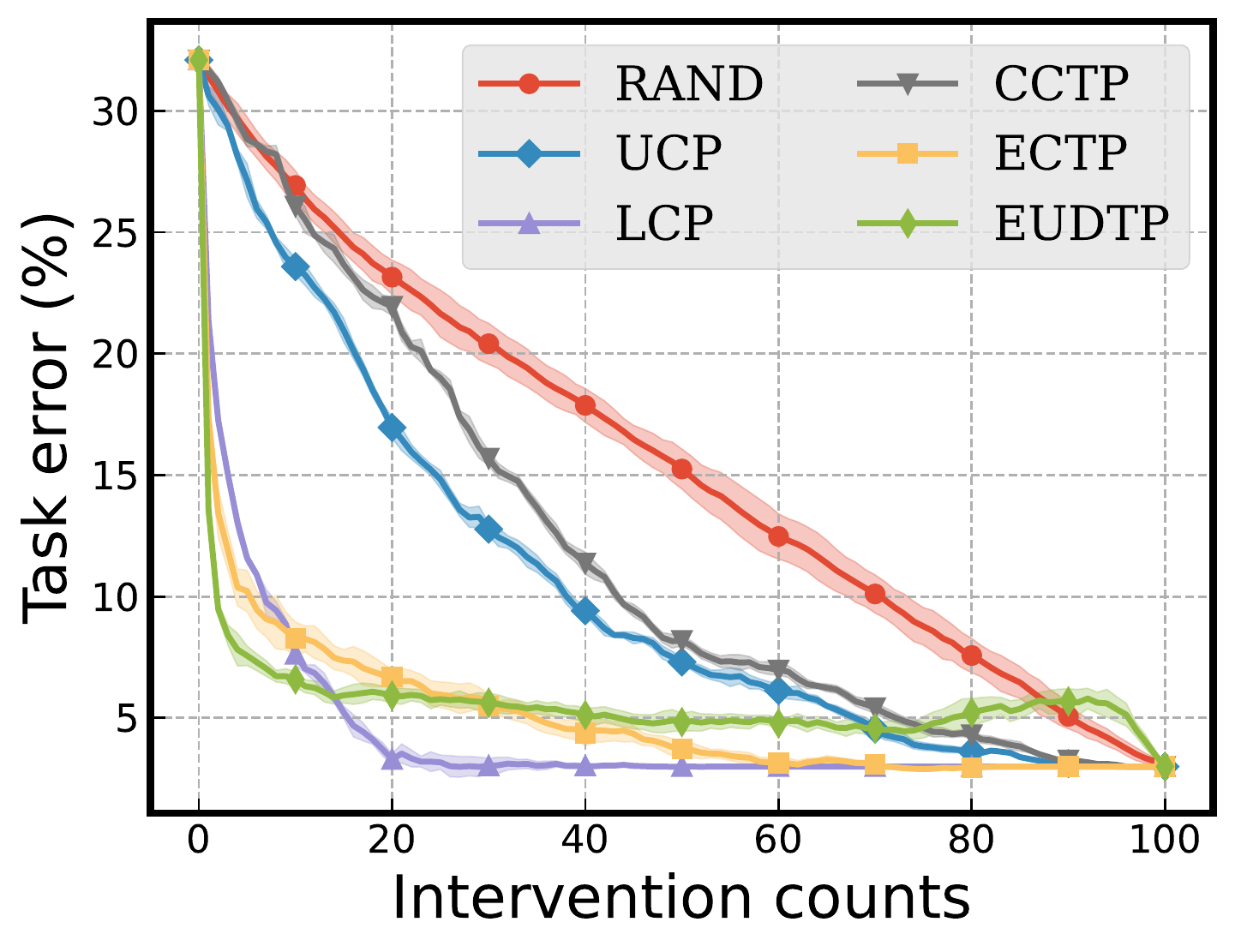}
    \caption{\textsc{jnt + p}}
    \label{fig:synthetic_result_training_jntp}
  \end{subfigure}%
  \caption{
    Comparison between concept selection criteria using different training strategies for the Synthetic.
    For \textsc{jnt, jnt + p}, we present the results when $\lambda=0.1$.
  }
  \label{fig:synthetic_result_training}
\end{figure*}

\begin{figure*}[!th]
  \begin{subfigure}{0.16\linewidth}
    \centering
    \includegraphics[width=\linewidth]{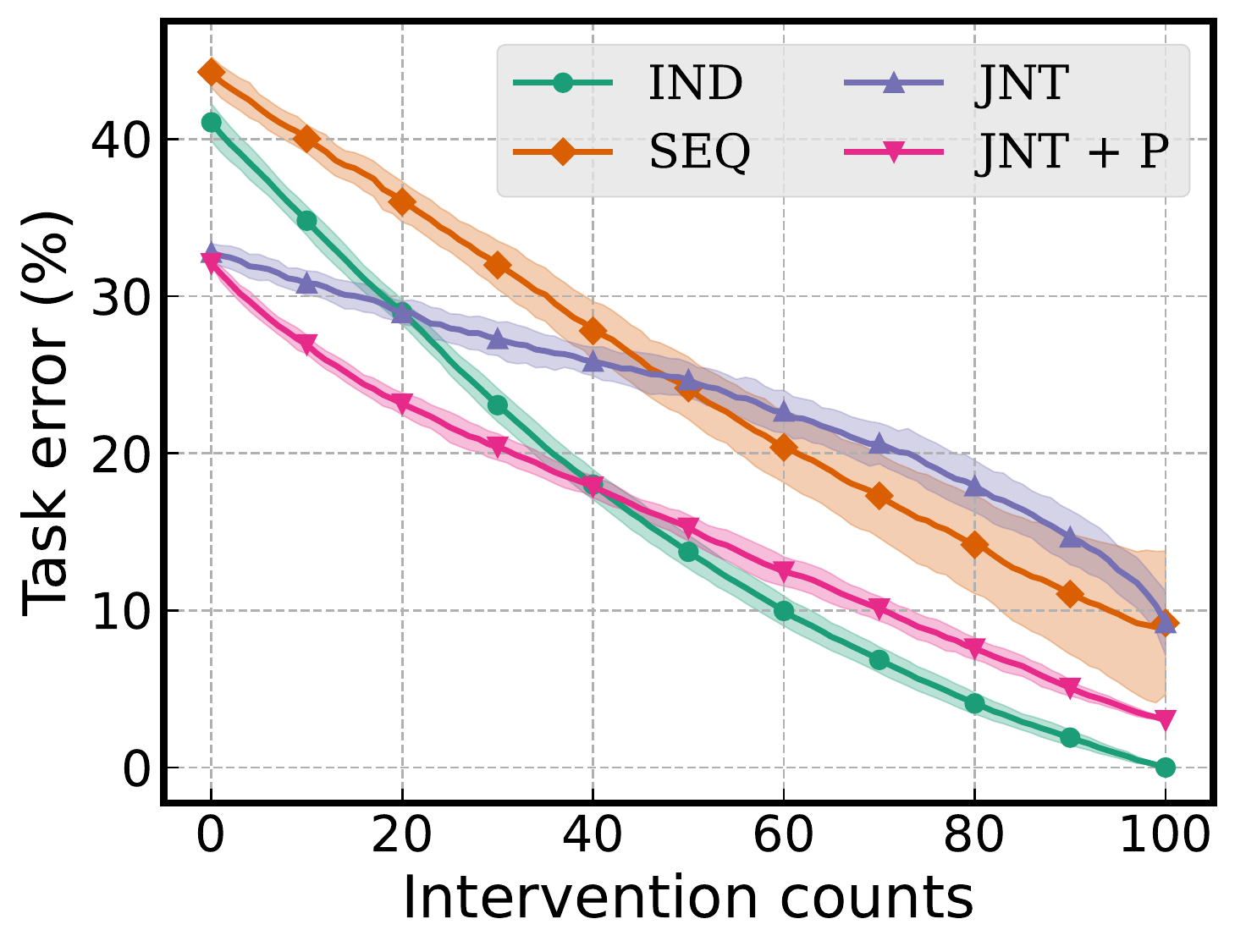}
    \caption{\textsc{rand}}
    \label{fig:synthetic_training_rand}
  \end{subfigure}%
  \begin{subfigure}{0.16\linewidth}
    \centering
    \includegraphics[width=\linewidth]{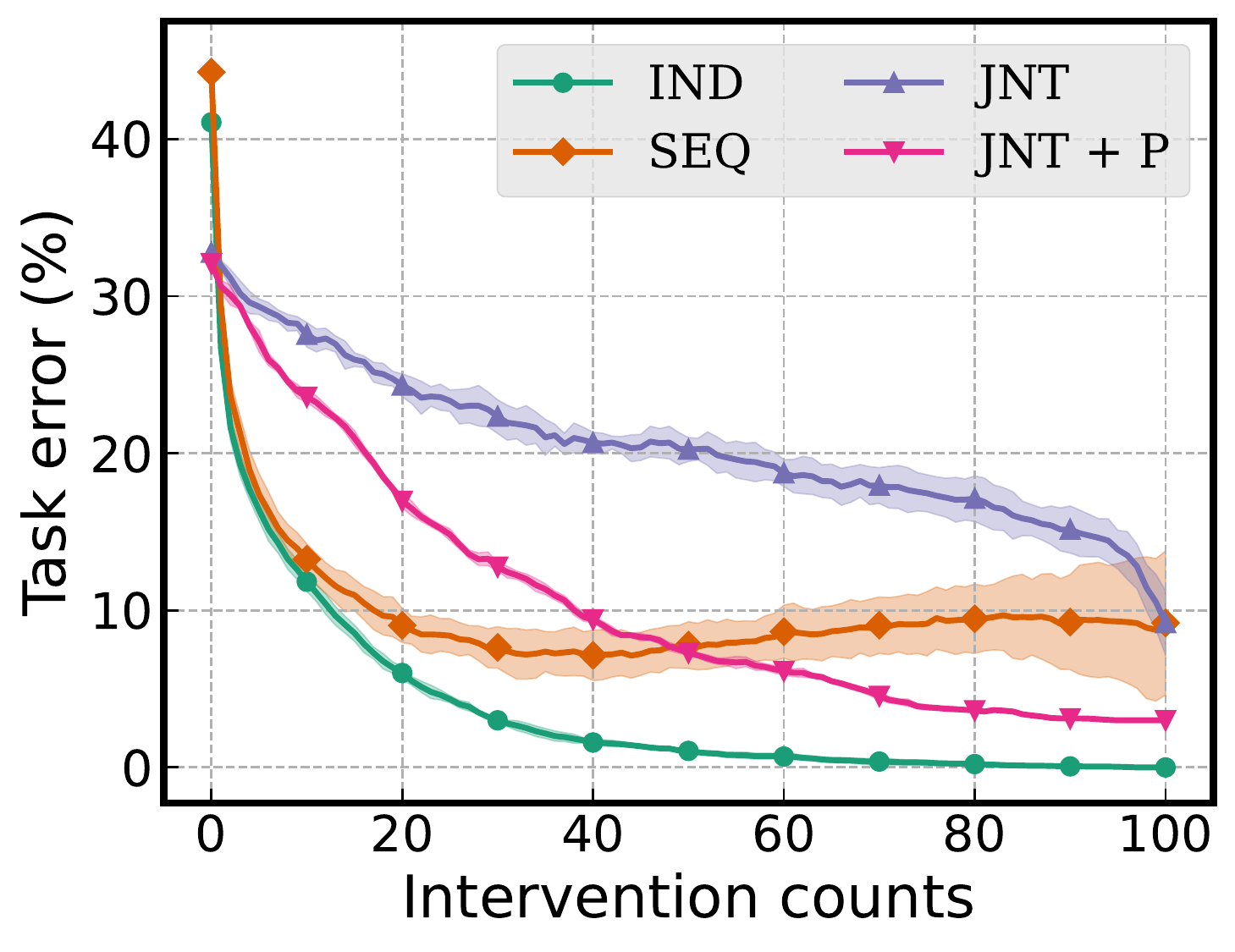}
    \caption{\textsc{ucp}}
    \label{fig:synthetic_training_ucp}
  \end{subfigure}%
  \begin{subfigure}{0.16\linewidth}
    \centering
    \includegraphics[width=\linewidth]{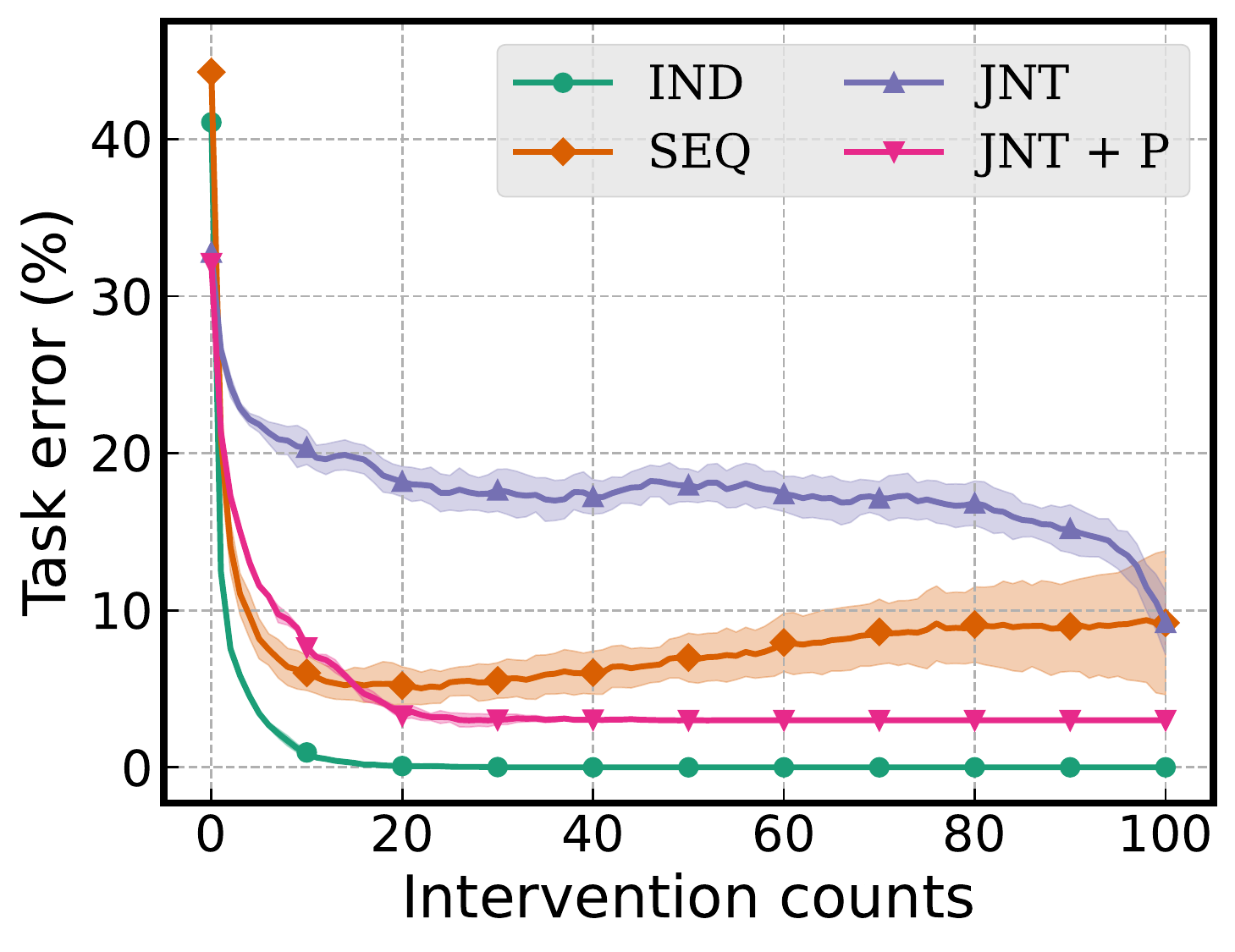}
    \caption{\textsc{lcp}}
    \label{fig:synthetic_training_lcp}
  \end{subfigure}%
  \begin{subfigure}{0.16\linewidth}
    \centering
    \includegraphics[width=\linewidth]{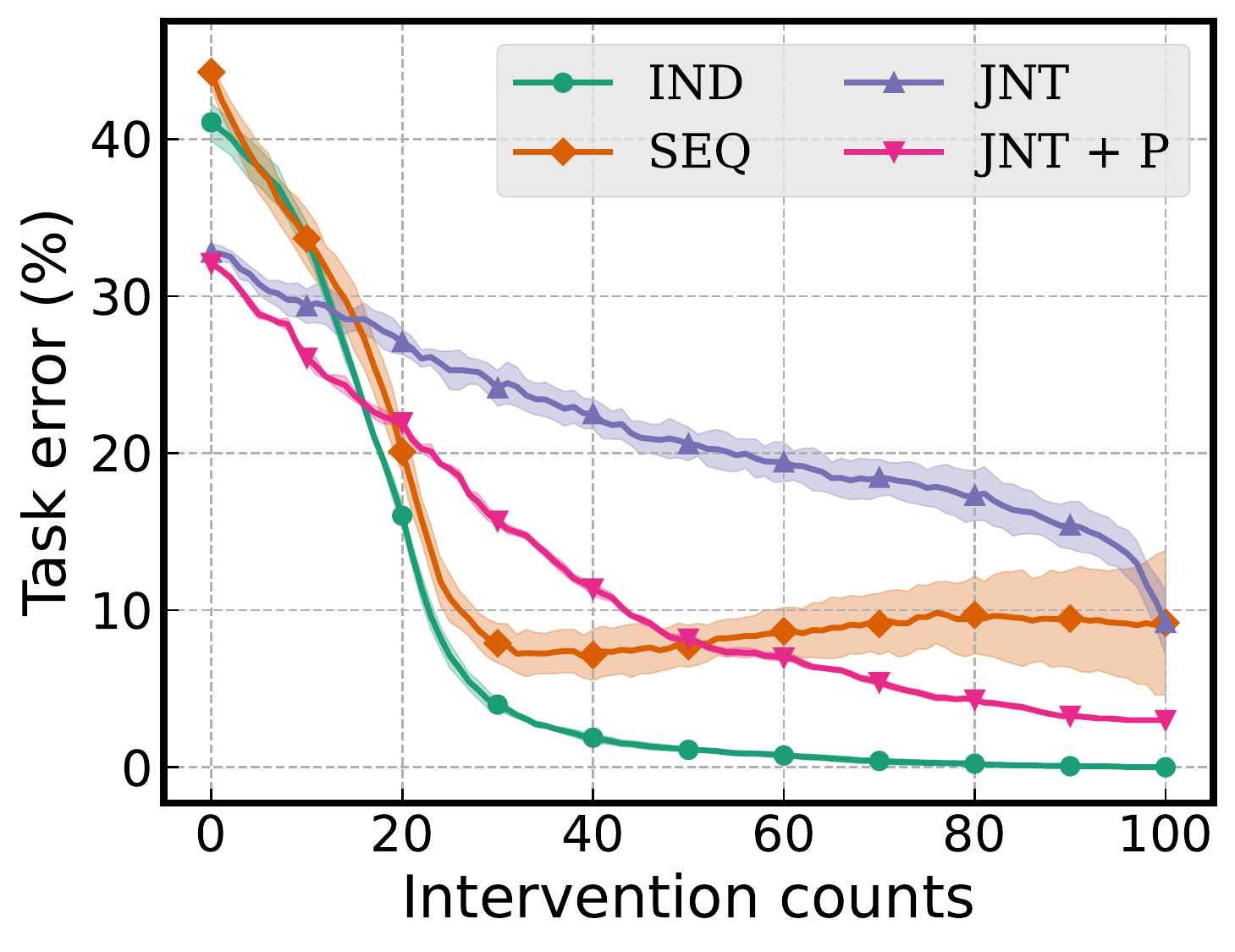}
    \caption{\textsc{cctp}}
    \label{fig:synthetic_training_cctp}
  \end{subfigure}%
  \begin{subfigure}{0.16\linewidth}
    \centering
    \includegraphics[width=\linewidth]{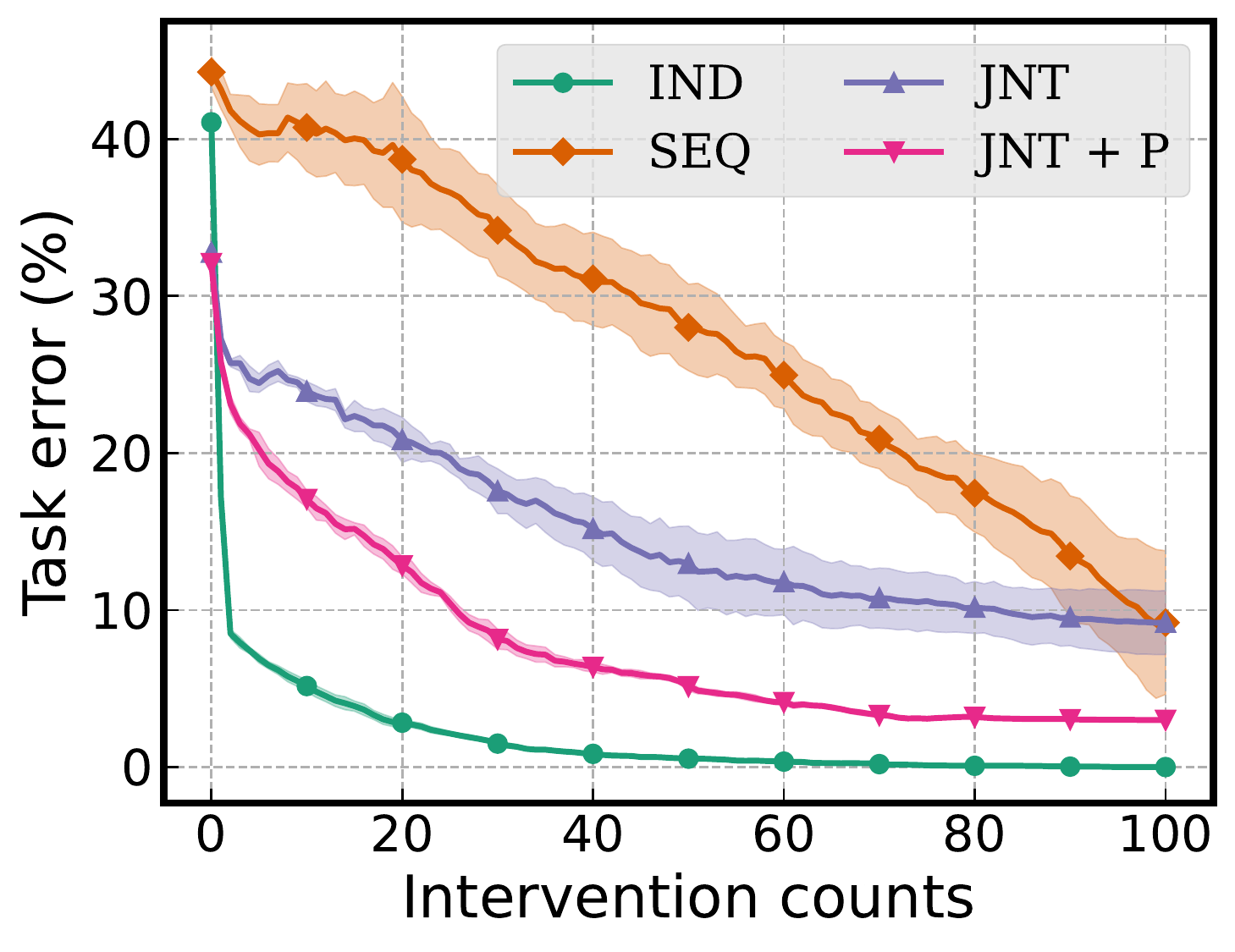}
    \caption{\textsc{ectp}}
    \label{fig:synthetic_training_ectp}
  \end{subfigure}%
  \begin{subfigure}{0.16\linewidth}
    \centering
    \includegraphics[width=\linewidth]{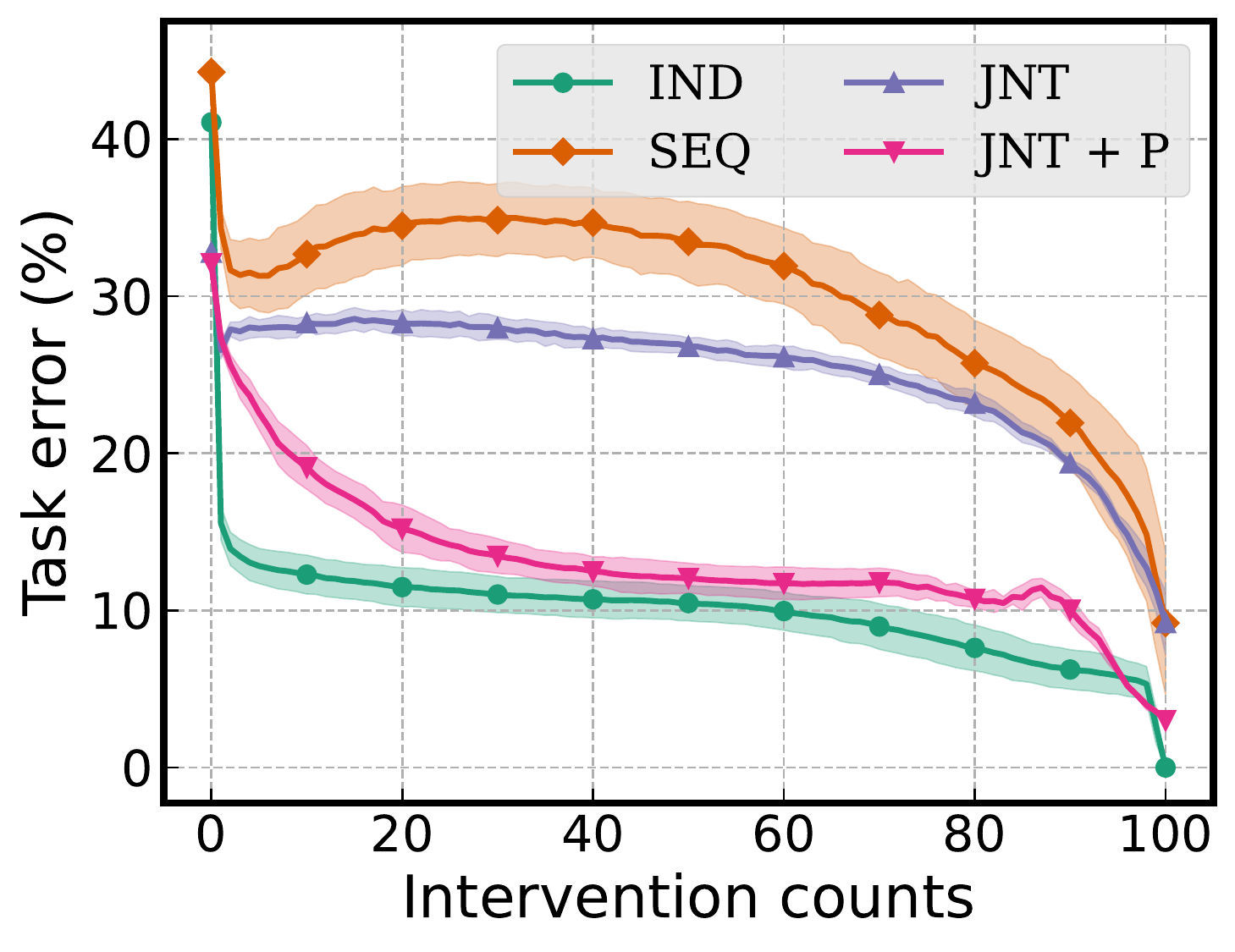}
    \caption{\textsc{eudtp}}
    \label{fig:synthetic_training_eudtp}
  \end{subfigure}%
  \caption{
    Comparison between different training strategies for a fixed concept selection criterion for the Synthetic.
    }
  \label{fig:synthetic_training_comparison}
\end{figure*}

The results for the CUB dataset are presented in \cref{fig:cub_result_training}.
Note that \textsc{eudtp} becomes even less effective than \textsc{rand} in \textsc{seq} and \textsc{jnt}.
For the synthetic datasets, \textsc{eudtp} also becomes much less effective as in the CUB dataset (see \cref{fig:synthetic_result_training}).
Note that when using \textsc{jnt} or \textsc{jnt+p} training schemes, \textsc{lcp} may not be the best choice as the target predictor $f$ is not trained with the ground-truth concept values and thus rectifying the concept with the highest prediction loss does not always guarantee the decrease in the task error.
Comparisons between different training strategies for a fixed concept selection criterion in the CUB and Synthetic are presented in \cref{fig:cub_training_comparison,fig:synthetic_training_comparison}.

\section{More Results on the Effect of Conceptualization Methods on Intervention}

\label{sec:results-others-conceptualization}

\begin{figure}[!th]
  \begin{subfigure}{0.16\linewidth}
    \centering
    \includegraphics[width=\linewidth]{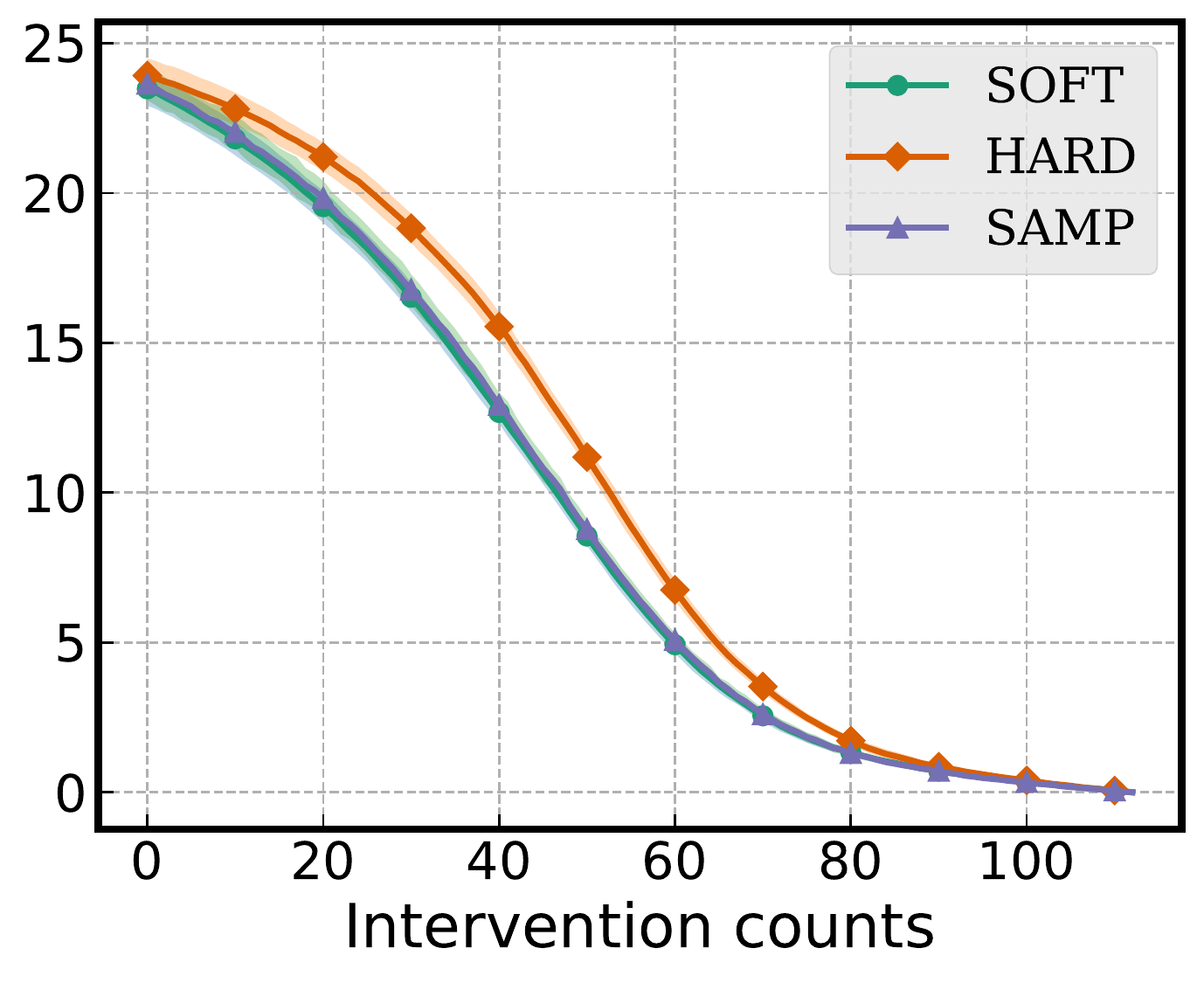}
    \caption{\textsc{rand}}
    \label{fig:cub_conceptualization_ind_rand}
  \end{subfigure}
  \begin{subfigure}{0.16\linewidth}
    \centering
    \includegraphics[width=\linewidth]{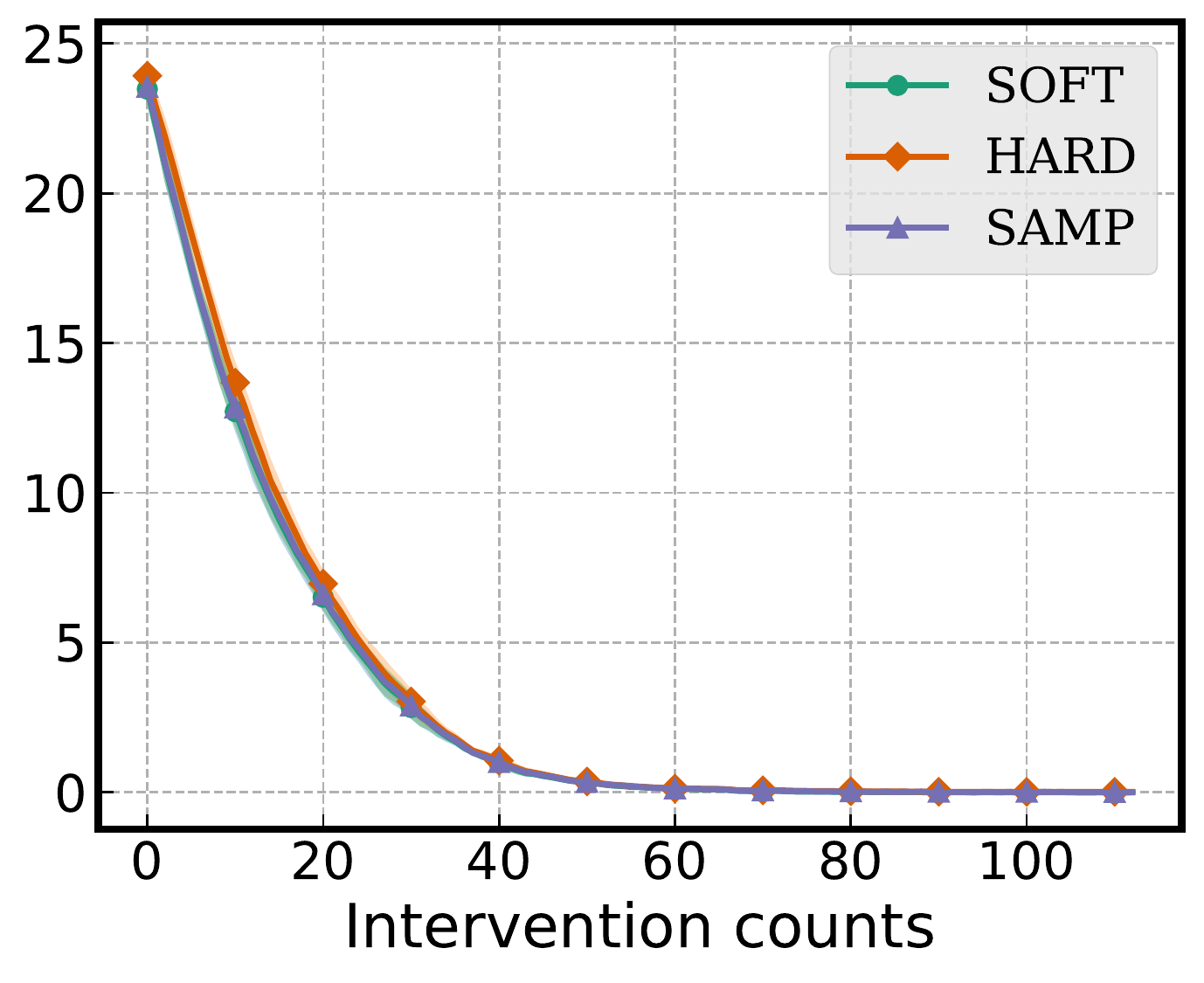}
    \caption{\textsc{ucp}}
    \label{fig:cub_conceptualization_ind_ucp}
  \end{subfigure}
  \begin{subfigure}{0.16\linewidth}
    \centering
    \includegraphics[width=\linewidth]{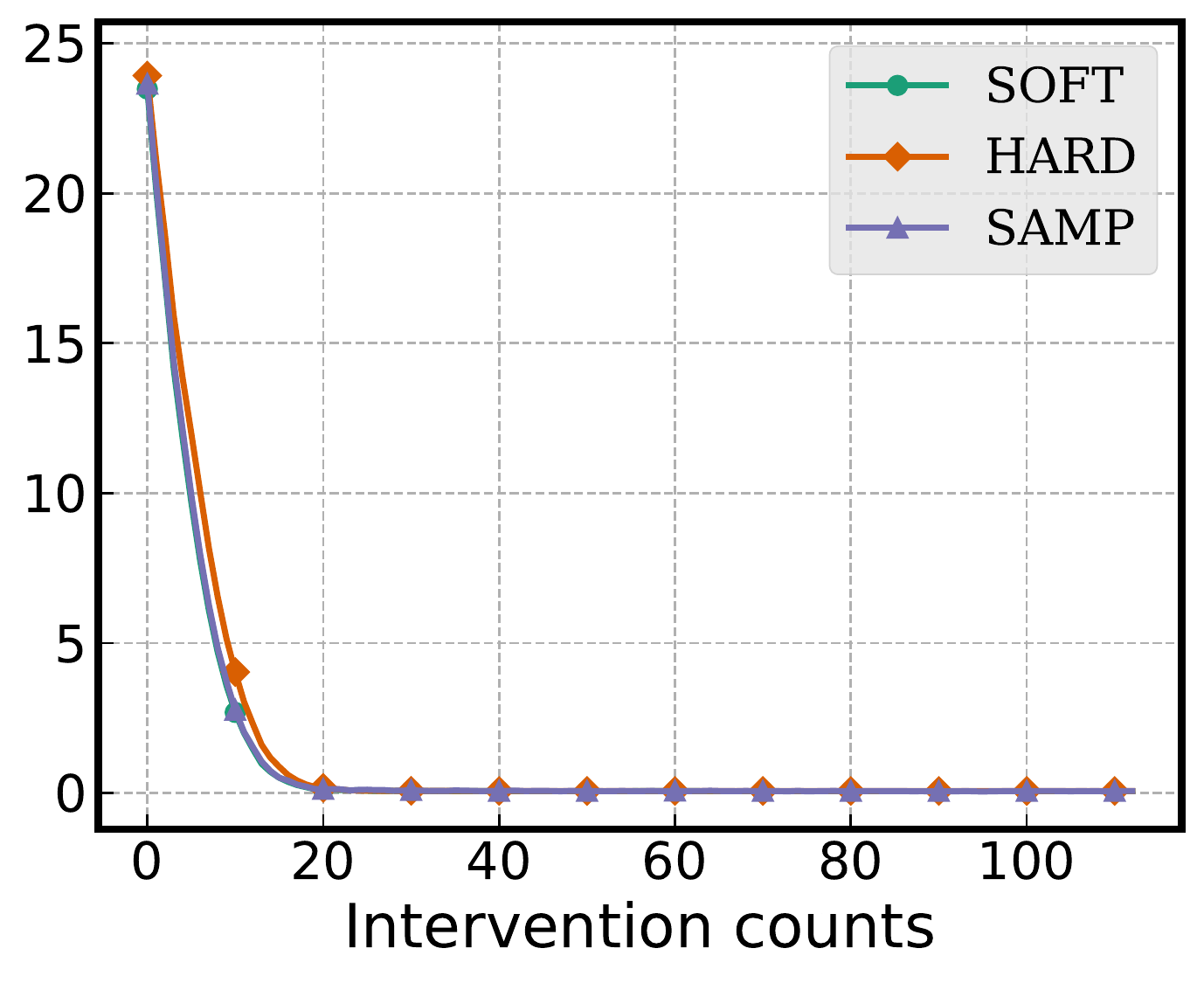}
    \caption{\textsc{lcp}}
    \label{fig:cub_conceptualization_ind_lcp}
  \end{subfigure}
  \begin{subfigure}{0.16\linewidth}
    \centering
    \includegraphics[width=\linewidth]{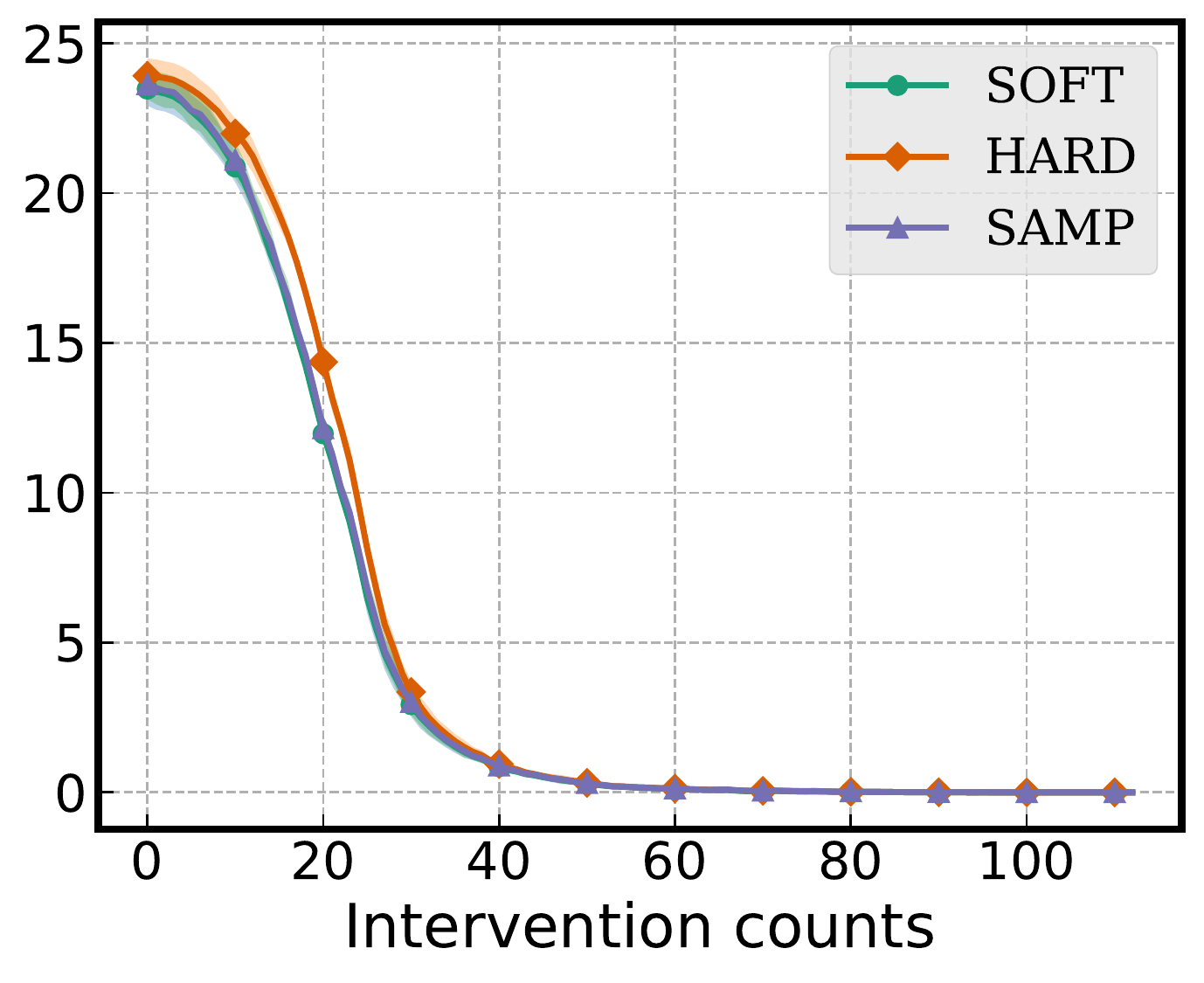}
    \caption{\textsc{cctp}}
    \label{fig:cub_conceptualization_ind_cctp}
  \end{subfigure}
  \begin{subfigure}{0.16\linewidth}
    \centering
    \includegraphics[width=\linewidth]{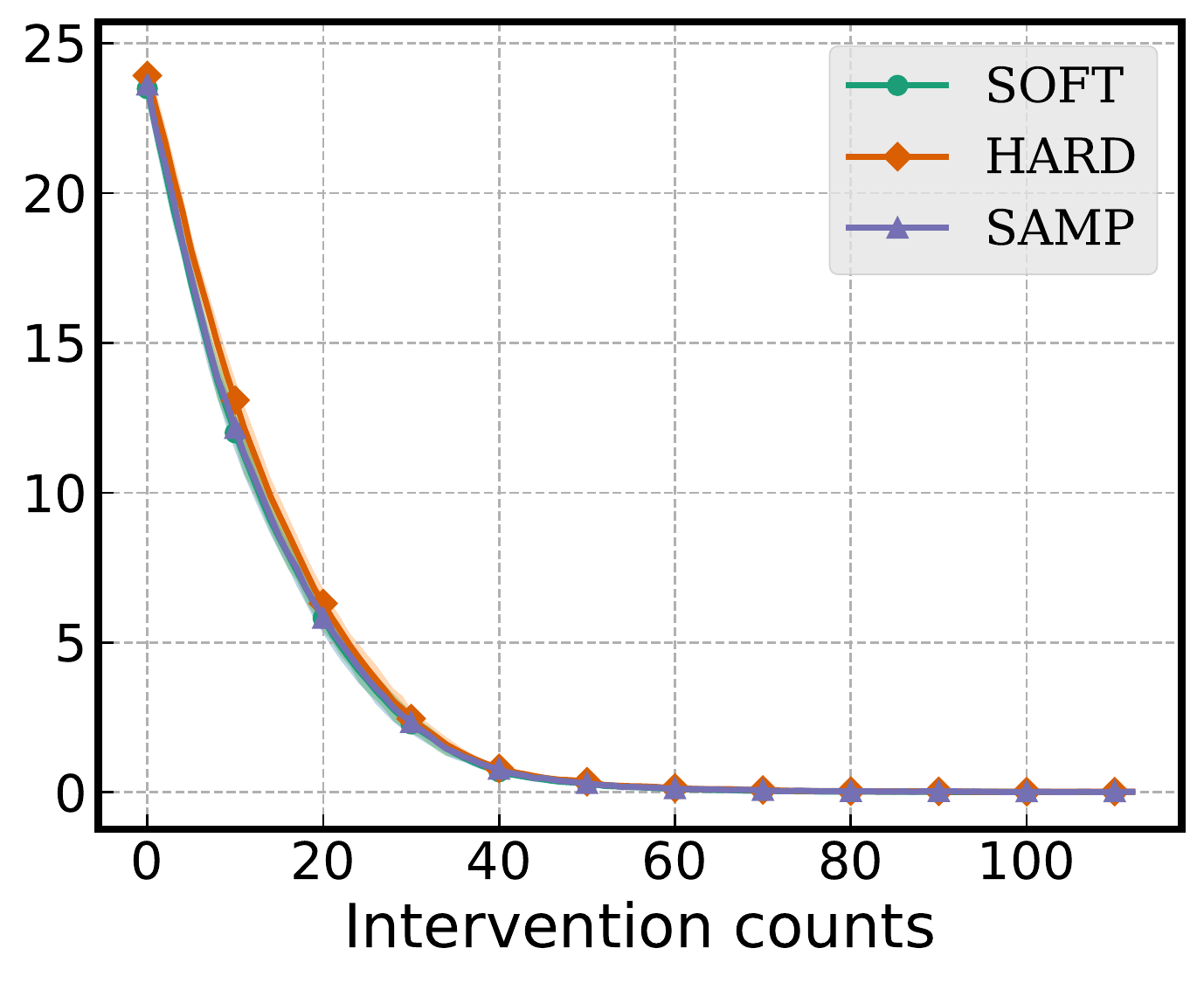}
    \caption{\textsc{ectp}}
    \label{fig:cub_conceptualization_ind_ectp}
  \end{subfigure}
  \begin{subfigure}{0.16\linewidth}
    \centering
    \includegraphics[width=\linewidth]{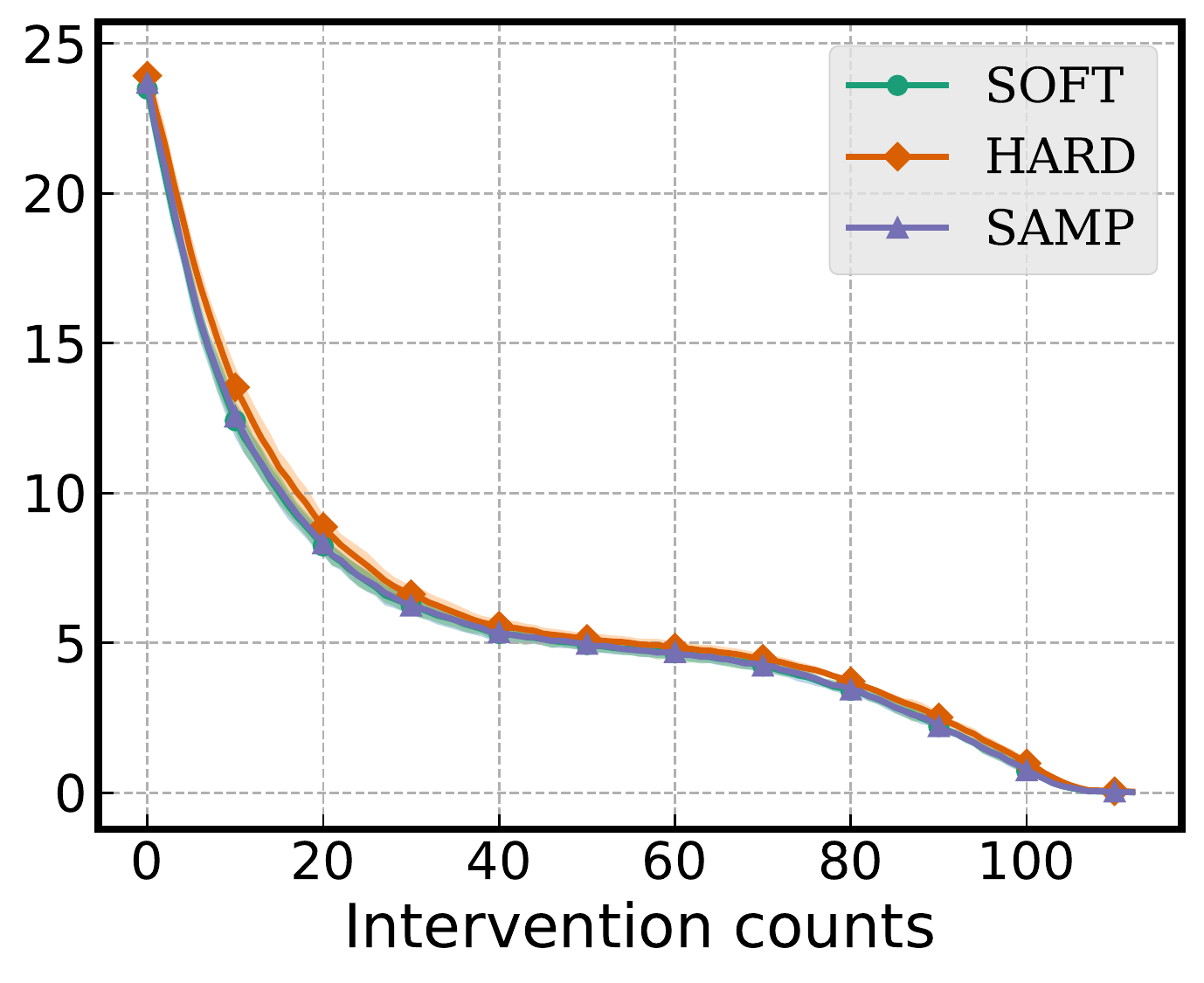}
    \caption{\textsc{eudtp}}
    \label{fig:cub_conceptualization_ind_eudtp}
  \end{subfigure}
  \caption{
    Intervention results under different conceptualization methods using various concept selection criteria.
    Here, we used \textsc{ind} training strategy for the CUB.
  }
  \label{fig:cub_conceptualization_ind}
\end{figure}

\begin{figure}[!th]
  \begin{subfigure}{0.16\linewidth}
    \centering
    \includegraphics[width=\linewidth]{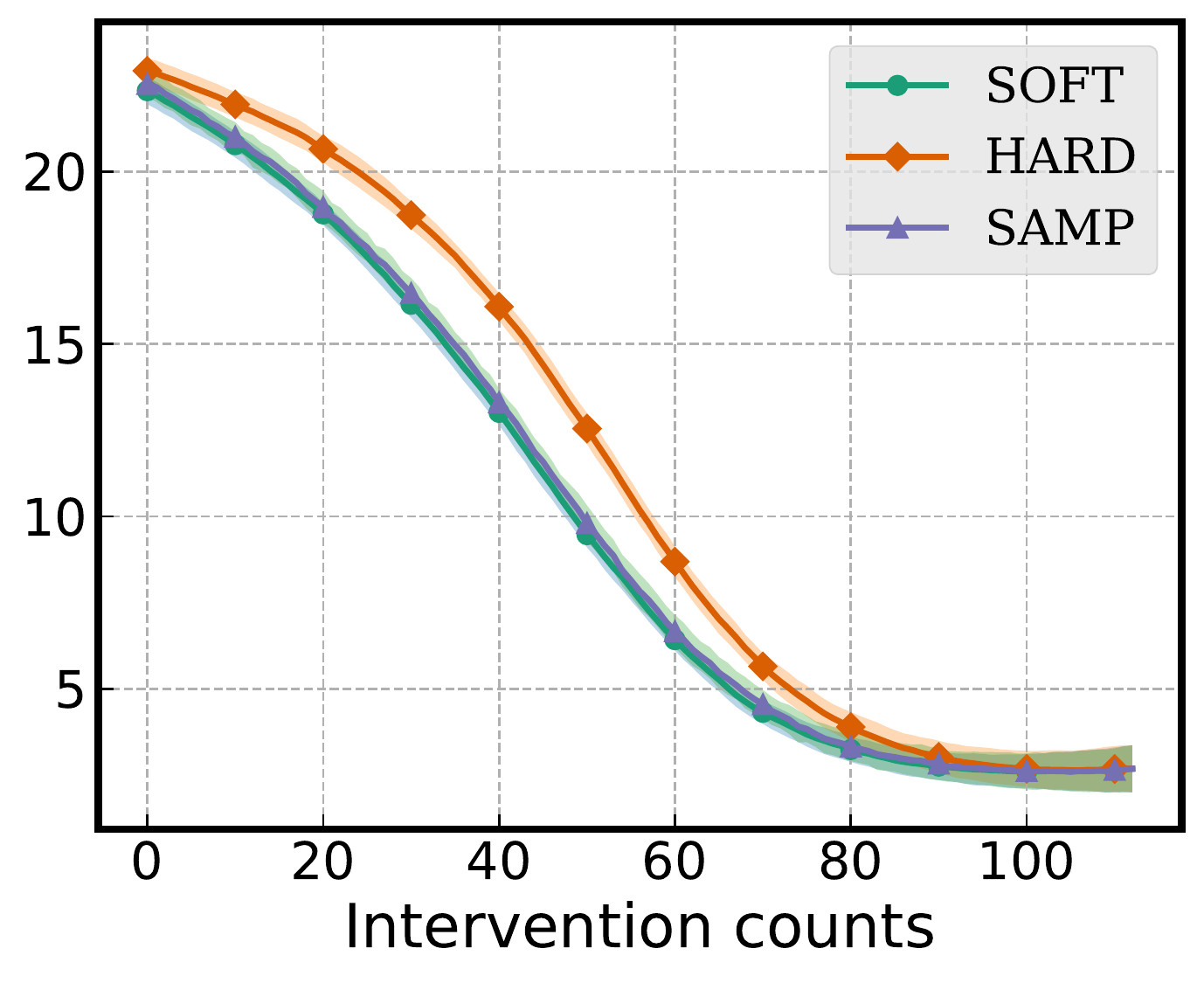}
    \caption{\textsc{rand}}
    \label{fig:cub_conceptualization_jntp_rand}
  \end{subfigure}
  \begin{subfigure}{0.16\linewidth}
    \centering
    \includegraphics[width=\linewidth]{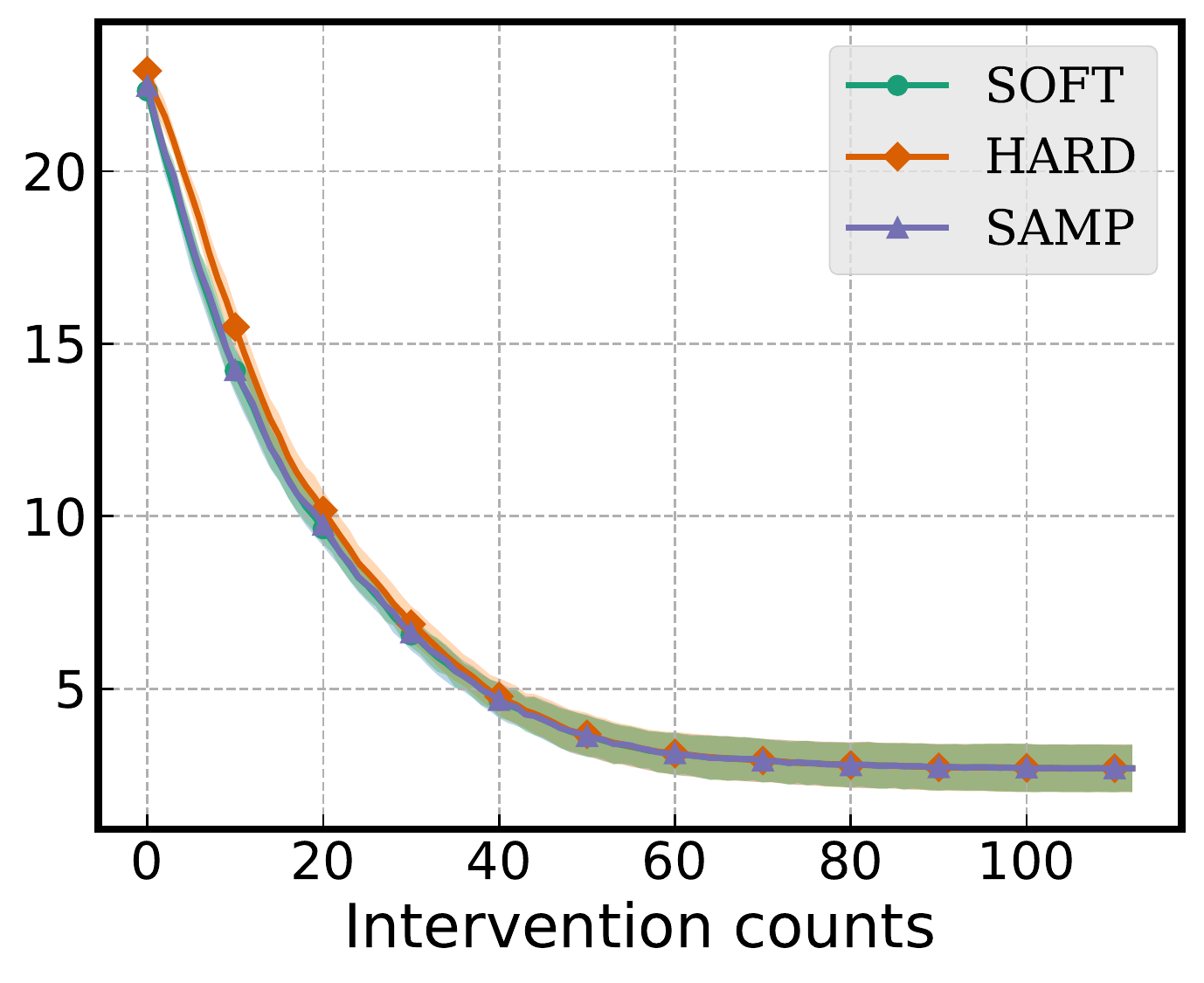}
    \caption{\textsc{ucp}}
    \label{fig:cub_conceptualization_jntp_ucp}
  \end{subfigure}
  \begin{subfigure}{0.16\linewidth}
    \centering
    \includegraphics[width=\linewidth]{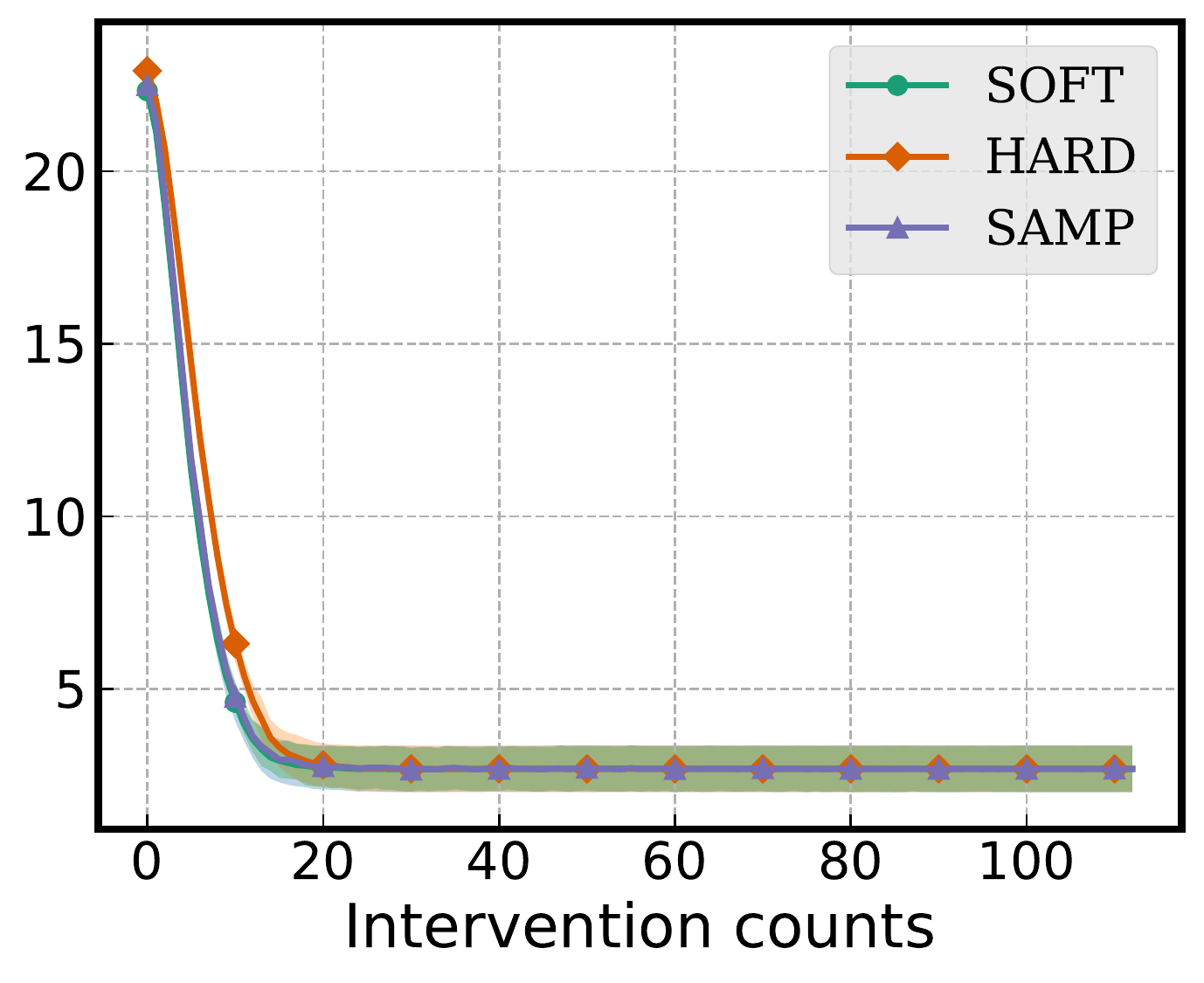}
    \caption{\textsc{lcp}}
    \label{fig:cub_conceptualization_jntp_lcp}
  \end{subfigure}
  \begin{subfigure}{0.16\linewidth}
    \centering
    \includegraphics[width=\linewidth]{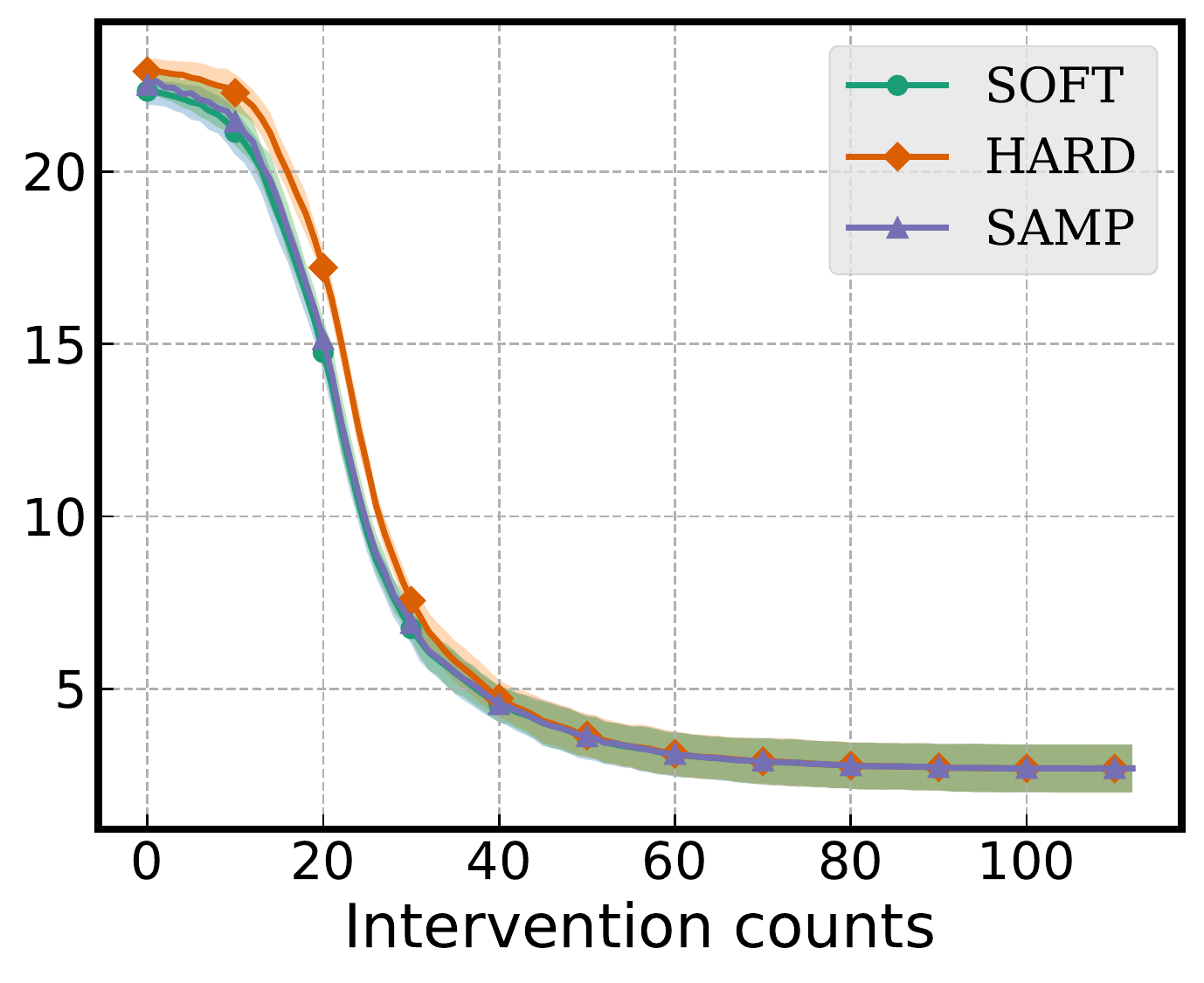}
    \caption{\textsc{cctp}}
    \label{fig:cub_conceptualization_jntp_cctp}
  \end{subfigure}
  \begin{subfigure}{0.16\linewidth}
    \centering
    \includegraphics[width=\linewidth]{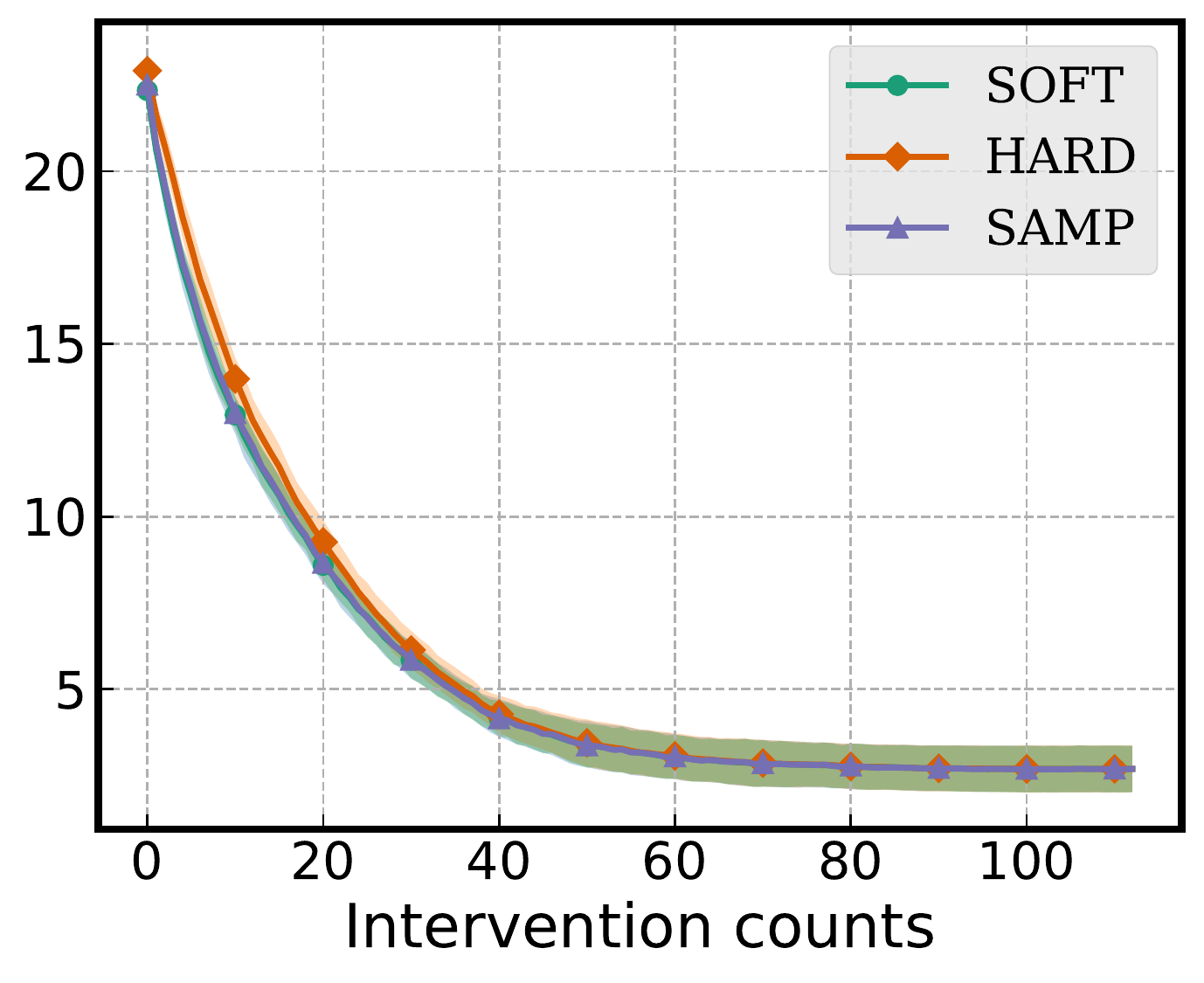}
    \caption{\textsc{ectp}}
    \label{fig:cub_conceptualization_jntp_ectp}
  \end{subfigure}
  \begin{subfigure}{0.16\linewidth}
    \centering
    \includegraphics[width=\linewidth]{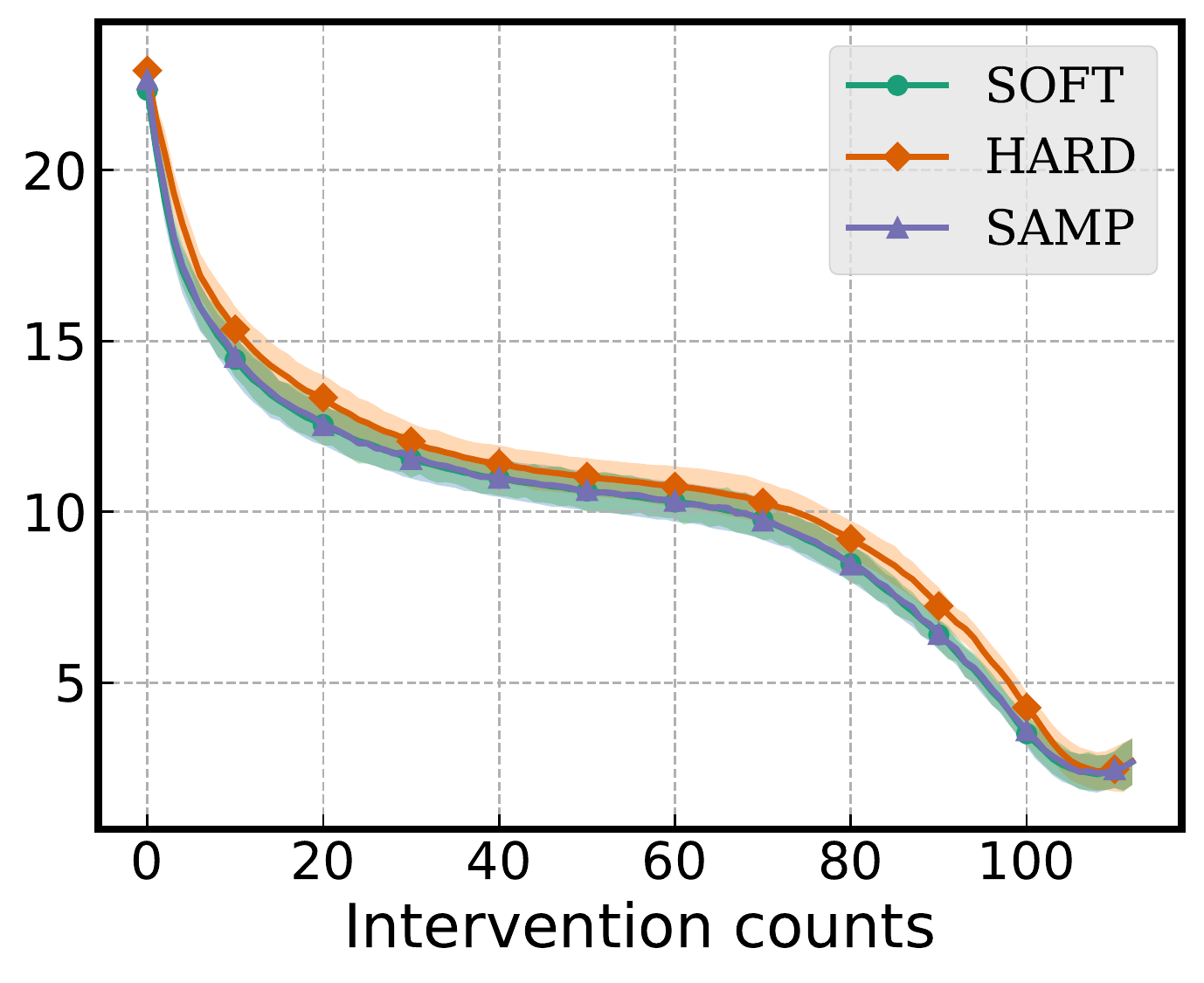}
    \caption{\textsc{eudtp}}
    \label{fig:cub_conceptualization_jntp_eudtp}
  \end{subfigure}
  \caption{
    Intervention results under different conceptualization methods using various concept selection criteria.
    Here, we used \textsc{jnt + p} training strategy for the CUB.
  }
  \label{fig:cub_conceptualization_jntp}
\end{figure}

\begin{figure}[!th]
  \begin{subfigure}{0.16\linewidth}
    \centering
    \includegraphics[width=\linewidth]{figures/skincon/conceptualization_rand.pdf}
    \caption{\textsc{rand}}
    \label{fig:skincon_conceptualization_ind_rand}
  \end{subfigure}
  \begin{subfigure}{0.16\linewidth}
    \centering
    \includegraphics[width=\linewidth]{figures/skincon/conceptualization_ucp.pdf}
    \caption{\textsc{ucp}}
    \label{fig:skincon_conceptualization_ind_ucp}
  \end{subfigure}
  \begin{subfigure}{0.16\linewidth}
    \centering
    \includegraphics[width=\linewidth]{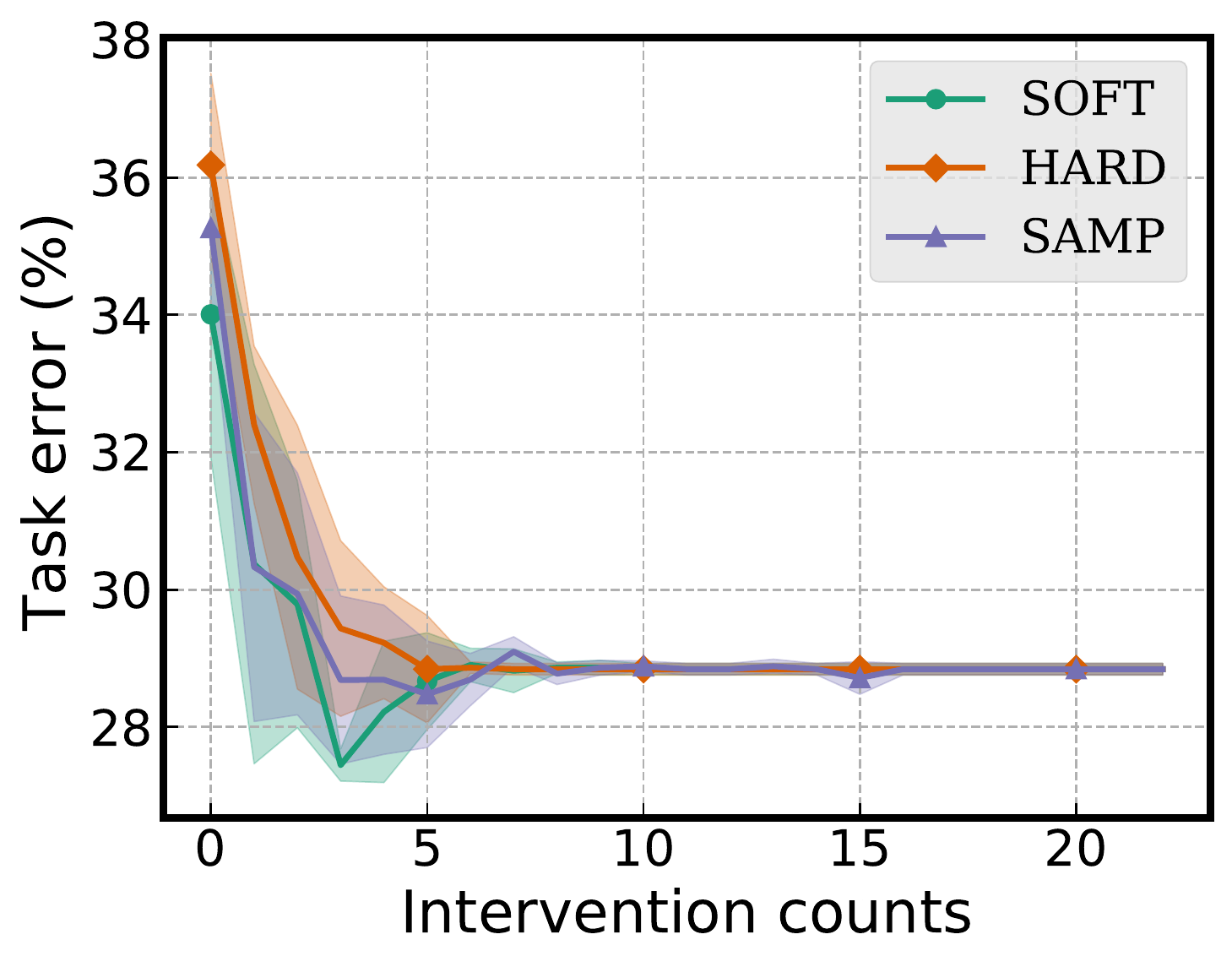}
    \caption{\textsc{lcp}}
    \label{fig:skincon_conceptualization_ind_lcp}
  \end{subfigure}
  \begin{subfigure}{0.16\linewidth}
    \centering
    \includegraphics[width=\linewidth]{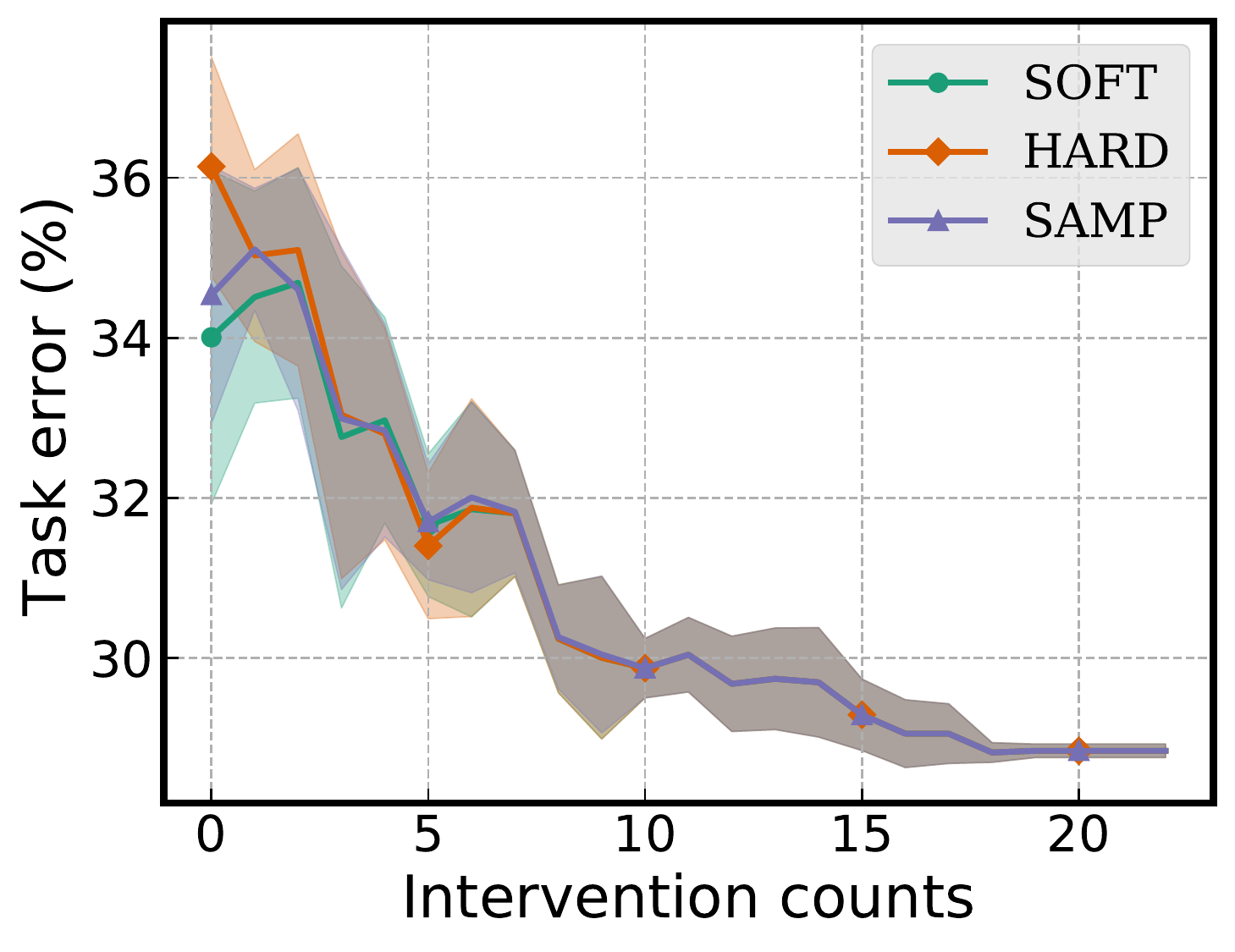}
    \caption{\textsc{cctp}}
    \label{fig:skincon_conceptualization_ind_cctp}
  \end{subfigure}
  \begin{subfigure}{0.16\linewidth}
    \centering
    \includegraphics[width=\linewidth]{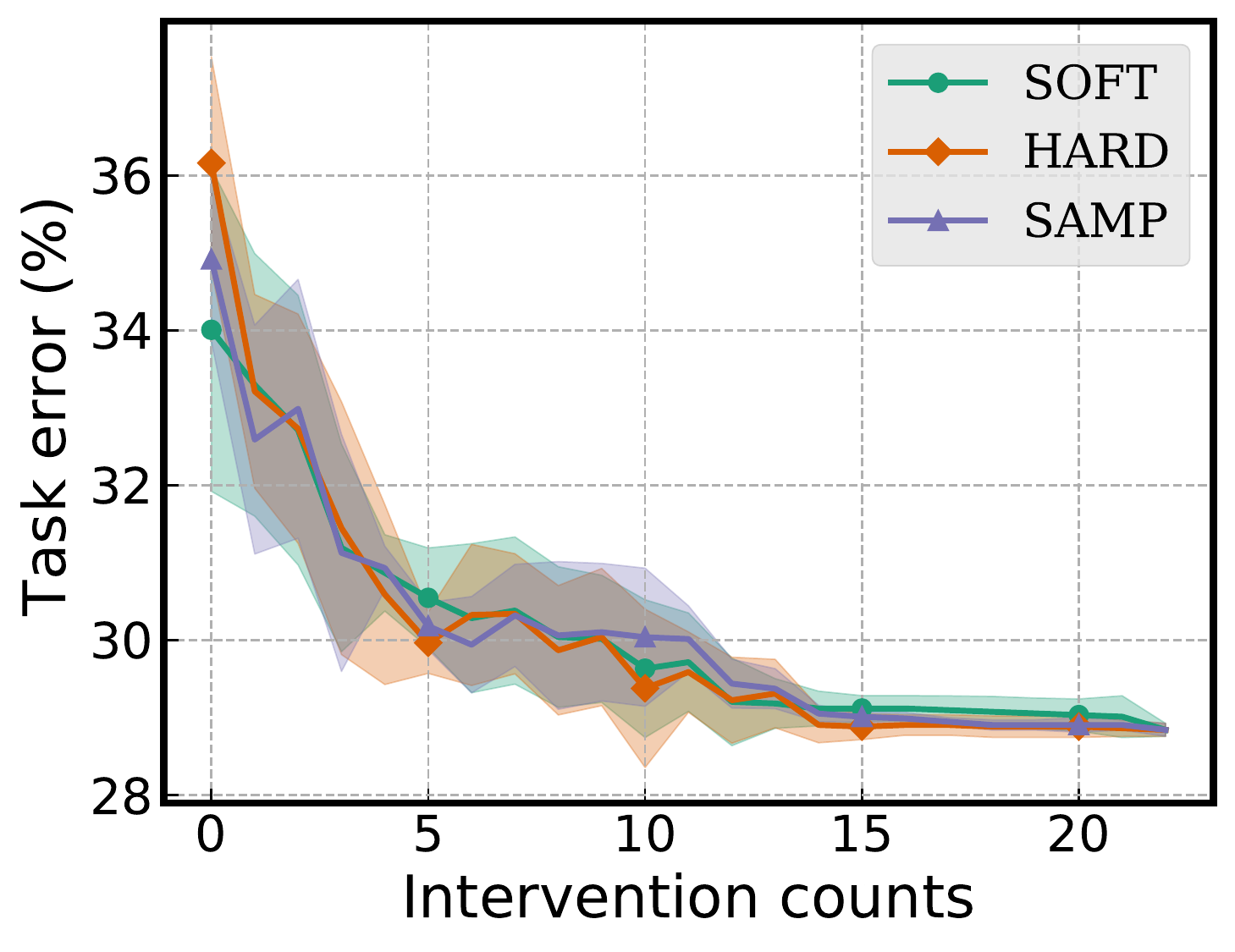}
    \caption{\textsc{ectp}}
    \label{fig:skincon_conceptualization_ind_ectp}
  \end{subfigure}
  \begin{subfigure}{0.16\linewidth}
    \centering
    \includegraphics[width=\linewidth]{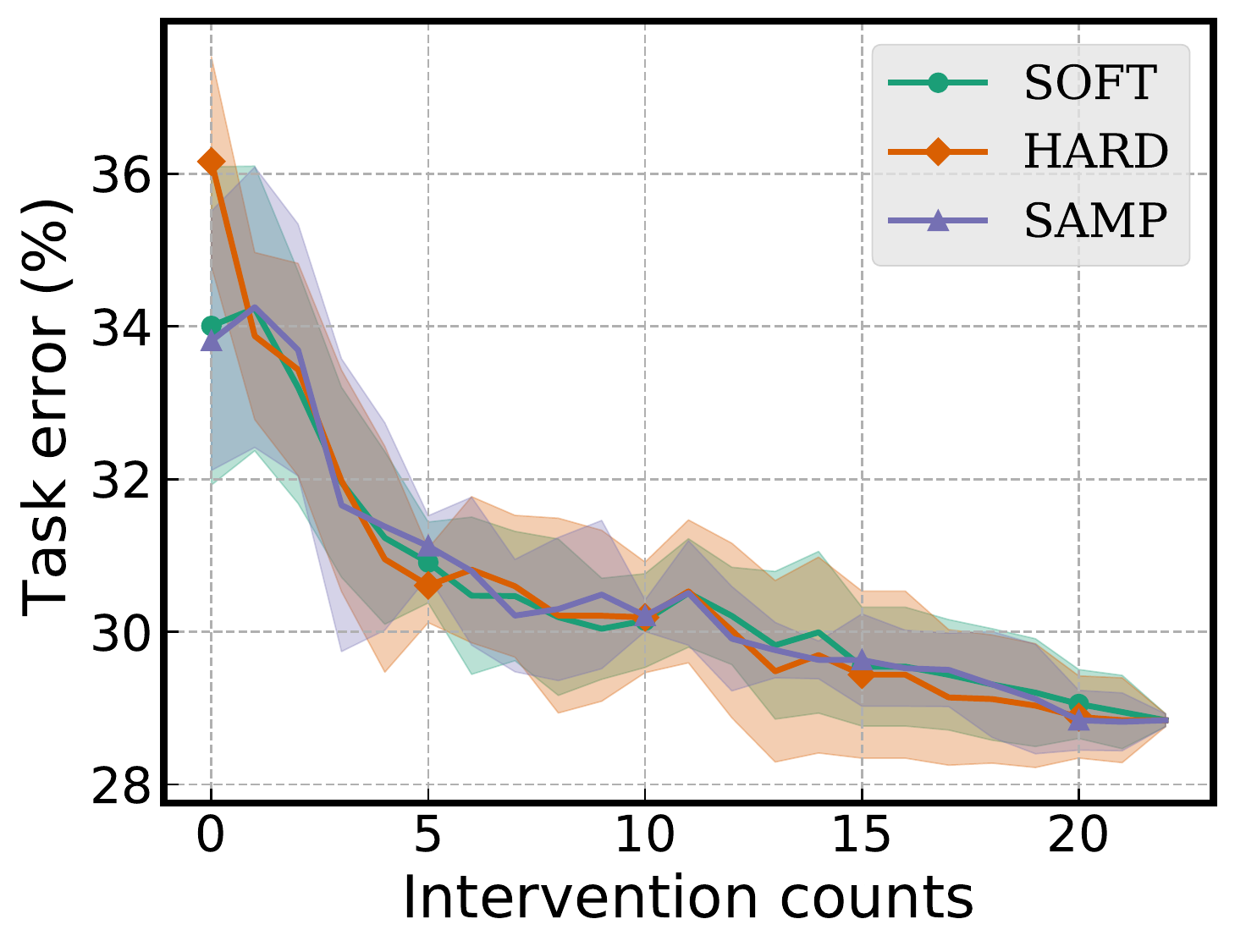}
    \caption{\textsc{eudtp}}
    \label{fig:skincon_conceptualization_ind_eudtp}
  \end{subfigure}
  \caption{
    Intervention results under different conceptualization methods using other concept selection criteria.
    Here, we used \textsc{ind} training strategy for the SkinCon.
  }
  \label{fig:skincon_conceptualization}
\end{figure}

\begin{figure}[!th]
  \begin{subfigure}{0.16\linewidth}
    \centering
    \includegraphics[width=\linewidth]{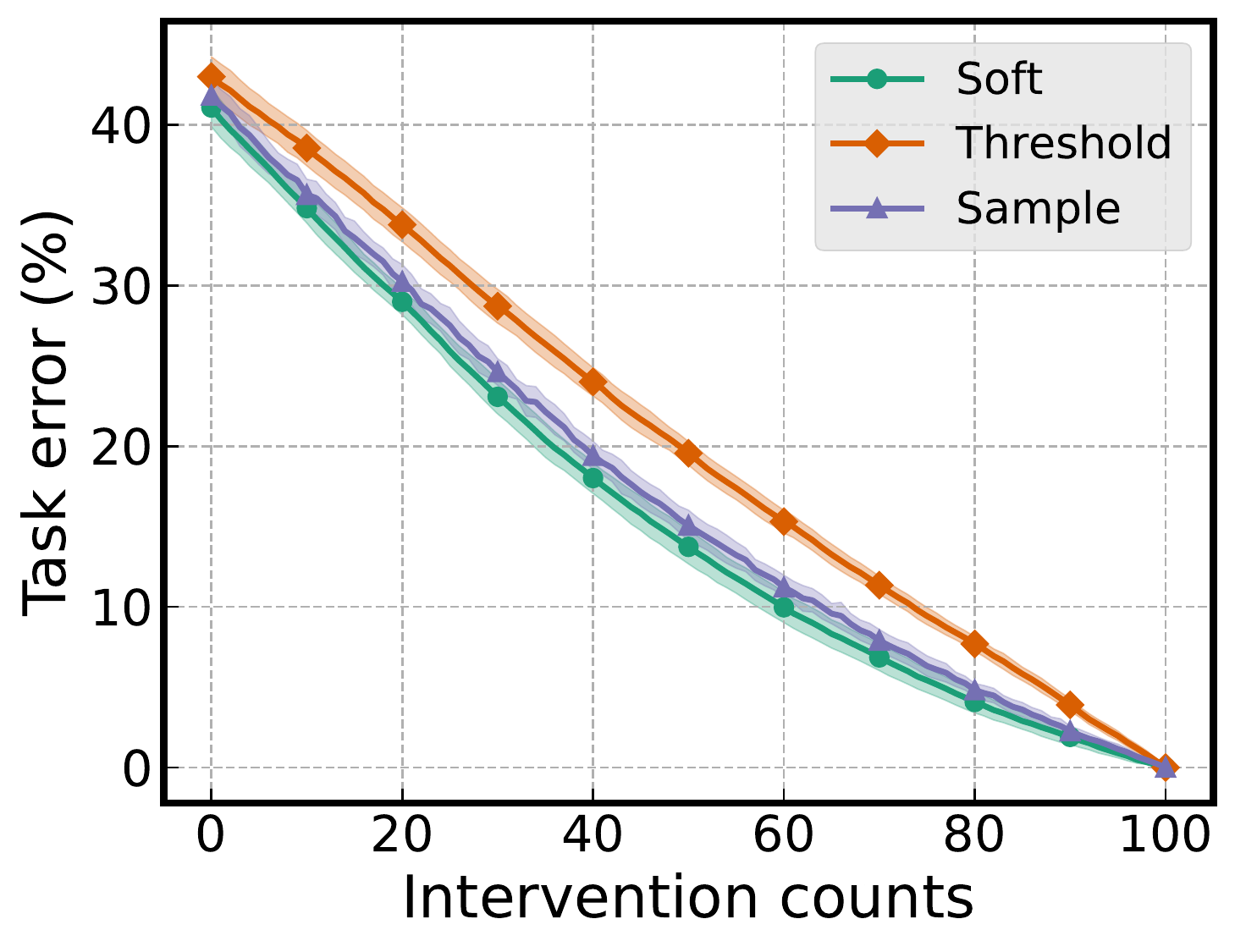}
    \caption{\textsc{rand}}
    \label{fig:synthetic_conceptualization_ind_rand}
  \end{subfigure}
  \begin{subfigure}{0.16\linewidth}
    \centering
    \includegraphics[width=\linewidth]{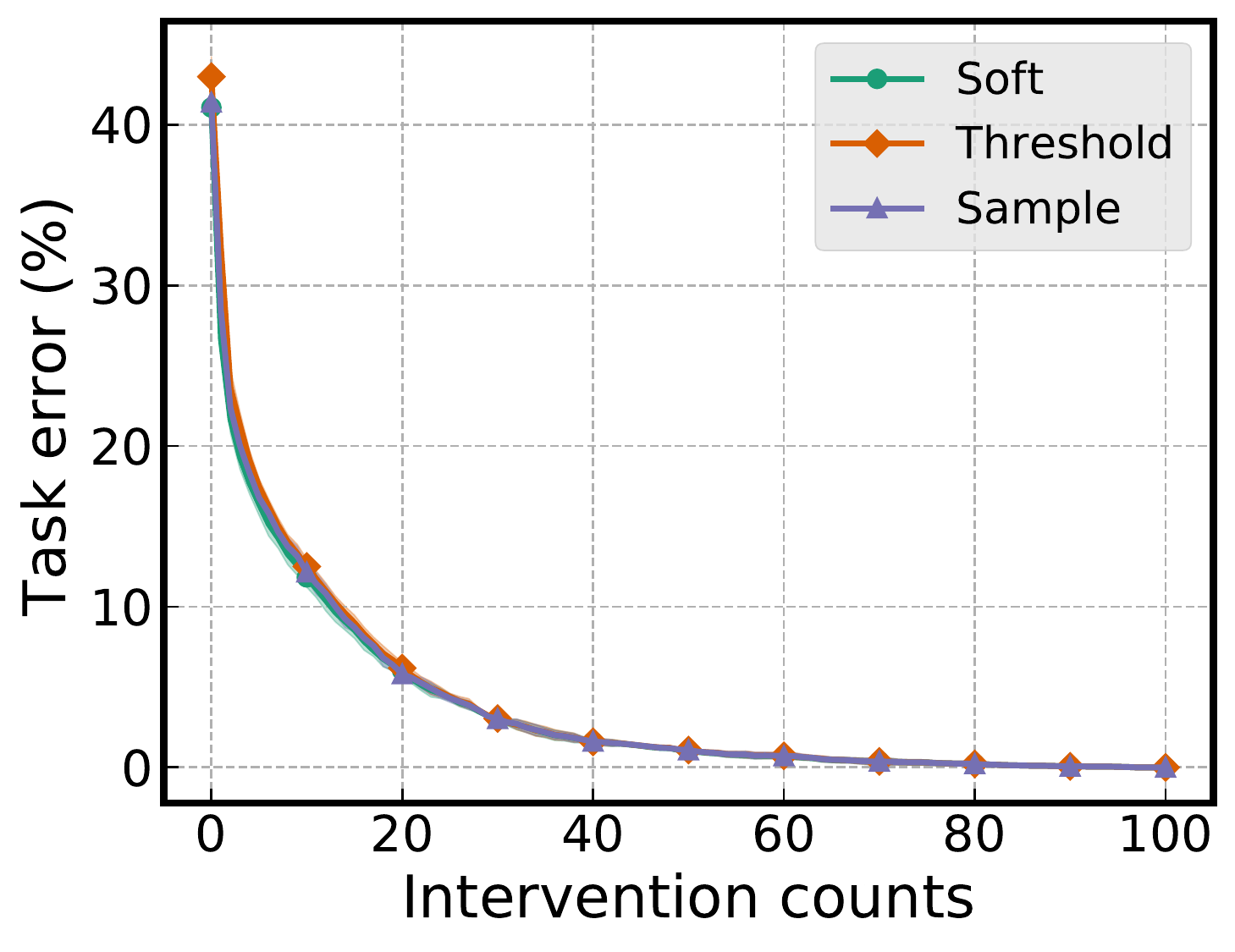}
    \caption{\textsc{ucp}}
    \label{fig:synthetic_conceptualization_ind_ucp}
  \end{subfigure}
  \begin{subfigure}{0.16\linewidth}
    \centering
    \includegraphics[width=\linewidth]{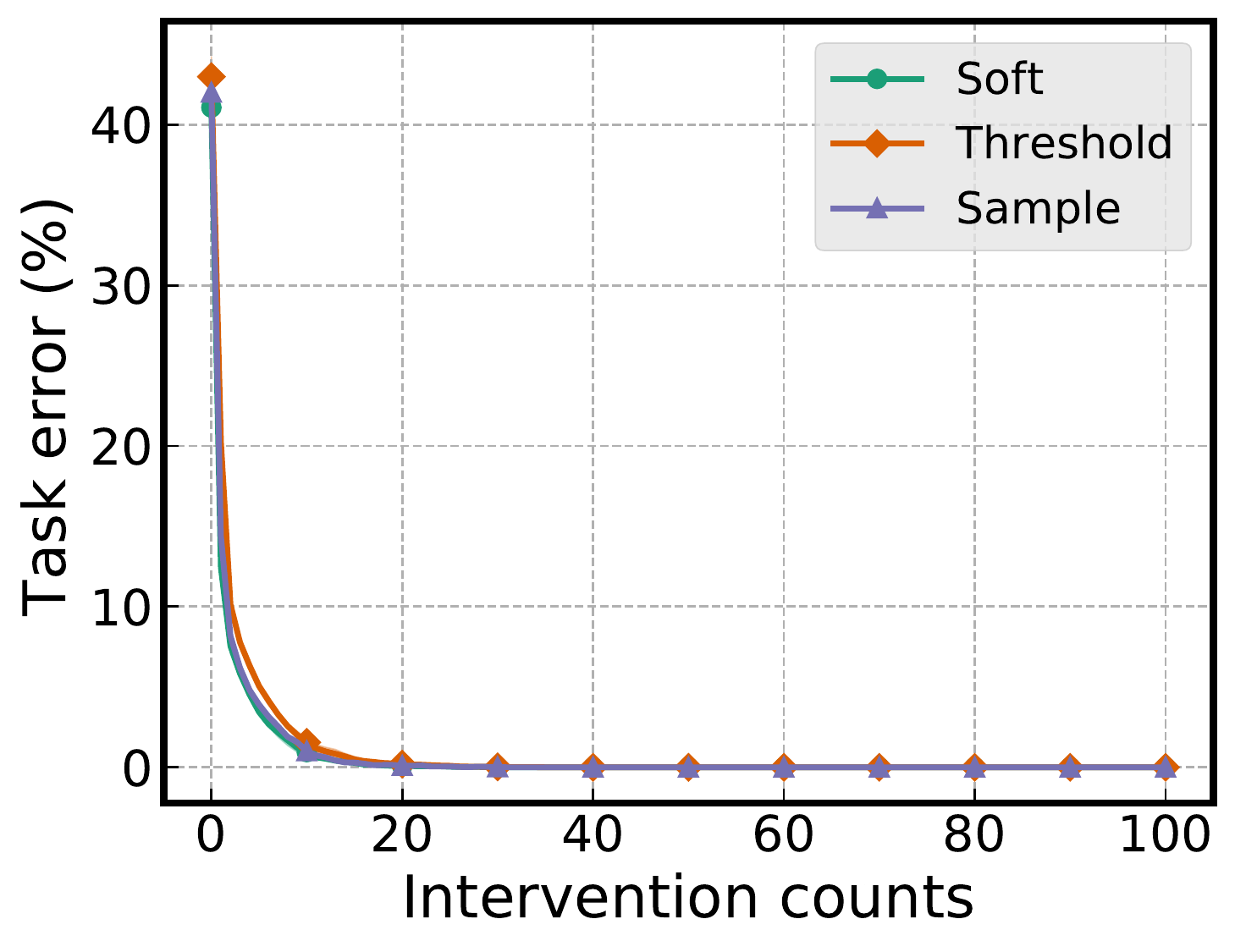}
    \caption{\textsc{lcp}}
    \label{fig:synthetic_conceptualization_ind_lcp}
  \end{subfigure}
  \begin{subfigure}{0.16\linewidth}
    \centering
    \includegraphics[width=\linewidth]{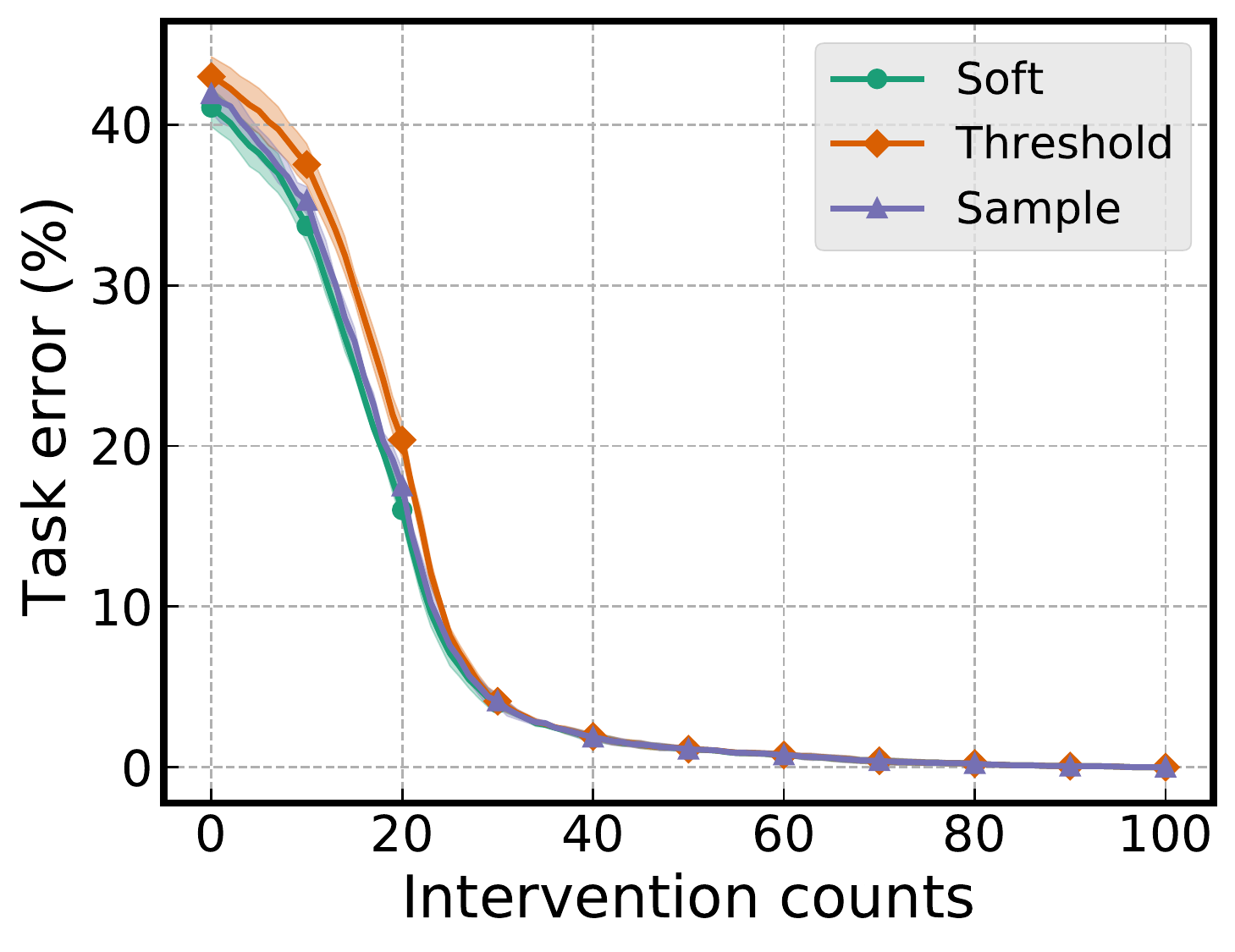}
    \caption{\textsc{cctp}}
    \label{fig:synthetic_conceptualization_ind_cctp}
  \end{subfigure}
  \begin{subfigure}{0.16\linewidth}
    \centering
    \includegraphics[width=\linewidth]{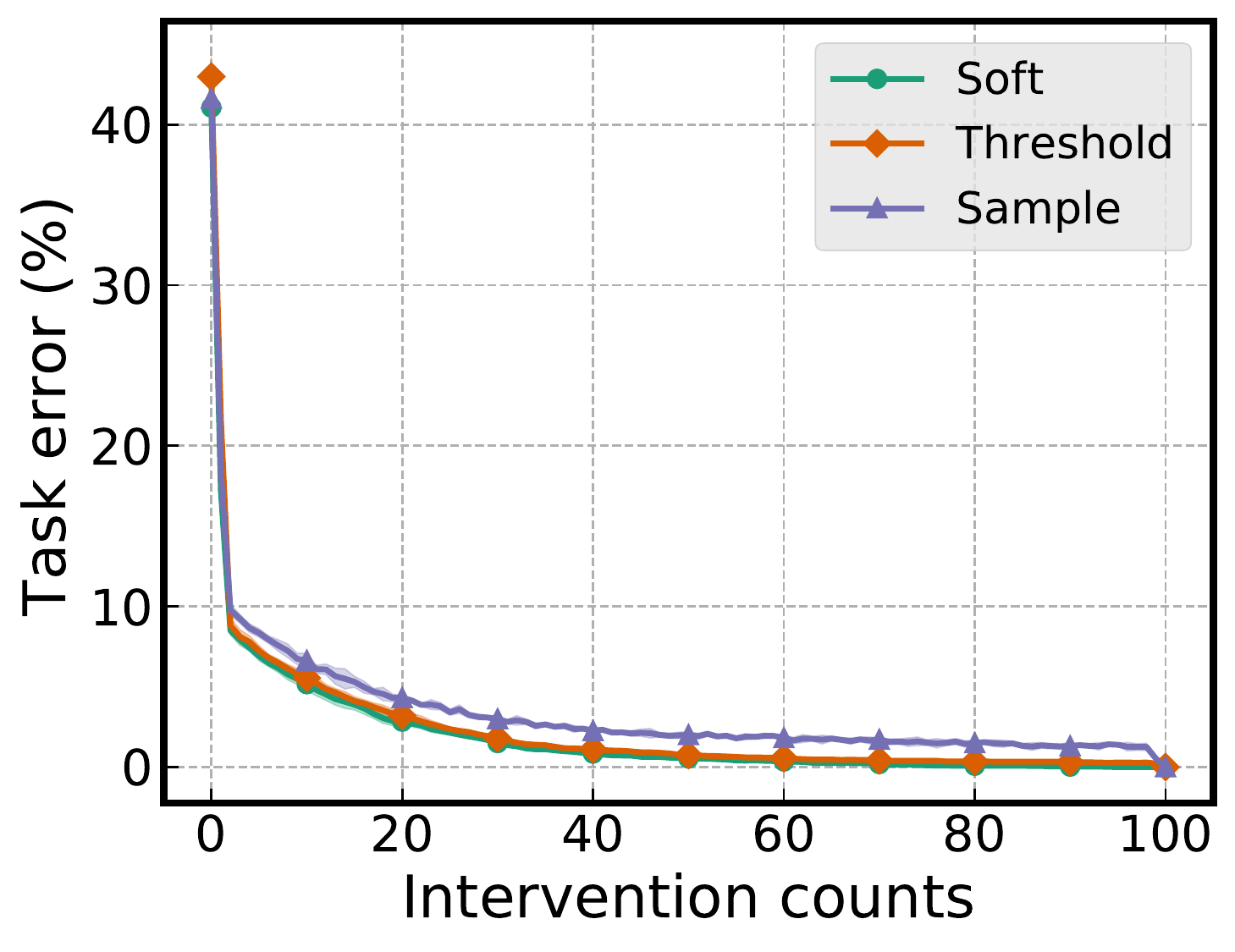}
    \caption{\textsc{ectp}}
    \label{fig:synthetic_conceptualization_ind_ectp}
  \end{subfigure}
  \begin{subfigure}{0.16\linewidth}
    \centering
    \includegraphics[width=\linewidth]{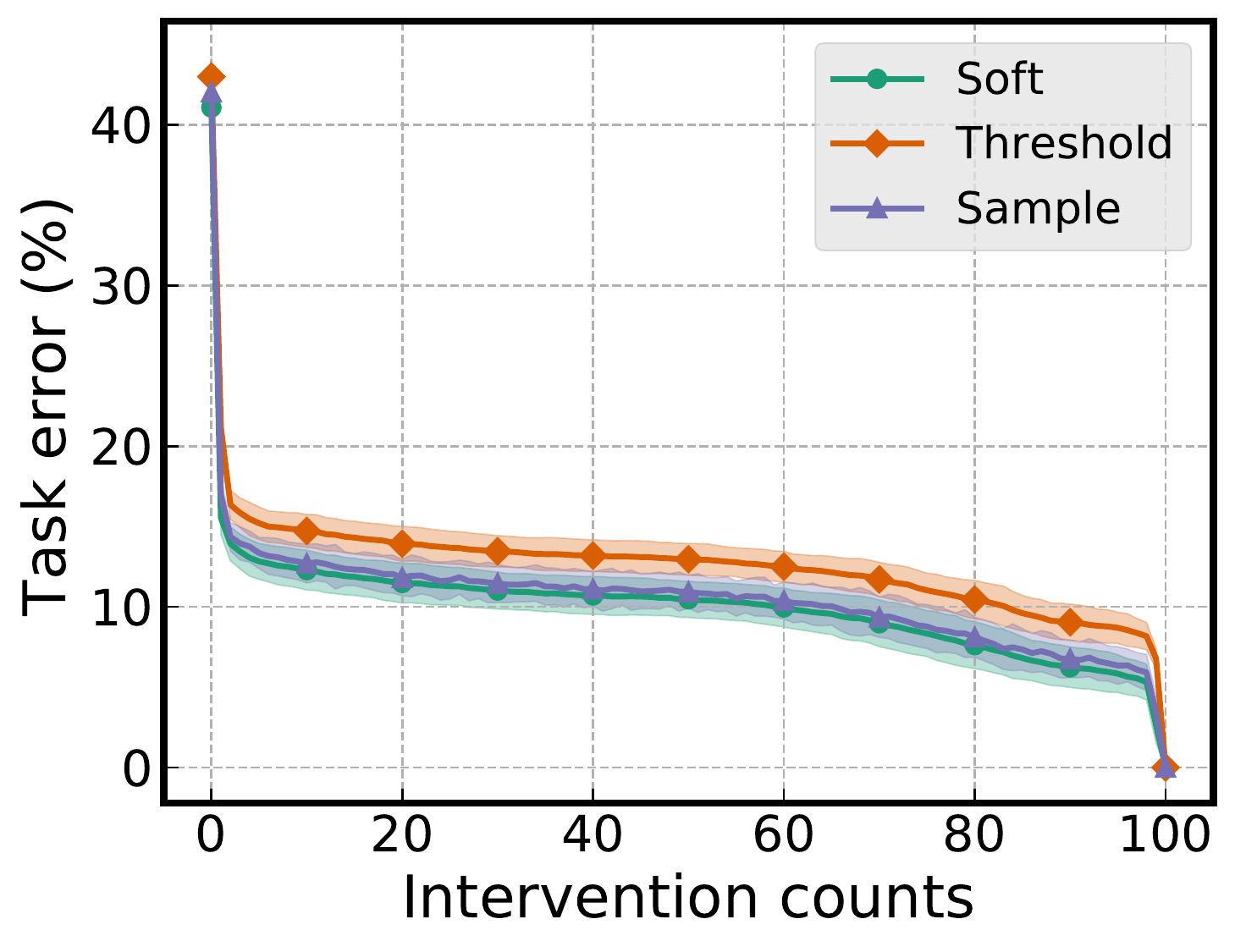}
    \caption{\textsc{eudtp}}
    \label{fig:synthetic_conceptualization_ind_eudtp}
  \end{subfigure}
  \caption{
    Intervention results under different conceptualization methods using various concept selection criteria.
    Here, we used \textsc{ind} training strategy for the synthetic dataset.
  }
  \label{fig:synthetic_conceptualization_ind}
\end{figure}

\begin{figure}[!th]
  \begin{subfigure}{0.16\linewidth}
    \centering
    \includegraphics[width=\linewidth]{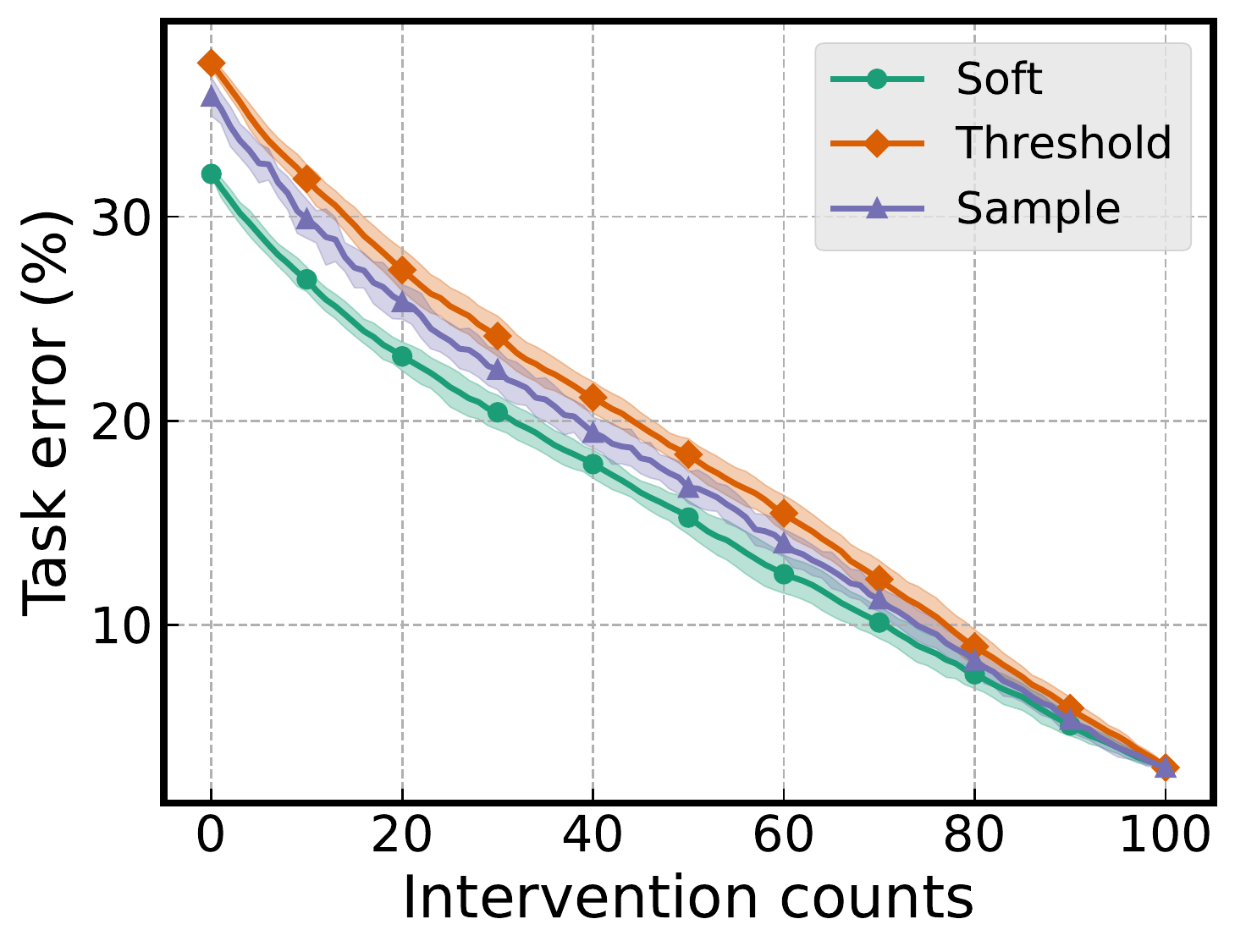}
    \caption{\textsc{rand}}
    \label{fig:synthetic_conceptualization_jntp_rand}
  \end{subfigure}
  \begin{subfigure}{0.16\linewidth}
    \centering
    \includegraphics[width=\linewidth]{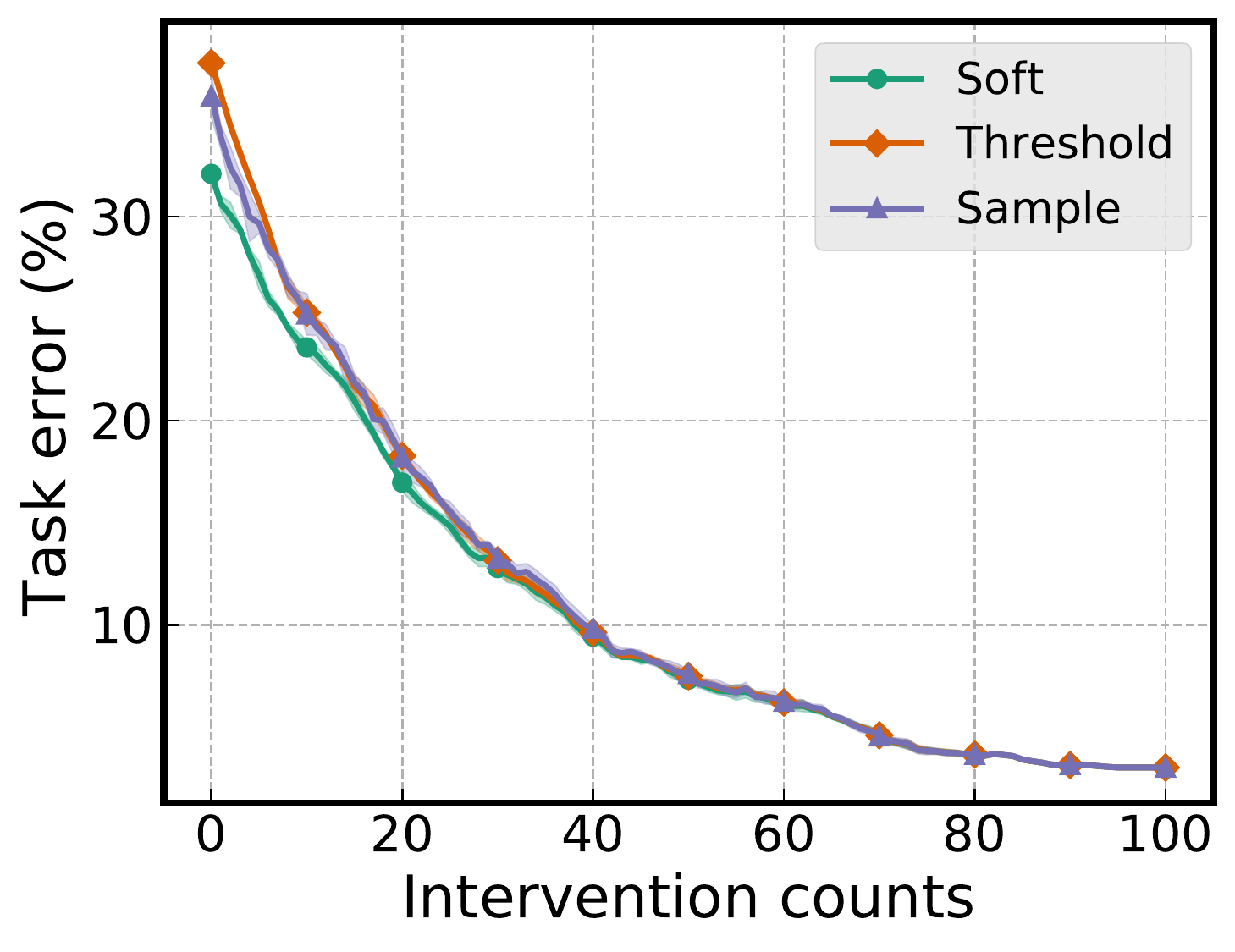}
    \caption{\textsc{ucp}}
    \label{fig:synthetic_conceptualization_jntp_ucp}
  \end{subfigure}
  \begin{subfigure}{0.16\linewidth}
    \centering
    \includegraphics[width=\linewidth]{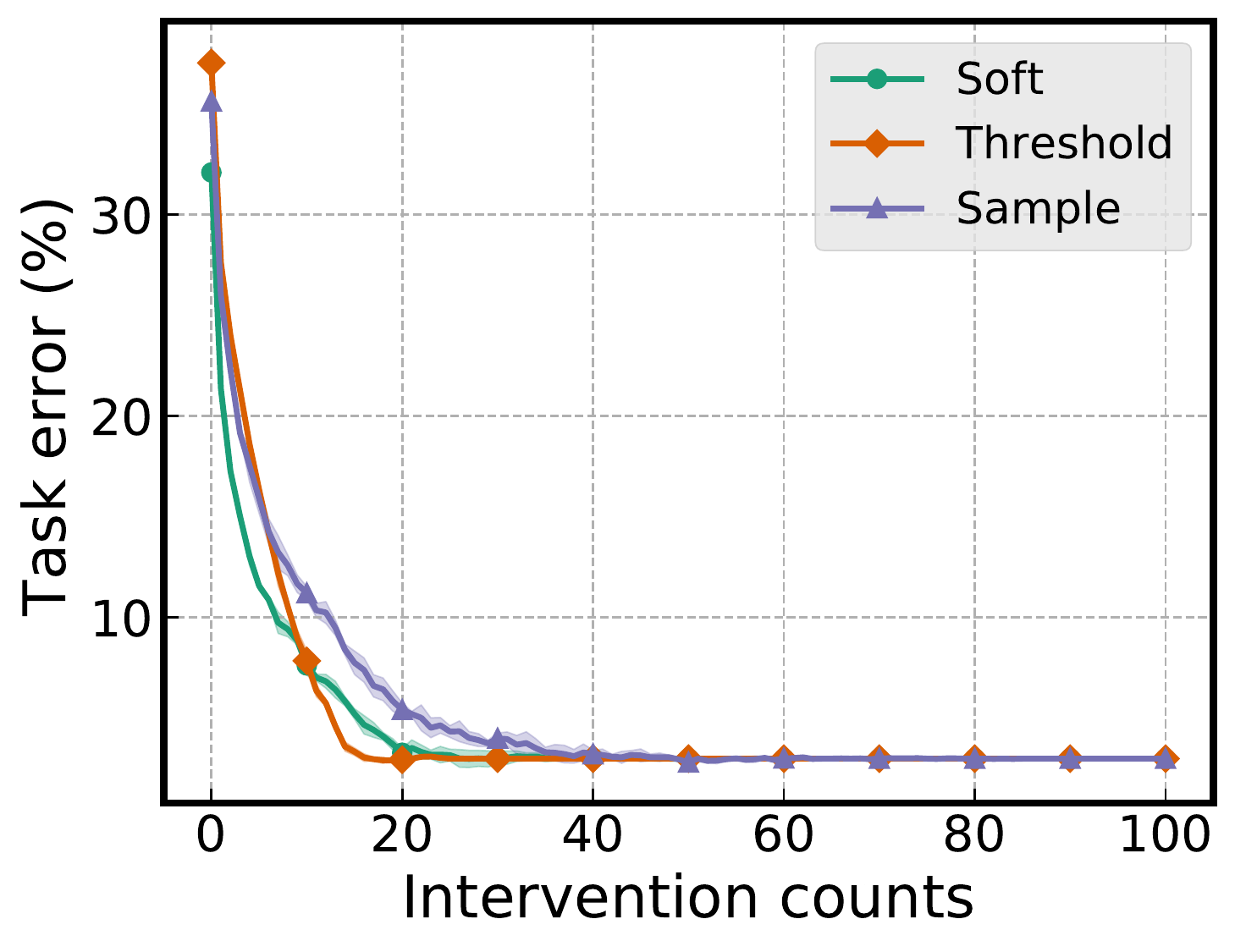}
    \caption{\textsc{lcp}}
    \label{fig:synthetic_conceptualization_jntp_lcp}
  \end{subfigure}
  \begin{subfigure}{0.16\linewidth}
    \centering
    \includegraphics[width=\linewidth]{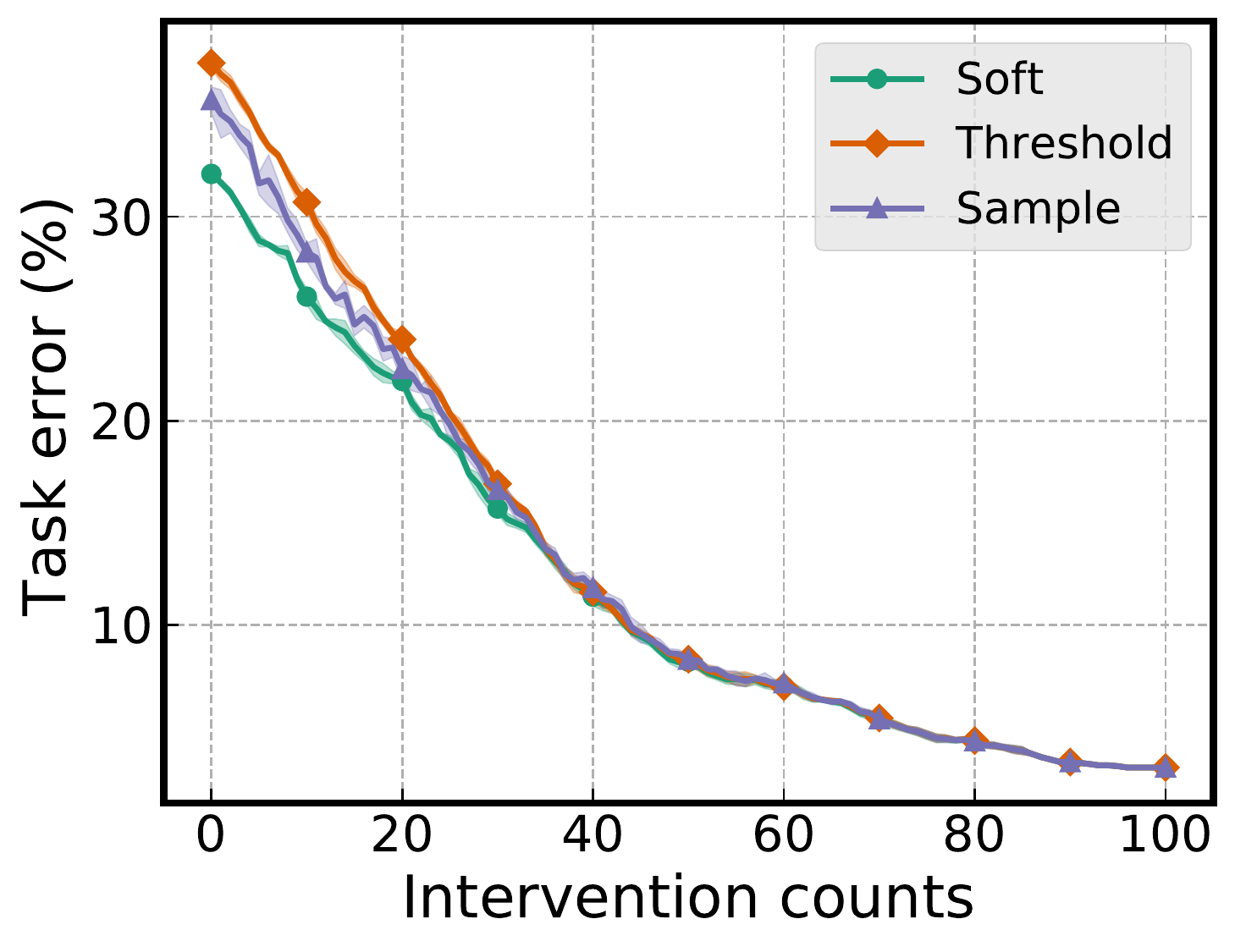}
    \caption{\textsc{cctp}}
    \label{fig:synthetic_conceptualization_jntp_cctp}
  \end{subfigure}
  \begin{subfigure}{0.16\linewidth}
    \centering
    \includegraphics[width=\linewidth]{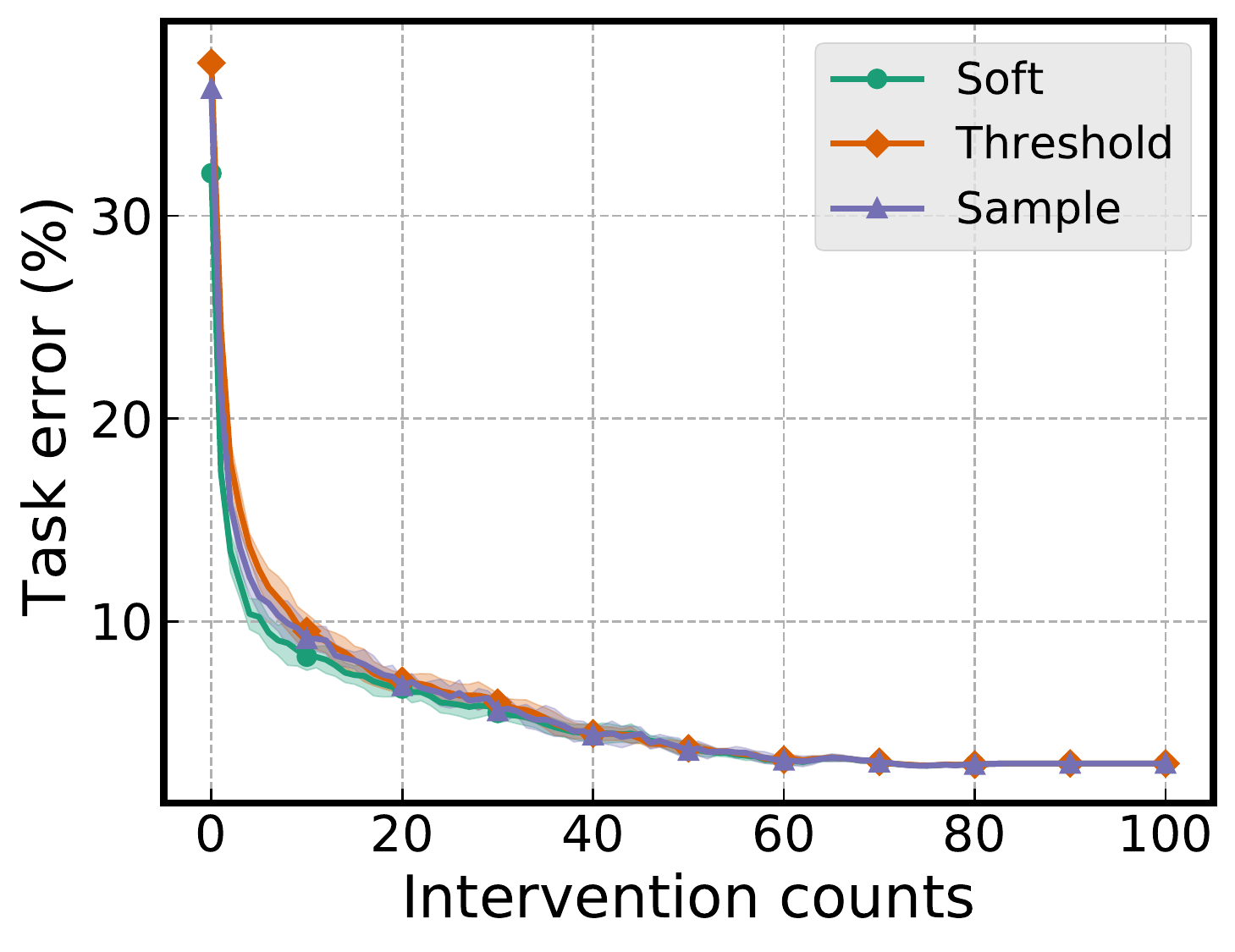}
    \caption{\textsc{ectp}}
    \label{fig:synthetic_conceptualization_jntp_ectp}
  \end{subfigure}
  \begin{subfigure}{0.16\linewidth}
    \centering
    \includegraphics[width=\linewidth]{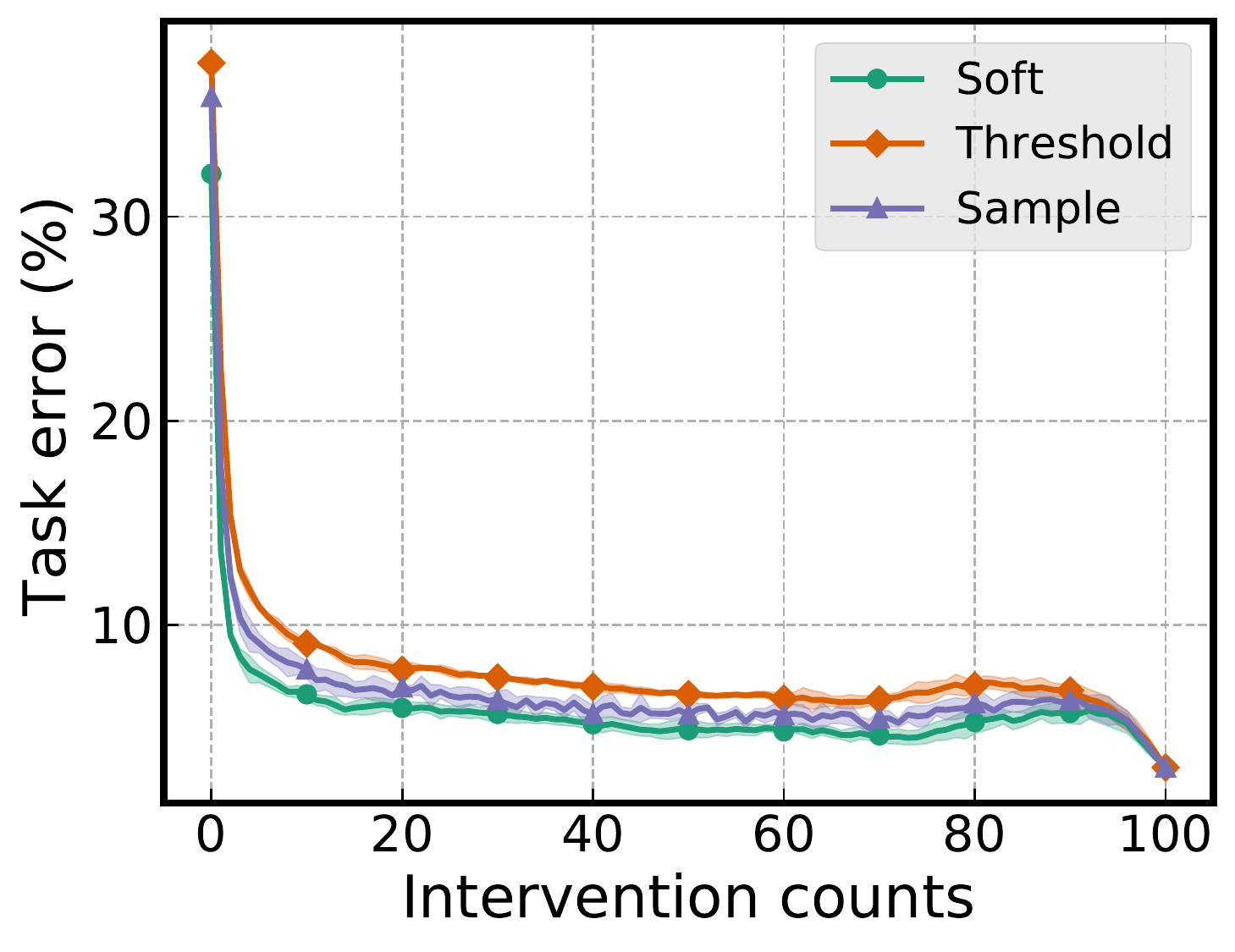}
    \caption{\textsc{eudtp}}
    \label{fig:synthetic_conceptualization_jntp_eudtp}
  \end{subfigure}
  \caption{
    Intervention results under different conceptualization methods using various concept selection criteria.
    Here, we used \textsc{jnt + p} training strategy for the synthetic dataset.
  }
  \label{fig:synthetic_conceptualization_jntp}
\end{figure}

Across all the datasets and concept selection criteria, utilizing effective criteria can reduce the gap between different conceptualization strategies much faster than \textsc{rand} criterion as seen in \cref{fig:cub_conceptualization_ind,fig:cub_conceptualization_jntp,fig:skincon_conceptualization,fig:synthetic_conceptualization_ind,fig:synthetic_conceptualization_jntp}.

\section{More Results on the Effect of Data on Intervention}

\label{sec:results-others-dataset}

\begin{figure}[!th]
\centering
  \begin{subfigure}{0.22\linewidth}
    \centering
    \includegraphics[width=\linewidth]{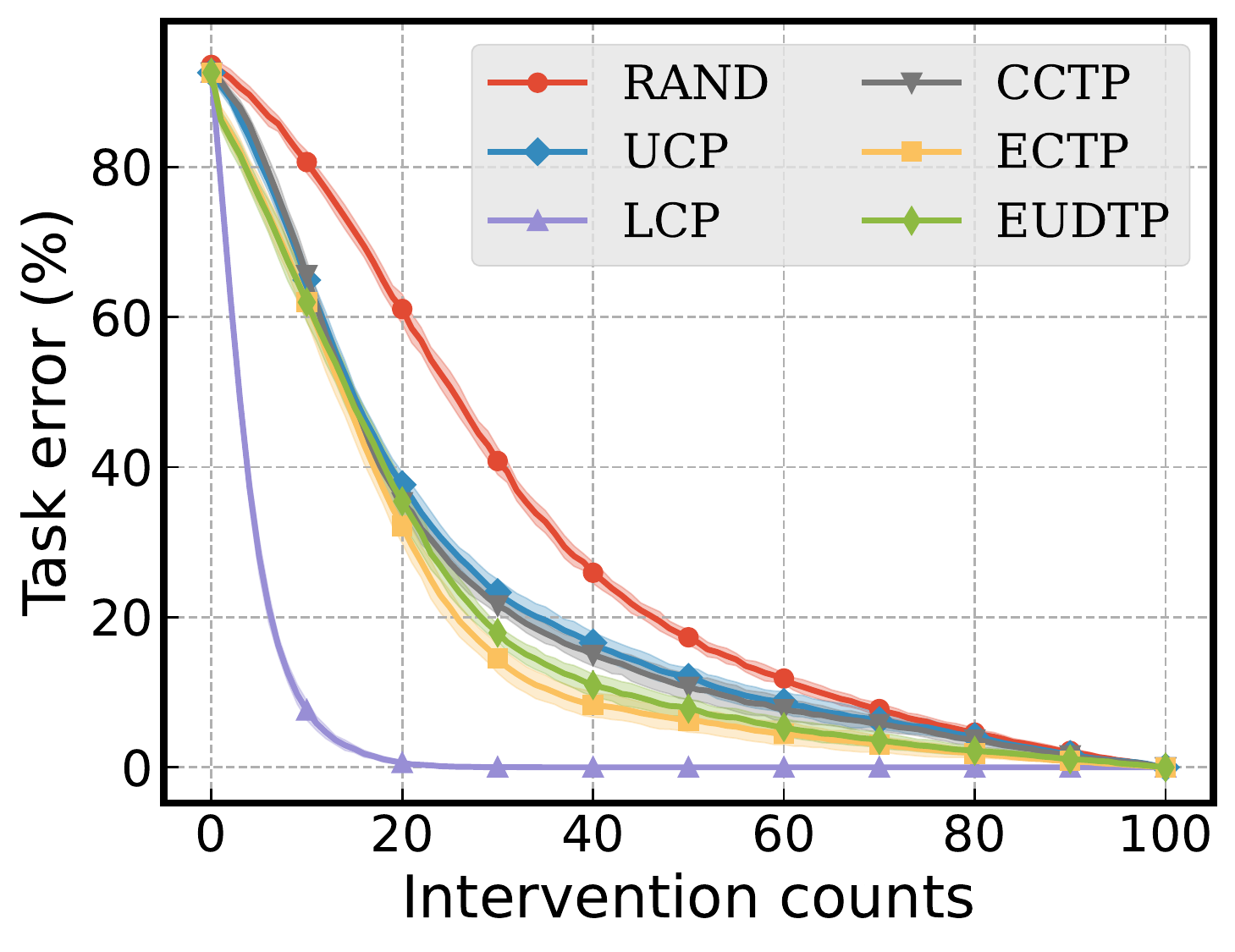}
    \caption{Data with extremely high input noise}
    \label{fig:synthetic_result_inputnoise2.0}
  \end{subfigure}
  \hspace*{10mm}
  \begin{subfigure}{0.22\linewidth}
    \centering
    \includegraphics[width=\linewidth]{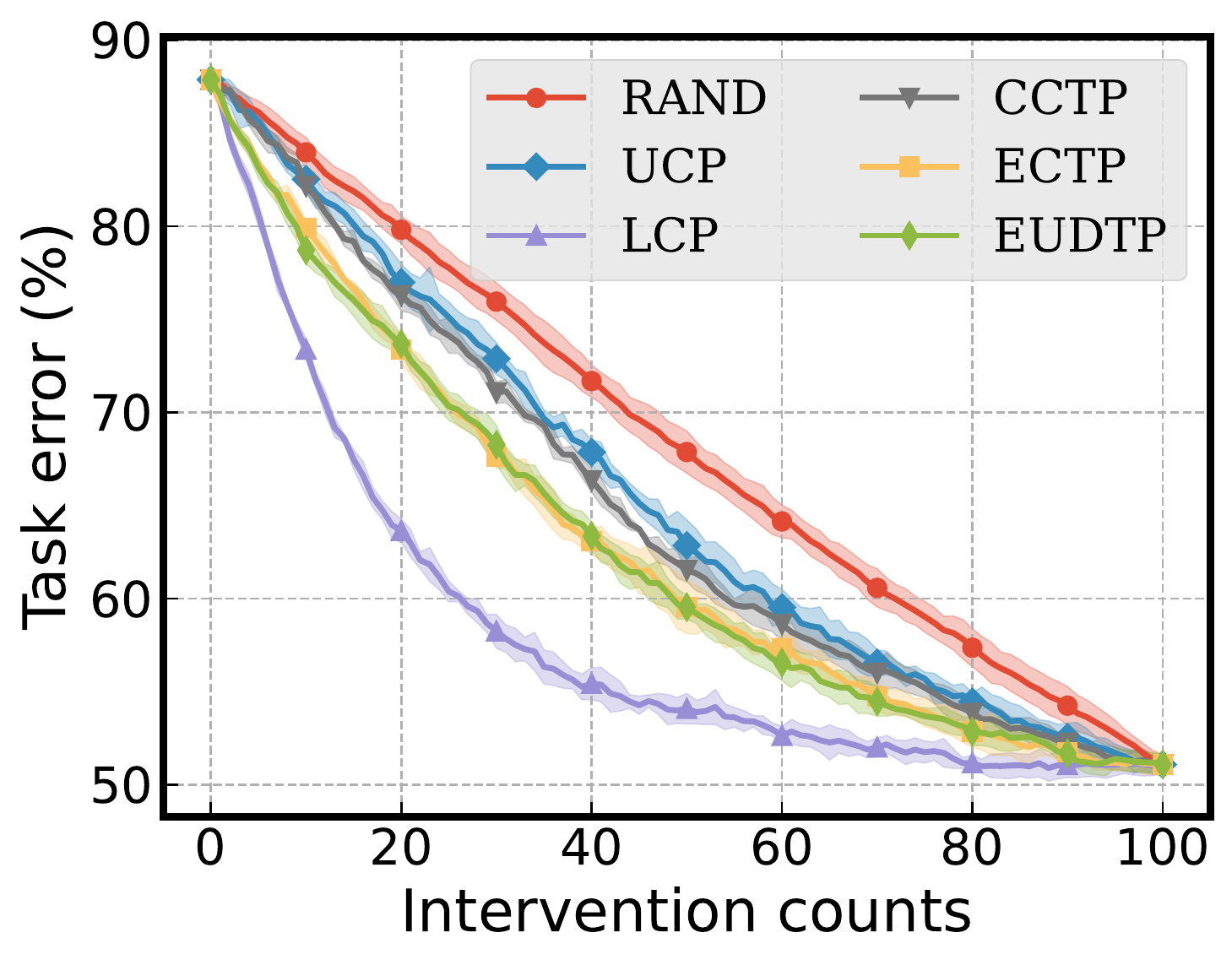}
    \caption{Data with extremely high concept diversity}
    \label{fig:synthetic_result_diversity3.0}
  \end{subfigure}
  \caption{Intervention results on the data with extremely high input noise (variance of $2.0$) or concept diversity (perturbation probability of $30\%$) respectively.
    In these cases, the proposed concept selection criteria work less effectively.}
    \label{fig:synthetic_result_extreme}
\end{figure}

\begin{figure}[!th]
\centering
  \begin{subfigure}{0.22\linewidth}
    \centering
    \includegraphics[width=\linewidth]{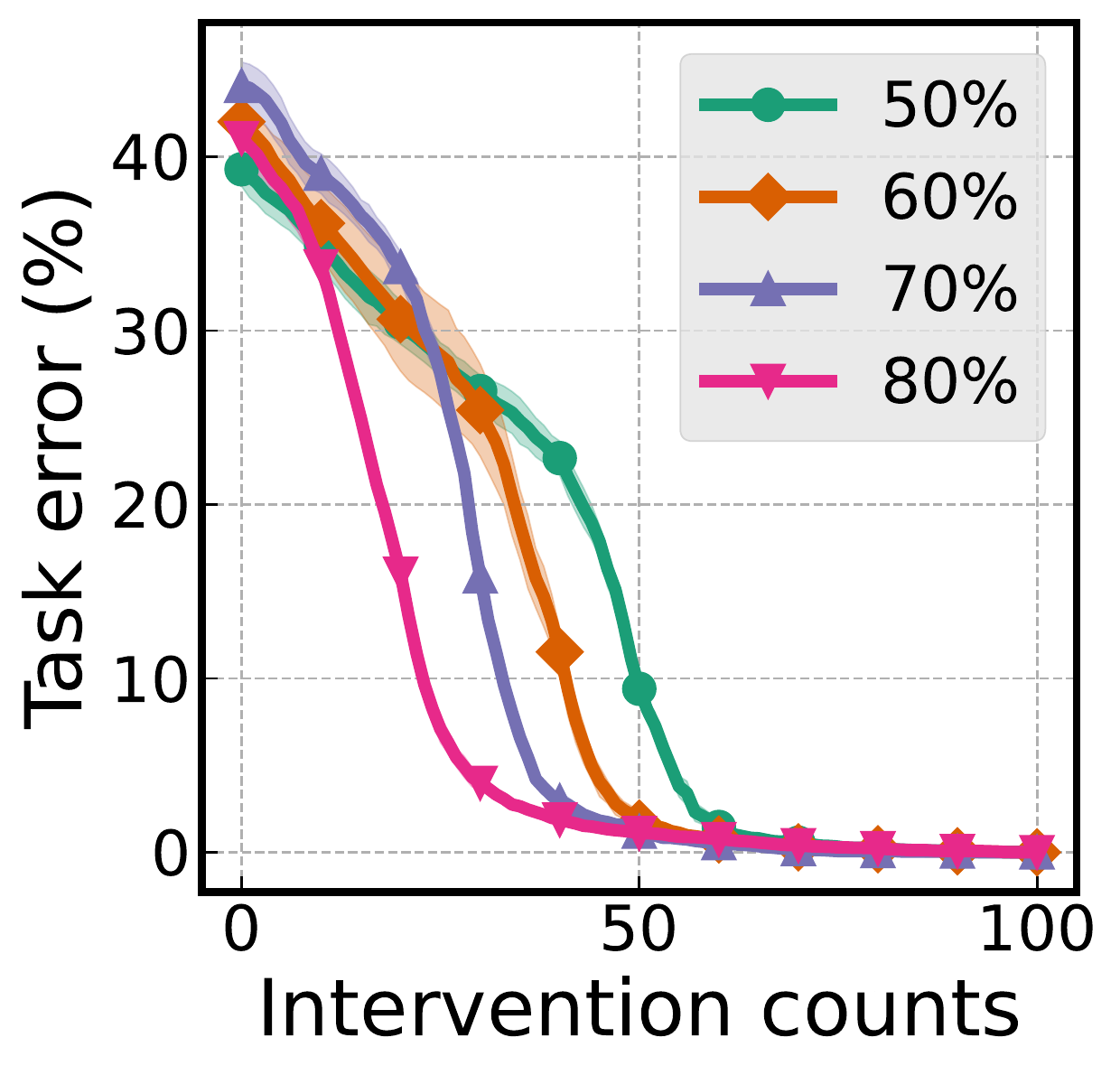}
    \caption{\textsc{cctp} with different concept sparsity levels}
    \label{fig:synthetic_sparsity_cctp}
  \end{subfigure}
  \hspace*{10mm}
  \begin{subfigure}{0.22\linewidth}
    \centering
    \includegraphics[width=\linewidth]{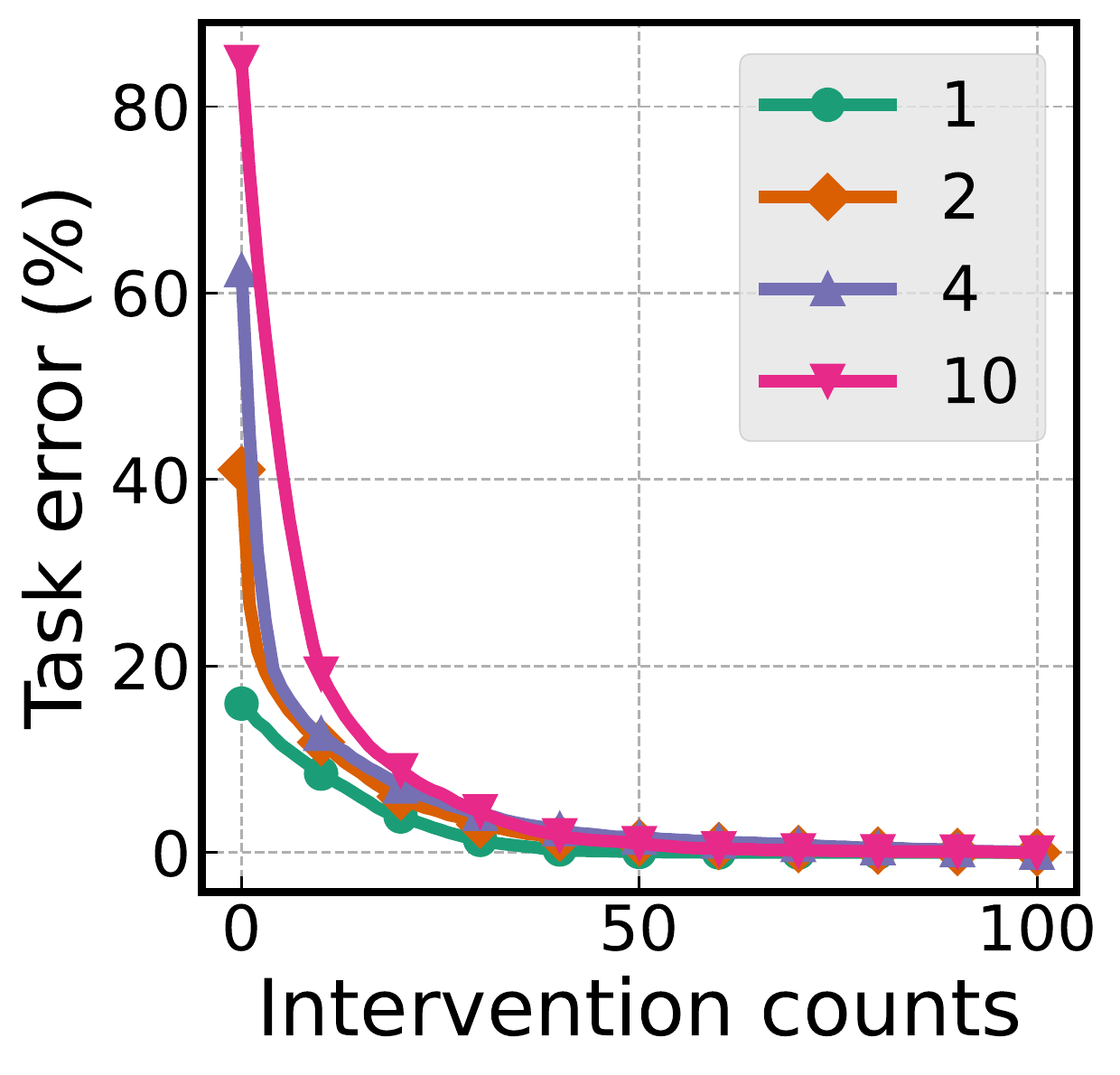}
    \caption{\textsc{ucp} with different sub-group sizes}
    \label{fig:synthetic_similarity_ucp}
  \end{subfigure}
  \caption{
    (a)
    \textsc{cctp} becomes more effective with a higher concept sparsity level.
    (b)
    Final task error increases, but intervention becomes more effective with larger sub-group sizes.
  }
  \label{fig:synthetic_characteristics_others}
\end{figure}

We find that intervention on data with extremely high input noise or extremely high diversity makes developed concept selection criteria less effective in general with a larger gap from \textsc{lcp} (see \cref{fig:synthetic_result_extreme}).
Specifically, \textsc{ucp} becomes less effective than other criteria in these cases.
We assume that concept prediction uncertainty is rather uncorrelated with concept prediction loss when the concept predictor $g$ achieves very low accuracy.

We also evaluate the effect of concept sparsity levels, \ie, probability of each concept having value $0$, using \textsc{cctp} criterion.
Note that intervention becomes less effective as the sparsity level gets closer to $50\%$ as seen in \cref{fig:synthetic_sparsity_cctp}.
To understand why, recall that this criterion aggregates the contribution of each concept to the target label prediction.
When the sparsity level is high and most concepts have value $0$, target prediction is determined by only a few concepts and \textsc{cctp} can work effectively by first intervening on the concept with the highest contribution.
In contrast, as the level gets closer to $50\%$, target prediction is determined by almost half of the concepts and contribution on target prediction becomes no longer a discriminative feature of the concepts, thus decreasing the effectiveness of the criterion.
Furthermore, we observe that the final task error increases but intervention becomes more effective with a large sub-group size $\gamma$ (see \cref{fig:synthetic_similarity_ucp}).
Specifically, we need $12$ intervention counts to decrease the task error by half for the data with $\gamma = 1$, but correcting $5$ concepts achieve the same effect for $\gamma = 10$.
This is because intervention can decrease the task error much faster for mis-classified examples by distinguishing from similar classes when $\gamma$ is large.

\section{More Results on Fairness of Majority Voting}

\label{sec:results-others-fairness}

\begin{figure}[!th]
\centering
  \begin{subfigure}{0.24\textwidth}
    \includegraphics[width=\linewidth]{figures/cub/mv_rand.pdf}
    \caption{\textsc{rand}}
    \label{fig:cub_mv_rand}
  \end{subfigure}%
  \hspace*{10mm}
  \begin{subfigure}{0.24\textwidth}
    \includegraphics[width=\linewidth]{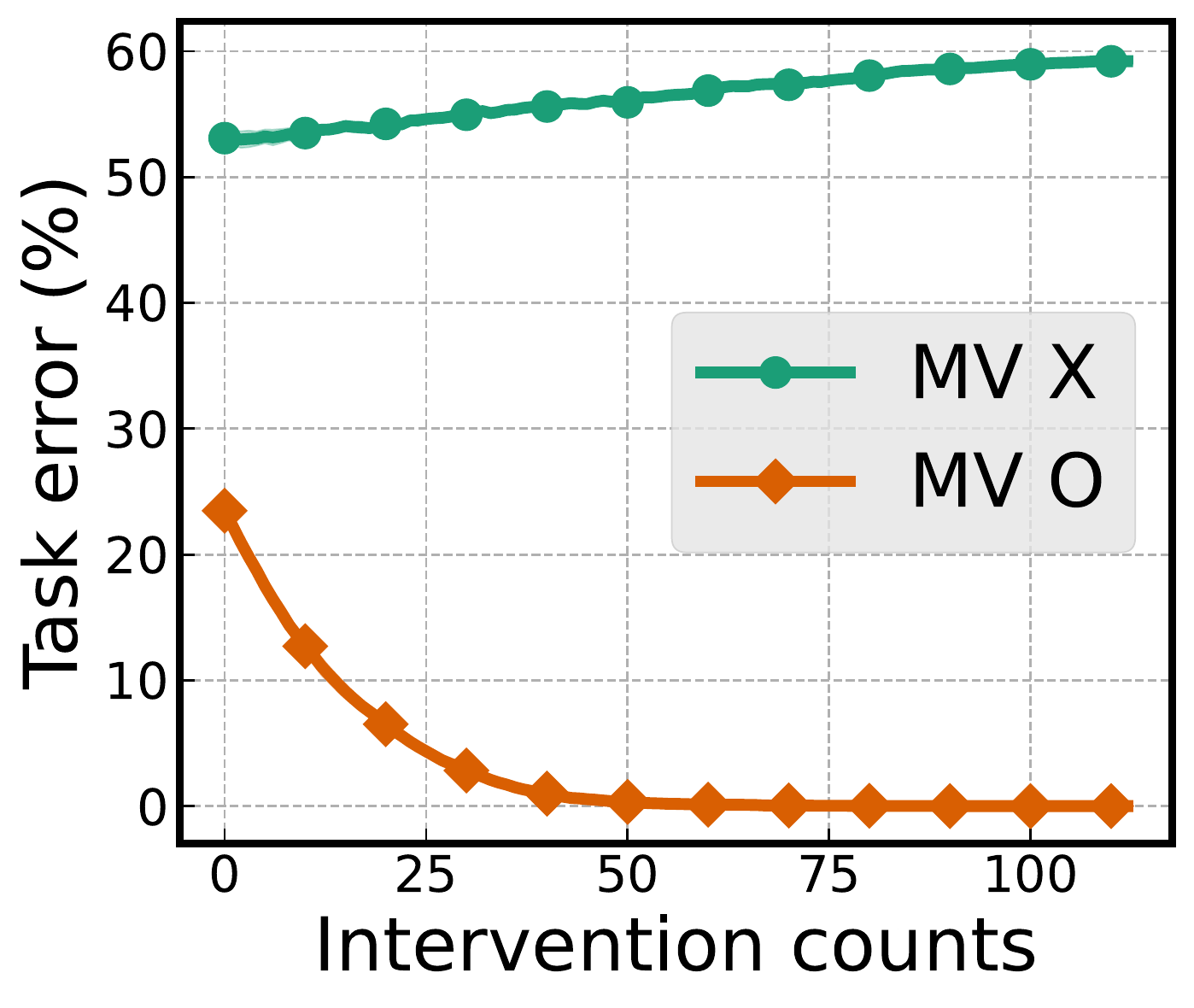}
    \caption{\textsc{ucp}}
    \label{fig:cub_mv_ucp}
  \end{subfigure}%
  \hspace*{10mm}
  \begin{subfigure}{0.24\linewidth}
    \includegraphics[width=\linewidth]{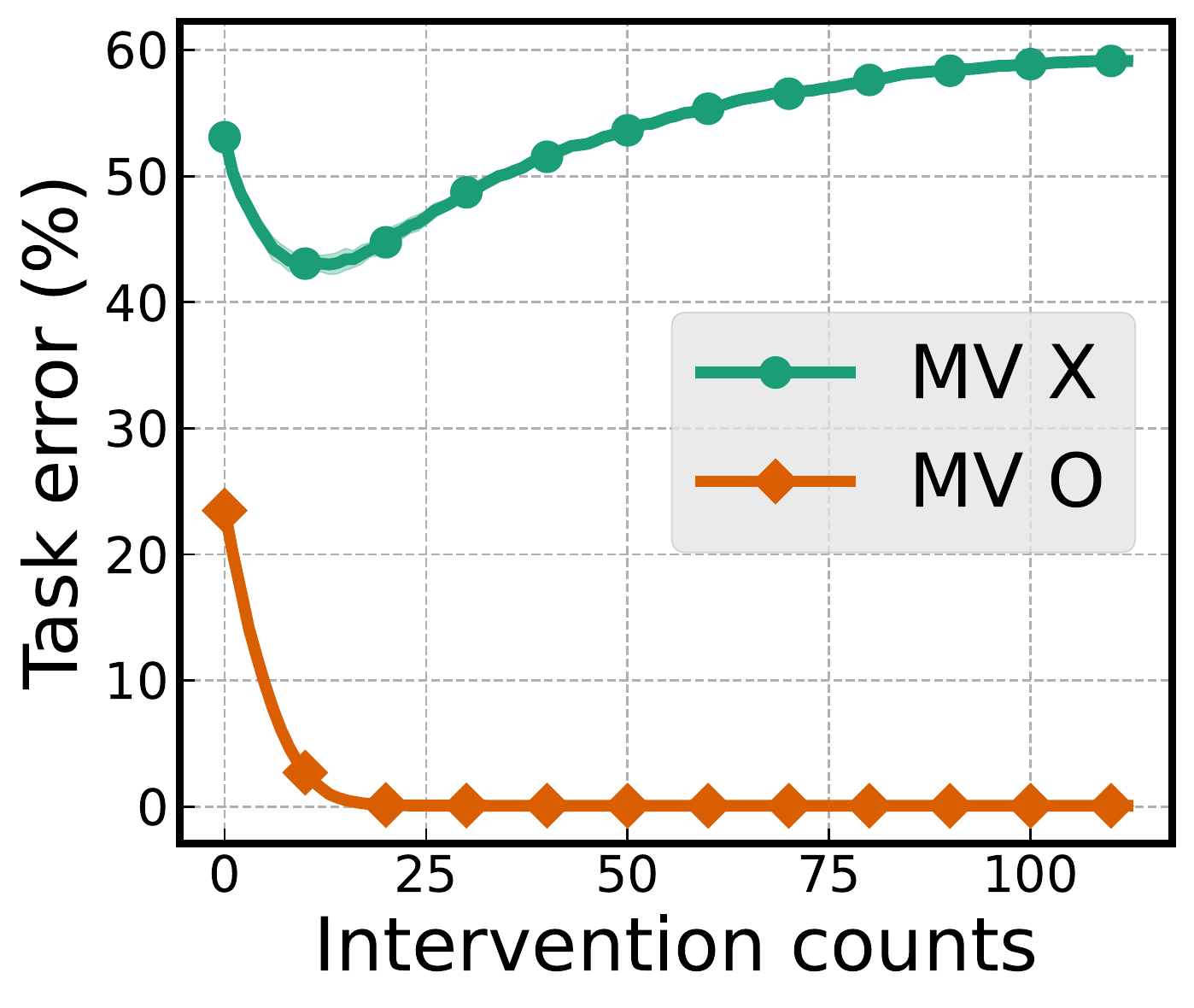}
    \caption{\textsc{lcp}}
    \label{fig:cub_mv_lcp}
  \end{subfigure}
  \caption{
    Comparison of test-time intervention results with and without using majority voting.
    }
  \label{fig:cub_mv}
\end{figure}

When we do not use majority voting on the CUB dataset, intervention rather increases the task error as seen in \cref{fig:cub_mv}.
Specifically, intervention does not decrease task error at all with \textsc{rand, ucp}.
Even with \textsc{lcp} criterion, intervention does not reduce the task error as much as when we use majority voting, and the error rather starts to increase after about $10$ concepts intervened.
See \cref{app:implementation} for the training details.

\end{document}